# Housing Potential Common Data Model and City Digital Twin

April 27, 2026

Megan Katsumi, Mark Fox, Anderson Wong, Divnoor Chatha

Urban Data Research Centre

School of Cities

University of Toronto

Contact: megan.katsumi@utoronto.ca

## Executive Summary

This report is intended for HICC project partners, along with the wider community of housing potential stakeholders—including researchers, municipal planners, and developers—who may seek to adopt or implement the common data model. The purpose of this report is to communicate the results of the "Common Data Model for Housing Potential Analysis and Visualization" project. Key outcomes of this work are the specification of requirements for a housing potential data model, definition of the Housing Potential Common Data Model, demonstration of how the model is implemented to create a City Digital Twin for housing, and development of a pilot application that leverages the digital twin. The development process also led to the identification of technical and administrative challenges and corresponding mitigation strategies, and recommendations to promote the adoption of the model.

The goal of this project was to ascertain the data requirements to support housing potential analysis and develop a common data model – the Housing Potential Common Data Model (HPCDM) – that will serve as a standard to support the integration and interoperability of housing potential related data across a variety of diverse datasets. The model is intended to enable the consistent evaluation of the degree of housing potential of a particular site. The requirements were gathered in Phase 1 of the project through the specification of use cases and curation of a catalogue of (360+) housing potential datasets.

Phases 2 and 3 focused on development of the model. Evaluation results demonstrated that the model addressed 68 of the 71 requirements derived from the housing potential use cases; the remaining three requirements were omitted because they fell outside of the scope of the core model. The model's representation ability was also assessed with the specification of mappings for relevant datasets in Toronto, Halifax and Vancouver. Finally, the model was validated through review by a panel of experts.

Phase 4 focused on implementation. To demonstrate the practical utility of the model, a knowledge graph-based City Digital Twin (CDT) was developed to integrate the housing potential data for the City of Toronto. This CDT serves as the data source for a pilot housing potential dashboard, illustrating how the model can be applied to enable real-world tools for housing potential analysis. This phase also served to provide insights into challenges and considerations for adoption of the model.

This report recommends three key areas of focus to promote adoption of the standard: (1) education – efforts must focus not only on promoting awareness of the standard, but understanding of the model's content and how it can be used; (2) resources – providing immediate value for data that is formalized using the model (e.g., in the form of an HPCDM-based tool); and (3) support – sharing resources to support users in understanding the model as well as streamlining development (e.g. with software libraries).

# Acknowledgements

This work was made possible with the support of Housing Infrastructure Communities Canada. Resources to support the digital twin and dashboard implementation were contributed by the Digital Research Alliance of Canada and Tata Consultancy Services. We also gratefully acknowledge the work of Chloe Loh, Chiara Nuvoloni, Jenny Wei, and Doris Tian for their contributions in surveying and cataloguing relevant data.

# Table of Contents

# 1 Introduction

"Housing Potential" refers to the capacity of a specific area, parcel of land, or existing building to accommodate new residential development or additional housing units. Research into housing potential generally focuses on mechanisms that might "unlock" housing potential, often looking at a single mechanism such as zoning changes, land banks, tax relief, etc. But to truly determine the potential for housing, one must evaluate a location from multiple perspectives, including zoning, land and building costs, population characteristics and needs, access to services such as transportation, education, food, health, etc. These pieces of information often exist in silos: they are spread across a multitude of datasets (if available at all), and do not share a common data model. In many cases this makes it difficult to integrate or link the data, and to represent a more holistic and consistent assessment of housing potential within and across communities.

The goal of the “Common Data Model for Housing Potential Analysis and Visualization” project was to ascertain the data requirements to support housing potential analysis and develop a common data model (the Housing Potential Common Data Model (HPCDM)) that will serve as a standard to support the integration and interoperability of housing potential related data across a variety of diverse datasets. This is intended to enable the consistent evaluation of the degree of housing potential of a particular site. While the data model itself does not perform the actual *analysis* to produce an assessment of housing potential, it enables a consistent representation of the various factors that serve as input to such an assessment, such as physical and regulatory constraints, available services and infrastructure and so on. The model is itself agnostic as to *how* housing potential should be assessed. This is important to ensure that it can be applied generally by different stakeholders in different contexts. Instead, it supports consistent assessment by providing a common language through which data may be described, integrated, and compared.

To demonstrate the capabilities of the model, a knowledge graph-based City Digital Twin (CDT) was developed to integrate the housing potential data for the City of Toronto. A City Digital Twin is a “…a virtual representation of a dynamic city that allows users to exploit the interrelationships between physical infrastructure, the natural environment, and people of a city scale.” [1].

In this project, the Toronto CDT was created using a knowledge graph-based data model. A knowledge graph-based CDT is one that is implemented using a graph-based data model of nodes and edges that supports the interconnection or “linking” of data sources into a single view. *Knowledge* graph implies the inclusion of semantics rather than only raw data points. This addresses the issue of data silos, allowing users to understand how information from different sources relates to one-another. This is achieved through the annotation of the content of the graph with a well-defined vocabulary. The vocabulary used for the Toronto CDT is the HPCDM. This supports a deep understanding of the data, alignment of data from heterogeneous sources, and a common language to access (query) the data.

The project was structured in four phases: Phase 1 [2] involved the identification of use case-based requirements for the HPCDM, and the curation of housing-related data sources across Canada into a central data catalogue within the Canadian Urban Data Catalogue[1]. The use cases and requirements were identified in cooperation with an advisory panel of experts. They would serve as a guide for the development and evaluation of the model in the subsequent phases. The catalogue has grown to include 360+ datasets and can be expected to continue to grow as work continues in the future. The data model was developed, based on international standards for city data [3] [4] [5], in Phases 2 and 3 [6], [7] and evaluated against the specified requirements. Its general suitability across different municipalities and data models was demonstrated through the specification of example mappings from data sources in Toronto, Vancouver and Halifax, into the HPCDM. Finally, this report focuses on Phase 4, the focus of which is the implementation of the HPCDM in a city digital twin.

---

[1] https://data.urbandatacentre.ca/catalogue/?q=tags%3A%22Housing+Potential%22

In this final phase, a knowledge graph-based city digital twin was implemented using the HPCDM. A pilot interface was then developed on top of the digital twin, with the goals of: (1) demonstrating what is enabled with the adoption and implementation of the HPCDM – both the digital twin and the data that it integrates and the kind of applications that it is capable of supporting, and (2) providing insight into considerations for implementation and adoption of the HPCDM that may be shared to benefit potential users of the HPCDM. The overarching goal of this project is the development of a common data model for housing, the purpose of this phase is to complement the model with a concrete example of how it may be implemented in practice, and in doing so provide useful insights to support its adoption.

This report describes the collective outcomes of all four project phases and highlights insights to encourage adoption of the model. It is structured as follows: Section 2 provides some background on housing potential, knowledge graphs and city digital twins; Section 3 describes the process of identifying the requirements for the model; Section 4 presents the HPCDM including details on its evaluation; Section 5 explains how the model was implemented to create a city digital twin knowledge graph and describes the development of a pilot housing potential tool as an interface to the digital twin; Sections 6 and 7 provides general guidelines for implementation, and considerations for adoption of the HPCDM; Section 8 highlights the challenges and limitations for its application; and finally, Section 9 concludes with the identification of key areas to be pursued in future work.

# 2 Requirements for a Common Data Model

To understand the requirements for a common data model for housing potential requires an understanding of the kind of questions that stakeholders need to ask of the CDT and the data that needs to be represented. This informs the scope of the model as well as the ways in which the concepts are represented. To achieve this, the project began with the development of a set of housing potential Use Cases and the curation of a catalogue of housing potential-related datasets.

The data requirements were drawn from Use Cases, informed by a panel of experts in housing potential analysis. A Use Case describes a specific scenario. It identifies who is performing the housing potential evaluation, their role in the organization they belong to, and the goals they have. Critically, it explains *how* they go about performing the evaluation. These requirements were elaborated with “competency questions” (CQs) that specify the questions a supporting database or decision support system should have to have the capability to answer (i.e., competence) to enable the Use Cases.

To answer these questions, the relevant data must be available and transformed and integrated into the common data model. The creation of a catalogue of housing potential data is a first step toward this. The catalogue identifies the real-world datasets that may be used to inform housing potential analysis. In doing so, it provides an immediate benefit to stakeholders by providing a single point of access for relevant datasets. This data curation effort began with a broad view of all data that could be used to inform the various perspectives of housing potential. Once the Use Cases and CQs were defined, they were used to guide the data collection effort in a targeted manner, as well as to identify gaps in the available data. In later stages of the project, the catalogue was be used to identify datasets to be mapped and integrated into the common data model. The mapping of this data serves to demonstrate and validate the design of the common data model and is later used to enable the implementation of the City Digital Twin.

## 2.1 Housing Potential Data Catalogue

The data catalogue was created using the Canadian Urban Data Catalogue (CUDC)[2]. The purpose of the CUDC is to provide awareness of the vast array of Canadian data by providing an open catalogue of Canadian datasets. It catalogues both open and closed datasets, along with data accessible via web services. The CUDC is a living artifact with entries continuously updated via bulk imports and manual entry.

### 2.1.1 Curation methodology

The search for relevant datasets was conducted in two phases. The preliminary phase began in advance of the specification of use cases and CQs. The curators[3] performed a broad search using keywords related to key aspects of housing potential: zoning and land use, development feasibility, redevelopment opportunities, infrastructure and services, and environmental and community constraints. In addition, questions related to these topics (in the context of housing potential) were formulated, and keywords were extracted from these questions to expand the search. Initially focused within the CUDC, which at the time contained over 44,000 Canadian datasets[4]. The search was then expanded to other government data portals, including municipal open data portals and other organizations such as the Canada Mortgage and Housing Corporation and Environment Canada.

The second, more targeted, phase of search was informed by the specification of requirements described in the previous section. The CQs enabled a more precise identification of necessary data and highlighted gaps in the existing data catalogue. In this phase, the CUDC was re-examined to ensure that no datasets were overlooked, followed by a consultation of municipal, provincial, and federal data sources (specifically, those not yet imported into the CUDC). With a more precise understanding of the requirements, the curators also consulted topic-specific databases (e.g., Statistics Canada for demographic data, CMHC for housing market data). To build a comprehensive catalogue of data, the search began with a geographic focus on Ontario (primarily, Toronto). The search was then expanded to other municipalities and provinces – specifically with a focus on Halifax and Vancouver.

### 2.1.2 Data Curation Results

While data that is relevant for housing potential analysis is remarkably diverse, most identified datasets fall into the following four categories:

- **Land boundaries:** Land boundary datasets define the geographic areas for housing potential assessment. These primarily consist of municipal parcel data showing property divisions, neighborhood boundary definitions, municipal jurisdictional boundaries, and regional planning districts. These datasets establish the spatial framework upon which all housing potential analysis depends. This data is often included with other data (e.g., part of a zoning dataset), but can also be published on its own such as with the Urban Boundary 2031 - Peel Region dataset.

[2] https://data.urbandatacentre.ca/

[3] Data curation work was performed by Chloe Loh, Chiara Nuvoloni, Jenny Wei and Doris Tian

[4] This number has since grown to over 85,000.

- **Land use:** Land use datasets include representations of current and planned policies (zoning). In addition, there are datasets on the actual, physical use (e.g., buildings, parks) of properties as well as potential future use (e.g., development applications). Together, these datasets provide crucial context for understanding what is currently built, what is permitted, and what changes may be coming. One example is the Zoning By-law dataset for the City of Toronto.
- **Development feasibility:** These datasets capture information related to the practical aspects of housing development. They include legal constraints beyond zoning such as heritage designations and environmental protections, infrastructure considerations such as road networks and utilities, economic indicators including housing market reports and construction costs, and geological factors like flood plains and soil conditions. These datasets collectively help determine where development is physically possible and economically viable. One example is the Housing Market Information Portal, which provides current housing market data for Canada.
- **Quality of housing:** This category includes a wide range of datasets that can be used to assess the livability and desirability of potential housing locations. There are datasets on environmental factors like air quality, and data on locations of essential services and amenities to support the assessment of the neighbourhood. One example is the Pedestrian infrastructure dataset, which provides insight into the walkability of an area.

### 2.1.3 Quantitative Summary

A quantitative summary of the findings, aggregated across several key attributes, is provided below. More detailed observations on data coverage may be found in Section 3.3 on data mappings. This summary may be used to inform the continued curation effort as to what resources (organizations, types of data) may be further examined for additional resources. Table 1 provides counts of the top ten tags found in the data catalogue. Tags are terms that are added to the datasets to provide a characterisation of the dataset. Each dataset may contain multiple tags. Tags may be generated automatically on dataset import or defined manually via user input. There is no restriction on the tags that may be specified. In contrast, Table 2 provides a listing of the top ten “Domain/Topic” labels defined within the catalogue. The Domain/Topic field provides a general characterisation of the dataset and is intended to be selected from a restricted set of terms, defined within the CUDC. Each dataset may have only one Domain/Topic, though a considerable number of datasets did not have a defined Domain/Topic. A wide variety of domains can be observed; this is unsurprising given the varied nature of relevant data. Table 3 summarizes the top ten areas that the datasets cover which, as expected, is currently focused on Toronto and Ontario. Here too, a significant number of datasets were missing location metadata. Table 4 provides the total number of datasets by authoring organization. These results are summarized in Table 5, which provides a count based on organization type.

*Table 1: Summary of top 10 tags[5] found in the data catalogue.*

| Tags | Count |
|---:|---:|
| Economics and Industry | 33 |
| Construction | 26 |
| Government information | 22 |
| Planning Development | 20 |
| Community infrastructures | 19 |
| Boundaries | 18 |
| Development Feasibility | 17 |
| Land Use | 15 |
| Economy and Business | 14 |
| Government and Politics | 13 |
| Nature and Environment | 11 |

*Table 2: Top 10 topics covered by the data catalogue.*

| Topic | Count |
|---:|---:|
| Infrastructure | 52 |
| Housing | 40 |
| Boundaries | 29 |
| Economy | 25 |
| Developments and infrastructure, Construction | 19 |
| Geoscientific Information | 42 |
| Permits and licenses | 11 |
| Construction | 9 |
| Transportation | 7 |
| City government | 5 |

*Table 3: Top 10 Locations described by the catalogue.*

| Area | Count |
|---:|---:|
| Vancouver | 70 |
| Halifax | 39 |
| Ontario | 22 |

[5] Omits the "Housing Potential" tag as well as tags that are not topic-related (e.g. "Halifax Open Data", "table").

| | |
|---|---|
| Toronto | 22 |
| Brampton | 15 |
| Canada | 14 |
| Peel Region | 12 |
| Oakville | 10 |
| British Columbia | 5 |
| Mississauga | 5 |

*Table 4: Top 10 Organizations publishing data found in the catalogue.*

| **Organization** | **Count** |
|---|---|
| City of Vancouver | 73 |
| Halifax Regional Municipality | 38 |
| Halifax Data Mapping and Analytics Hub | 34 |
| City of Toronto Open Data | 29 |
| Government of Ontario | 28 |
| Statistics Canada | 22 |
| City of Brampton | 15 |
| Canada Mortgage Housing Corporation | 13 |
| City of Toronto | 13 |
| Peel Region | 10 |

*Table 5: Summary of organization types publishing data found in the catalogue*

| **Organization Type** | **Organization Count** | **Dataset Count** |
|---|---|---|
| Municipal | 11 | 222 |
| Provincial | 9 | 59 |
| Federal | 8 | 34 |
| Regional | 2 | 11 |
| NGO (Non-Commercial) | 4 | 14 |
| Commercial | 2 | 5 |

### 2.1.4 Dataset Listing

The complete dataset catalogue may be accessed on the CUDC using the tag "Housing Potential".

## 2.2 Use Cases and Competency Questions

The data requirements were drawn from Use Cases that were developed in collaboration with a panel of experts in housing potential analysis. A Use Case describes a specific scenario. It identifies who is performing the housing potential evaluation, their role in the organization they belong to, and the goals they have. Critically, it explains *how* they go about performing the evaluation. Each Use Case was then used to formulate Competency Questions (CQs). These specify the questions a supporting database or decision support system must be capable of answering to enable the use case. The CQs serve to guide the design of the ontology by identifying the required concepts and intended semantics with greater precision. Then, during the evaluation stage, the formalization of the CQs (along with anticipated answers) using the HPCDM serves as an important part of the model's verification and validation.

The Use Cases and their associated CQs are detailed below. The CQs are divided into three main categories: (1) land use and zoning, (2) development feasibility, and (3) development desirability. The HPCDM is designed to define the core concepts across all three categories and is presented in Section 3. Note that some housing market CQs were identified but excluded from the scope of the model. Representation of the market is important, but should be pursued as a separate, specialized extension to the HPCDM.

### 2.2.1 Use Case 1

**Personae**: Affordable Housing advocate

Joe is an advocate for the creation of more affordable housing in the City. His role is to identify parcels of vacant land that can be used for building affordable housing. He must identify vacant land, but also any restrictions on its use for housing.

**CQs:**

**Land Use/Zoning**

1. Where in the city does there exist vacant parcels of land?
    a. What is the size of the parcel?
    b. What is the perimeter of the parcel?
2. Who owns parcel x?
3. What use is parcel x zoned for, e.g., residential (single family, multifamily), commercial, mixed, industrial?
    a. What is it currently being used for?
    b. What is the current density of the neighbourhood?

**Development Feasibility**

4. Does there exist any issues with the parcel, such as prior use (e.g., gas station, industrial waste) requiring remediation?

5. Is the parcel accessible directly by road?
    a. Fire and emergency access
6. Is the parcel serviced?
    a. Water
    b. Sewage
    c. Electricity
    d. Natural gas
7. What regulations exist that limit construction on the parcel?
    a. Setbacks from road or adjacent parcels?
    b. What is the FSR (Floor Surface Ratio)?
    c. What height restriction exists?
    d. Minimum lot size?
    e. Shape restrictions?
8. How closely does recent development in the area follow the above regulations? Does it demonstrate a different set of practical restrictions?
9. How close are burdensome facilities (industrial plants, burdensome services, roads of supra-local significance)
10. What is the capacity for services to absorb increases in population?
    a. Schools
    b. Water
    c. Sewage
    d. Transport
11. What are the demographics for the neighbourhood:
    a. Population
    b. Annual income
    c. Rooms per home
    d. Land per home
    e. Market rent

**Development desirability (quality)**

12. What is the amenity score for the parcel? 15 minutes to:
    a. Transportation
    b. Schools
    c. Food

d. Medical services
    e. Parks
    f. Libraries
13. What support services exist for lower income families?
    a. Community centre/programmes
    b. Credit union/banking
    c. Harm reduction
    d. Meal programmes
    e. Senior services/home care
14. Does the neighbourhood exhibit any stress/dereliction?

### 2.2.2 Use Case 2

**Personae**: Housing Department Planner

Alice is a planner in the housing department at the City. She is responsible for evaluating large, mixed use development plans. Her role requires her to evaluate whether sufficient services are in place for the development and whether additional services are required.

It also requires her to consider how the development impacts the objectives and policies laid out by the city.

**CQs:**

**Land Use/Zoning**

For a parcel of land:

15. What is the current land use?
16. How many dwellings occupy the parcel?
17. How many residents occupy the parcel?
18. What is the current zoning?

**Development Feasibility**

19. What is the demand for housing in the area/city?
    a. Broken down by income
    b. Broken down by number of bedrooms
    c. Broken down by …
20. What are the banks, CMHC, and other funders willing to finance?
21. What is the average time to receive construction approvals?
22. What constraints exist?
    a. Debt?

b. Rent thresholds?

23. How do you want to approach development?
    a. Joint development with partners?

For a parcel of land:

24. What are the government policies that impact the potential of a parcel?
    a. Taxation?
    b. Density?
    c. Mixed use?
25. What is the cost of the parcel?
26. What is the cost of developing the parcel?
27. What is the opportunity cost of not developing the parcel?
28. What public utilities currently service the area? What is the capacity?
    a. Water
    b. Gas
    c. Power
    d. Waste disposal
    e. Road network
    f. Public transit
    g. Fire and emergency services

**Development desirability (quality)**

29. What social services are in the area? What is their capacity?
    a. Schools
    b. Parks
    c. Medical facilities and pharmacies
    d. Child care, Meal programs
    e. Libraries
30. How liveable will the space be?
    a. Green space?
    b. Density?

### 2.2.3 Use Case 3

**Personae**: Urban Planner

Natalie is an urban planner at the City. She is responsible for developing land use plans. This requires an investigation into aggregate availability and use of land in the city. Her role requires her to consider how land use impacts the different objectives and policies laid out by the city. Currently, she is tasked with increasing the supply of housing to meet the demand forecasted for the next 20 years, while maintaining or increasing the greenspace in each neighbourhood and increasing access to transit.

**CQs:**

**Land Use/Zoning**

31. What publicly owned parcels of land are available for current or future development?
32. For a parcel of land:
    a. What is the current land use?
    b. What is the current zoning?
    c. What is the difference between current zoning and future uses?
    *Note: this can be reinterpreted as a combination of 32b above "What is the current zoning" and "What is the planned zoning" or even "What is the planned land use" (i.e., approved projects)*
    d. How many dwellings occupy the parcel?
    e. How many residents occupy the parcel?

**Development Feasibility**

33. What is the aggregate demand for housing? Broken down by:
    a. Number of bedrooms
    b. Size of unit
    c. Income levels
34. What are the priority growth areas?
35. For a parcel of land:
    a. What public utilities currently service the area? What is the capacity? What are the future expansion plans?
        i. Water & Sewage
        ii. Solid Waste
        iii. Gas
        iv. Electricity
        v. Road network
        vi. Public transit
        vii. Fire and emergency services
    b. What is the aggregate amount of housing that can be built over the next n years?
36. What assumptions are made about land and construction costs?
37. What is the zoned capacity for a parcel of land?

a. Could this differ from the practical capacity? How do stated zoning capacities (or other properties) differ with the capacities (or other properties) of built projects over past X years?

**Development desirability (quality)**

38. What social infrastructure are in place and what are their planned expansions over the next 20 years?
    a. Schools
    b. Parks
    c. Medical and pharmacies
    d. Childcare
    e. Food programs
    f. Harm reduction services
    g. Libraries

### 2.2.4 Use Case 4

**Personae**: Developer

Ben is a developer at Acme Corp. He is responsible for identifying and assessing potential opportunities for housing development projects in the City of Toronto. For any potential parcel of land, he must consider whether development is feasible from a legal standpoint (i.e., zoning), and if so, whether it will be desirable from an occupant standpoint. He must also consider whether the project will be profitable: what is the expected market value for each dwelling unit, and what additional costs will need to be incurred beyond the construction of the building(s) itself (e.g., connection to public services)?

**CQs:**

**Land Use/Zoning**

39. What is the zoning for a parcel of land?
40. What is the zoning of the nearby parcels?
41. Is it currently occupied by any building(s)?
    a. If so, what is the age and condition of the building?
    b. Does the building have occupants?
42. What parcels of land are available for purchase, or already owned by Acme Corp?

**Development Feasibility**

43. What is the value of the land?
44. What is the average income of the neighbourhood?
45. Is the land already serviced by public utilities? If so, what is the current capacity? Are there plans (by the city) to change the capacity in the future?

a. Water
   b. Waste disposal
   c. Road network
   d. Public transit
   e. Fire and emergency services access?

46. What are alternative building structures that optimize cost and number of houses produced, based on constraints such as zoning, building codes, etc.
47. What are the banks, CMHC, and other funders willing to finance?
48. What is the average time to receive construction approvals?

**Development desirability (quality)**

49. What amenities are nearby?
    a. Schools
    b. Parks
    c. Medical and pharmacies
    d. Childcare
    e. Supermarkets
    f. Libraries

### 2.2.5 Use Case 5

**Personae**: Government Land Office

Hilda has been tasked with identifying federally owned land and buildings that can be repurposed for affordable housing. She needs to identify properties that are vacant and underused. She then needs to determine whether any of these properties are a good location for housing.

**CQs**:

**Land Use/Zoning**

Vacant Land CQs

50. What vacant land does the federal government own? Which part of government owns it?
51. What is the current use of the land?
52. What is the land zoned for?

Occupied Land CQs

53. What buildings does the federal government own?
54. Is the building vacant? Underutilized?
55. Who owns the building?
    *Note: this is a variation of 53, above.*

56. What is the current use of the building?
57. What is the building zoned for?

**Development Feasibility**

Vacant Land CQs

58. Is the land accessible?
59. Is the land serviced?
60. What limitations for building exist?

Occupied Land CQs

61. Is the building accessible?
62. Is the building serviced?
63. What limitations for replacing the building exist? For example, historical designation, NIMBY.

**Development desirability (quality)**

Vacant Land CQs

64. Is the vacant land close to a population centre?
65. Is the land close to amenities?
66. Is it close to employment opportunities?

Occupied Land CQs

67. Is the building close to amenities?
68. Is it close employment opportunities?

69. For either occupied or vacant land - is the land subject to any environmental risks (e.g. flooding)?[6]

# 3 Housing Potential Common Data Model

This section describes the development and design of the HPCDM. The model – described and formalized in Section 3.2 – is available in an owl encoding at http://ontology.eil.utoronto.ca/HPCDM.owl.

## 3.1 Related Work

When undertaking the development of a common data model, it is particularly important to consider the reuse of existing models and standards. When possible, reuse serves to reduce the required design work, while also facilitating compatibility of the model with datasets and applications from a wider range of communities. Therefore, where appropriate, existing data

[6] This question was identified for inclusion after the initial Use Case development effort.

models—particularly standards—will be reused, and relevant components extracted to form parts of the HPCDM.

### 3.1.1 Survey of Relevant Data Models

As part of this project, a review of relevant data models was conducted. The survey focused on the identification of city data standards but also included some relevant ontologies. The following provides a summary of each model, focusing on the intended use and (relevant) scope. Appendix B provides a detailed table to indicate the coverage of each data model, with respect to the identified CQs.

#### *3.1.1.1 ISO/IEC 5087*

The ISO/IEC 5087 Series is a multi-part standard intended to provide a data model for city services, to support interoperability within and across cities. Two parts of the standard have been published, and development of additional parts is ongoing. The ontologies are encoded in OWL, and the intended focus on city services is defined in a broad sense to include areas such as transportation, public health, housing, and energy. As illustrated in Figure 1, ISO/IEC 5087-1 [8] defines foundational concepts for city data, such as: parthood, city units, time, change, location, activity, recurring events, resources, agents, organization structure, and agreements. ISO/IEC 5087-2 [9] extends the foundational concepts and introduces the core concepts that are common across city services, such as: classification codes, infrastructure, transportation infrastructure, buildings, land use, persons, city residents, households, city organizations, contracts, bylaws, contact information, and sensors. While the standards documents themselves are copyrighted by ISO and not freely available, the encodings of the data models *are* publicly available and so may be used and shared freely.

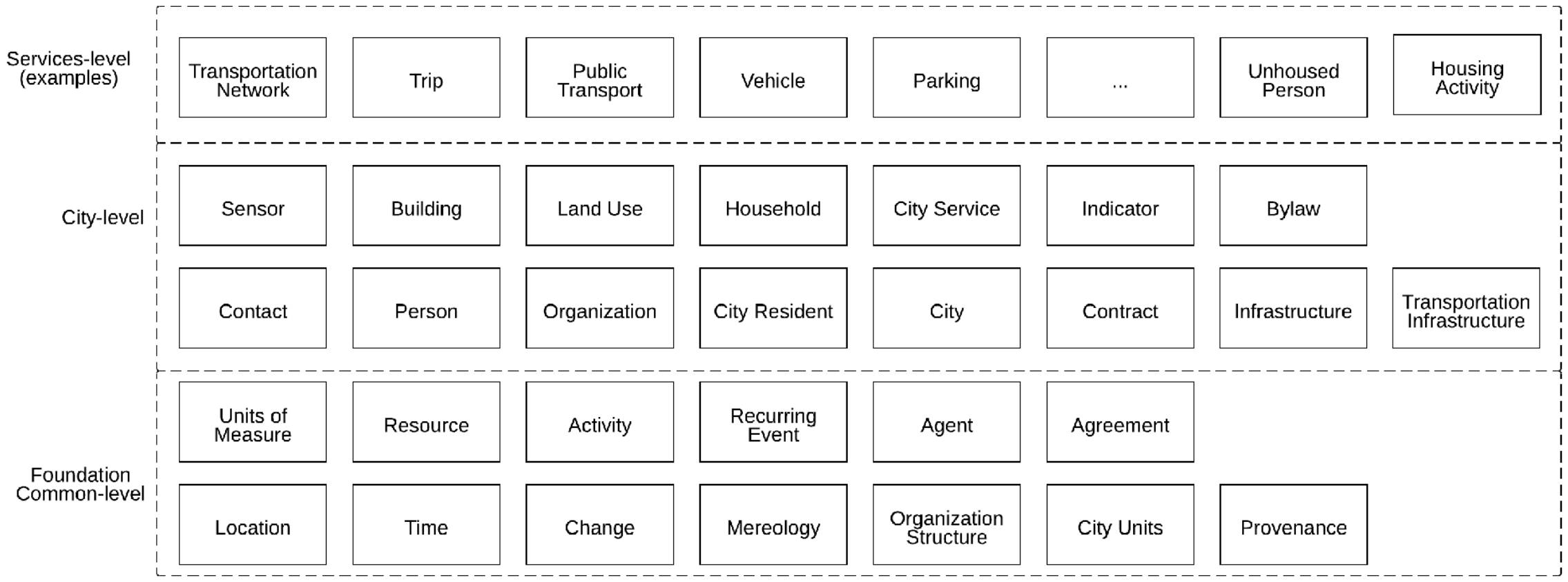

*Figure 1: Overview of patterns included in the foundation-level (ISO/IEC 5087-1) and city-level (ISO/IEC 5087-2) of the ISO/IEC 5087 Series. Patterns at the services level (for additional parts of the standard) are included as examples.*

#### *3.1.1.2 ISO/IEC 21972*

ISO/IEC 21972:2020 Information technology — Upper level ontology for smart city indicators [10] specifies a data model that defines the core concepts necessary to represent (city) indicators, such as: indicators, quantities, units of measure, populations, and parameters such as mean and sum. The standard serves as a normative for the ISO/IEC 5087-1 definition of units of measure for cities. In the context of housing potential, this standard is useful for capturing and integrating various quantities and their associated measures (e.g., parcel and building dimensions, infrastructure capacities). The representation of city indicators is also important for comparing various metrics used to characterize areas of a city. As with the ISO/IEC 5087 Series, while the standard documentation is owned by ISO and not freely available, the data model encodings are open to the public.

#### *3.1.1.3 CityGML 3.0*

CityGML [11] is an open, conceptual model and exchange format for 3D city models, widely used for integrating, storing, and exchanging urban geospatial data in smart cities and digital twins. It is jointly published by the Open Geospatial Consortium (OGC) and the International Standards Organization (ISO), with encodings in UML, GML/XML and JSON. Its scope covers visualization elements (e.g. pertaining to object appearance/rendering), versioning, time series (sensor) data, spatial representations, infrastructure (visualization focused), land use (combined with land cover), and transportation.

#### *3.1.1.4 LandInfra*

A representation of facilities and the land on which they're constructed, LandInfra [12] is intended to support infrastructure asset management and planning by providing a standard data model to support the exchange of information for land and civil engineering infrastructure. It is published by the Open Geospatial Consortium (OGC), with encodings in UML and GML/XML. Its scope covers a representation of datasets, physical infrastructure, projects, spatial representations, transportation, surveys (observations and sensors), land cover, and land use (focusing on various types of land divisions).

#### *3.1.1.5 INSPIRE*

The INSPIRE data models are designed to support integration of spatial data, with a focus on environmental applications. The models are formalized in UML and divided into 34 themes; with the following of relevance: cadastral parcels [13], transport networks [14], buildings [15], land use [16] , and utility and government services [17].

#### *3.1.1.6 SAREF*

The Smart Applications REFerence Ontology [18] is a data model published by the European Telecommunications Standards Institute (ETSI). It is comprised of a set of ontologies, intended to support interoperability across sectors, with a focus on the Internet of Things (IoT). Though the focus on devices is not relevant, SAREF includes ontologies for the following area of interest for housing potential: SAREF-core, SAREF4ENVI (environment), SAREF4BLDG (building), SAREF4CITY (smart city), SAREF4WATER (water). Review of the contents of the models reveals that the IoT focused use cases influence the definition of some relevant terms (e.g. the SAREF4ENVI defines a Building as a kind of Perimeter) in ways that may not be appropriate for housing potential.

#### *3.1.1.7 ISO 19152*

ISO 19152 Geographic information — Land Administration Domain Model (LADM) is an international standard intended to support interoperability and exchange of land administration data (and often, to support the design of land administration platforms). The standard is owned by ISO and not freely available, however OGC has indicated interest in working on an open, implementation version of the standard[7]. The model is encoded on UML and is divided into several parts, of interest for housing potential are: ISO 19152-1:2024 [19], Generic conceptual model, ISO 19152-2:2025 Land registration [20], ISO 19152-4:2025 Valuation information [21], and ISO 19152-5 Spatial plan information (under publication) [22]. The model covers several relevant topics however the models are only available through purchase of the documentation. In this case, publicly available overviews [23] may be used to inform design decisions, and to identify approximate mappings to key classes.

#### *3.1.1.8 ISO 37105*

ISO 37105:2019 [24] specifies a descriptive framework for a city including an associated foundational ontology of the anatomical structure of a city or community. Its aim is to provide "a common language for the description of cities" to support the comparison of cities and sharing of city solutions (in particular, working toward achieving the United Nations Sustainable Development (UNSDG) Goal 11: Make cities and human settlements inclusive, safe, resilient and sustainable). The standard and its content are copyrighted by ISO and not freely available, however as it is based on the City Anatomy Ontology (CAO) [25], this earlier artifact may be considered in its stead.

The CAO was developed by the City Protocol Society with the intent of providing a foundational, holistic view of the city. The ontology is formalized in OWL-DL, and it focuses on the description of a city as a system of systems. The parts of a city system are divided by structure, interactions, and society. The structure system includes a representation of the environment, built domain, and infrastructure; the interactions system describes the interactions between structure and society, including various "functions" such as education and transportation; the society system focuses on citizens and organizations in the city. The model does not appear to include the geospatial relationships between the various system elements.

#### *3.1.1.9 KM4City*

The KM4City ontology [26] aims to support integration of smart city data and tools. Its primary focus appears to be on supporting the creation of city dashboards (charts/maps to support monitoring activities). It is encoded in OWL and covers the following topics: administration (areas), "street guide" (road network), points of interest, public transport, sensors, time and the Internet of Things (IoT). While the points of interest module covers many relevant topics (e.g. education and utilities), the definition of these classes is limited to a service taxonomy.

#### *3.1.1.10 OntoZoning (OZ), OntoPlanningRegulations (OPR), OntoBuildableSpace (OBS)*

The OntoZoning (OZ) [27], OntoPlanningRegulations (OPR), and OntoBuildableSpace (OBS) [28] ontologies have been developed to capture planning regulations and their implications on buildable space. The ontologies are encoded in OWL. While they have been designed and implemented for Singapore, they define topics that have general use for housing potential assessment. The ontologies cover concepts related to land use, built form regulations, "buildable space", and related measures.

---

[7] https://www.ogc.org/requests/ogc-to-form-land-administration-domain-model-standards-working-group-public-comment-sought-on-charter/

#### 3.1.1.11 Global City Indicator (GCI) Ontologies

The Global City Indicator (GCI) Ontologies refer to a series of ontologies (under development) that extend ISO/IEC 21972 to represent the city indicators from each theme of ISO 37120. Currently, the GCI Ontology series addresses the following indicator themes relevant to the housing potential use cases: education [29], energy [30], fire and emergency [31], healthcare [32], public safety [33], recreation [34], shelter [35], solid waste [36], transportation [37], and water and sanitation [38].

#### 3.1.1.12 Census Ontology

The Canadian Census Ontology [39] was created to support the definition of Canadian Census of Population data. The intent of the ontology is to enable the unambiguous definition of data across census years to support understanding and interoperability of results. The ontology defines the concept of a Characteristic, which is an indicator used to represent statistical information about various segments of the population. In terms of housing potential, the Canadian Census Ontology provides the terminology required to describe demographic information related to various areas in the city.

#### 3.1.1.13 City Digital Twin Ontology

Earlier work on the development of a city digital twin has resulted in a City Digital Twin Ontology that is based upon to the ISO/IEC 5087 series of standards. The City Digital Twin Ontology extends these standards with specialized concepts required to capture various data of interest. Some of the data that has been captured thus far intersects with the concepts required by the housing potential use case – specifically the representation of city services. An earlier implementation of the digital twin focused on complete community analysis, therefore the City Digital Twin Ontology includes a representation of various city services and amenities such as parks and hospitals.

### 3.1.2 Selection of Relevant Data Models

Several standards and existing ontologies were identified for reuse to form the basis of the zoning component of the HPCDM, in particular: ISO/IEC 5087-1, ISO/IEC 5087-2, ISO/IEC 21972, OZ, OPR, OBS. The survey conducted identifies a high degree of coverage for all of these models, in particular for the Land Use and Zoning theme (the focus of the HPCDM Core). The OZ, OPR, OBS ontologies' models are complementary to the content defined by the ISO/IEC 5087 series, in that the latter does not provide a detailed formulation of zoning regulations. Other city data models exhibited a comparatively lower level of coverage, and much of the identified scope (relevant to zoning) overlapped with what is already contained in the six selected ontologies.

In addition, several models were selected for reuse to develop the HPCDM beyond zoning and land use. The Census Ontology was reused directly as it provides the terminology necessary to describe the census characteristics (their values, and the location to which they apply) that are needed to describe the population demographics of an area. The City Digital Twin ontology was also reused directly – it extends the ISO/IEC 5087 series of standards (and so is consistent with the core model) to define various categories of organizations and the services they provide in a city. While the Global City Indicator (GCI) Ontologies are closely related to several areas included in the HPCDM, they are not all suitable for reuse. The Education, Fire and emergency, and Healthcare ontologies define indicators and concepts that may be used to support the definition of the associated service capacities. However, beyond this the GCI ontologies define domain elements and indicators that are not relevant for the current scope of the HPCDM. The INSPIRE model also defines concepts related to buildings, utilities and services; however, relative to the defined requirements there is no significant new content required beyond what is specified in the ISO/IEC 5087 series of standards. For this reason, it is not reused directly to formulate the contents of the HPCDM.

Finally, the LandInfra model includes a detailed specification of land divisions with a perspective on ownership that will be useful to consider for future extensions to the HPCDM but is not required for the current scope.

## 3.2 Model Specification

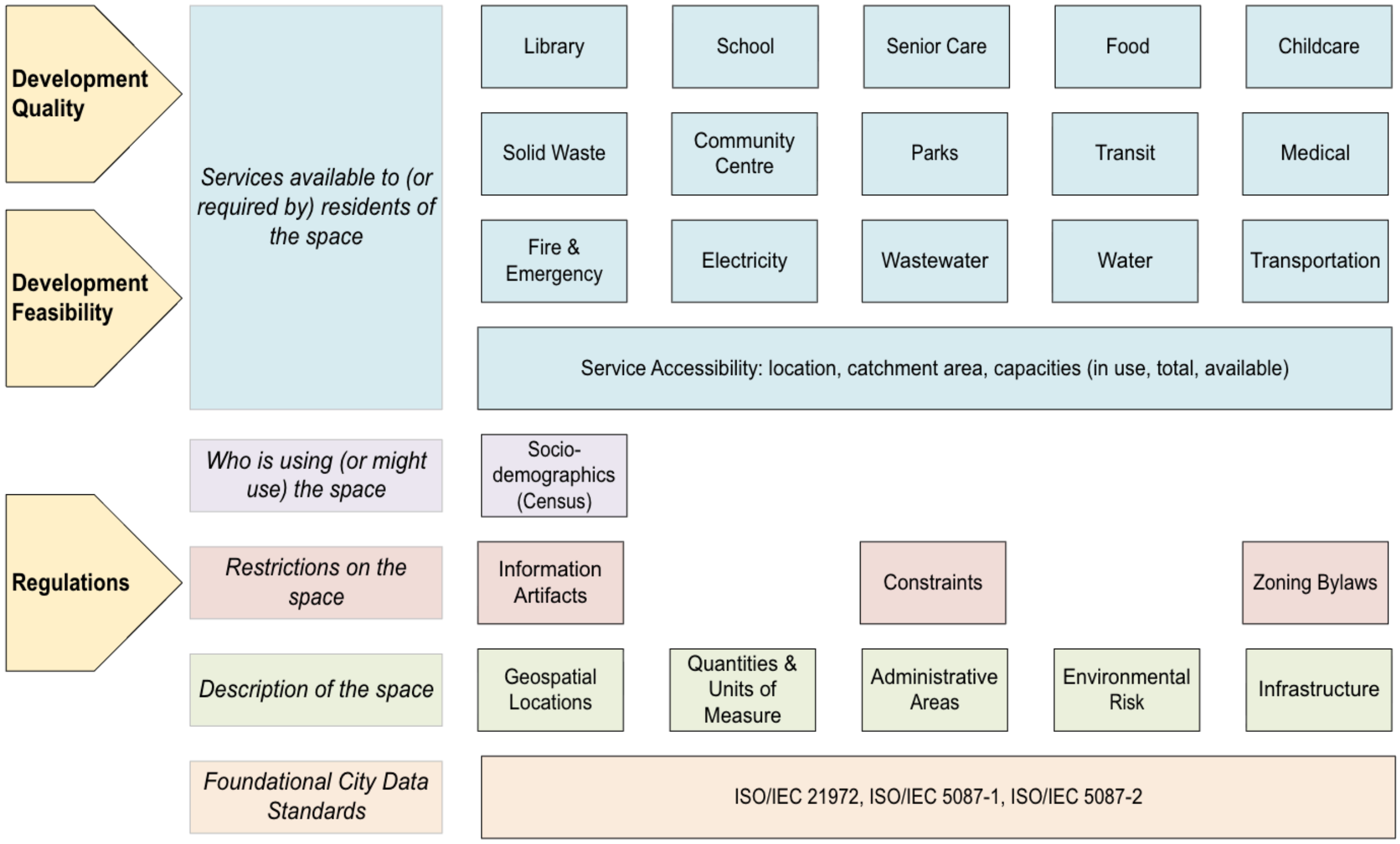

*Figure 2: Overview of the Housing Potential Common Data Model (HPCDM).*

The HPCDM is broken down into theme-specific patterns that have been designed to support navigation and modular use. The following patterns are defined, as depicted in Figure 2, in addition to the content imported from existing standards:

- Quantity Constraints
- Administrative Area
- Zoning
- Building
- Population Demographics
- Service Accessibility
- Fire & Emergency
- Electricity
- Solid Waste

- Water
- Wastewater
- Transportation
- Transit
- Childcare
- Community Centre
- Library
- School
- Parks
- Medical
- Food
- Senior Services
- Environmental Risk

The core of the model is depicted in the lower layers. It corresponds to a representation of the space, and the restrictions that apply to it (e.g. zoning). This is extended with a representation of the users of the space (population demographics), and the various services that are required by and/or available to (potential) residents at various areas of the city.

The patterns are summarized with the use of diagrams that use the following notation: white rectangular boxes to denote classes, lines between boxes to denote object properties, shaded boxes to denote individuals (instances of classes), and datatype values denoted with plain text.

Classes are defined using the following table format, which is a simplification of description logic (DL) where the first column identifies the class name, the second column its properties, and the third column indicates the restrictions on the property (i.e., how many instances of the property can or should occur for a particular class). Value restrictions are specified according to Manchester Syntax, as defined by the W3C. For example, Table 6 specifies that Agent is a subclass of the intersection of (Person or Organization) and can be a memberOf of zero or more Organizations.

*Table 6—Example formalization of the Agent class*

| Class | Property | Value Restriction |
|---|---|---|
| Agent | rdfs:subClassOf | Person or Organization |
| | memberOf | only Organization |

Terms reused from existing standards and ontologies are prefixed with the appropriate namespace to indicate their origin. The following prefixes are used:

- 5087-1: ISO/IEC 5087-1 (note that this is shorthand to cover a number of individual, pattern-specific namespaces introduced in the ISO/IEC 5087-1 encoding)
- 5087-2: ISO/IEC 5087-2 (note that this is shorthand to cover a number of individual, pattern-specific namespaces introduced in the ISO/IEC 5087-2 encoding)
- i72: ISO/IEC 21972
- cdt: City Digital Twin Ontology
- census: (generic) Census Ontology
- cacensus: Canadian Census Ontology (extends the generic Census Ontology)

- oz: OntoZoning (OZ) ontology
- opr: OntoPlanningRegulation (OPR) ontology
- obs: OntoBuildableSpace (OBS) ontology
- gcife: http://ontology.eil.utoronto.ca/GCI/ISO37120/FireEmergency.owl#
- gciofb: Global City Indicator Fire & Emergency Response Ontologies - Fire Brigade Organization Ontology[8]
- gcir: Global City Indicator Recreation Ontology[9]
- gcie: Global City Indicator Education Ontology[10]
- gcih: Global City Indicator Health Ontology[11]

Terms without prefixes are to be interpreted as defined in the HPCDM Ontology (prefix hp: http://ontology.eil.utoronto.ca/HPCDM/).

### 3.2.1 City Indicator and Quantity Pattern (ISO/IEC 21972)

Many of the patterns introduced in the HPCDM involve the representation of quantities (e.g. measured attributes of some object such as a building) as well as indicators that are defined to capture certain qualities of a city. To support this, the HPCDM reuses the ISO/IEC 21972:2020 standard for an upper-level ontology for smart city indicators. Key elements of this ontology are reviewed here to support understanding of how they are used in other patterns. The ontology introduces the following key classes:

- i72:Quantity: represents the thing being measured. Examples include height, area, or speed. It has the following key property:
    - i72:hasValue: specifies a i72:Measure of the quantity. A single quantity may have multiple measures.
- i72:Measure: represents a specific measurement of a quantity. It has the following key properties:
    - i72:hasNumericalValue: specifies the numeric value of the measure.
    - i72:hasUnit: specifies the unit that the measure is specified in (e.g. “metres”).
- i72:Indicator: a type of i72:Quantity that describes a city. It has the following key properties:
    - i72:for_city: identifies the city that the indicator describes.
    - i72:for_time_interval: identifies the interval in time that the indicator describes (e.g. the population size in 2021).
- i72:Parameter: a type of i72:Quantity that represents quantities that can be defined for a population. Here, population is defined in the statistical sense as some collection of objects. Examples include cardinality (population size), and averages and sums of some attribute of the population’s members (e.g. income of a population of residents).

---

[8] http://ontology.eil.utoronto.ca/GCI/FireEmergency/FBOrganization.owl

[9] http://ontology.eil.utoronto.ca/GCI/Recreation/GCI-Recreation.owl

[10] http://ontology.eil.utoronto.ca/GCI/Education/GCI-Education.owl

[11] http://ontology.eil.utoronto.ca/GCI/Health/GCI-Health.owl

The following terms are introduced in the HPCDM as extensions to the city indicator ontology:

- RatioQuantity: a type of i72:Quantity that is defined as a ratio of two other quantities. For example, “firefighters per capita” is a ratio of some quantity of a count of firefighters to a count of resident population. Note that RatioQuantity is not currently in ISO/IEC 21972, though it is expected to be proposed in the next revision. If it is added, it is expected that this will be an owl:EquivalentClass. It has the following properties:
    - i72:numerator: specifies the i72:Quantity that is the numerator of the ratio.
    - i72:numerator: specifies the i72:Quantity that is the denominator of the ratio.
- DifferenceQuantity: a type of i72:Quantity that is defined as the difference between two i72:Quantity objects. It has the following properties:
    - i72:term_1: specifies the first term (the minuend).
    - i72:term_2: specifies the second term (the subtrahend).
- forLocation: this property associates a quantity with a particular location. ISO/IEC 21972 defines a i72:forCity property, however due to its focus on *city* indicators, it does not define a property for the association of a quantity to a general location.

The following instances of the i72:Unit_of_measure class are defined to support the formalization of the various capacity measures included in the model:

- kilovolt_ampere
- square_foot
- storeys
- avg_inpatients_daily_per_bed
- cubic_metre_per_year
- kilometers_per_hour
- person_per_day
- sites_per_person
- square_metre_per_person
- tonnes_per_year
- vehicles_per_hour

#### *3.2.1.1 Formalization*

Formalization of the pattern is specified in Table 7 and Table 8.

*Table 7: Formalization of class extensions to ISO/IEC 21972:2020*

| Class | Property | Value Restriction |
|---|---|---|
| RatioQuantity | rdfs:subClassOf | i72:Quantity |
| | i72:numerator | only i72:Quantity |
| | i72:denominator | only i72:Quantity |
| DifferenceQuantity | rdfs:subClassOf | i72:Quantity |
| | i72:term_1 | only i72:Quantity |
| | i72:term_2 | only i72:Quantity |

*Table 8: Formalization of property extensions to ISO/IEC 21972:2020*

| Property | Characteristic | Value (if applicable) |
|---|---|---|
| forLocation | rdfs:domain | i72:Quantity |

### 3.2.2 Information Artifacts

In the context of housing potential analysis, information artifacts are important as they represent sources of much of the content of interest, such as bylaws and plans in the context of zoning. This pattern includes only the level of detail required to specify the basic terms involved in the description of an information artifact. It introduces the following class:

- InformationArtifact: represents a piece of information. Note that this is distinct from the physical artifact (e.g. a piece of paper, a hard drive) that the information is encoded on. It has the following properties:
  - 5087-1:hasIdentifier: specifies an identifier for the information artifact.
  - 5087-1:hasName: specifies a name for the information artifact.
  - 5087-1:hasDescription: specifies a description for the information artifact.
  - 5087-1:wasAttributedTo: specifies the 5087-1:Agent (e.g. organization or person) responsible for creating the artifact.

Examples of information artifacts include legislation (e.g. zoning documents) and plans. In extensions of the HPCDM, this class may be extended with other types of artifacts as required, such as development proposals.

#### *3.2.2.1 Formalization*

The pattern is formalized in Table 9.

*Table 9: Formalization of the Information Artifact Pattern*

| Class | Property | Value Restriction |
|---|---|---|
| InformationArtifact | 5087-1:hasIdentifier | only xsd:string |
| | 5087-1:hasName | only xsd:string |
| | 5087-1:hasDescription | only xsd:string |
| | 5087-1:wasAttributedTo | only 5087-1:Agent |

### 3.2.3 Quantity Constraint Pattern

Quantities of interest for central objects of housing potential analysis (e.g., building dimensions or lot sizes) vary across application and municipality. It is also likely that the *definition* of such quantities may vary from region to region. Instead of attempting to include an exhaustive set of such definitions, this pattern provides the framework to extend the model with definitions for individual municipalities and applications. The foundational class for these definitions is Quantity, as defined in ISO/IEC 21972. Its definition enables a level of detail that supports automated calculation and regulation compliance checking. Comparisons and even interoperability between definitions between regions may also be achieved.

Beyond the specification of quantities of interest, a critical requirement is the identification and understanding of constraints that apply to these quantities (while other types of constraints exist in general, they are not the focus of this pattern). Often defined according to a particular regulation (e.g., part of a zoning bylaw, outlined in Section 3.2.5), these quantity constraints typically specify upper or lower limits on a quantity, though other formulations (e.g., "equivalent to a value", or "within x units of a value") are conceivable. An overview is depicted in Figure 3. Figure 4 illustrates an example of the relationship between Quantities (e.g. heights of buildings) and a Quantity Constraint (maximum building height). The following classes are defined for the Quantity Constraint pattern:

- QuantityConstraint: A QuantityConstraint refers to some restriction on a i72:Quantity. While quantities are commonly used to describe measured attributes of an object (e.g. area, height), the ISO/IEC 21972 ontology also includes a representation of i72:Parameters as types of i72:Quantity. Thus, QuantityConstraints may apply not only to attributes of individual objects but to parameters (e.g., minimums, maximums, averages) of object populations. The value specified by the constraint on the quantity is expressed with the Constraint Specification class.
  Note the distinction between a QuantityConstraint and a constraint that is specified on a Quantity class. For example, ISO/IEC 21972 allows for a specification of constraints on types of Quantities – for example, a class such as "Amenity Rating" could be defined such that its associated measures must always be between values of 0 and 5. This is distinct from the *specification* (Quantity Constraint) that a *particular* amenity rating(s) should be above 3 (for example).
  A QuantityConstraint has the following properties:
    - constrains: specifies the i72:Quantity that the constraint applies to.
    - i72:hasValue: identifies value of the constraint imposed as a ConstraintSpecification
    - definedIn: identifies the InformationArtifact that is the source of the constraint, (e.g. a bylaw or a plan)
    - definedAtTime: identifies the time (5087-1:Instant) at which the constraint was defined.
    - definedForTime: identifies time interval (5087-1:Interval) at which the constraint is meant to apply.
- ConstraintSpecification: A ConstraintSpecification represents an actual value of the limitation set by the QuantityConstraint. Like the i72:Measure class, this is formalized with the i72:hasNumericalValue and i72:hasUnit properties.
- QuantityAllowance: A QuantityAllowance is a QuantityConstraint that specifies an allowed amount (maximum) for the constrained quantity.
- QuantityRequirement: A QuantityRequirement is a QuantityConstraint that specifies a minimum amount for the constrained quantity.
- QuantityEquivalence: A QuantityEquivalence is a QuantityConstraint that specifies an exact required amount for the constrained quantity.
- Maximum: A specialization of the i72:Parameter class, Maximum represents the maximum of some Variable of a i72:Population.
- Minimum: A specialization of the i72:Parameter class, Minimum represents the minimum of some Variable of a i72:Population.

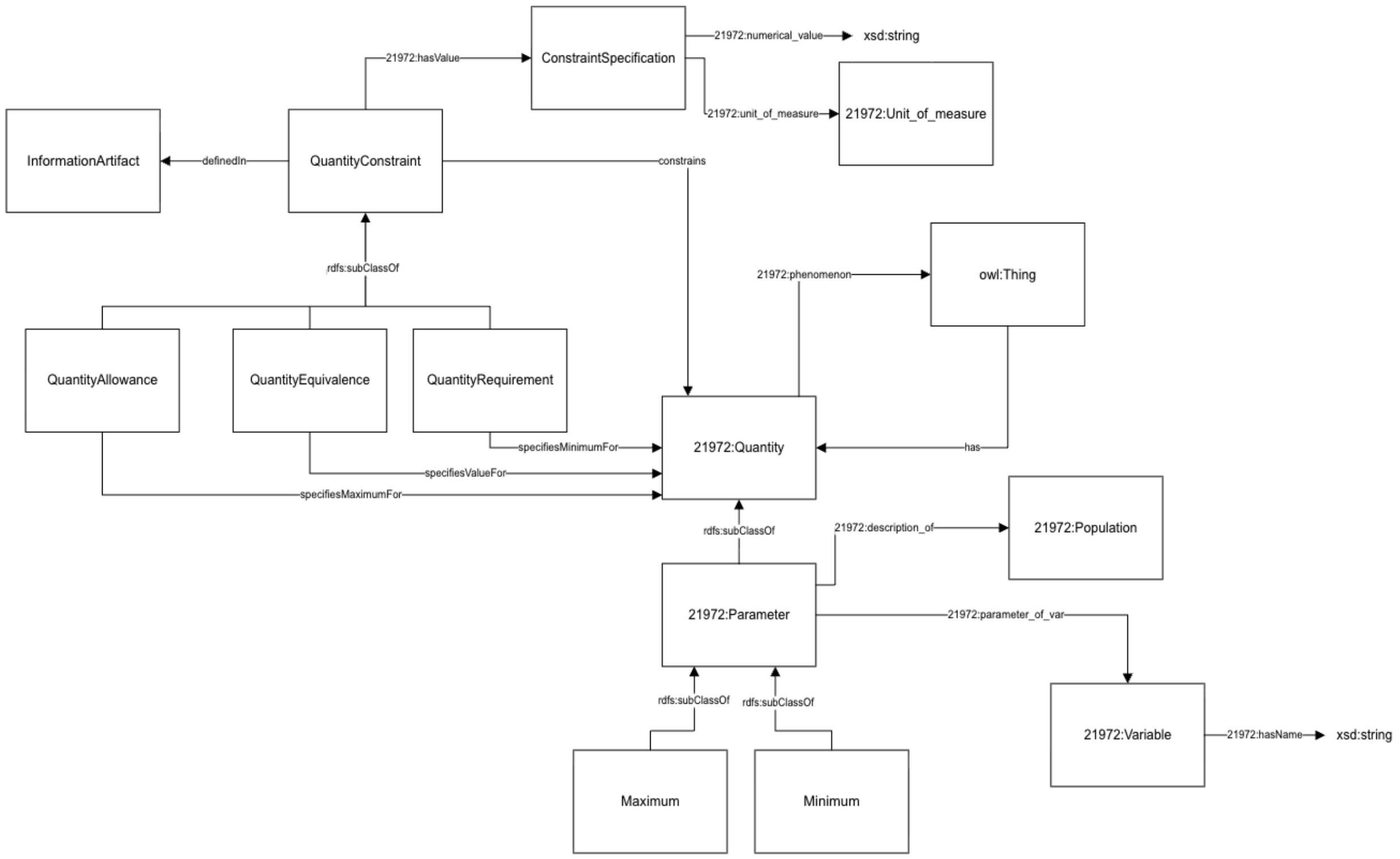

*Figure 3: Overview of the key classes and properties in the Quantity Constraint Pattern*

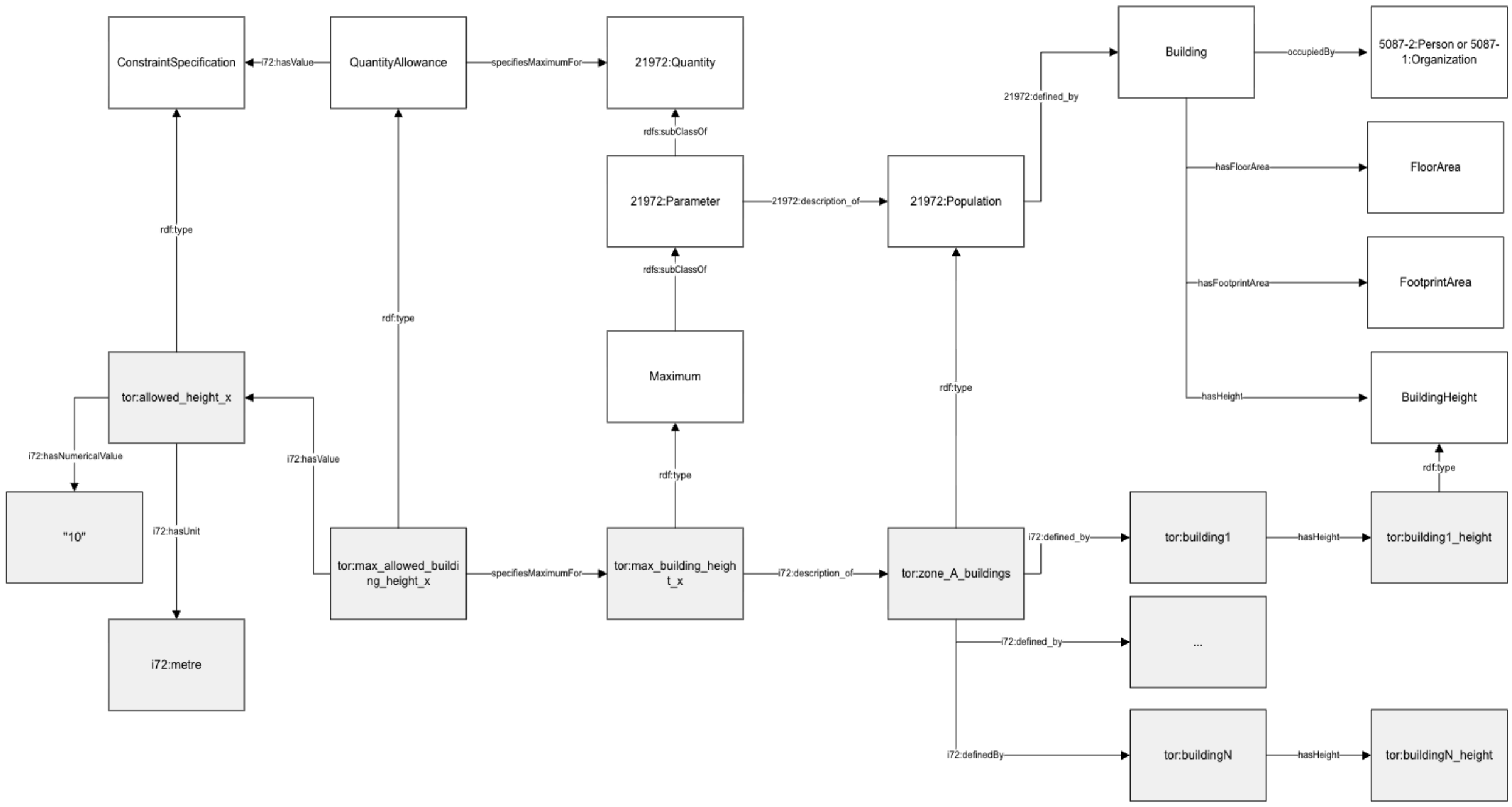

*Figure 4: Example relationship between a Quantity Constraint and individual Quantities*

#### *3.2.3.1 Formalization*

The pattern is formalized in Table 10 and Table 11.

*Table 10: Formalization of the classes in Quantities and Constraints Pattern*

| Class | Property | Value Restriction |
|---|---|---|
| QuantityConstraint | i72:hasValue | only ConstraintSpecification |
| | constrains | only i72:Quantity |
| | definedIn | only InformationArtifact |
| | definedAtTime | only 5087-1:Instant |
| | definedForTime | only 5087-1:Interval |
| QuantityAllowance | rdfs:subClassOf | QuantityConstraint |
| | specifiesMaximumFor | only i72:Quantity |
| QuantityRequirement | rdfs:subClassOf | QuantityConstraint |
| | specifiesMinimumFor | only i72:Quantity |
| QuantityEquivalence | rdfs:subClassOf | QuantityConstraint |
| | specifiesValueFor | only i72:Quantity |
| ConstraintSpecification | i72:hasNumericalValue | exactly 1 xsd:string |
| | i72:hasUnit | exactly 1 i72:Unit_of_measure |
| Maximum | rdfs:subClassOf | i72:Parameter |
| | maximumOf | exactly 1 i72:Population |
| Minimum | rdfs:subClassOf | i72:Parameter |
| | minimumOf | exactly 1 i72:Population |

*Table 11: Formalization of the properties in the Quantity Constraint pattern*

| Property | Characteristic | Value (if applicable) |
|---|---|---|
| i72:phenomenon | rdfs:inverseOf | has |
| specifiesMaximumFor | owl:subpropertyOf | constrains |
| specifiesMinimumFor | owl:subpropertyOf | constrains |
| maximum_of | owl:subpropertyOf | i72:description_of |
| minimum_of | owl:subpropertyOf | i72:description_of |

### 3.2.4 Administrative Area Pattern

For housing potential analysis to identify property that is available for purchase and/or (re-)development, there is a basic need to represent areas of land. Beyond the geospatial representation, key attributes of these areas of land are how they are being used, the nature of the land division (e.g., an arbitrary area of land to be assessed as opposed to an individual parcel), and who owns it. These characteristics are formalized in the Administrative Area Pattern. Figure 5 illustrates the key classes and properties.

The ISO/IEC 5087-2 standard defines a Land Use pattern designed to be extended with different land use classification systems. This is achieved with the use of the Code class that enables a single instance of Land Use to be associated to multiple "codes" from different land use classification systems. Additional information on the Code Pattern is provided in Appendix A. Note that the land use that is directly associated with a LandArea in ISO/IEC 5087-2 defines the *current* land use associated with an area. This is distinct from the land use that is defined based on the zoning for an area. Land use defined according to zoning regulations is addressed in the Zoning Pattern. While ISO/IEC 5087-2 introduces a generic "LandArea" class, more specific types of areas are needed to interpret various housing related data. There is a need to identify the individual areas of land referred to by land governance policies and land registry systems: lots and parcels, respectively. This is discussed in greater detail below.

This pattern also refers to some key terms from the OZ ontology. This is required to align the related classes, supporting the reuse of the OPR and OBS ontologies in the Zoning Pattern (described in the following section). The resulting class definitions are summarized in Table 12.

- AdministrativeArea: a specialization of 5087-2:LandArea that defines properties and subclasses that are relevant for housing potential. It inherits the properties of an associated (geospatial) location, land use, and area. In addition, it has the following properties:
    - 5087-1:partOf: identifies the 5087-2:LandAreas that fully contains the AdministrativeArea. The 5087-1:partOf property is a generalization of the proper part relationship, therefore all areas are, at least, part of themselves.
    - hasArea: specifies the 5087-1:Area of the AdministrativeArea. All 5087-2:LandAreas should have some 5087-1:Area, though it is not a required property.
    - hasFrontage: specifies the Frontage of the AdministrativeArea. As frontage refers to a measure of "width" along a street or road, not all AdministrativeAreas will have a Frontage.
    - hasFSI: specifies the FloorSpaceIndex (FSI), also known as floor area ratio, of the area. FSI is calculated as a ratio of the total built area to the land area. In order for an area to have a (non-zero) FSI value, it must have some built area.
    - hasPerimeter: specifies the perimeter of the area as a 5087-1:Length quantity.
    - inverse(occupies): identifies a 5087-2:InfrastructureElement(s) that occupies (is built on) the area.
    - hasDevelopmentClassification: identifies a LandDevelopmentClassification for the parcel, if any.
    - zonedAsType: specifies what zoning types have been defined for the area. This is based on the geospatial intersection of the area's location with the locations that zoning types that are defined for.

- zonedForConstraint: specifies what constraints have been defined for the area. This is based on the geospatial intersect of the area with the location that constraints are defined to apply to.
- servicedBy: identifies a service(s) that is available at the area. Determination of availability may be based on catchment areas, service radii, or other criteria.

- Parcel: a parcel (sometimes "cadastral parcel") defines a unit of ownership of some area of land. A Parcel is distinct from a Lot, though they are closely related. A Parcel is referenced when addressing questions of ownership and taxation, whereas Lots are typically referred to when describing land use regulations. Parcels may be "created from" lots when they are registered in the land title system (e.g. when a lot is divided according to a plan) – *however,* there is not always a 1:1 relationship between the two objects. A Parcel may contain multiple Lots, and a Lot may contain multiple Parcels. It is possible that a Parcel(s) is not wholly contained by a Lot (or vice versa). A Land Parcel has the following properties:
  - 5087-1:hasIdentifier: land parcels have (cadastral) identifiers, unique (within a jurisdiction), that may be referenced to define ownership.
  - ownership: identifiers the 5087-1:Organization or 5087-1:Agent identified as the owner of the parcel.
  - createdFromLot: if applicable, identifies the Lot that a Parcel was created from.
- Lot: a specialization of oz:Plot that integrates the OZ definition of a Plot with the more generic 5087-2:LandArea class employed in this model. A Lot is a division of land that is created to capture land governance policies (i.e., around zoning and development). It is distinct (disjoint) from, though closely related to a Parcel. In a typical development workflow, a Parcel will be created *based on* a defined Lot. A Lot has the following properties:
  - 5087-1:hasIdentifier: specifies the Lot's identifier, unique (within a jurisdiction), that may be referenced to define land governance policies.
  - definedInPlan: identifies the Plan that the Lot is defined in.
  - allowsUse: identifies a type of Land Use that is permitted on the Lot.
  - mayAllowUse: identifies a type of Land Use that *may* be permitted (e.g., subject to some constraints or additional approvals on the Lot.
  - doesNotAllowUse: identifies a type of Land Use that is prohibited on the Lot.
- LandUse: refers to the general classification of the land use (permitted, planned or current) occurring on an area of land. It is typically defined at a high level (e.g. residential, commercial), though the granularity of land use types is not restricted by the HPCDM. The Code class is used to support the use and/or alignment of any classification systems. Land Use is a specialization of 5087-1:LandUse and OZ:LandUse that integrates the two models. In the OZ ontology, the definition of Land Use is intended (though not restricted) to focus on more granular land use types/activities, however the HPCDM Land Use includes both generalized classifications (such as residential) and more specific land use "types" (e.g. single-unit dwelling, library). To capture this, the Land Use class introduces the following property:
  - includesLandUse: identifies more specific land uses that are included in a more general Land Use classification. This property may be used identify the relationships between different land uses, for example to relate more specific types of use (e.g. "single-unit dwelling") with more general classifications (e.g.,

"Residential"). It may also be used to determine regulation compliance (e.g., given a specialized type of land use, to determine if it is included in a more general permitted/prohibited class).

- LandDevelopmentClassification: refers to a classification of the land from the perspective of potential development. It is related to past land uses and the impacts or risks on redevelopment. Common examples are: "greenfield", "greyfield", and "brownfield". As there is no standard, Canada wide classification in use, LandDevelopmentClassifications are defined using the 5087-2:hasCode property to incorporate classifications from different sources.
- Frontage: Frontage is a kind of 5087-1:Length quantity. In general, it refers to the street facing length of an area of land (AdministrativeArea). The way that Frontage is measured can vary, therefore this class may be extended to capture these distinctions (e.g., TorontoLotFrontage, HalifaxLotFrontage).
- FloorSpaceIndex: Floor Space Index (FSI) is a subclass of RatioQuantity. In general, it is defined as the numerator is the total built floor area and the denominator is the area of the land (AdministrativeArea) it is a property of. Distinctions between how the FSI is calculated may be defined with extensions to this class, as required.

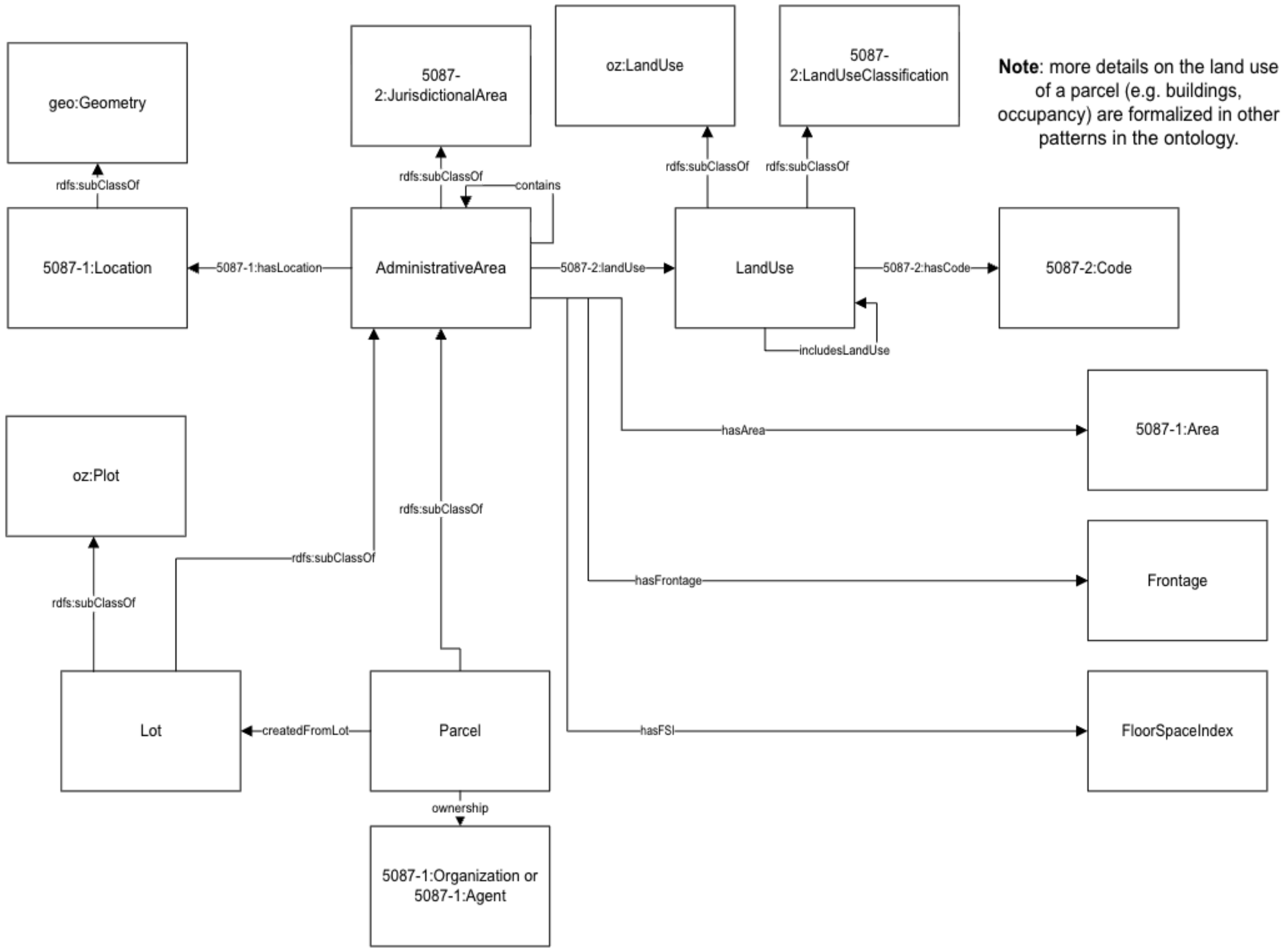

*Figure 5: Key classes and properties in the Administrative Area Pattern.*

#### 3.2.4.1 Formalization

The pattern is formalized in Table 12 and Table 13.

*Table 12: Definition of classes in the Administrative Area Pattern*

| Class | Property | Value Restriction |
| --- | --- | --- |
| AdministrativeArea | rdfs:subClassOf | 5087-2:JurisdictionalArea |
| | 5087-2:landUse | only LandUse |
| | 5087-1:partOf | some AdministrativeArea |
| | 5087-1:hasLocation | only 5087-1:Location |
| | hasArea | only 5087-1:Area |
| | hasFrontage | only Frontage |
| | hasFSI | only FloorSpaceIndex |
| | hasPerimeter | only 5087-1:Length |
| | hasDevelopmentClassification | only LandDevelopmentClassification |
| | inverse(occupies) | only 5087-2:InfrastructureElement |
| | zonedAsType | only ZoningType |
| | zonedForConstraint | only QuantityConstraint |
| | servicedBy | only Service |
| Parcel | rdfs:subClassOf | AdministrativeArea |
| | 5087-1:hasIdentifier | some xsd:string |
| | ownership | some 5087-1:Organization or 5087-1:Agent |
| | createdFromLot | only Lot |
| Lot | rdfs:subClassOf | AdministrativeArea |
| | rdfs:subClassOf | oz:Plot |
| | 5087-1:hasIdentifier | some xsd:string |
| | definedInPlan | only Plan |
| | allowsUse | only LandUse |
| | mayAllowUse | only LandUse |
| | doesNotAllowUse | only LandUse |
| 5087-2:LandUseClassification | 5087-2:hasCode | only 5087-2:Code |
| LandUse | rdfs:subClassOf | 5087-2:LandUseClassification |
| | rdfs:subClassOf | oz:LandUse |
| | includesLandUse | only LandUse |
| Frontage | rdfs:subClassOf | 5087-1:Length |
| FloorSpaceIndex | rdfs:subClassOf | i72:RatioQuantity |
| | i72:numerator | only i72:Sum |
| | i72:denominator | only 5087-1:Area |
| LandDevelopmentClassification | 5087-2:hasCode | only 5087-2:Code |

*Table 13: Definition of properties in the Administrative Area pattern.*

| Property | Characteristic | Value (if applicable) |
|---|---|---|
| hasArea | owl:subpropertyOf | 5087-2:hasArea |
| | owl:subpropertyOf | has |
| hasFrontage | owl:subpropertyOf | has |
| hasFSI | owl:subpropertyOf | has |
| includesLandUse | owl:TransitiveProperty | - |

### 3.2.5 Zoning Pattern

While several city data standards include some aspect of zoning, it is typically only specified by way of an indication of current or planned land use. Zoning bylaws go beyond land use to include restrictions on parcels and the infrastructure they contain. To capture this kind of information, the HPCDM reuses the OntoZoning (OZ), OntoPlanningRegulations (OPR) and OntoBuildableSpace (OBS) ontologies described in the previous section. The HPCDM also incorporates the ISO/IEC 21972 ontology to more precisely define zoning regulations, many of which may be formulated as indicators (e.g. ratios, averages). Figure 6 illustrates an overview of the key classes and properties.

- ZoningBylaw: defined as a specialization of 5087-2:Bylaw and a kind of InformationArtifact. Bylaws are defined in the Bylaw Pattern in ISO/IEC 5087-2, and this representation is extended in the HPCDM to more precisely define *zoning* bylaws and the ways in which they influence the housing potential of different areas in the city. A ZoningBylaw has the following properties:
  - 5087-1:hasProperPart: refers to a ZoningBylawPart that makes up some part of the Bylaw.
  - definesZoningType: identifies the ZoningType(s) defined by the zoning bylaw.
  - definesRegulation: identifies the Regulation(s) defined by the zoning bylaw. Regulations may be associated directly with a zoning bylaw, or with its parts. If a regulation is defined in a zoning bylaw part, it is also defined in any information artifacts (e.g. the ZoningBylaw) that it is a part of
- ZoningBylawPart: A subclass of InformationArtifact. The identification of a part within a Zoning Bylaw is important to allow for more precise references – it enables the deconstruction of zoning bylaw text into parts. ZoningBylawPart provides a generalization of the more specific types of parts defined in ISO/IEC 5087-2 (e.g., Clause, Definition). A part could be a chapter, a section, a definition, or even an arbitrary selection of (continuous) lines within the text. It has the following property:
    - 5087-1:hasPart: identifies a ZoningBylawPart that is part of the ZoningBylawPart (e.g., a section within a chapter).
- Regulation: defined as a specialization of opr:Regulation[12], a Regulation identifies some requirement or allowance specified (in a bylaw) on how some area(s) of the city may be used and developed. It is defined for some area of land (in some cases the entire city, though often a specific zone). Many regulations specify either the characterization of zoning types (what is or isn't permitted in a certain zone) or the application of a zoning types to an area (i.e., this area is of type X). However, regulations may also define constraints on the attributes of infrastructure (often, buildings) or even on the parcels themselves (e.g., minimum area), independent of a particular zoning type. A Regulation has the following properties:
    - definedIn: identifies a ZoningBylaw, ZoningBylawPart, or Plan that the Regulation is defined in. If something is defined in an information artifact, then it is also defined in any information artifacts that it (the defining artifact) is a part of.
    - regulationDefinedIn: a subproperty of definedIn that applies only for regulations; the inverse of definesRegulation.
    - opr:forZoningType: inherited from oz:ZoningType, identifies a Zoning Type that the Regulation applies to. For example, a particular regulation on height limitations may be defined for any buildings within a residential zone. Note that as discussed previously, not all regulations will specify an associated zoning type.
    - forLandUse: identifies a LandUse type that the Regulation applies to. Regulations may be defined to apply for certain types of land use, independent of zoning type. For example, in the City of Vancouver, a parcel used for a school must have a minimum frontage of 20.1 m.
    - definedFor: identifies zero or more 5087-2:LandAreas that the regulation applies to. This property is a generalization of opr:appliedTo as it allows for the specification of a general area of land (i.e., the "zone" that a regulation applies to).
    - designatesZoningType: identifies a ZoningType that is designated (for some land area) by a Regulation.
    - specifiesConstraint: identifies a QuantityConstraint that is defined by the Regulation. Multiple Quantity Constraints may be defined by an individual regulation (e.g., constraints on lot frontage and on FSI).
    - onPopulation: identifies the i72:Population that the regulation applies to. Note that the class i72:Population doesn't refer to a population in the typical sense of inhabitants of a particular region, it is defined to refer to any collection of objects, e.g. the population of parcels within a particular area or the population of

[12] While note included in the recent publications on OPR, correspondence with the authors has confirmed the introduction of such a class to allow for generalized use of the ontology outside of Singapore.

buildings that meet certain criteria. The ISO/IEC 21972 ontology provides the terminology to explicitly capture the criteria for Population membership. Since multiple QuantityConstraints may be specified for a single Regulation, it may apply to multiple Populations. These values may be specified or inferred (based on the specified constraints) for this property.
    - appliesTo: describes a relationship between a regulation and an area. This property is a generalization of definedFor; it captures both the direct and indirect relationship of a regulation being designated for some area. For example, in some cases a regulation is defined to apply based on zoning type – it can be inferred that such regulations apply to any area that is designated with that zoning type.

Different types of Regulation may be identified, such as land use regulations, regulations on parcels' characteristics (e.g. minimum area), regulations on infrastructure characteristics (e.g. maximum height). No requirements for such a classification have been identified, so these regulation types are not defined in the HPCDM.

- ZoningType: defined as a specialization of oz:ZoningType, a ZoningType defines various combinations (and amounts) of land use (e.g. Residential, Commercial Residential, Open Space) permitted in a particular area. They are also used as the basis for other regulations – for example, a series of regulations on lot size, density, building height and so on may be defined for a particular zoning type. Zoning Types have the following properties:
    - definedIn: identifies a ZoningBylaw or BylawPart that the ZoningType is defined in.
    - 5087-1:hasIdentifier: specifies (at most one) identifier for the Zoning Type, if defined.
    - 5087-1:hasName: specifies the name of the Zoning Type
    - 5087-1:hasDescription: specifies a description of the Zoning Type
    - oz:allowsUse: inherited from OZ, identifies a LandUse type that is included in the zoned area.
    - oz:mayAllowUse: inherited from OZ, identifies a LandUse type that may (under certain conditions or subject to review) be included in the zoned area.
    - oz:doesNotAllowUse: inherited from OZ, identifies a LandUse type that is prohibited in the zoned area.
    - hasMinUseQuantum: identifies a minimum Use Quantum required for the Zoning Type. A Zoning Type may have zero or more Use Quantums, each is defined with respect to a particular type of Land Use. Note that this property is distinct from (a generalization of) oz:hasMinUseQuantum, as the OntoZoning property applies only to related LandUse to Use Quantums.
    - hasMaxUseQuantum: identifies a minimum Use Quantum required for the Zoning Type. A Zoning Type may have zero or more Use Quantums, each is defined with respect to a particular type of Land Use. Note that this property is distinct from (a generalization of) oz:hasMaxUseQuantum, as the OntoZoning property applies only to related LandUse to Use Quantums.
    - subZoningType: identifies a specialization of a Zoning Type. This allows for the definition of a hierarchy of Zoning Types (e.g., Residential, Residential Detatched, …). This is useful because bylaws may define some regulations with respect to a general zone, and some with respect to a specialization.

- definesZoningException: identifies one or more exceptions (Regulations) that supersede some part of the usual regulations associated with the ZoningType.
  - zoningTypeDesignatedBy: identifies a regulation(s) that make a designation of this zoning type. This property is introduced to support implementation of the property chaining rule to define appliesTo.
- UseQuantum: defined as a specialization of oz:UseQuantum. A Use Quantum corresponds to the amount of use permitted (minimum or maximum). Instead of defining a Use Quantum exclusively for an individual Land Use object, in the HPCDM, a Use Quantum is defined according to some ZoningType and then associated to a type of Land Use.
  - forLandUse: identifies the LandUse type that the UseQuantum applies to.
  - oz:hasValue: identifies the use limitation as an i72:Quantity (typically, a percent).
- Plan: a type of InformationArtifact that specifies some intended (future) state for a municipality. The Plan class is currently minimally defined in the scope of the HPCDM. Its main role is to capture the source of various regulations and distinguish between actual and proposed policies. Plans may define (new) lot divisions, land use zoning, and other regulations. Plans may or may not be adopted at some point after their specification. A Plan has the following properties:
  - proposedFor: identifies the time at which the plan is proposed for adoption.
- hasZone: The zoning pattern also introduces the property hasZone to identify any zoning types that apply to a particular AdministrativeArea class. This property is similar to the oz:hasZone property defined in the OZ ontology, however the OZ property is restricted to the OZ classes oz:Plot and oz:ZoningType and so cannot be reused here. The hasZone property defined in the HPCDM may be defined and inferred through object property chaining, based on the existence of Regulations establishing the zoning types.
- oz:hasValue and i72:hasValue: it is our assessment that these properties are equivalent. We adopt the use of i72:hasValue in this model to maintain consistency and coherence with the use of the ISO/IEC 21972 throughout the rest of the model.
- oz:hasNumericalValue and i72:hasNumericalValue: it is our assessment that these properties are equivalent. We adopt the use of i72:hasNumericalValue in this model to maintain consistency and coherence with the use of the ISO/IEC 21972 throughout the rest of the model.

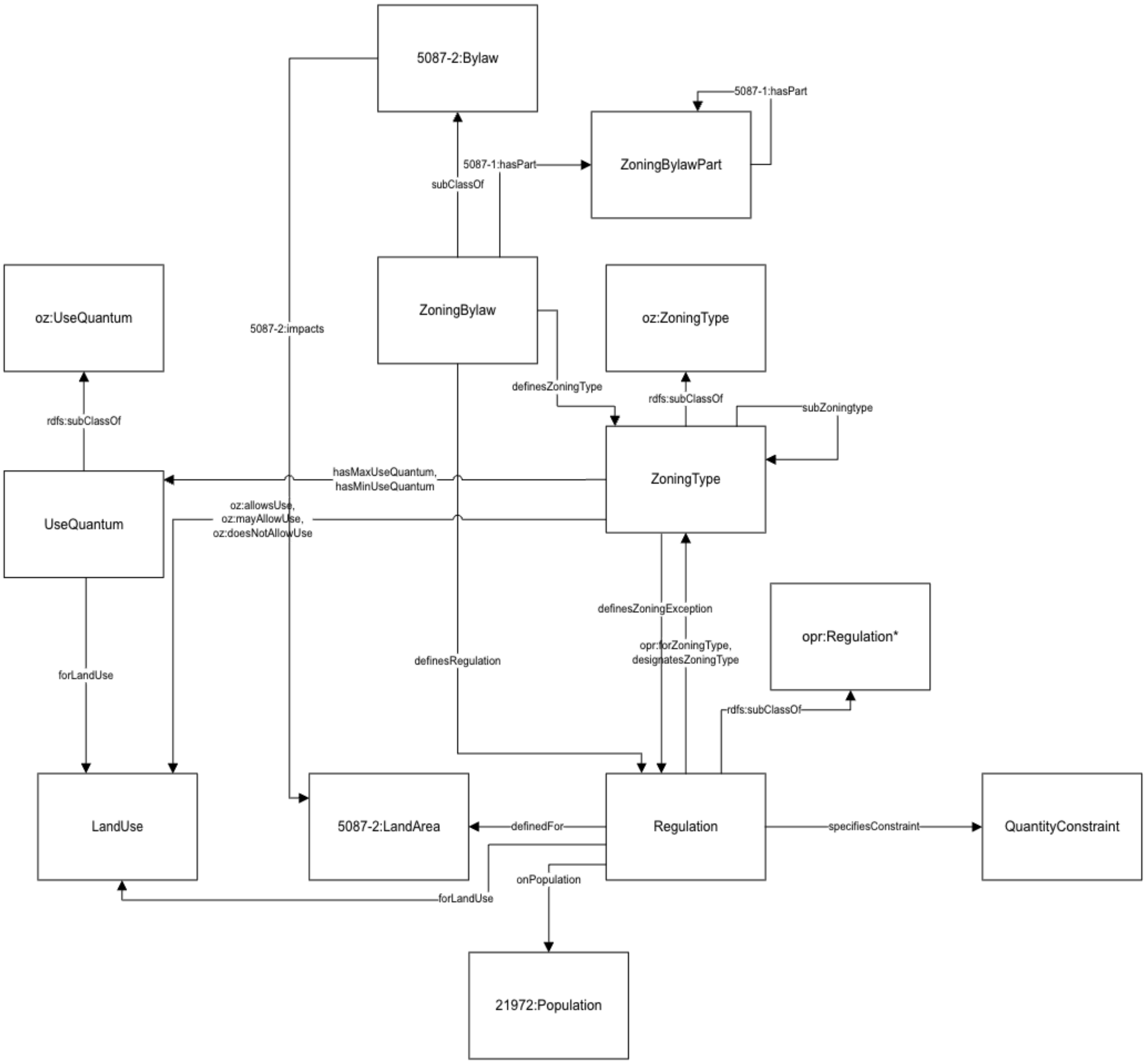

*Figure 6: Overview of the key classes and properties in the Zoning Pattern*

#### *3.2.5.1 Formalization*

The pattern is formalized in Table 14 and Table 15.

*Table 14: Definition of classes in the Zoning Pattern*

| Class | Property | Value Restriction |
|---|---|---|
| ZoningBylaw | rdfs:subClassOf | 5087-2:Bylaw |
| | 5087-1:hasProperPart | only ZoningBylawPart |
| | definesRegulation | only Regulation |
| | definesZoningType | only ZoningType |
| ZoningBylawPart | 5087-1:hasIdentifier | only xsd:string |
| | 5087-1:hasName | only xsd:string |
| | 5087-1:hasDescription | only xsd:string |
| | 5087-1:hasProperPart | only ZoningBylawPart |
| Regulation | rdfs:subClassOf | opr:Regulation |
| | definedIn | only ZoningBylaw or ZoningBylawPart or Plan |
| | opr:forZoningType | only ZoningType |
| | forLandUse | only LandUse |
| | definedFor | some AdministrativeArea |
| | designatesZoningType | only ZoningType |
| | onPopulation | only i72:Population |
| | specifiesConstraint | only QuantityConstraint |
| | appliesTo | only AdminsitrativeArea |
| ZoningType | rdfs:subClassOf | oz:ZoningType |
| | definedIn | only ZoningBylaw or ZoningBylawPart |
| | 5087-1:hasIdentifier | only xsd:string |
| | 5087-1:hasName | only xsd:string |
| | 5087-1:hasDescription | only xsd:string |
| | subZoningType | only ZoningType |
| | hasMaxUseQuantum | only UseQuantum |
| | hasMinUseQuantum | only UseQuantum |
| | oz:allowsUse | only LandUse |
| | oz:mayAllowUse | only LandUse |
| | oz:doesNotAllowUse | only LandUse |
| | definesZoningException | only Regulation |
| | zoningTypeDesignatedby | only Regulation |
| UseQuantum | rdfs:subClassOf | oz:UseQuantum |
| | forLandUse | only LandUse |
| | i72:hasValue | only i72:Quantity |
| AdministrativeArea | hasZone | only ZoningType |
| Plan | rdfs:subClassOf | InformationArtifact |
| | proposedFor | only time:TemporalEntity |

*Table 15: Definition of properties in the Zoning Pattern*

| Object Property | Characteristic | Value (if applicable) |
|---|---|---|
| inverse(definedFor) o designatesZoningType | owl:subPropertyOf | hasZone |
| regulationDefinedIn | owl:subPropertyOf | definedIn |
| regulationDefinedIn | owl:inverseOf | definesRegulation |
| definedFor | owl:subPropertyOf | appliesTo |
| opr:forZoningType o hp:zoningTypeDesignatedBy o hp:definedFor | owl:subPropertyOf | appliesTo |
| 5087-1:hasProperPart[13] o definesRegulation | owl:subPropertyOf | definesRegulation |
| definedIn o 5087-1:properPartOf | owl:subPropertyOf | definedIn |

#### 3.2.5.2 Planned versus actual characteristics

The Zoning Types (and regulations) described in this pattern may be associated to a bylaw to represent historical or in-effect zoning, or a Plan object (currently minimally defined) to capture proposed zoning. Land use may also be defined directly for a particular parcel to capture the current state; similarly, while regulations may define restrictions on land or built form characteristics, the HPCDM also supports the definition of the *actual* characteristics. This is important to enable a comparison of the existing built form with the stated allowances and requirements. There may be situations where the current state is in violation of the bylaws – it is important to be able to identify these discrepancies. It is also necessary to be able to consider the potential for *additional* built form, which requires understanding the difference between the current state and the regulations.

### 3.2.6 Building Pattern

Buildings are the infrastructure that provide housing and thus they represent a critical concept for housing potential. Zoning regulations place restrictions on the kinds of buildings that can be built. The use cases also point to a need to describe the existing buildings. The representation of buildings in the HPCDM is based on the Building Pattern defined in ISO/IEC 5087-2. Figure 7 illustrates the key classes and properties. The following classes are introduced:

- Building: A Building is a structure with a roof and walls. The HPCDM Building class extends the 5087-2:Building class to capture additional properties that are relevant for housing potential:
    - occupies: identifies an AdministrativeArea that the Building occupies.

---

[13] Note that hasProperPart should be defined as transitive, however this is restricted in OWL (since it is used in cardinality restrictions, and that makes it a non-simple). To implement the transitive definition will require the addition of a custom ruleset.

- occupiedBy: identifies a 5087-2:Person or 5087-1:Organization that occupies the building.
    - hasFootprintArea: identifies the FootprintArea of the Building's footprint. A Building must have some FootprintArea.
    - hasFloorArea: identifies the FloorArea of the Building. This refers to the total floor area for the building.
    - hasHeight: identifies the BuildingHeight for the Building.
    - hasSetback: identifies the setback of the Building as a BuildingSetback object.
    - hasCondition: specifies the condition of the building as a BuildingCondition object. BuildingCondition may be defined according to different classification systems using the 5087-2:hasCode property.
    - ownership: identifies a 5087-2:Person or 5087-2:Organization that owns the building.
    - hasNumberOfRooms: specifies the number of rooms in the Building as an xsd:nonNegativeInteger.
- BuildingUnit: A BuildingUnit refers to a part of a Building that is physically separate (i.e. has its own entrance and its own identifier within the building). It extends the 5087-2:BuildingUnit class with the following additional properties, relevant for housing potential:
    - 5087-2:use: identifies the use(s) of the Building Unit, defined as a 5087-2:BuildingUse (note: this property is already defined for 5087-2:Building).
    - hasFloorArea: identifies the FloorArea of the BuildingUnit. A BuildingUnit should have some FloorArea.
    - hasNumberOfRooms: specifies the number of rooms in the BuildingUnit as an xsd:nonNegativeInteger.
- DwellingUnit: represents a specialization of a Building or Building Unit that is intended to provide a living accommodation for a household. This is defined with a restriction on the 5087-2:use property that defines a "residential_occupancy" use for the DwellingUnit. This individual may be connected to any existing building use classification system through the 5087-2:hasCode class. The class may also be extended with specific conditions may also be added to further capture the definition (and any variances across municipalities) as required.
- BuildingFootprintArea: is a subclass of 5087-1:Area. In general, it is defined as the area of a Building's footprint. Differences in the definition of a building's footprint may arise in different municipalities and applications; these may be captured with extensions to this class, as required.
- FloorArea: is a subclass of 5087-1:Area. In general, it refers to the total floor area of the subject. Differences in the definition of total floor area (e.g., what floors and parts are included) may vary, and this class may be extended to capture these distinctions as required.
- BuildingHeight: is a subclass of 5087-1:Length. It refers to the height of a building, which can be measured in stories as well as units of measure typically associated with length. The precise definition of BuildingHeight may vary between municipalities (i.e., the definitions of the reference points for the measurement), so this class may be extended to capture these distinctions.

- BuildingSetback: a subclass of 5087-1:Length that generally refers to the distance from the face of the building to the street.

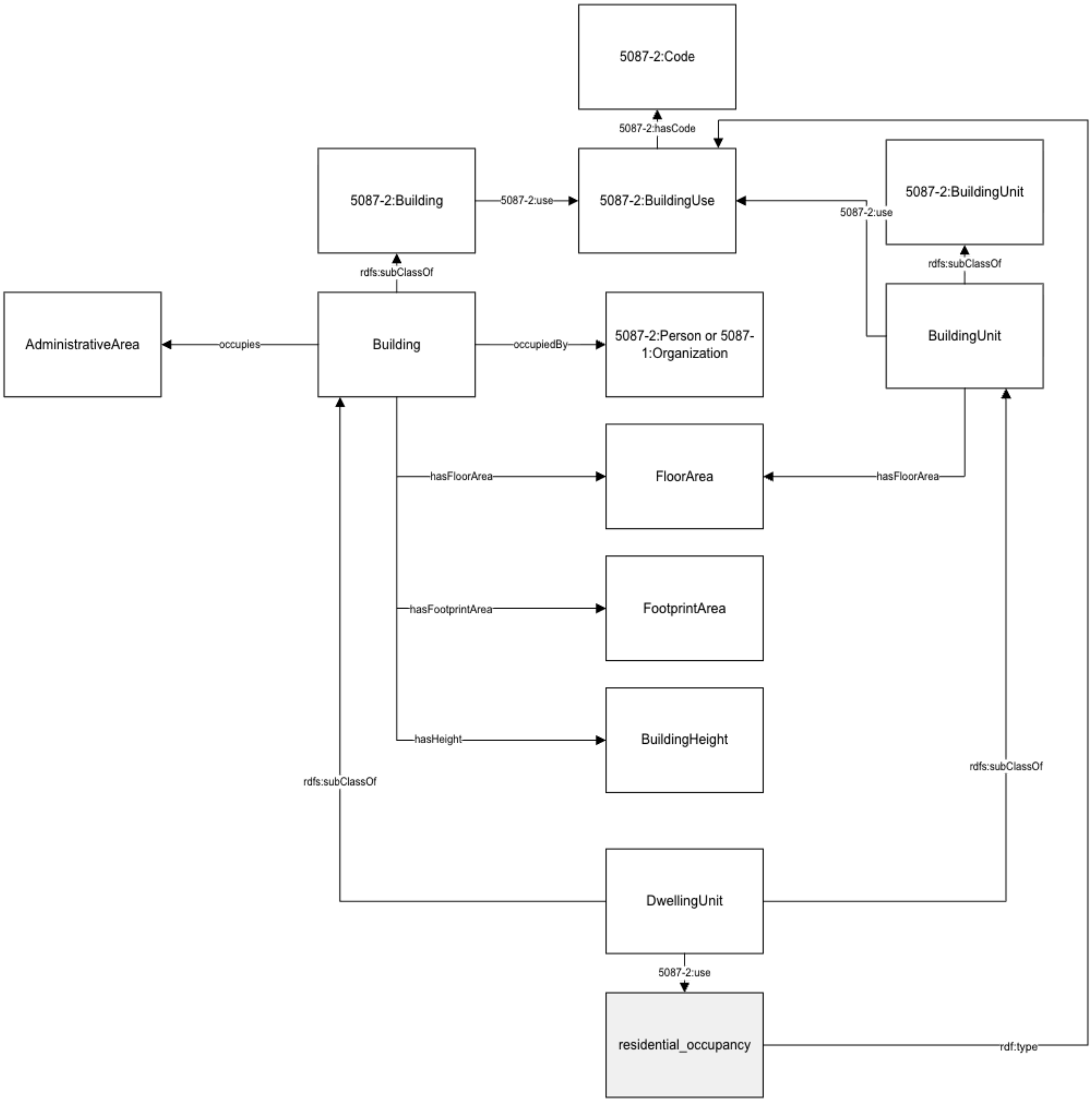

*Figure 7: Key classes and properties in the Building Pattern*

### *3.2.6.1 Formalization*

The pattern is formalized in Table 16 and Table 17.

*Table 16: Definition of classes in the Building Pattern*

| Class | Property | Value Restriction |
|---|---|---|
| Building | rdfs:subClassOf | 5087-2:Building |
| | occupies | only 5087-2:LandArea |
| | occupiedBy | only 5087-2:Person or 5087-1:Organization |
| | hasFootprintArea | only BuildingFootprintArea |
| | hasFloorArea | only FloorArea |
| | hasHeight | only BuildingHeight |
| | hasSetback | only BuildingSetback |
| | hasCondition | only BuildingCondition |
| | ownership | only 5087-2:Person or 5087-1:Organization |
| | hasNumberOfRooms | only xsd:nonNegativeInteger |
| BuildingUnit | rdfs:subClassOf | 5087-2:BuildingUnit |
| | occupiedBy | only 5087-2:Person or 5087-1:Organization |
| | 5087-2:use | only 5087-2:BuildingUse |
| | hasFloorArea | only FloorArea |
| | ownership | only 5087-2:Person or 5087-1:Organization |
| | hasNumberOfRooms | only xsd:nonNegativeInteger |
| DwellingUnit | rdfs:subClassOf | Building or BuildingUnit |
| | 5087-2:use | value residential_occupancy |
| BuildingFootprintArea | rdfs:subClassOf | 5087-1:Area |
| FloorArea | rdfs:subClassOf | 5087-1:Area |
| BuildingHeight | rdfs:subClassOf | 5087-1:Length |
| BuildingCondition | 5087-2:hasCode | only 5087-2:Code |

*Table 17: Definition of properties in the Building Pattern.*

| **Object Property** | **Characteristic** | **Value (if applicable)** |
|---|---|---|
| hasFootprintArea | owl:subPropertyOf | has |
| hasFloorArea | owl:subPropertyOf | has |
| hasHeight | owl:subPropertyOf | has |
| | owl:subPropertyOf | 5087-2:hasBuildingHeight |

### 3.2.7 Population Demographics Pattern

The Population Demographics pattern focuses on the description of households, the residents that comprise them, and the residences that they occupy. This kind of information is useful in providing an understanding of the area that the parcel(s) is located in (and is also required to describe housing goals and requirements). Specifically, this pattern is designed to support answering questions from the use cases, such as the following:

- What is the current density of the neighbourhood? (CQ 3b)
- What is the average rooms per home in the neighbourhood? (CQ 11c)
- What is the average land per home in the neighbourhood? (CQ 11d)
- How many residents occupy the parcel? (CQ 17)
- What is the population of the neighbourhood? (CQ 11a)
- What is the average annual income (household) of the neighbourhood? (CQ 11b)

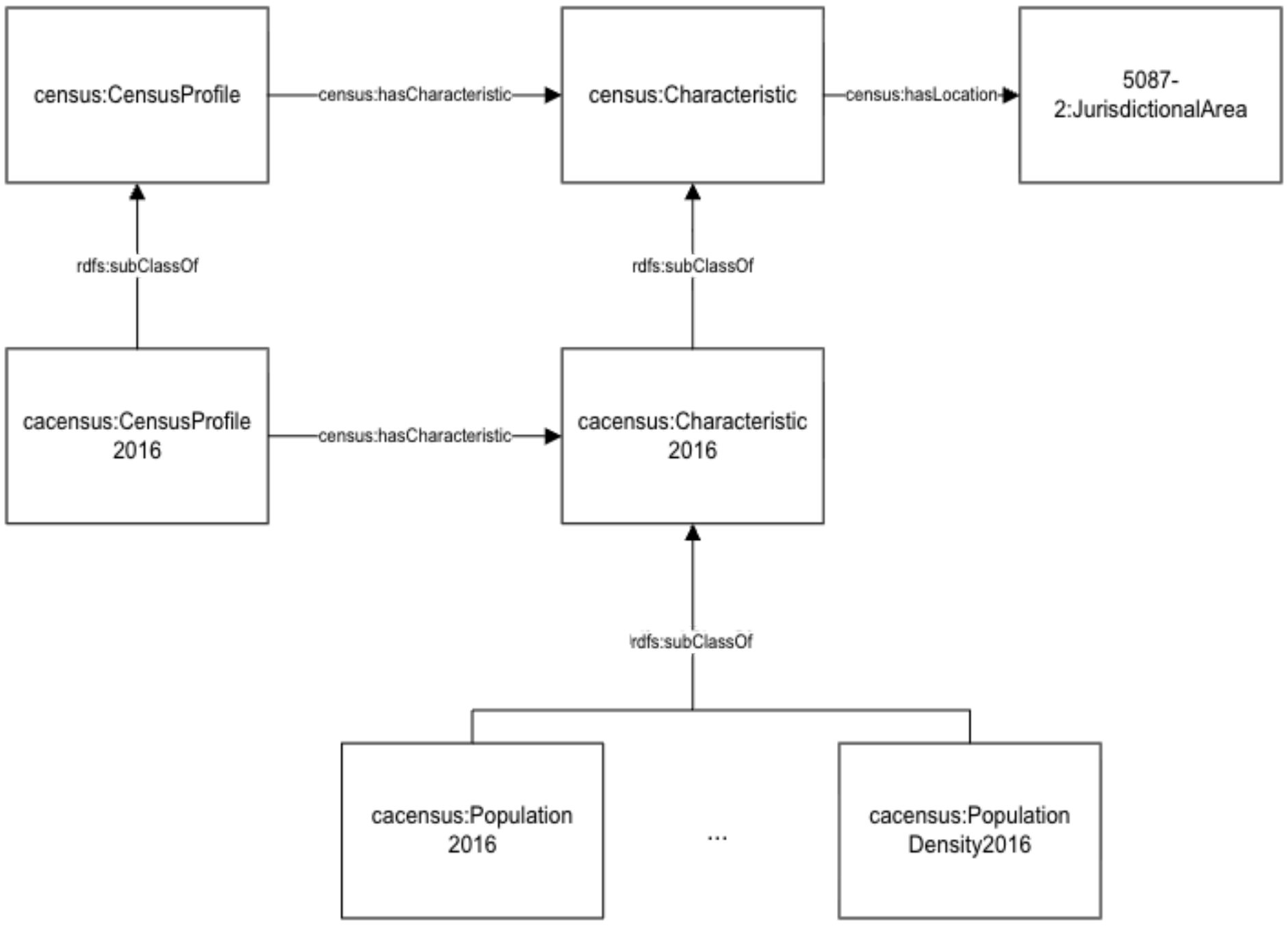

*Figure 8: Overview of the Canadian Census Ontology*

Many of these questions refer to data that is captured by the Canadian Census. The Canadian Census Ontology [39], identified in Section 3.1 forms the basis of this pattern. It provides a representation of demographic data, addressing many of the above requirements. The core classes of the ontology are census:CensusProfile, census:Characteristic, and census:CensusTract (a specialization of 5087-2:JurisdictionalArea). A Census Profile refers to the census data describing a geographic area, for a particular census year. Each profile is associated with many Characteristics – indicators that specify data about a particular area or population (e.g. average number of rooms per dwelling). Characteristics are defined for Census Tracts – which correspond to the area that the data describes. Separate Characteristic classes are defined for each census year, as definitions may change from one census to the next. An overview of the Census Ontology representation is illustrated in Figure 8.

Of the Characteristics defined for the 2016 census, the following capture demographic data relevant for the analysis of housing potential: census:Population2016, census:TotalPrivateDwellings2016, census:PopulationDensity2016, census:AverageNumberOfRoomsPerDwelling2016, and census:AverageAfterTaxIncome25Sample2016.

Note that the current Census Ontology includes characteristics from the 2016 Canadian Census of Population. It may be extended to capture data from the 2021 and future censuses as required.

### 3.2.8 Service Accessibility

Housing potential analysis requires a representation of both physical and soft city infrastructure that focuses on service area/access locations and service capacity (current and future). This pattern defines the core classes and properties necessary to capture this information. A range of different infrastructure services are relevant for the assessment of housing potential, from road networks to education to emergency services. Each is specified with its own pattern, defined as an extension to the core specified here. Additional types of services may also be defined as extensions to the core, as required for specific applications.

The basic requirements for services of interest for housing potential involve questions of the form:

- Does the area of land have access to the service?
- How much (available/planned/total) capacity does the service have?

The definition of *accessibility* to a service varies between services. Some define accessibility in terms of a service (catchment) area, whereas for others accessibility may be determined based on proximity to the service site *(Is the parcel on the road network? Is it a reasonable distance from a library?).*

The definition of the capacity of a service varies depending on the nature of the service. Capacities may also be measured or approximated in different ways, even for a single service; for example, while the capacity of solid waste services may be defined in terms of tonnes per year, it could also be captured as the number of households serviced per year. In addition, to consider *available* capacity requires a measure of what is currently being used as well as what the total capacity is.

The HPCDM reuses and extends the definition of Service specified in ISO/IEC 5087-2. In addition, it refers to the concepts of Resources and Activities from ISO/IEC 5087-1 to derive a representation of capacity information for a service. These extensions are outlined in Figure 9.

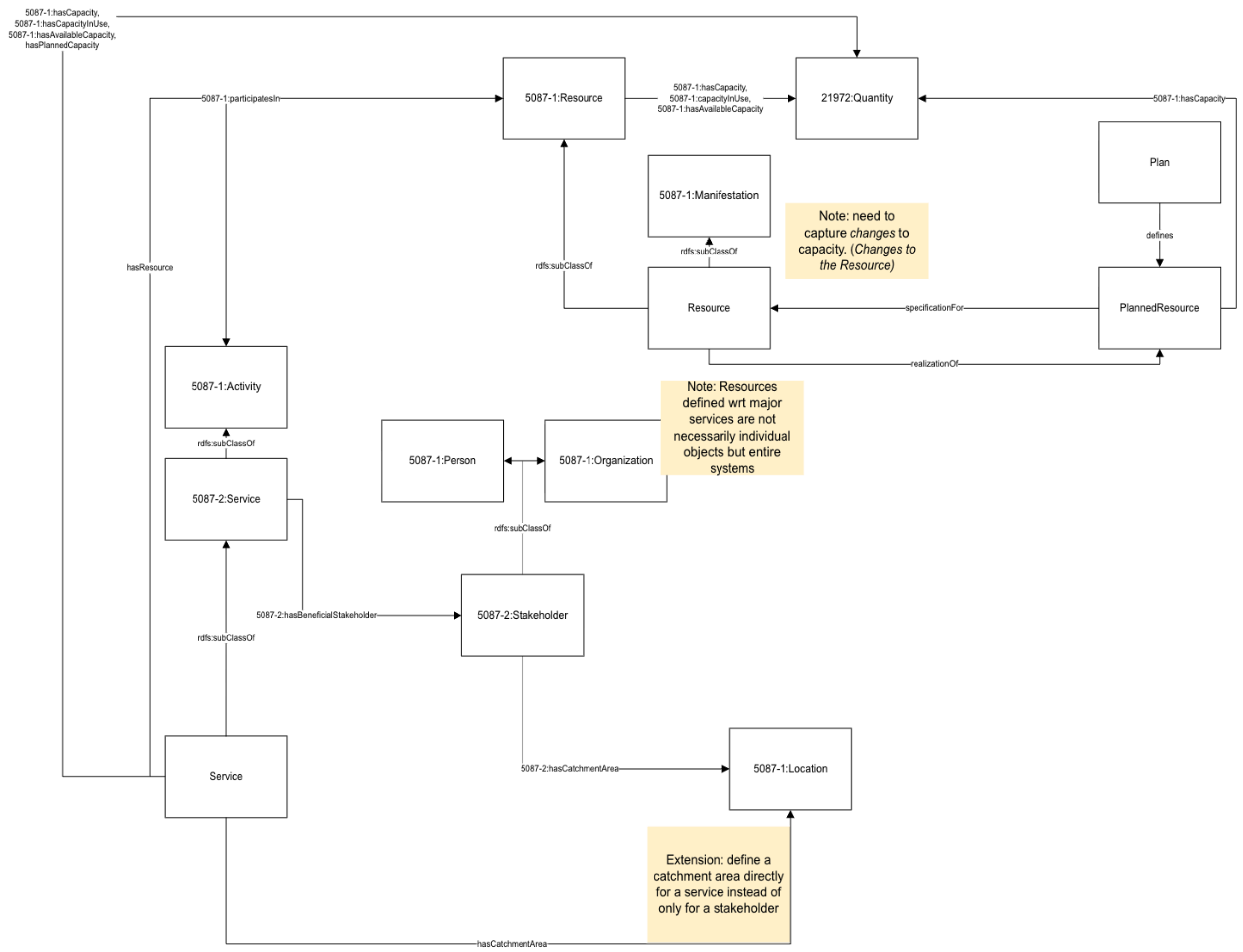

*Figure 9: Reuse of terms from ISO/IEC 5087-2 and ISO/IEC 5087-1.*

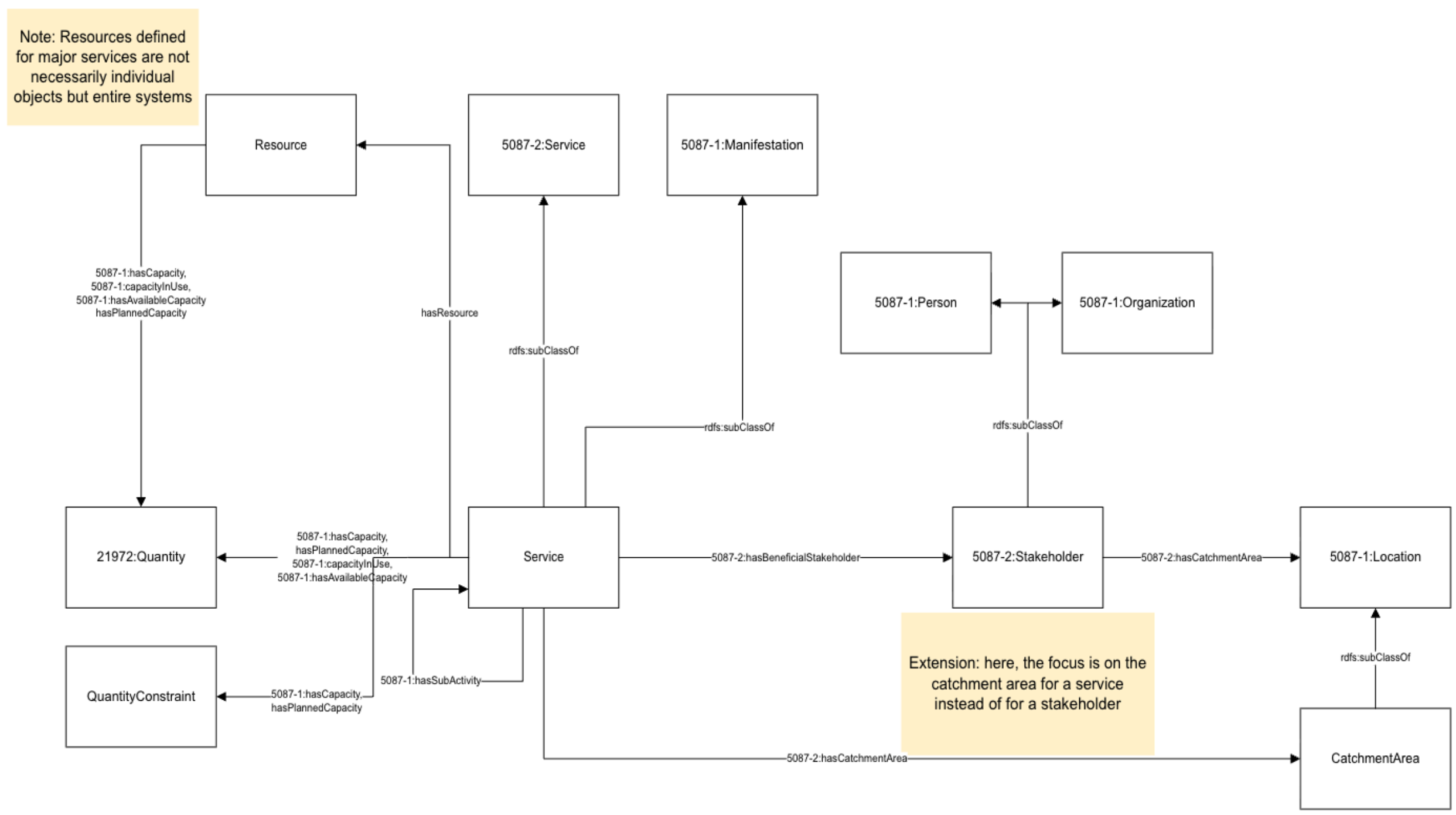

*Figure 10: Overview of the Service Accessibility Pattern*

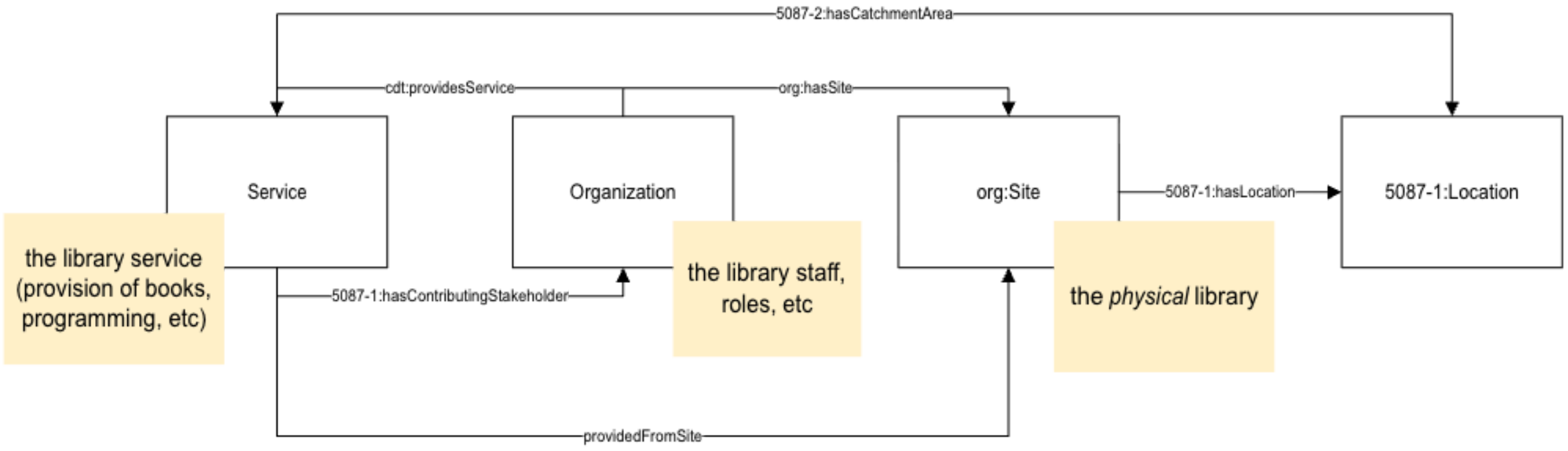

*Figure 11: Relationship between Services, Organizations, and Sites.*

The classes needed to define services in the context of housing potential are depicted in Figure 10 and Figure 11 and described as follows:

- Service: Service defined as a specialization of the city Service class introduced in ISO/IEC 5087-2. It refers, generally, to an activity that is provided to produce some output to its clients. While Services *often* refer to activities performed or offered by the city, they also include offerings from other organizations such as education services offered by universities or colleges or medical services offered by a pharmacy. The Service class in the HPCDM introduces some additional properties that are important for housing potential use cases. It is also defined as a subclass of 5087-1:Manifestation to capture the changes to the properties of a Service (in particular, its capacity) over time. Service has the following properties (a few are inherited from ISO/IEC 5087-2 but included for reference):

- 5087-2:hasCatchmentArea: identifies a Catchment Area that defines the location that the service is provided within. Not all services have formally defined catchment areas. For example, a supermarket serves any customer who comes in to the store (though its delivery service may be restricted to a particular catchment area). The catchment area of a service may be inferred based on the catchment area of its beneficial stakeholders (as defined in ISO/IEC 5087-2), if specified. Note that service provision within the defined catchment area is not strict. For example, in the case of emergency services the catchment areas define a location for the primary responder, however other responders may service the area as needed (e.g., based on availability).
- providedFromSite: identifies the 5087-1:Site(s) at which the Service is provided. Note that this includes situations where the service is provided *at* the site, and situations where the service *originates* at the site (e.g. is dispatched from) but is provided elsewhere. Sites are important to understand proximity to services when assessing the housing potential of an area.
- providedToParcel: identifies a Parcel that the service is provided or delivered to. In some cases, a service may be delivered directly to a parcel of land (e.g., for utilities). In many other cases, the provision of a service is provided to (residents of) the parcel may be inferred based on some defined catchment area.
- hasResource: identifies a Resource involved in the provision of the service. This property is the inverse of the 5087-1:participatesIn property, which associated Resources to the Activities they are involved in (Service is a subclass of 5087-2:Service and, indirectly, 5087-1:Activity). While this property is not directly required by the use cases, it is necessary to support potential inferences related to capacity based on the resources of a service.
- 5087-1:hasCapacity: identifies the capacity of a Service as a i72:Quantity or QuantityConstraint. In some cases, the total capacity may be defined based on characteristics of the service (e.g., enrollment spaces, physical infrastructure limitations) – these correspond to i72:Quantity specifications. However, in other cases, this type of measurement may not be feasible or practical. As an example, consider the capacity of Fire and Emergency Services. A measure of the total number of calls a that a fire station could respond to per day could be approximated, however this may not be very informative from a planning perspective. Instead, in such cases the total capacity may be defined as a kind of *target* maximum use (e.g. a level of service such as number of firefighters per 1000 residents) – in these cases the capacity is not a quantity of the service itself, but a QuantityConstraint on some aspect of the service.
- 5087-1:capacityInUse: identifies the capacity of a Service that is currently in use as an i72:Quantity.
- 5087-1:availableCapacity: identifies the capacity of a Service that is available (based on the total capacity and the capacity in use) as a DifferenceQuantity or QuantityConstraint. Available capacity is defined as the difference between the (total) capacity, and capacity in-use of a service.

  - hasPlannedCapacity: specifies a planned capacity for the Service as a QuantityConstraint. It is a constraint (defined in some plan) on the *future* capacity of the service.
  - hasSubService: a subproperty of 5087-1:hasSubActivity, it identifies a more granular Service that is part of the (larger) Service. This allows for the reference to both a broad service like *the power service* or *the solid waste service*, and a decomposition of the service as required to define capacities and availability with more precision, in terms of specific areas, or specific components of the service.
  - hasServiceRadius: specifies the radius from a service site from which the service may be delivered. This may be defined individually or at the class level.
- Utility: a subclass of Service defined to identify the category of essential services delivered through physical infrastructure. Examples of utility services include electricity and water.
- LowIncomeSupportService: Low income support services represent a category of services, rather than a single type of service. Examples of low-income support services could include food banks, job training centres, but also resources that serve the general public such as libraries and community centres. Other services may be identified as LowIncomeSupportServices in extensions to the model.
- Resource: a subclass of 5087-1:Resource. It introduces the following, additional property:
  - hasPlannedCapacity: identifies a planned capacity for a resource (e.g., due to planned improvements or modifications).
- CatchmentArea: a specialization of 5087-1:Location. It is extended with the following properties:
  - 5087-1:hasIdentifier: specifies an identifier assigned to the area as an xsd:string. Identifiers may be assigned within a particular system.
  - 5087-1:hasName: specifies a name for the catchment area as an xsd:string.
  - 5087-1:hasDescription: specifies a description for the catchment area as an xsd:string.
- ResidentPopulationSize: represents the number of residents (in a particular area). This class is useful for many types of services because the capacity of city services is often specified as some measure *per capita.* To represent these types of measures requires the specification of the value as a ratio of some quantity to the population in an area. It is defined as a subclass of i72:Cardinality that is the cardinality of a ResidentPopulation, where ResidentPopulation is a Population defined by objects from the 5087-2:CityResident class.

*A note on capacities*: there are often multiple ways in which the capacity of a service may be measured or specified. The service patterns define a general capacity class along with common definitions of capacity for the service. These general capacity classes are designed to be extended to capture the capacity quantities, beyond those defined in this model, as required.

#### *3.2.8.1 Formalization*

The pattern is formalized in Table 18 and Table 19.

*Table 18: Formalization of classes in the Service Accessibility pattern*

| Class | Property | Value Restriction |
|---|---|---|
| Service | rdfs:subClassOf | 5087-2:Service |
| | rdfs:subClassOf | 5087-1:Manifestation |
| | 5087-2:hasCatchmentArea | only CatchmentArea |
| | providedFromSite | only 5087-1:Site |
| | providedToParcel | only Parcel |
| | hasResource | only 5087-1:Resource |
| | 5087-1:hasCapacity | only (i72:Quantity or QuantityConstraint) |
| | 5087-1:capacityInUse | only i72:Quantity |
| | 5087-1:hasAvailableCapacity | only (DifferenceQuantity or QuantityConstraint) |
| | hasPlannedCapacity | only (i72:Quantity or QuantityConstraint) |
| | hasSubService | only Service |
| | hasServiceRadius | some 5087-1:Length |
| Utility | rdfs:subClassOf | Service |
| LowIncomeSupportService | rdfs:subClassOf | Service |
| Resource | rdfs:subClassOf | 5087-1:Resource |
| | hasPlannedCapacity | only (i72:Quantity or QuantityConstraint) |
| CatchmentArea | rdfs:subClassOf | 5087-1:Location |
| | 5087-1:hasIdentifier | only xsd:string |
| | 5087-1:hasName | only xsd:string |
| | 5087-1:hasDescription | only xsd:string |
| ResidentPopulationSize | rdfs:subClassOf | i72:Cardinality |
| | i72:cardinality_of | exactly 1 ResidentPopulation |
| ResidentPopulation | i72:definedBy | only 5087-2:CityResident |

*Table 19: Formalization of properties in the Service Accessibility pattern*

| Object Property | Characteristic | Value (if applicable) |
|---|---|---|
| hasSubService | owl:subPropertyOf | 5087-1:hasSubActivity |

### 3.2.9 Emergency Services

The emergency services pattern currently focuses on the formalization of fire prevention services. It may be extended to capture other emergency response services (e.g., paramedics, police) in the future. The emergency services pattern extends the service accessibility pattern. It is also informed by the design of the GCIO representation of fire and emergency services indicators [40], introducing a specification of capacity/use as a ratio of the number of firefighters to the size of the population. The GCIO indicators are defined specifically at the city level and cannot be reused as indicators for varying levels of administrative areas; however, the quantities and terms used to define the indicators can be reused to capture the required populations in the same way.

The pattern, illustrated in Figure 12, is comprised of the following classes:

- FireEmergencyService: a type of Service that involves the immediate provision assistance (via skilled personnel and sometimes specialized equipment) in response to situations that pose a serious risk to health, property, or the environment. It has the following properties, in addition to those inherited from the general Service class:
  - 5087-2:hasCatchmentArea identifies the location that the emergency service is responsible for. An Emergency Service must be responsible for some location.
  - 5087-1:hasCapacity: identifies the EmergencyServicesCapacity of the service. Various measures of capacity may be defined as appropriate and available. As an example, it may be recommended that there be a ratio of 1 firefighter per 1000 residents. This ratio could be specified as the capacity of the service.
  - 5087-1:capacityInUse: identifies the capacity in use of the fire and emergency service. Again, due to the nature of the service this doesn't refer to the number of firefighters or equipment engaged in firefighting at any given time, but rather an estimate of how well the existing resources are able to cover the typical demands of the population. Again, a ratio of firefighters to the actual population is a possible measure.
  - 5087-1:hasAvailableCapacity: identifies the available capacity of the fire and emergency service. This should be formulated as the difference (DifferenceQuantity) between the defined capacity and the capacity in use.
  - hasPlannedCapacity: identifies a planned EmergencyServicesCapacity of the service.
- EmergencyServicesCapacity: a type of i72:Quantity or QuantityConstraint that represents the capacity of an emergency service. Various measures of capacity may be defined as appropriate and available.
- MinFirefightersPerPopulation: One way of defining the capacity of fire and emergency services is in terms of a limit on the ratio of firefighters to population. As an example, it may be recommended that there be a ratio of 1 firefighter per 1000 residents. This ratio would be defined as the capacity of the service. It is a subclass of QuantityConstraint (as it is not an actual quantity of the service, but a limit imposed by some external source). It does not have a defined numerator and denominator, as it is a constraint on the ratio, not a ratio itself. It has the following properties:
  - i72:hasUnit: specifies the i72:population_ratio_unit as the unit of measure for the capacity

- FirefightersPerPopulation: one means of specifying the capacity in use. It is a type of EmergencyServicesCapacity. It is the actual measure of firefighters per population, defined analogously to the indicator specified in the GCIO Fire and Emergency Response Ontology: a ratio indicator with a numerator of the number of fulltime firefighter, and a denominator of the population (in this case, of any area, not just the entire city). It has the following properties:
    - i72:denominator: identifies the ResidentPopulationSize of the area that the ratio is defined for.
    - i72:numerator: identifies the FullTimeFirefighterCount in the area that the ratio is defined for.
    - i72:hasUnit: specifies the i72:population_ratio_unit as the unit of measure for the capacity
    - forLocation: identifies the 5087-1:Location (e.g., the catchment area) that the ratio is defined for.
- AvailableFirefightersPerPopulation: is a kind of QuantityConstraint that represents the available capacity of the service, as the difference between the capacity and the capacity in use. In other words, what change to the service use is possible without exceeding its defined capacity? It has the following property:
    - i72:term_2: specifies the MinFirefightersPerPopulation capacity of the service.
    - i72:term_1: specifies the FirefightersPerPopulation capacity in use of the service.
    - i72:hasUnit: specifies the i72:population_ratio_unit as the unit of measure for the capacity
    - forLocation: identifies the 5087-1:Location (e.g., the catchment area) that the ratio is defined for.
- FullTimeFirefighterCount: is a cardinality class that represents the size of a population of firefighterters, defined by the class gcife:10.1_FullTimeFirefighter_OpationalStaff_population

Note that the representation of services in the HPCDM allows for an aggregation of the service at the city-wide level (e.g. “Toronto Fire Services, with a catchment area of the entire city and 84 fire stations”) as well as a decomposition of the service to the site level (e.g., “Station 355 Fire Services, comprised of a single fire station, responsible for the Spadina-Fort York area in the city).

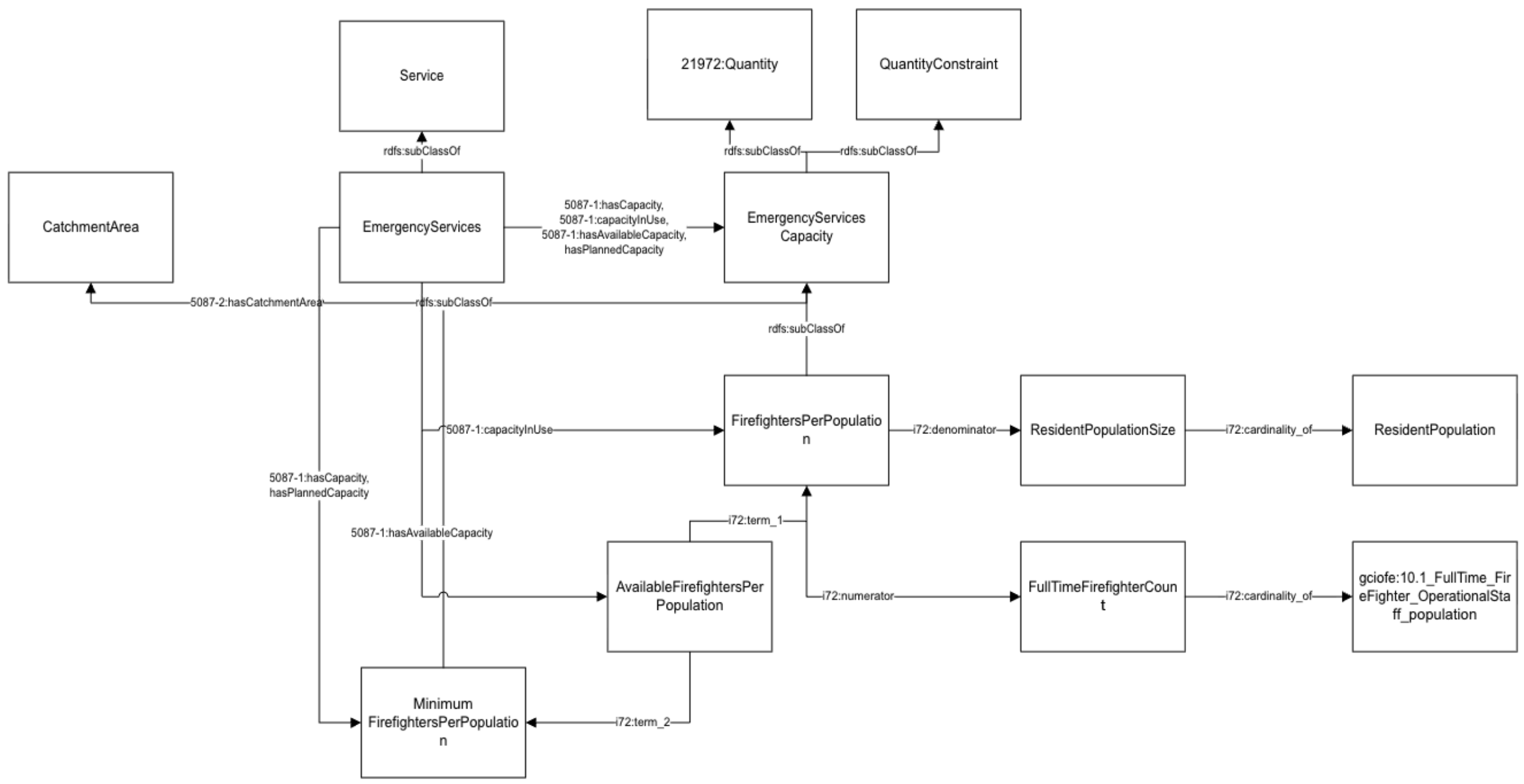

*Figure 12: Overview of the Fire and Emergency Service Access Pattern*

### *3.2.9.1 Formalization*

The pattern is formalized in Table 20

*Table 20: Formalization of classes in the Fire and Emergency Service Access Pattern*

| Class | Property | Value Restriction |
|---|---|---|
| FireEmergencyService | rdfs:subClassOf | Service |
| | 5087-2:hasCatchmentArea | some CatchmentArea |
| | 5087-1:hasCapacity | only EmergencyServicesCapacity |
| | 5087-1:capacityInUse | only EmergencyServicesCapacity |
| | 5087-1:hasAvailableCapacity | only EmergencyServicesCapacity |
| | hasPlannedCapacity | only EmergencyServicesCapacity |
| EmergencyServicesCapacity | rdfs:subClassOf | i72:Quantity or QuantityConstraint |
| MinFirefightersPerPopulation | rdfs:subClassOf | EmergencyServicesCapacity |
| | rdfs:subClassOf | QuantityConstraint |
| | i72:hasUnit | value i72:population_ratio_unit |
| FirefightersPerPopulation | rdfs:subClassOf | EmergencyServicesCapacity |
| | rdfs:subClassOf | RatioQuantity |
| | i72:numerator | exactly 1 FullTimeFirefighterCount |
| | i72:denominator | exactly 1 ResidentPopulationSize |
| | i72:hasUnit | value i72:population_ratio_unit |
| | forLocation | only 5087-1:Location |
| AvailableFirefightersPerPopulation | rdfs:subClassOf | EmergencyServicesCapacity |
| | rdfs:subClassOf | QuantityConstraint |
| | i72:hasUnit | value i72:population_ratio_unit |
| | i72:term_1 | exactly 1 FirefightersPerPopulation |
| | i72:term_2 | exactly 1 MinFirefightersPerPopulation |
| | forLocation | only 5087-1:Location |
| FullTimeFirefighterCount | rdfs:subClassOf | i72:Cardinality |
| | i72:cardinality_of | Exactly 1 gcife:10.1_FullTimeFirefighter_OpationalStaff_population |

### 3.2.10 Electricity

This pattern, depicted in Figure 13, reuses and extends the general Service Accessibility pattern to capture accessibility to electricity services. It defines the following classes:

- ElectricService: represents the service that delivers electricity to customers (residential and commercial) from some power generation service. It is defined as a Utility and has the following properties:
  - hasCatchmentArea: identifies service areas within which connections to the network exist (or are assumed to exist). Parcels within a catchment area are assumed to be serviced by the power network.
  - 5087-1:hasCapacity: identifies the ElectricServiceCapacity that indicates how much of the service can be used (in total).
  - 5087-1:hasAvailableCapacity: identifies the ElectricServiceCapacity that indicates how much of the service is available to be used (i.e., considering what is already in use).
  - 5087-1:capacityInUse: identifies the ElectricServiceCapacity that indicates how much of the service is currently in use.
  - hasPlannedCapacity: identifies a planned ElectricServiceCapacity that indicates how much of the service could be used (in total).
- ElectricServiceCapacity: is an i72:Quantity or QuantityConstraint that represents a measure of the capacity of the power service (total, in use, or available).

- ElectricalLoad: a specialization of ElectricServiceCapacity and a i72:Quantity. The electrical load capacity specifies the (peak) amount of electricity delivered to an area. It is a measure of the capacity in use of a service.
- ElectricalLoadCapacity: a specialization of ElectricServiceCapacity and a i72:Quantity. The electrical load capacity specifies the amount of electricity that *can be* delivered to an area (i.e., based on the local grid infrastructure). It is one possible measure of the capacity of an electrical service.
- AvailableElectricalCapacity: a specialization of ElectricServiceCapacity and a i72:Quantity. The available electrical load capacity specifies the amount of electricity *available to be* delivered to an area (i.e., based on the capacity and what's currently being used). It is a subclass of DifferenceQuantity as it can be defined as the difference between the electrical load capacity and the peak electrical load.

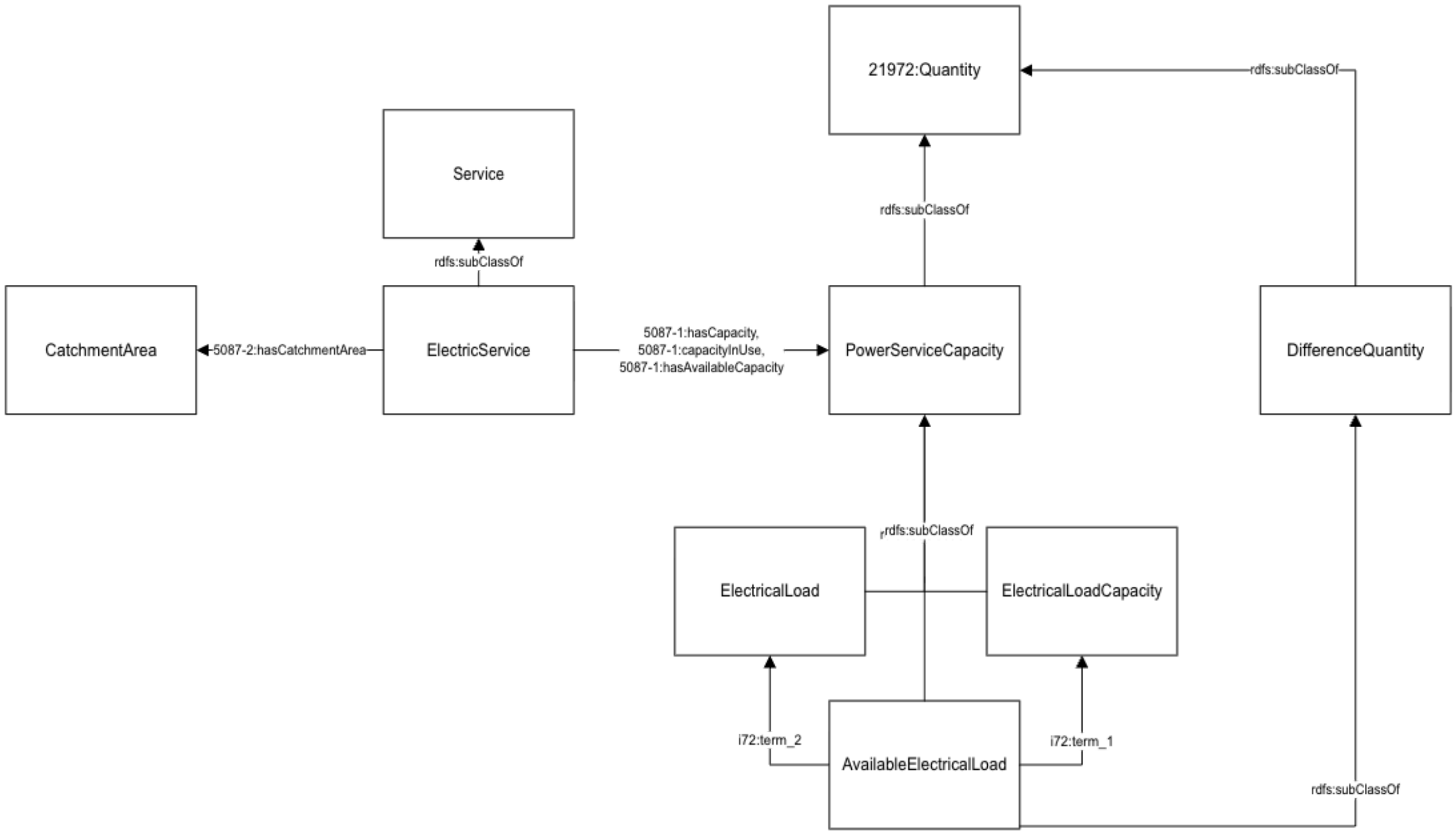

*Figure 13: Overview of the Electricity Service Access Pattern*

### *3.2.10.1 Formalization*

The pattern is formalized in Table 21.

*Table 21: Formalization of classes in the Electricity Service Access Pattern*

| Class | Property | Value Restriction |
|---|---|---|
| ElectricService | rdfs:subClassOf | Utility |
| | 5087-1:hasCapacity | only ElectricServiceCapacity |
| | 5087-1:capacityInUse | only ElectricServiceCapacity |
| | 5087-1:hasAvailableCapacity | only ElectricServiceCapacity |
| | hasPlannedCapacity | only ElectricServiceCapacity |
| ElectricServiceCapacity | rdfs:subClassOf | i72:Quantity or QuantityConstraint |
| ElectricalLoad | rdfs:subClassOf | ElectricServiceCapacity |
| | rdfs:subClassOf | i72:Quantity |
| | i72:hasUnit | value hp:kilovolt_ampere |
| | forLocation | only 5087-1:Location |
| ElectricalLoadCapacity | rdfs:subClassOf | ElectricServiceCapacity |
| | rdfs:subClassOf | i72:Quantity |
| | i72:hasUnit | value hp:kilovolt_ampere |
| | forLocation | only 5087-1:Location |
| AvailableElectricalCapacity | rdfs:subClassOf | ElectricServiceCapacity |
| | rdfs:subClassOf | i72:Quantity |
| | i72:hasUnit | value hp:kilovolt_ampere |
| | i72:term_1 | exactly 1 ElectricalLoadCapacity |
| | i72:term_2 | exactly 1 ElectricalLoad |
| | forLocation | only 5087-1:Location |

### 3.2.11 Solid Waste

This pattern, depicted in Figure 14, reuses and extends the general Service Accessibility pattern to capture accessibility to solid waste services. It defines the following classes:

- SolidWasteService: refers to service that collects solid waste (garbage, recycling, organics) from homes.
    - hasCatchmentArea: all solid waste services have some defined CatchmentArea(s) that are serviced by their collection routes.
    - 5087-1:hasCapacity: based on the current resources (trucks, transfer stations), a solid waste service will have some capacity to collect and process waste. This is specified as a SolidWasteServiceCapacity.
    - 5087-1:capacityInUse: specifies the current use of the solid waste service as a SolidWasteServiceCapacity.
    - 5087-1:hasAvailableCapacity: based on the current resources and the current levels of service, a solid waste service will have some (possibly zero or even a deficit) capacity to collect and process *additional* waste. This is specified as a

SolidWasteServiceCapacity and should be the difference between the corresponding values of 5087-1:hasCapacity and 5087-1:capacityInUse.
    - hasPlannedCapacity: identifies a planned capacity to collect and process waste, specified as a SolidWasteServiceCapacity.
- SolidWasteServiceCapacity: a i72:Quantity or QuantityConstraint that defines a measure of the service's use/availability. This capacity can be captured in different ways, for example based on rates of collection or total waste volume.
- WasteProcessingRate: a type of SolidWasteServiceCapacity. This RatioQuantity specifies the rate at which waste is processed by the service (often, in tonnes per year). It is a possible measure of the capacity in use of a service. It is a ratio of TotalWasteProcessedMass to the 5087-1:Duration (over which it is processed).
- MaxWasteProcessingRate: a type of SolidWasteServiceCapacity. It is a i72:Quantity that specifies the maximum rate at which waste *can be* processed by the service (often, in tonnes per year).
- AvailableWasteProcessingRate: a type of SolidWasteServiceCapacity. This Quantity indicates the surplus waste processing capacity (if any). It is the difference between the corresponding values of the MaxWasteProcessingRate and the WasteProcessingRate for a service.
- TotalWasteProcessedMass: a measure of (the mass of) waste processed by the solid waste service during some time period.

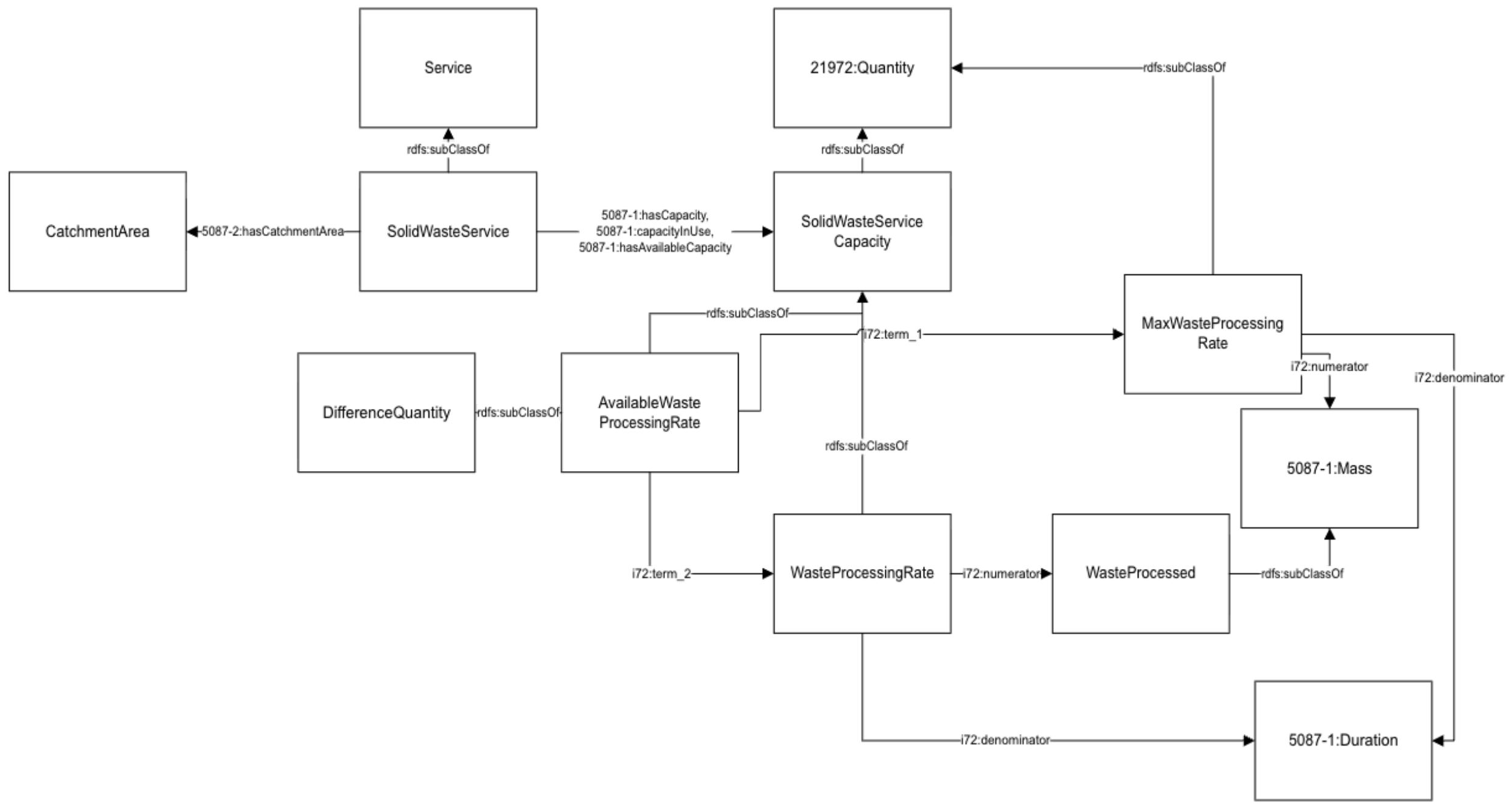

*Figure 14: Overview of the Solid Waste Service Access Pattern*

### 3.2.11.1 Formalization

The pattern is formalized in Table 22.

*Table 22: Solid Waste Service Access Pattern Class formalization*

| Class | Property | Value Restriction |
|---|---|---|
| SolidWasteService | rdfs:subClassOf | Service |
| | 5087-1:hasCapacity | only SolidWasteServiceCapacity |
| | 5087-1:capacityInUse | only SolidWasteServiceCapacity |
| | 5087-1:hasAvailableCapacity | only SolidWasteServiceCapacity |
| | hasPlannedCapacity | only SolidWasteServiceCapacity |
| SolidWasteServiceCapacity | rdfs:subClassOf | i72:Quantity or QuantityConstraint |
| WasteProcessingRate | rdfs:subClassOf | SolidWasteServiceCapacity |
| | rdfs:subClassOf | RatioQuantity |
| | forLocation | only 5087-1:Location |
| | i72:numerator | exactly 1 TotalWasteProcessedMass |
| | i72:denominator | 5087-1:Duration |
| MaxWasteProcessingRate | rdfs:subClassOf | SolidWasteServiceCapacity |
| | rdfs:subClassOf | RatioQuantity |
| | i72:numerator | exactly 1 i72:Mass |
| | i72:denominator | 5087-1:Duration |
| | forLocation | only 5087-1:Location |
| AvailableWasteProcessingRate | rdfs:subClassOf | SolidWasteServiceCapacity |
| | rdfs:subClassOf | DifferenceQuantity |
| | i72:term_1 | exactly 1 MaxWasteProcessingRate |
| | i72:term_2 | exactly 1 WasteProcessingRate |
| | forLocation | only 5087-1:Location |
| TotalWasteProcessedMass | rdfs:subClassOf | 5087-1:Mass |

### 3.2.12 Water Distribution

This pattern, depicted in Figure 15, reuses and extends the general Service Accessibility pattern to capture accessibility to water services. It defines the following classes:

- WaterService: refers to the service that provides water to homes and businesses. It is defined as a Utility and has the following properties:

- 5087-1: hasCapacity: identifies a WaterDistributionCapacity that indicates how much water the service is capable of providing, in total.
  - 5087-1:hasAvailableCapacity: identifies a WaterDistributionCapacity that indicates how much additional water the service is capable of providing.
  - 5087-1:capacityInUse: identifies a WaterDistributionCapacity that indicates how much water the service is currently.
  - hasPlannedCapacity: identifies a planned WaterDistributionCapacity that indicates how much water the service would be capable of providing, in total.
- WaterDistributionCapacity: a measure of the water service's capacity (total, in use, or available).
- WaterDistributionRate: a type of WaterDistributionCapacity. It is a i72:Quantity that defines the total volume of water distributed over some time period. It is a possible measure of the capacity in-use of the water service.
- MaxWaterDistributionRate: a type of WaterDistributionCapacity. It is a i72:Quantity that represents the maximum, total volume of water that can be distributed by the system over some time period. It is a possible measure of the total capacity of a water service.
- AvailableWaterDistributionRate: a type of WaterDistributionCapacity. It is a DifferenceQuantity that represents the water distribution rate that is "available", i.e., how much more water may be distributed until the system is at its capacity.
- WaterDistributionVolume: a measure of the total amount of water distributed during some time period.

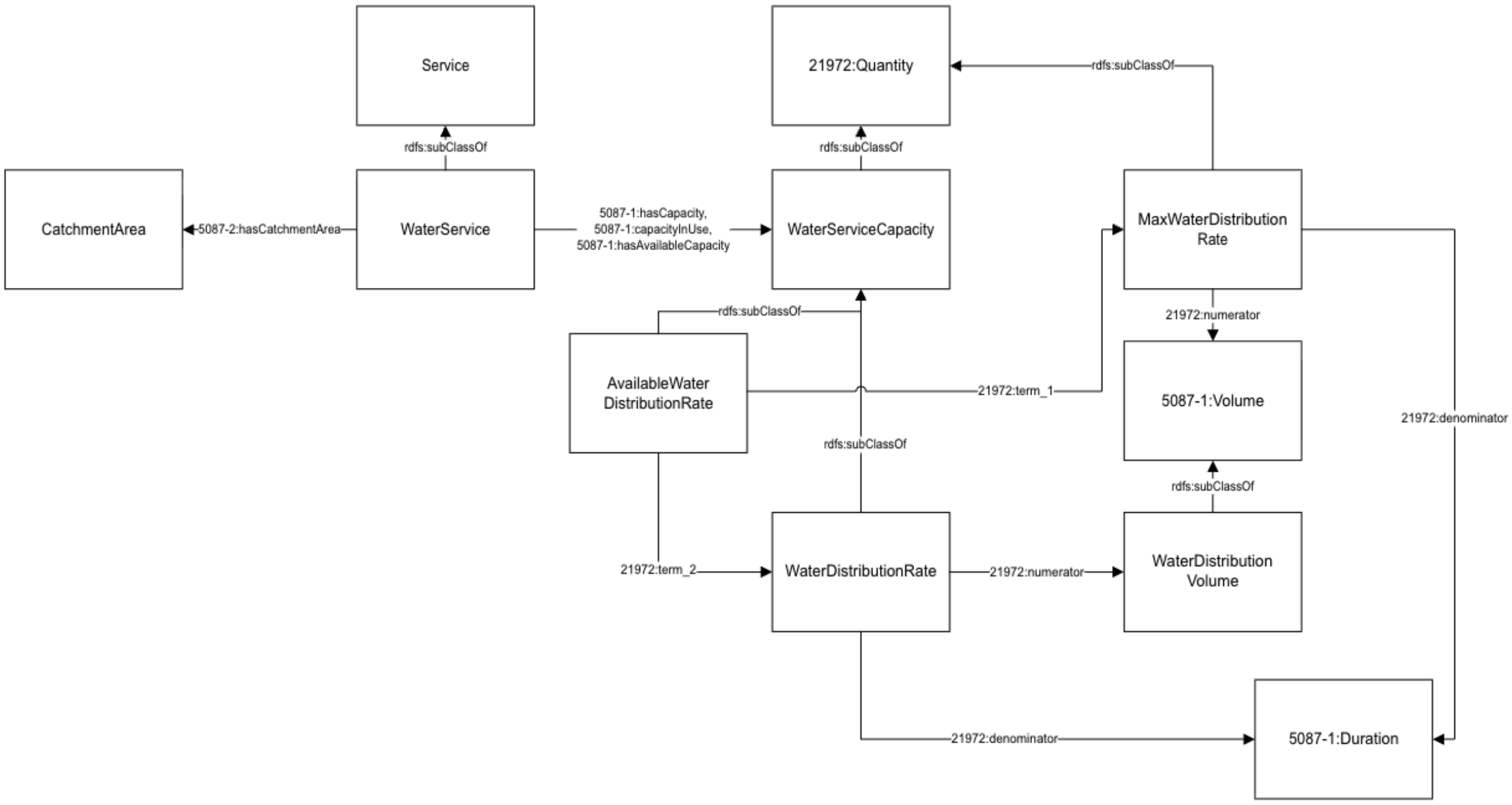

*Figure 15: Overview of the Water Distribution Service Access Pattern*

### *3.2.12.1 Formalization*

The pattern is formalized in Table 23.

*Table 23: Formalization of Water Distribution Service Access Pattern classes*

| Class | Property | Value Restriction |
|---|---|---|
| WaterService | rdfs:subClassOf | Utility |
| | 5087-1:hasCapacity | only WaterDistributionCapacity |
| | 5087-1:capacityInUse | only WaterDistributionCapacity |
| | 5087-1:hasAvailableCapacity | only WaterDistributionCapacity |
| | hasPlannedCapacity | only WaterDistributionCapacity |
| WaterDistributionCapacity | rdfs:subClassOf | i72:Quantity or QuantityConstraint |
| WaterDistributionRate | rdfs:subClassOf | WaterDistributionCapacity |
| | rdfs:subClassOf | RatioQuantity |
| | forLocation | only 5087-1:Location |
| | i72:numerator | exactly 1 WaterDistributionVolume |
| | i72:denominator | 5087-1:Duration |
| MaxWaterDistributionRate | rdfs:subClassOf | WaterDistributionCapacity |
| | rdfs:subClassOf | RatioQuantity |
| | forLocation | only 5087-1:Location |
| | i72:numerator | exactly 1 Volume |
| | i72:denominator | 5087-1:Duration |
| AvailableWaterDistributionRate | rdfs:subClassOf | WaterDistributionCapacity |
| | rdfs:subClassOf | DifferenceQuantity |
| | i72:term_1 | exactly 1 MaxWaterDistributionRate |
| | i72:term_2 | exactly 1 WaterDistributionRate |
| | forLocation | only 5087-1:Location |
| WaterDistributionVolume | rdfs:subClassOf | 5087-1:Volume |

### 3.2.13 Wastewater

This pattern, depicted in Figure 16, reuses and extends the general Service Accessibility pattern to capture accessibility to wastewater services. It defines the following classes:

- WastewaterService: refers to the service that processes wastewater from homes and businesses. It is defined as a type of Utility and has the following properties:

- 5087-1:hasCapacity: identifies a WastewaterServiceCapacity that specifies the total amount of water that can be processed by the service.
- 5087-1:hasAvailableCapacity: identifies a WastewaterServiceCapacity that specifies the amount of water that can be processed by the service, in addition to what is currently being processed.
- 5087-1:capacityInUse: identifies a WastewaterServiceCapacity that specifies the amount of water that is being processed by the service.
- hasPlannedCapacity: identifies a planned WastewaterServiceCapacity that specifies the total amount of water that could be processed by the service.

- WastewaterServiceCapacity: a measure how much wastewater the service can process (total, in use, or available).
- WaterProcessingRate: a type of WastewaterServiceCapacity. It is a i72:Quantity that defines the total volume of water processed over some time period. It is a possible measure of the actual or available capacity of the water service.
- MaxWaterProcessingRate: a type of WastewaterServiceCapacity. It is a QuantityConstraint that defines the maximum volume of water (per time) that can be processed over some time period.
- AvailableWaterProcessingRate: a DifferenceQuantity that identifies the waste processing rate that is available in the system. It is defined as the difference between the MaxWaterProcessingRate and the WaterProcessingRate.
- WaterProcessedVolume: a measure of the total amount of water processed during some time period.

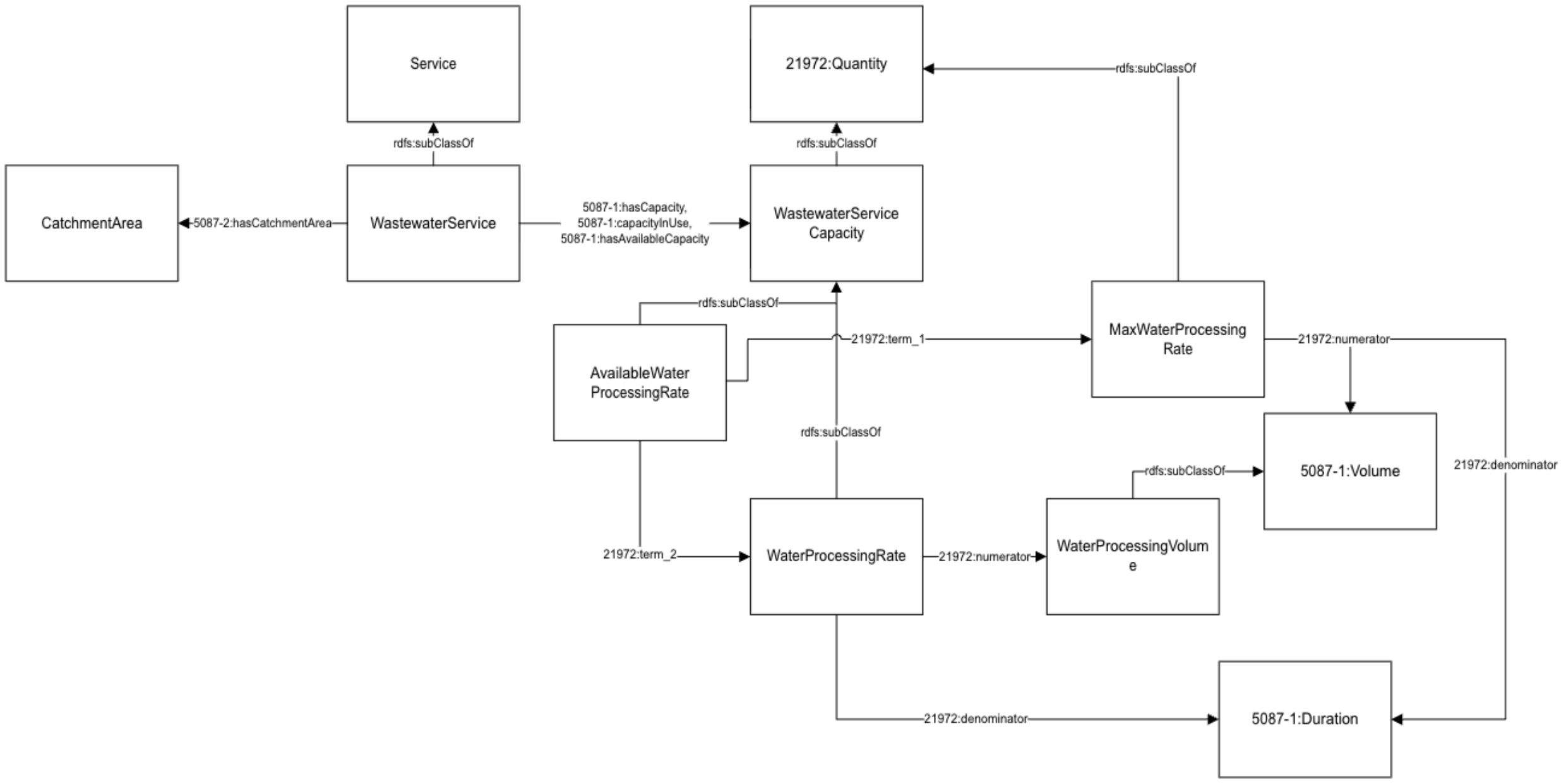

*Figure 16: Overview of the Wastewater Service Access Pattern*

### 3.2.13.1 Formalization

The pattern is formalized in Table 24.

*Table 24: Formalization of classes in the Wastewater Service Access Pattern*

| Class | Property | Value Restriction |
|---|---|---|
| WastewaterService | rdfs:subClassOf | Utility |
| | 5087-1:hasCapacity | only WastewaterServiceCapacity |
| | 5087-1:capacityInUse | only WastewaterServiceCapacity |
| | 5087-1:hasAvailableCapacity | only WastewaterServiceCapacity |
| | hasPlannedCapacity | only WaterDistributionCapacity |
| WastewaterServiceCapacity | rdfs:subClassOf | i72:Quantity or QuantityConstraint |
| WaterProcessingRate | rdfs:subClassOf | WastewaterServiceCapacity |
| | rdfs:subClassOf | RatioQuantity |
| | forLocation | only 5087-1:Location |
| | i72:numerator | exactly 1 WaterProcessedVolume |
| | i72:denominator | 5087-1:Duration |
| MaxWaterProcessingRate | rdfs:subClassOf | WastewaterServiceCapacity |
| | rdfs:subClassOf | RatioQuantity |
| | forLocation | only 5087-1:Location |
| | i72:numerator | exactly 1 5087-1:Volume |
| | i72:denominator | 5087-1:Duration |
| AvailableWaterProcessingRate | rdfs:subClassOf | WastewaterServiceCapacity |
| | rdfs:subClassOf | DifferenceQuantity |
| | i72:term_1 | exactly 1 MaxWaterProcessingRate |
| | i72:term_2 | exactly 1 WaterProcessingRate |
| | forLocation | only 5087-1:Location |
| WaterProcessedVolume | rdfs:subClassOf | 5087-1:Volume |

### 3.2.14 Transportation Network

Access to transportation often refers to the road (vehicle) network, however it should also consider other modes of transportation and the infrastructure required. The transportation network pattern represents transportation services as being provided from these various types of networks (road network, bike path, etc.). This pattern, depicted in Figure 17, reuses and extends the general Service Accessibility pattern to capture accessibility to the transportation network. It defines the following classes:

- TransportationNetworkService: refers to the service of enabling the movement of people and goods. It has the following properties:
  - providedFromSite: identifies the TransportationSite(s) that a service is provided from.
  - 5087-1:hasCapacity: identifies a TransportationCapacity that specifies the total volume (e.g. of persons, vehicles) that can be moved through the service.
  - 5087-1:hasAvailableCapacity: identifies a TransportationCapacity that specifies the additional available volume (e.g. of persons, vehicles) that can be moved through the service.
  - 5087-1:capacityInUse: identifies a TransportationCapacity that specifies the volume (e.g. of persons, vehicles) that is currently moved through the service.
  - hasPlannedCapacity: identifies a planned TransportationCapacity that specifies the total volume (e.g. of persons, vehicles) that could be moved through the service.
- TransportationSite: defined as a subclass of 5087-2:TravelledWaySegment or 5087-2:TravelledWayLink or 5087-2:TravelledWay. As defined in ISO/IEC 5087-2, TravelledWay, TravelledWayLink, and TravelledWaySegment classes define the transportation network at varying levels of granularity (from highest to lowest, respectively). The classes refer to the physical infrastructure where the transportation takes place. This includes road networks but also footpaths and bicycle paths. TransportationSite is a subclass of 5087-2:TravelledWaySegment (the superclass that includes 5087-2:RoadSegments). Note that 5087-2:TravelledWaySegment has a property 5087-2:networkType that allows for the specification of the mode of travel permitted on the segment.
- TransportationCapacity: a generalized representation of the capacity of the service (current/total/available). It is a subclass of i72:Quantity or QuantityConstraint. Note that TransportationCapacity often directly correlates with the capacities of a TransportationSite (e.g., a road segment) as these elements are the key resources that enable transportation on the network.
- VehicleThroughputRate: a kind of TransportationCapacity. It is a Ratio Quantity of the number of vehicles that pass through the network to the duration of time that the travel takes place. It is a possible measure of the in use capacity of a TransportationNetworkService. (consider: alternatives could include, e.g. bicycle or pedestrian)
- AvailableVehicleThroughputRate: a kind of TransportationCapacity. It is a DifferenceQuantity that specifies the difference between the VehicleThroughputRate and the MaxVehicleThroughputRate as an indication of the available capacity of the service.
- MaxVehicleThroughputRate: a kind of TransportationCapacity. It is a Ratio Quantity that represents the maximum potential VehicleThroughput for a TransportationNetworkService.

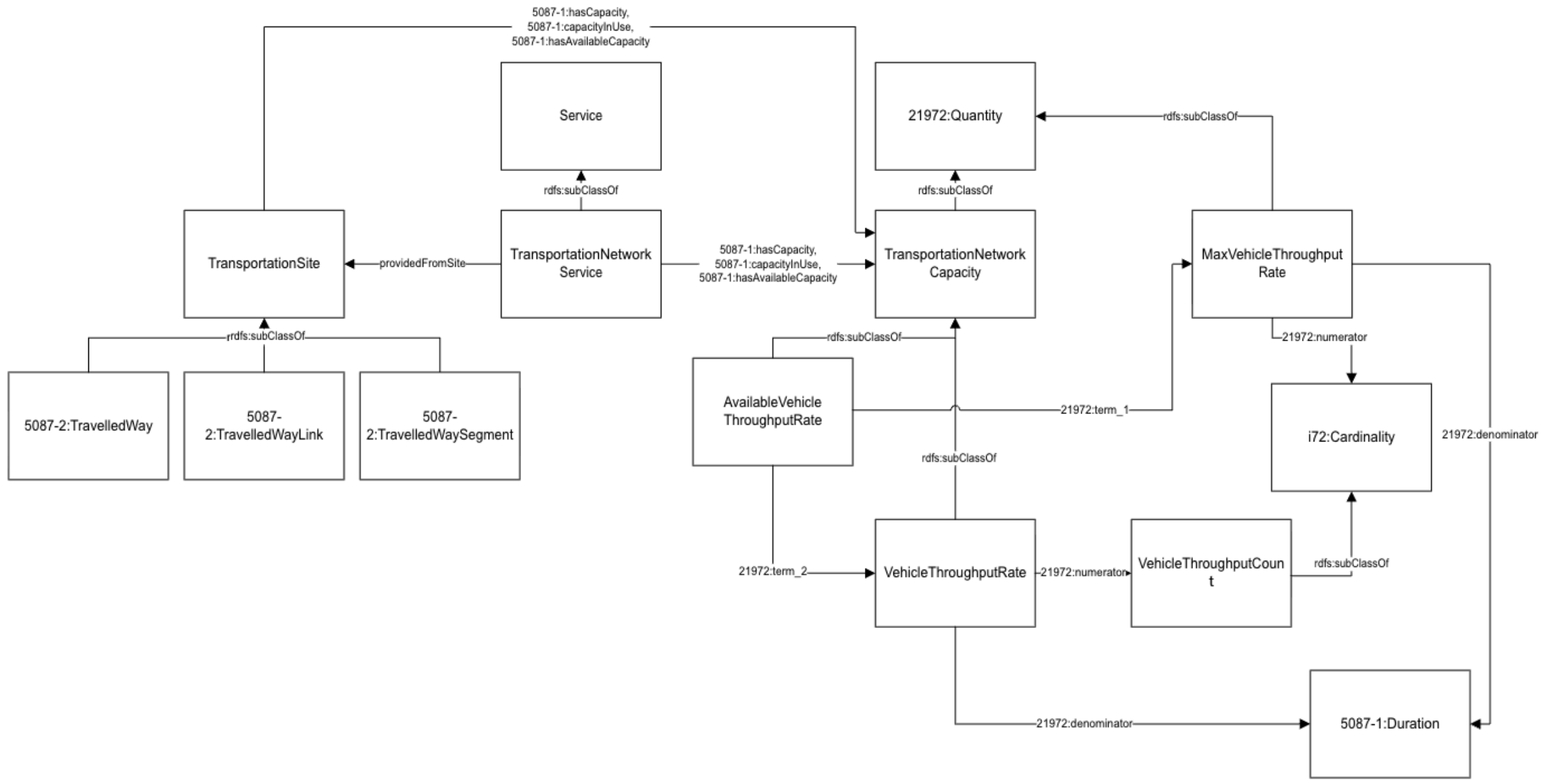

*Figure 17: Overview of the Transportation Network Service Access Pattern*

### *3.2.14.1 Formalization*

The pattern is formalized in Table 25.

*Table 25: Formalization of classes in the Transportation Network Service Access Pattern*

| Class | Property | Value Restriction |
|---|---|---|
| TransportationNetworkService | rdfs:subClassOf | Service |
| | providedFromSite | only TransportationSite |
| | 5087-1:hasCapacity | only TransportationCapacity |
| | 5087-1:capacityInUse | only TransportationCapacity |
| | 5087-1:hasAvailableCapacity | only TransportationCapacity |
| | hasPlannedCapacity | only TransportationCapacity |
| TransportationSite | rdfs:subClassOf | 5087-2:TravelledWaySegment or 5087-2:TravelledWayLink or 5087-2:TravelledWay |
| TransportationCapacity | rdfs:subClassOf | i72:Quantity or QuantityConstraint |
| VehicleThroughputRate | rdfs:subClassOf | TransportationCapacity |
| | rdfs:subClassOf | RatioQuantity |
| | forLocation | only 5087-1:Location |
| | i72:numerator | exactly 1 VehicleThroughputCount |
| | i72:denominator | exactly 1 5087-1:Duration |
| MaxVehicleThroughputRate | rdfs:subClassOf | TransportationCapacity |
| | rdfs:subClassOf | RatioQuantity |
| | forLocation | only 5087-1:Location |
| | i72:numerator | exactly 1 i72:Cardinality |
| | i72:denominator | exactly 1 5087-1:Duration |
| AvailableVehicleThroughputRate | rdfs:subClassOf | TransportationCapacity |
| | rdfs:subClassOf | DifferenceQuantity |
| | i72:term_1 | exactly 1 MaxVehicleThroughputRate |
| | i72:term_2 | exactly 1 VehicleThroughputRate |
| | forLocation | only 5087-1:Location |
| VehicleThroughputCount | rdfs:subClassOf | i72:Cardinality |

### 3.2.15 Transit

A public transit service can be considered at the various levels of detail. Access to services is generally assessed based on proximity to transit stops. This pattern, illustrated in Figure 18, reuses and extends the general Service Accessibility pattern to capture accessibility to public transit services. It defines the following classes:

- PublicTransitService: represents a service that enables the movement of people through publicly accessible routes and vehicles. It may be defined at the city-, region-, or route-level. It has the following properties:
    - providedFromSite: identifies the cdt:TransitStop(s) that the transit service can be accessed from.
    - 5087-1: hasCapacity: identifies a TransitCapacity that indicates the total capacity of the service.
    - 5087-1:hasAvailableCapacity: identifies a TransitCapacity that indicates the available capacity of the service (i.e., the difference between the total capacity and the capacity in use).
    - 5087-1:capacityInUse: identifies a TransitCapacity that indicates the capacity of the service that is currently in use.
    - hasPlannedCapacity: identifies a planned TransitCapacity that indicates a total planned capacity of the service.

- TransitCapacity: a measure of how much can be transported by the public transit service.
- PassengerThroughputRate: A Quantity that indicates the volume of passengers moved by the transit system in a particular time period.
- MinPassengerThroughputRate: a QuantityConstraint that indicates the minimum acceptable rate of passengers moved by the service per unit of time.
- AvailablePassengerThroughputRate: A DifferenceQuantity that specifies the difference between the PassengerThroughputRate and the MinPassengerThroughputRate as an indication of the available capacity of the service.
- TotalPassengerThroughputCount: a measure of the number of passengers moved by the service.

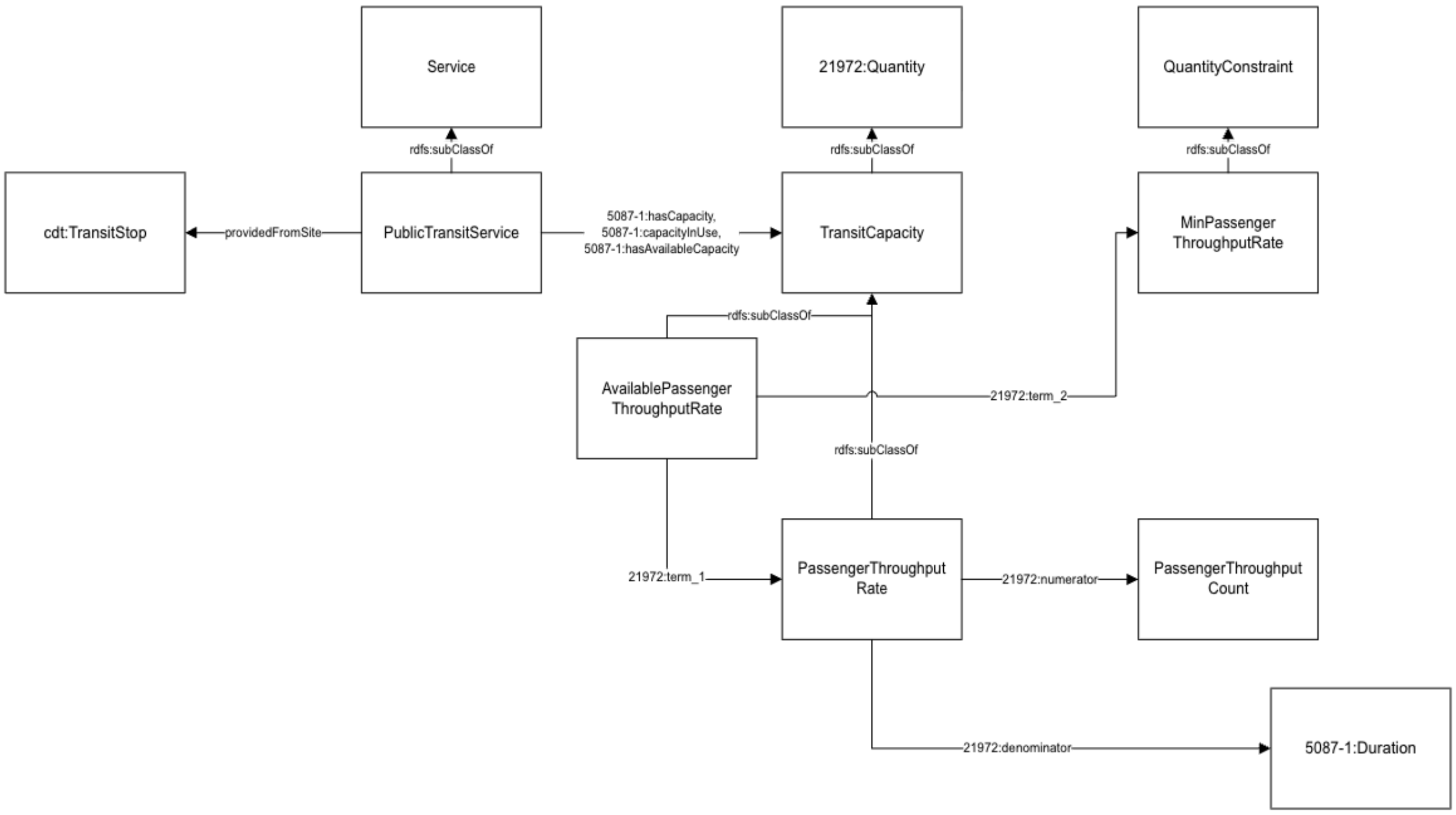

*Figure 18: Overview of the Transit Service Access Pattern*

### 3.2.15.1 Formalization

The pattern is formalized in Table 26.

*Table 26: Formalization of classes in the Transit Service Access Pattern*

| Class | Property | Value Restriction |
|---|---|---|
| PublicTransitService | rdfs:subClassOf | Service |
| | providedFromSite | only cdt:TransitStop |
| | 5087-1:hasCapacity | only TransitCapacity |
| | 5087-1:capacityInUse | only TransitCapacity |
| | 5087-1:hasAvailableCapacity | only TransitCapacity |
| | hasPlannedCapacity | only TransitCapacity |
| TransitCapacity | rdfs:subClassOf | i72:Quantity or QuantityConstraint |
| PassengerThroughputRate | rdfs:subClassOf | TransitCapacity |
| | rdfs:subClassOf | RatioQuantity |
| | forLocation | only 5087-1:Location |
| | i72:numerator | exactly 1 PassengerThroughputCount |
| | i72:denominator | exactly 1 5087-1:Duration |
| MinPassengerThroughputRate | rdfs:subClassOf | TransitCapacity |
| | rdfs:subClassOf | QuantityConstraint |
| | forLocation | only 5087-1:Location |
| AvailablePassengerThroughputRate | rdfs:subClassOf | TransitCapacity |
| | rdfs:subClassOf | DifferenceQuantity |
| | i72:term_1 | exactly 1 PassengerThroughputRate |
| | i72:term_2 | exactly 1 MinPassengerThroughputRate |
| | forLocation | only 5087-1:Location |
| PassengerThroughputCount | rdfs:subClassOf | i72:Cardinality |

### 3.2.16 Childcare

This pattern, illustrated in Figure 19, reuses and extends the general Service Accessibility pattern to capture accessibility to childcare services. It defines the following classes:

- ChildcareService: refers to the professional supervision and caretaking of children. Childcare services are considered separately from school services. They have the following properties:
  - providedFromSite: identifies the ChildcareSite that a childcare service is provided from.
  - 5087-1:hasCapacity: identifies the capacity of the childcare service. This is specified as a ChildcareServiceCapacity.
  - 5087-1:hasAvailableCapacity: identifies the available capacity of the childcare service. This is specified as a ChildcareServiceCapacity.
  - 5087-1:capacityInUse: identifies the childcare service's capacity in use. This is specified as a ChildcareServiceCapacity.
  - hasPlannedCapacity: identifies a planned capacity of the childcare service. This is specified as a ChildcareServiceCapacity.

- ChildcareSites: identifies the physical premise where the childcare service is provided.
- ChildcareServiceCapacity: an i72:Quantity or QuantityConstraint that captures a measure of the capacity of a childcare service.
- ChildcareEnrollmentSize: a specialization of ChildcareServiceCapacity and i72:Quantity. It provides a count of the current population enrolled at the childcare service. It may be used as a measure of the childcare service's capacity in use.
- ChildcareEnrollmentPopulation: a i72:Population of persons (children) enrolled in the childcare service.
- ChildcareEnrollmentSpaces: a specialization of ChildcareServiceCapacity. It is an i72:Quantity that specifies a count of the current number of spaces (i.e., spots for enrollment) for the childcare service. It may be used as a measure of the childcare service's total capacity.
- ChildcareEnrollmentSpacePopulation: a i72:Population of enrollment spaces at a childcare service.
- ChildcareAvailableEnrollmentSpaces: a DifferenceQuantity that represents the available spaces for enrollment at a childcare service.

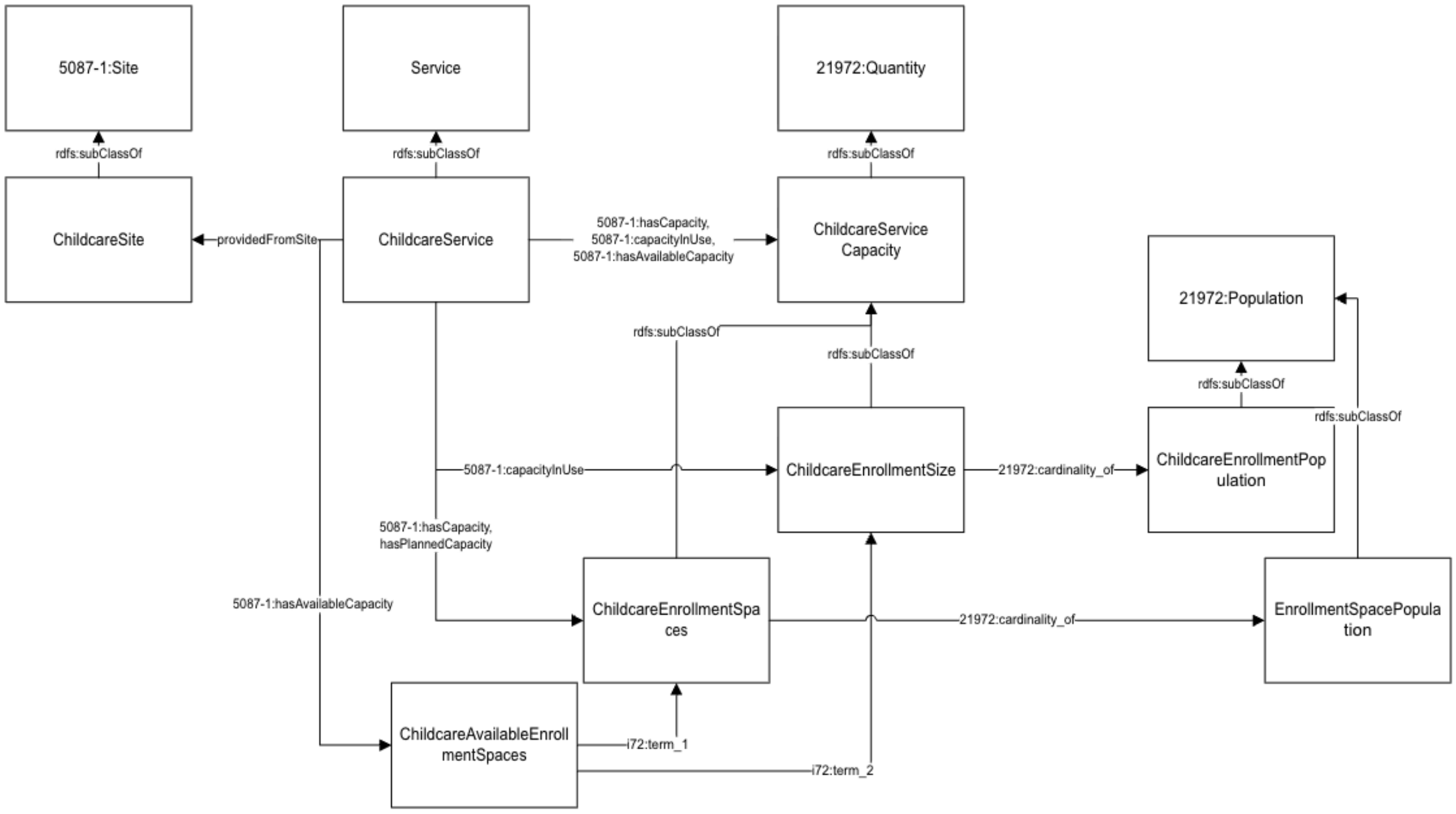

*Figure 19: Overview of Childcare Service Access Pattern*

### *3.2.16.1 Formalization*

The pattern is formalized in Table 27.

*Table 27: Formalization of classes in the Childcare Service Access Pattern*

| Class | Property | Value Restriction |
|---|---|---|
| ChildcareService | rdfs:subClassOf | Service |
| | providedFromSite | only ChildcareSite |
| | 5087-1:hasCapacity | only ChildcareServiceCapacity |
| | 5087-1:capacityInUse | only ChildcareServiceCapacity |
| | 5087-1:hasAvailableCapacity | only ChildcareServiceCapacity |
| | hasPlannedCapacity | only ChildcareServiceCapacity |
| ChildcareServiceCapacity | rdfs:subClassOf | i72:Quantity or QuantityConstraint |
| ChildcareSite | rdfs:subClassOf | 5087-1:Site |
| ChildcareEnrollmentSize | rdfs:subClassOf | ChildcareServiceCapacity |
| | rdfs:subClassOf | i72:Cardinality |
| | i72:cardinality_of | exactly 1 ChildcareEnrollmentPopulation |
| ChildcareEnrollmentSpaces | rdfs:subClassOf | ChildcareServiceCapacity |
| | rdfs:subClassOf | i72:Cardinality |
| | i72:cardinality_of | ChildcareEnrollmentSpacePopulation |
| ChildcareAvailableEnrollmentSpaces | rdfs:subClassOf | ChildcareServiceCapacity |
| | rdfs:subClassOf | DifferenceQuantity |
| | i72:term_1 | exactly 1 ChildcareEnrollmentSpaces |
| | i72:term_2 | exactly 1 ChildcareEnrollmentSize |
| ChildcareEnrollmentPopulation | rdfs:subClassOf | i72:Population |
| ChildcareEnrollmentSpacePopulation | rdfs:subClassOf | i72:Population |

### 3.2.17 Community Centre

This pattern, illustrated in Figure 20, reuses and extends the general Service Accessibility pattern to capture accessibility to community centre services. It defines the following classes:

- CommunityCentreService: represents the service provided by community centres such as access to facilities and various programming. Different community centres may provide different types of these services, but they can all be classified as a general CommunityCentreService (or part of such a general service). CommunityCentreService is a subclass of LowIncomeSupportService. It has the following properties:
  - providedFromSite: identifies a CommunityCentreSite where the service is provided.

- 5087-1:hasCapacity: identifies the capacity of the community centre service. This is specified as a CommunityCentreCapacity.
    - 5087-1:hasAvailableCapacity: identifies the available capacity of the community centre service. This is specified as a CommunityCentreCapacity.
    - 5087-1:capacityInUse: identifies the community centre service's capacity in use. This is specified as a CommunityCentreCapacity.
    - hasPlannedCapacity: identifies a planned capacity of the community centre service. This is specified as a CommunityCentreCapacity.
- CommunityCentreSite: a specialization of 5087-1:Site that represents the community centre (the facility) where the service is provided.
- CommunityCentreServiceCapacity: a specialization of i72:Quantity that captures a measure of the capacity of a community centre service. Client size (number of users of the centre) and client spaces (i.e., the size of population the centre was designed to service) are two possible measures of capacity. Other approximations such as square footage to nearby population could also be captured in extensions.
- CommunityCentreClientSize: an i72:Quantity that captures a measure of the population currently serviced by the community centre (i.e., the population within its catchment area). Client sizes and the service areas by which they are determined are likely to vary based on context (e.g., urban or rural) and size of the centre. It may be used as a measure of a service's capacity in-use.
- CommunityCentreClientSpaces: a QuantityConstraint that captures a measure of the maximum number of residents that a centre was designed to service (i.e., overall as opposed to a single point in time). It may be used as a measure of a service's total capacity.
- CommunityCentreAvailableSpaces: an i72:Quantity that captures a measure of the available spaces (if any) that a centre is capable of supporting. It is defined as the difference between the CommunityCentreClientSpaces and the CommunityCentreClientSize.

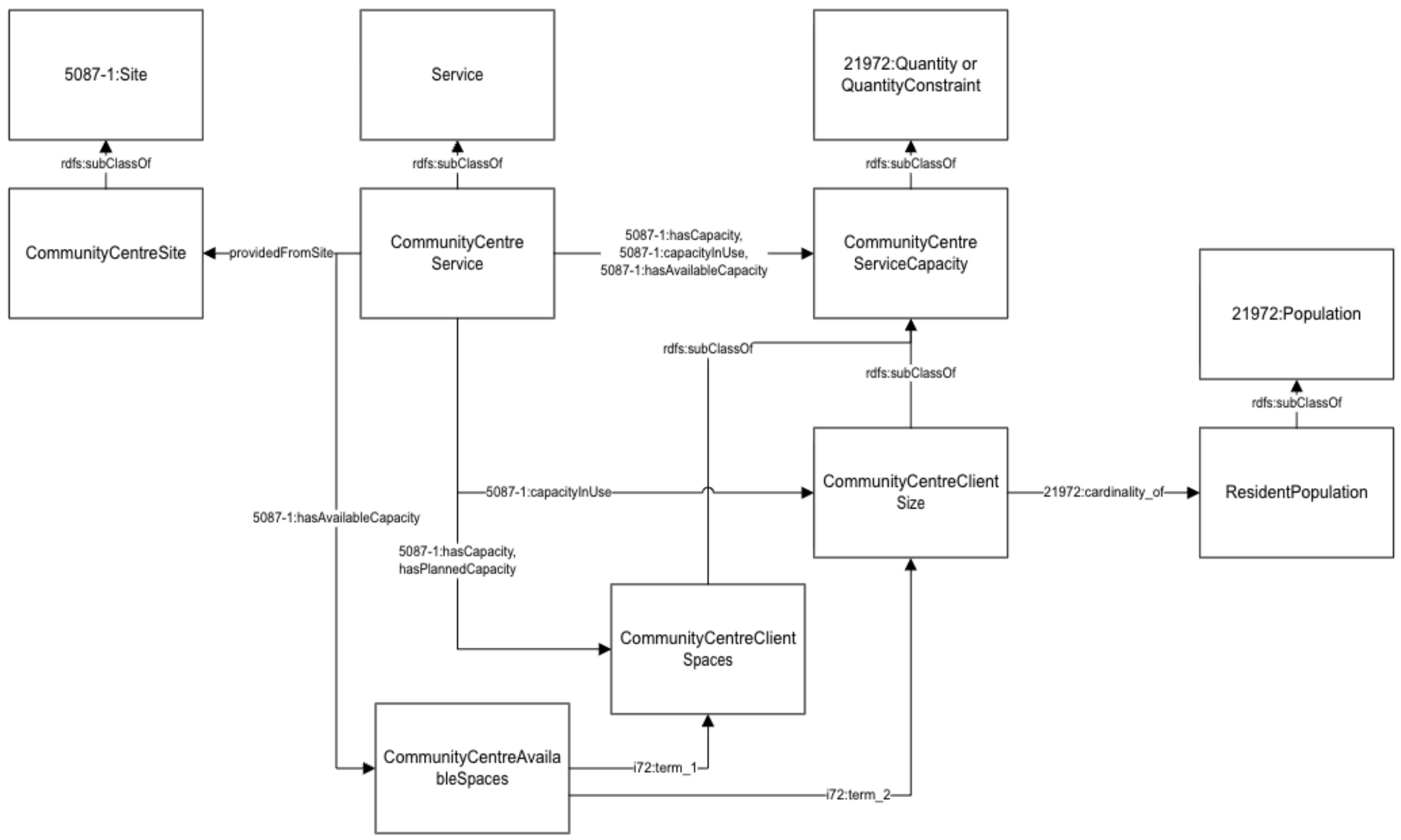

*Figure 20: Overview of the Community Centre Service Access Pattern*

### *3.2.17.1 Formalization*

The pattern is formalized in Table 28.

*Table 28: Formalization of classes in the Community Centre Service Access pattern*

| Class | Property | Value Restriction |
|---|---|---|
| CommunityCentreService | rdfs:subClassOf | Service |
| | providedFromSite | only CommunityCentreSite |
| | 5087-1:hasCapacity | only CommunityCentreCapacity |
| | 5087-1:capacityInUse | only CommunityCentreCapacity |
| | 5087-1:hasAvailableCapacity | only CommunityCentreCapacity |
| | hasPlannedCapacity | only CommunityCentreCapacity |
| CommunityCentreCapacity | rdfs:subClassOf | i72:Quantity or QuantityConstraint |
| CommunityCentreSite | rdfs:subClassOf | 5087-1:Site |
| CommunityCentreClientSize | rdfs:subClassOf | CommunityCentreCapacity |
| | rdfs:subClassOf | i72:Cardinality |
| | i72:cardinality_of | exactly 1 ResidentPopulation |
| | i72:hasUnit | value i72:population_cardinality_unit |
| CommunityCentreClientSpaces | rdfs:subClassOf | CommunityCentreCapacity |
| | rdfs:subClassOf | QuantityConstraint |
| | i72:hasUnit | value i72:population_cardinality_unit |
| CommunityCentreAvailableSpaces | rdfs:subClassOf | CommunityCentreCapacity |
| | rdfs:subClassOf | DifferenceQuantity |
| | i72:term_1 | exactly 1 CommunityCentreClientSpaces |
| | i72:term_2 | exactly 1 CommunityCentreClientSize |
| | i72:hasUnit | value i72:population_cardinality_unit |

### 3.2.18 Library

This pattern, illustrated in Figure 21, reuses and extends the general Service Accessibility pattern to capture accessibility to library services. It defines the following classes:

- LibraryService: represents the service provided by libraries such as access to books and programming. It is a subclass of cdt:LibraryService and LowIncomeSupportService and has the following properties:
  - providedFromSite: identifies the cdt:LibrarySite where the service is provided.
  - 5087-1:hasCapacity: identifies the capacity of the library in terms of its ability to provide a service (e.g., persons served as opposed to the physical capacity of the building). This is specified as a LibraryServiceCapacity.
  - 5087-1:hasAvailableCapacity: identifies the available capacity of the library service. This is specified as a LibraryServiceCapacity.
  - 5087-1:capacityInUse: identifies the library service's capacity in use. This is specified as a LibraryServiceCapacity.
  - hasPlannedCapacity: identifies a planned capacity of the library service. This is specified as a LibraryServiceCapacity.

- LibraryServiceCapacity: a type of i72:Quantity or QuantityConstraint that captures a measure of the capacity of a library service. ClientSize and ClientSpaces are two quantities that could be used to describe a library service's capacity. Alternative capacity measures that could be specified in an extension to the HPCDM include branch visits or items checked out.
- LibraryAreaPopulationRatio: a type of LibraryServiceCapacity that is the ratio of the total library floor area to the population it is intended to service (in its catchment area). It is an i72:Quantity with the following properties:
    - i72:numerator: identifies the LibraryAreaSum quantity to be used in the ratio. Specifically, it is the total floor area of the LibrarySites within the service's catchment area is provided.
    - i72:denominator: identifies the ResidentPopulationSize to be used in the ratio. Specifically, it is the resident population of the catchment area for the library.
- MinLibraryAreaPopulationRatio: a type of QuantityConstraint that specifies the minimum acceptable ratio of library area to number of residents (i.e., library area per capita) that can serve as a kind of measure for the capacity of an area. In other words, any value below that and the area will not be able to provide satisfactory access to library services.
- AvailableLibraryPopulationRatio: a QuantityConstraint that represents the maximum possible decrease in LibraryAreaPopulationRatio while maintaining an acceptable capacity use. It is calculated based on the difference between the LibraryAreaPopulationRatio and the MinLibraryAreaPopulationRatio.
- LibraryPopulation: an i72:Population that represents the libraries within a particular area.
- LibraryAreaSum: an i72:Sum that represents the total library floor area (from all libraries) in a particular area.

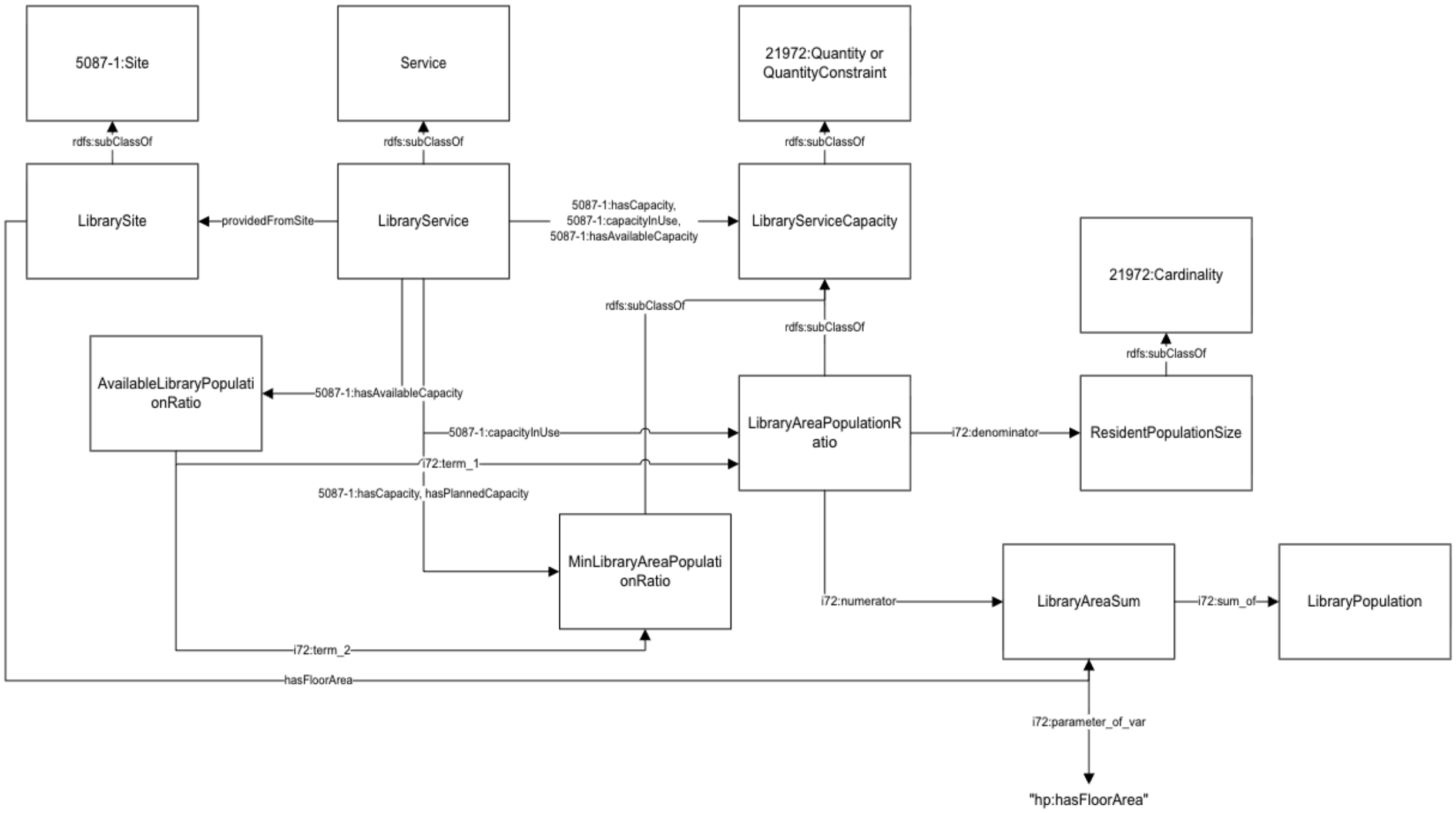

*Figure 21: Overview of the Library Service Access pattern*

#### *3.2.18.1 Formalization*

The pattern is formalized in Table 29.

*Table 29: Formalization of classes in the Library Service Access pattern*

| Class | Property | Value Restriction |
|---|---|---|
| LibraryService | rdfs:subClassOf | LowIncomeSupportService |
| | rdfs:subClassOf | cdt:LibraryService |
| | providedFromSite | only cdt:LibrarySite |
| | 5087-1:hasCapacity | only LibraryServiceCapacity |
| | 5087-1:capacityInUse | only LibraryServiceCapacity |
| | 5087-1:hasAvailableCapacity | only LibraryServiceCapacity |
| | hasPlannedCapacity | only LibraryServiceCapacity |
| LibraryServiceCapacity | rdfs:subClassOf | i72:Quantity or QuantityConstraint |
| LibraryAreaPopulationRatio | rdfs:subClassOf | LibraryServiceCapacity |
| | rdfs:subClassOf | RatioQuantity |
| | i72:numerator | exactly 1 LibraryAreaSum |
| | i72:denominator | exactly 1 ResidentPopulationSize |
| | i72:hasUnit | value square_metre_per_person |
| MinLibraryAreaPopulationRatio | rdfs:subClassOf | LibraryServiceCapacity |
| | rdfs:subClassOf | QuantityConstraint |
| | i72:hasUnit | value square_metre_per_person |
| AvailableLibraryPopulationRatio | rdfs:subClassOf | LibraryServiceCapacity |
| | rdfs:subClassOf | QuantityConstraint |
| | i72:term_1 | exactly 1 LibraryAreaPopulationRatio |
| | i72:term_2 | exactly 1 MinLibraryAreaPopulationRatio |
| | i72:hasUnit | value square_metre_per_person |
| LibraryAreaSum | rdfs:subClassOf | i72:Sum |
| | i72:sum_of | exactly 1 LibraryPopulation |
| | i72:parameter_of_var | only (i72:Variable and i72:has_Name value "hp:hasFloorArea") |
| LibraryPopulation | rdfs:subClassOf | i72:Population |
| | i72:defined_by | only cdt:LibrarySite |

### 3.2.19 School

This pattern, illustrated in Figure 22, reuses and extends the general Service Accessibility pattern to capture accessibility to school services. It defines the following classes:

- SchoolService: represents the service provided by schools, particularly the delivery of various education programs.
  - providedFromSite: identifies the SchoolSite from which the school (education) service is provided.
  - 5087-2:hasCatchmentArea: identifies the catchment area for the SchoolService.
  - 5087-1:hasCapacity: identifies the capacity of the school in terms of its ability to provide the service (e.g., students educated served as opposed to the physical capacity of the building). This is specified as a SchoolServiceCapacity.
  - 5087-1:hasAvailableCapacity: identifies the available capacity of the school service. This is specified as a SchoolServiceCapacity.
  - 5087-1:capacityInUse: identifies the school service's capacity in use. This is specified as a SchoolServiceCapacity.
  - hasPlannedCapacity: identifies a planned capacity of the school service. This is specified as a SchoolServiceCapacity.
- SchoolService has the following subclasses that distinguish between the service provided by different school levels. There are no distinctions in the definition of these classes for the scope of the HPCDM, however it is important to make the distinction between the two classes of service to accurately reflect their capacities and demand. There are also distinctions in attributes such as the way the services are evaluated and their beneficial stakeholders that are relevant for other applications.:
  - ElementarySchoolService
  - SecondarySchoolService
- SchoolSite: a Site (i.e., the facility) where a SchoolService is provided.
- SchoolServiceCapacity: a type of i72:Quantity or QuantityConstraint that captures a measure of the capacity of a school service.
- SchoolEnrollmentSize: is a SchoolServiceCapacity and an i72:Quantity that represents the number of students enrolled in a school. It may be used as a measure of the school service capacity in use.
- SchoolEnrollmentSpaces: is a SchoolServiceCapacity that represents the total number of enrollment spaces at a school. It can be used as a measure of the total school service capacity.
- SchoolAvailableEnrollmentSpaces: a DifferenceQuantity that represents the available spaces for enrollment at a school. It is defined as the difference between SchoolEnrollmentSpaces and SchoolEnrollmentSize.

- SchoolEnrollmentPopulation: represents the population of students enrolled at a particular school service (individual school, school grade(s)s, or set of schools depending on the service definition).
- SchoolEnrollmentSpacePopulation: represents the total enrollment spaces for a particular school service.

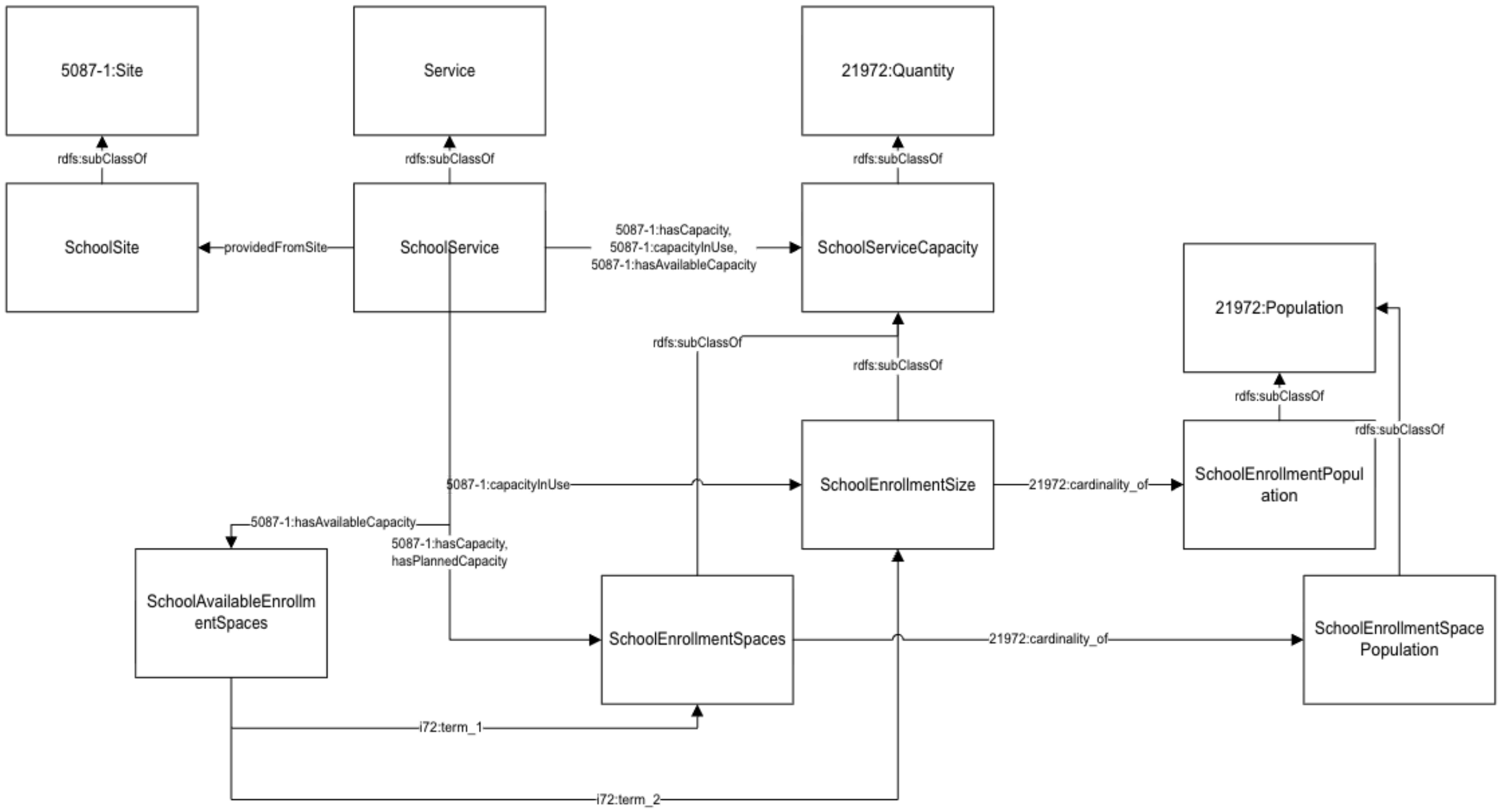

*Figure 22: Overview of the School Service Access pattern*

### *3.2.19.1 Formalization*

The pattern is formalized in Table 30.

*Table 30: Formalization of classes in the School Service Access pattern*

| Class | Property | Value Restriction |
|---|---|---|
| SchoolService | rdfs:subClassOf | Service |
| | providedFromSite | only SchoolSite |
| | 5087-1:hasCapacity | only SchoolServiceCapacity |
| | 5087-1:capacityInUse | only SchoolServiceCapacity |
| | 5087-1:hasAvailableCapacity | only SchoolServiceCapacity |
| | hasPlannedCapacity | only SchoolServiceCapacity |
| SchoolSite | rdfs:subClassOf | 5087-1:Site |
| ElementarySchoolService | rdfs:subClassOf | SchoolService |
| SecondarySchoolService | rdfs:subClassOf | SchoolService |
| SchoolServiceCapacity | rdfs:subClassOf | i72:Quantity or QuantityConstraint |
| SchoolEnrollmentSize | rdfs:subClassOf | SchoolServiceCapacity |
| | rdfs:subClassOf | i72:Cardinality |
| | i72:cardinality_of | exactly 1 SchoolEnrollmentPopulation |
| SchoolEnrollmentSpaces | rdfs:subClassOf | SchoolServiceCapacity |
| | rdfs:subClassOf | i72:Cardinality |
| | i72:cardinality_of | exactly 1 SchoolEnrollmentSpacePopulation |
| SchoolAvailableEnrollmentSpaces | rdfs:subClassOf | SchoolServiceCapacity |
| | rdfs:subClassOf | DifferenceQuantity |
| | i72:term_1 | exactly 1 SchoolEnrollmentSpaces |
| | i72:term_2 | exactly 1 SchoolEnrollmentSize |
| SchoolEnrollmentPopulation | rdfs:subClassOf | i72:Population |
| SchoolEnrollmentSpacePopulation | rdfs:subClassOf | i72:Population |

### 3.2.20 Parks

This pattern, illustrated in Figure 23, reuses and extends the general Service Accessibility pattern to capture accessibility to park services. It defines the following classes:

- ParkService: represents the service provided by parks such as access to recreation facilities and greenspace.
  - providedFromSite: identifies the gcir:Park from which the park service (e.g., facilities, greenspace) is provided.
  - hasCatchmentArea: identifies the intended catchment area for the park service.
  - 5087-1:hasCapacity: identifies the capacity of the park in terms of its ability to provide the service (e.g., persons who frequent the park, not necessarily the number of persons who can be physically be accommodated at once). This is specified as a ParkServicesCapacity.
  - 5087-1:hasAvailableCapacity: identifies the available capacity of the park service. This is specified as a ParkServicesCapacity.
  - 5087-1:capacityInUse: identifies the park service's capacity in use. This is specified as a ParkServicesCapacity.
  - hasPlannedCapacity: identifies a planned capacity of the park service. This is specified as a ParkServicesCapacity.

- ParkServiceCapacity : a specialization of i72:Quantity that captures a measure of the capacity of a park service. The isor:13.1 indicator is one quantity that may be used as a measure of capacity at the city level. It can also be used to inform the definition of a comparable property at the individual park level.
- RecreationAreaPopulationRatio: defined similarly to the park area per capita indicator (isor:13.1) in the Global City Indicator Recreation Ontology [41] represents a measure of total park area to the population size in a region. It has the following properties:
    - i72:denominator: identifies the ResidentPopulationSize used for the denominator of the ratio.
    - i72:numerator: identifies the RecreationAreaSum used for the numerator of the ratio.
    - forLocation: identifies the 5087-1:Location that the quantity is calculated for.
    - i72:hasUnit: the ratio is measured in hp:square_metres_per_person
- MinRecreationAreaPopulationRatio: a Quantity Constraint that represents the minimum acceptable ratio of recreation area to resident population in a particular area.
- AvailableRecreationAreaPopulationRatio: a QuantityConstraint that represents the maximum possible change to the RecreationAreaPopulationRatio before violating the defined capacity. A difference between RecreationAreaPopulationRatio and MinRecreationAreaPopulationRatio.
- RecreationAreaSum: a i72:Sum quantity that represents a sum of the recreation areas (defined with the gcir:hasFloorArea property), in the specified 5087-1:Location.

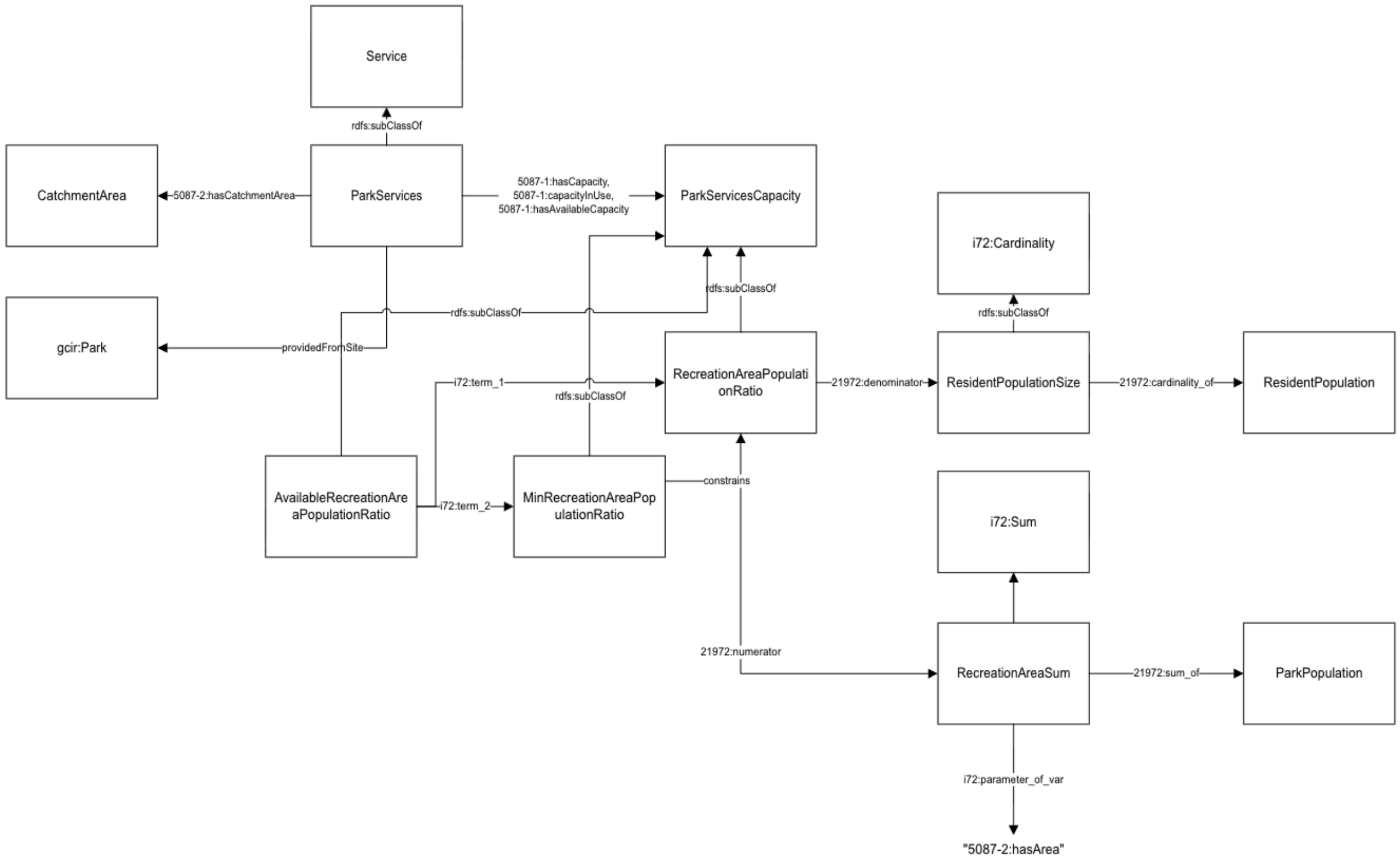

*Figure 23: Overview of the Park Service Access pattern*

### 3.2.20.1 Formalization

The pattern is formalized in Table 31.

*Table 31: Formalization of classes in the Park Service Access pattern.*

| Class | Property | Value Restriction |
|---|---|---|
| ParkService | rdfs:subClassOf | Service |
| | providedFromSite | only gcir:Park |
| | 5087-1:hasCapacity | only ParkServicesCapacity |
| | 5087-1:capacityInUse | only ParkServicesCapacity |
| | 5087-1:hasAvailableCapacity | only ParkServicesCapacity |
| | hasPlannedCapacity | only ParkServicesCapacity |
| ParkServiceCapacity | rdfs:subClassOf | i72:Quantity or QuantityConstraint |
| RecreationAreaPopulationRatio | rdfs:subClassOf | ParkServiceCapacity |
| | rdfs:subClassOf | RatioQuantity |
| | i72:numerator | exactly 1 RecreationAreaSum |
| | i72:denominator | exactly 1 ResidentPopulationSize |
| | forLocation | only Location |
| | i72:hasUnit | value square_metre_per_person |
| MinRecreationAreaPopulationRatio | rdfs:subClassOf | ParkServiceCapacity |
| | rdfs:subClassOf | QuantityConstraint |
| | i72:hasUnit | value square_metre_per_person |
| AvailableRecreationAreaPopulationRatio | rdfs:subClassOf | ParkServiceCapacity |
| | rdfs:subClassOf | QuantityConstraint |
| | i72:term_1 | exactly 1 RecreationAreaPopulationRatio |
| | i72:term_2 | exactly 1 MinRecreationAreaPopulationRatio |
| | i72:hasUnit | value square_metre_per_person |
| RecreationAreaSum | rdfs:subClassOf | i72:Sum |
| | i72:sum_of | exactly 1 ParkPopulation |
| | i72:parameter_of_var | only (i72:Variable and i72:has_Name value "http://ontology.eil.utoronto.ca/HPCDM/hasArea") |
| ParkPopulation | rdfs:subClassOf | i72:Population |
| | i72:defined_by | only gcir:Park |

### 3.2.21 Medical

The Medical Service Access pattern, illustrated in Figure 24 and Figure 25, is currently defined for hospitals but may be extended to capture pharmacies and clinics in the future. This pattern reuses and extends the general Service Accessibility pattern to capture accessibility to hospital services. It defines the following classes:

- HospitalService: represents the service provided by hospitals, such as access to specialized treatments and emergency healthcare. It has the following properties:
  - providedFromSite: identifies the HospitalSite from which the service is provided.
  - hasCatchmentArea: identifies the intended catchment area for the hospital service.
  - 5087-1:hasCapacity: identifies the capacity of the hospital in terms of its ability to provide services. This is specified as a HospitalServiceCapacity.
  - 5087-1:hasAvailableCapacity: identifies the available capacity of the hospital service. This is specified as a HospitalServiceCapacity.
  - 5087-1:capacityInUse: identifies the hospital service's capacity in use. This is specified as a HospitalServiceCapacity.
  - hasPlannedCapacity: identifies a planned capacity of the hospital service. This is specified as a HospitalServiceCapacity.
- HospitalEmergencyService: a type of Service that is offered by some hospitals (often referred to as "Emergency" or "Urgent Care"). It is distinguished from the larger HospitalService as its individual capacity is often of interest as it represents a critical capacity issue, with longer wait times having serious and immediate consequences. It has the following properties:
  - inverse(hasSubService): identifies the HospitalService that the HospitalEmergencyService is a part of.
  - 5087-1:hasCapacity: identifies the capacity of the hospital emergency service in terms of its ability to provide services. This is specified as a HospitalServiceCapacity.
  - 5087-1:hasAvailableCapacity: identifies the available capacity of the hospital emergency service. This is specified as a EmergencyCareServiceCapacity.
  - 5087-1:capacityInUse: identifies the hospital service's capacity in use. This is specified as a EmergencyCareServiceCapacity.
  - hasPlannedCapacity: identifies a planned capacity of the hospital emergency service. This is specified as a EmergencyCareServiceCapacity.
- HospitalServiceCapacity: a specialization of i72:Quantity that captures a measure of the capacity of a hospital service.
- EmergencyCareServiceCapacity: a specialization of i72:Quantity that captures a measure of the capacity of an emergency/urgent care service.

- HospitalBedPopulationRatio: a type of HospitalServiceCapacity that is defined as the ratio of the number of hospital beds to the population of the catchment area. It has the following properties:
  - i72:denominator: identifies the ResidentPopulationSize to be used in the ratio.
  - i72:numerator: identifies the HospitalBedCount to be used in the ratio.
  - forLocation: identifies the 5087-1:Location (e.g., the catchment area) that the ratio is defined for.
- MinHospitalBedPopulationRatio: a type of QuantityConstraint that specifies the *minimum* acceptable ratio of hospital beds per capita.
- AvailableHospitalBedPopulationRatio: a type of DifferenceConstraint that identifies the available beds per capita (i.e., before reaching the minimum acceptable rate).
- EmergencyBedPopulationRatio: similar to the HospitalBedPopulationRatio, it is a type of EmergencyCareServiceCapacity that is the ratio of the number of emergency care beds to the population of the catchment area. It has the following properties:
  - i72:denominator: identifies the ResidentPopulationSize to be used in the ratio.
  - i72:numerator: identifies the EmergencyBedCount to be used in the ratio.
  - forLocation: identifies the 5087-1:Location (e.g., the catchment area) that the ratio is defined for.
- MinEmergencyBedPopulationRatio: a type of QuantityConstraint that specifies the *minimum* acceptable ratio of emergency care beds per capita. It is a constraint on the EmergencyBedPopulationRatio.
- AvailableEmergencyBedPopulationRatio: a type of QuantityConstraint that identifies the available beds per capita (i.e., before reaching the minimum acceptable rate). It is defined as the difference between the EmergencyBedPopulationRatio and the MinEmergencyBedPopulationRatio.
- HospitalBedCount: is a i72:Cardinality that represents a count of the number of gcih:HospitalBeds in the location of interest.
- EmergencyBedCount: a i72:Cardinality of the number of EmergencyBeds in the location of interest.
- EmergencyBed: a subclass of gcih:HospitalBeds that captures only the beds used for temporary, acute care when patients are admitted to the emergency room.
- HospitalBedPopulation: is an i72:Population quantity that captures the population of hospital beds in a particular area (i.e., at a particular hospital). Its members are defined_by the gcih:HospitalBed class.
- HospitalEmergencyBedPopulation: is an i72:Population quantity that captures the population of hospital beds for emergency care in a particular area (i.e., at a particular hospital). Its members are defined_by the EmergencyHospitalBed class.

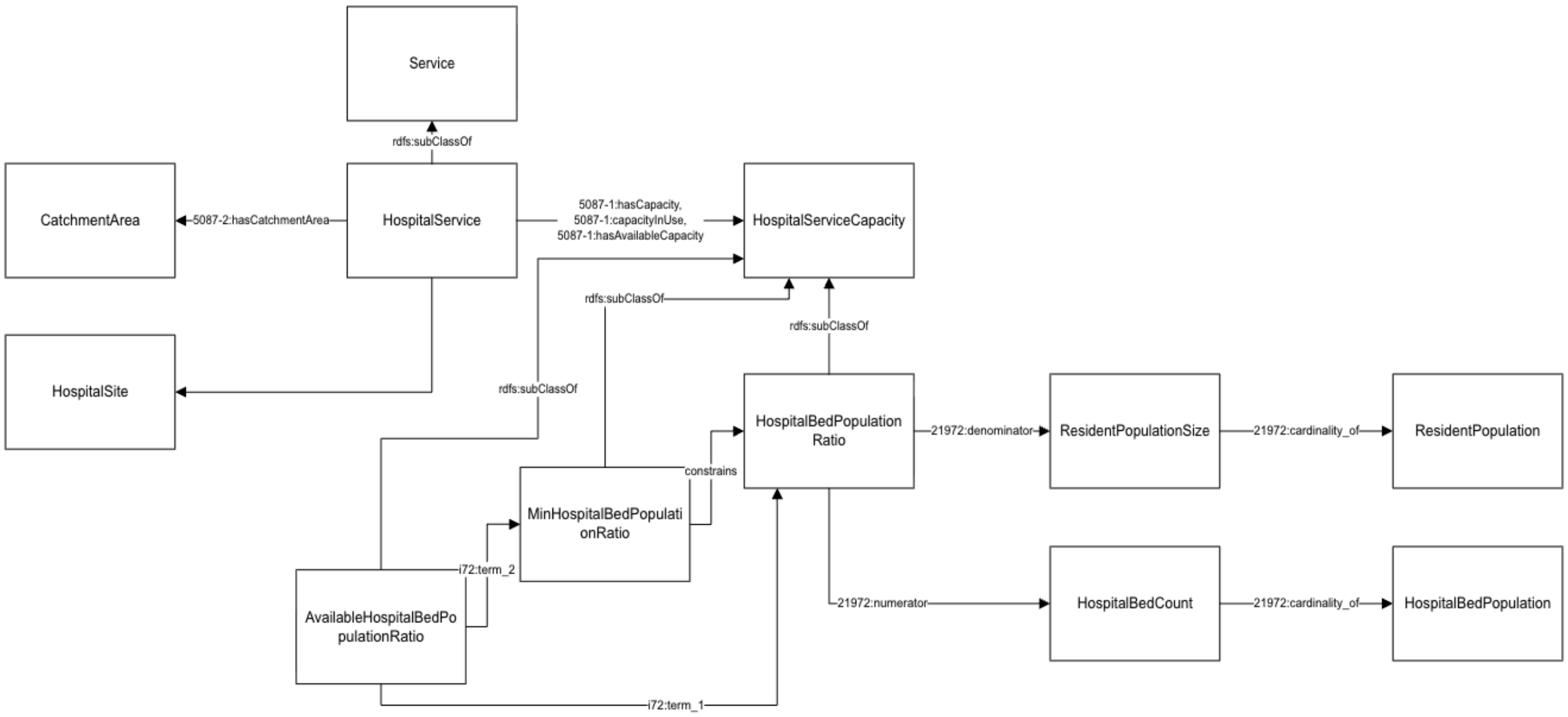

*Figure 24: Overview of the Hospital Service representation in the Medical Service Access pattern*

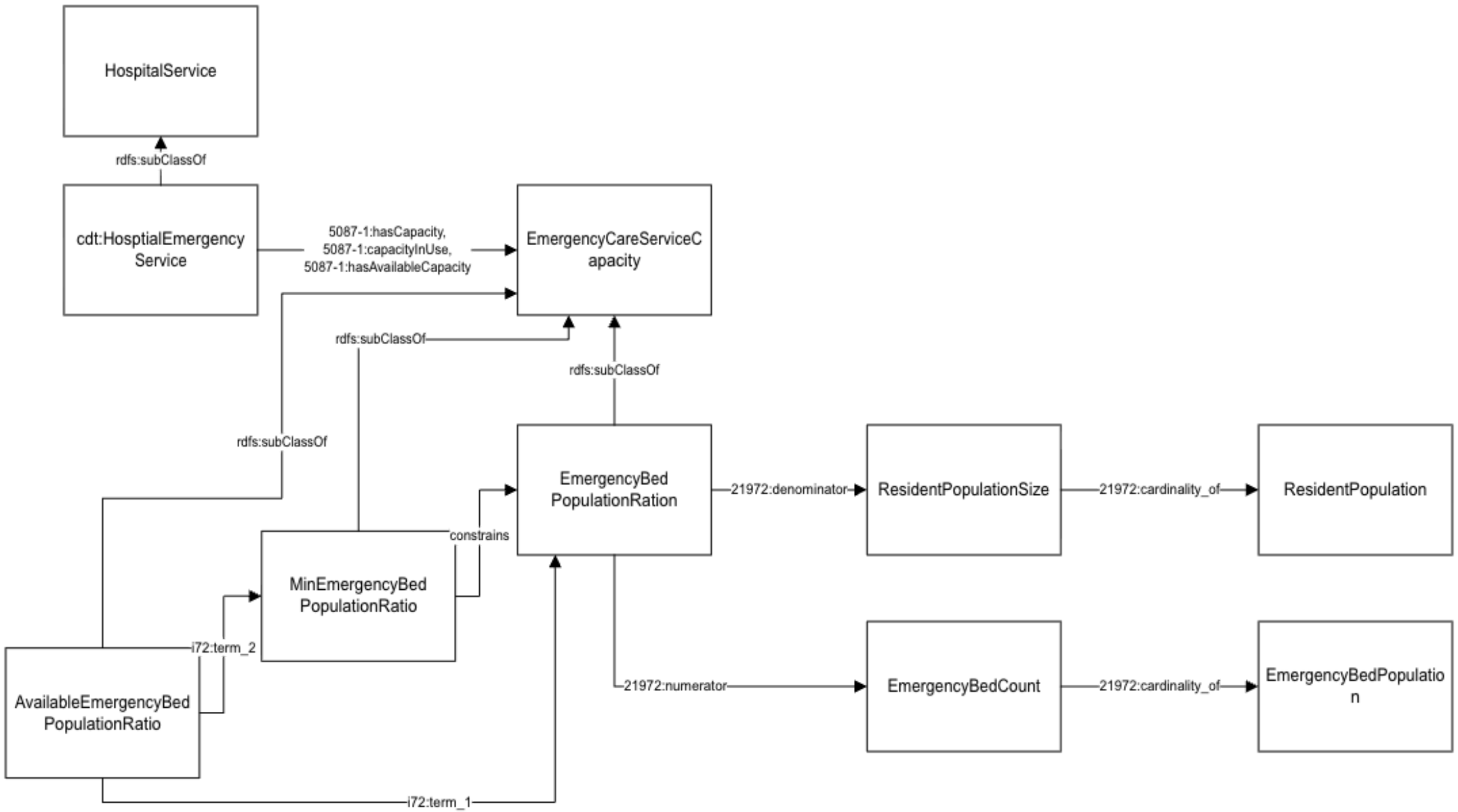

*Figure 25: Overview of the Emergency Service representation in the Medical Service Access pattern*

### 3.2.21.1 Formalization

The pattern is formalized in Table 32.

*Table 32: Formalization of classes in the Medical Service Access pattern*

| Class | Property | Value Restriction |
|---|---|---|
| HospitalService | rdfs:subClassOf | Service |
| | providedFromSite | only HospitalSite |
| | 5087-1:hasCapacity | only HospitalServiceCapacity |
| | 5087-1:capacityInUse | only HospitalServiceCapacity |
| | 5087-1:hasAvailableCapacity | only HospitalServiceCapacity |
| | hasPlannedCapacity | only HospitalServiceCapacity |
| HospitalSite | rdfs:subClassOf | 5087-1:Site |
| HospitalEmergencyService | rdfs:subClassOf | Service |
| | rdfs:subClassOf | cdt:HospitalEmergencyService |
| | inverse(hasSubService) | some HospitalService |
| | 5087-1:hasCapacity | only EmergencyCareServiceCapacity |
| | 5087-1:capacityInUse | only EmergencyCareServiceCapacity |
| | 5087-1:hasAvailableCapacity | only EmergencyCareServiceCapacity |
| | hasPlannedCapacity | only EmergencyCareServiceCapacity |
| HospitalServiceCapacity | rdfs:subClassOf | i72:Quantity or QuantityConstraint |
| EmergencyCareServiceCapacity | rdfs:subClassOf | i72:Quantity or QuantityConstraint |
| HospitalBedPopulationRatio | rdfs:subClassOf | HospitalServiceCapacity |
| | rdfs:subClassOf | RatioQuantity |
| | i72:numerator | exactly 1 HospitalBedCount |
| | i72:denominator | exactly 1 ResidentPopulationSize |
| | forLocation | only Location |
| | i72:hasUnit | value i72:population_ratio_unit |
| MinHospitalBedPopulationRatio | rdfs:subClassOf | HospitalServiceCapacity |
| | rdfs:subClassOf | QuantityConstraint |
| | i72:hasUnit | value i72:population_ratio_unit |
| AvailableHospitalBedPopulationRatio | rdfs:subClassOf | HospitalServiceCapacity |
| | rdfs:subClassOf | QuantityConstraint |
| | i72:term_1 | exactly 1 HospitalBedPopulationRatio |
| | i72:term_2 | exactly 1 MinHospitalBedPopulationRatio |
| | i72:hasUnit | value i72:population_ratio_unit |
| EmergencyBedPopulationRatio | rdfs:subClassOf | EmergencyCareServiceCapacity |
| | rdfs:subClassOf | RatioQuantity |
| | i72:numerator | exactly 1 EmergencyBedCount |
| | i72:denominator | exactly 1 ResidentPopulationSize |
| | forLocation | only Location |
| | i72:hasUnit | value i72:population_ratio_unit |
| MinEmergencyBedPopulationRatio | rdfs:subClassOf | EmergencyCareServiceCapacity |
| | rdfs:subClassOf | QuantityConstraint |
| | i72:hasUnit | value i72:population_ratio_unit |
| AvailableEmergencyBedPopulationRatio | rdfs:subClassOf | EmergencyCareServiceCapacity |
| | rdfs:subClassOf | QuantityConstraint |
| | i72:term_1 | exactly 1 EmergencyBedPopulationRatio |
| | i72:term_2 | exactly 1 MinEmergencyBedPopulationRatio |
| | i72:hasUnit | value i72:population_ratio_unit |
| HospitalBedCount | rdfs:subClassOf | i72:Cardinality |
| | i72:cardinality_of | only HospitalBedPopulation |
| HospitalBedPopulation | rdfs:subClassOf | i72:Population |
| | i72:defined_by | only gcih:HospitalBed |
| EmergencyBedCount | rdfs:subClassOf | i72:Cardinality |
| | i72:cardinality_of | HospitalEmergencyBedPopulation |
| HospitalEmergencyBedPopulation | rdfs:subClassOf | i72:Population |
| | i72:defined_by | only EmergencyBed |
| EmergencyHospitalBed | rdfs:subClassOf | gcih:HospitalBed |

### 3.2.22 Food

The Food Service Access pattern, illustrated in Figure 26, is currently defined for supermarkets but may be extended to capture other services (e.g. restaurants) in the future. This pattern reuses and extends the general Service Accessibility pattern to capture accessibility to supermarket services. It defines the following classes:

- SupermarketService: the sale of goods (food, in particular) to the public from a supermarket. It has the following properties:
  - providedFromSite: identifies the SupermarketSite from which the service is provided.
  - 5087-1:hasCapacity: identifies the capacity of the supermarket in terms of its ability to provide services (i.e., distinct from the physical capacity of the building). This is specified as a SupermarketServiceCapacity.
  - 5087-1:hasAvailableCapacity: identifies the available capacity of the supermarket service. This is specified as a SupermarketServiceCapacity.
  - 5087-1:capacityInUse: identifies the supermarket service's capacity in use. This is specified as a SupermarketServiceCapacity.
  - hasPlannedCapacity: identifies a planned capacity of the supermarket service. This is specified as a SupermarketServiceCapacity.
- SupermarketSite: refers to the physical place (i.e., the supermarket) where the service is provided.
- SupermarketServiceCapacity: a specialization of i72:Quantity that captures a measure of the capacity of a supermarket service.
- SupermarketsPopulationRatio: a subclass of SupermarketServiceCapacity. Describes the capacity of a supermarket as the ratio of the number of supermarkets to the population *in its catchment area*. It has the following properties:
  - i72:denominator: identifies the ResidentPopulationSize to be used in the ratio.
  - i72:numerator: identifies the SupermarketCount to be used in the ratio.
  - forLocation: identifies the 5087-1:Location (e.g., the catchment area) that the ratio is defined for.
- SupermarketCount: is an i72:Cardinality quantity that represents the number of supermarkets (i.e., the size of the supermarket population) in the location of interest. It is a i72:cardinality_of some SupermarketPopulation.

- SupermarketPopulation: is an i72:Population quantity that captures the population of supermarkets in a particular area. It is i72:defined_by the SupermarketSite class.
- MinSupermarketsPopulationRatio: a QuantityConstraint on a particular supermarket population ratio. It represents some recommended minimum value for a ratio of the number of supermarkets to resident population (i.e., supermarkets per capita) that can serve as a kind of measure for the capacity of an area. In other words, any value below that and the area will not be able to provide satisfactory access to supermarket (grocery) services.
- AvailableSupermarketsPopulationRatio: a QuantityConstraint that is calculated based on the difference between the SupermarketsPopulationRatio and the MinSupermarketsPopulationRatio.

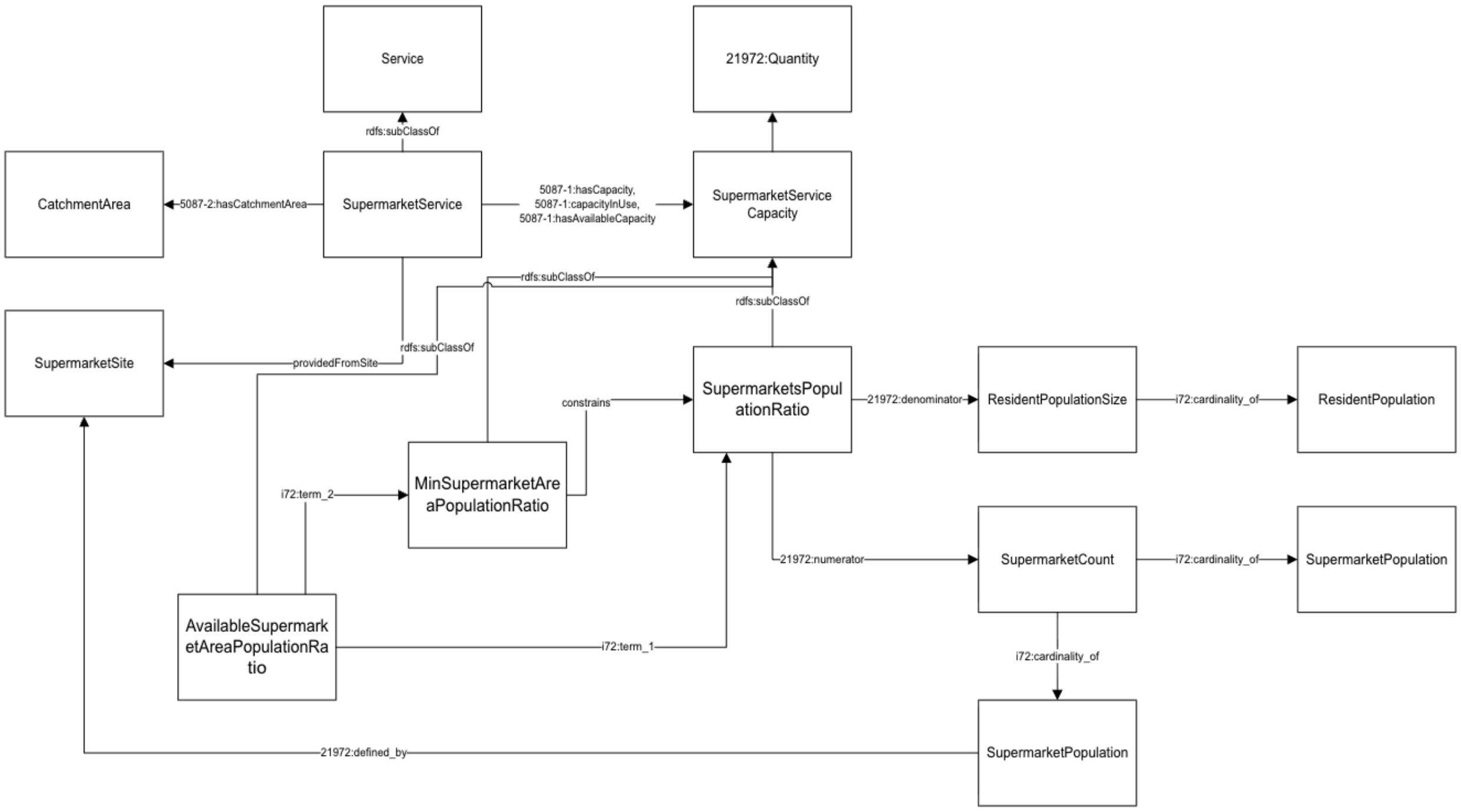

*Figure 26: Overview of the Food Service Access pattern*

### *3.2.22.1 Formalization*

The pattern is formalized in Table 35.

*Table 33: Formalization of classes in the Food Service Access pattern*

| Class | Property | Value Restriction |
|---|---|---|
| SupermarketService | rdfs:subClassOf | Service |
| | providedFromSite | only SupermarketSite |
| | 5087-1:hasCapacity | only SupermarketServiceCapacity |
| | 5087-1:capacityInUse | only SupermarketServiceCapacity |
| | 5087-1:hasAvailableCapacity | only SupermarketServiceCapacity |
| | hasPlannedCapacity | only SupermarketServiceCapacity |
| SupermarketSite | rdfs:subClassOf | 5087-1:Site |
| SupermarketServiceCapacity | rdfs:subClassOf | i72:Quantity or QuantityConstraint |
| SupermarketPopulationRatio | rdfs:subClassOf | SupermarketServiceCapacity |
| | rdfs:subClassOf | RatioQuantity |
| | i72:numerator | exactly 1 SupermarketCount |
| | i72:denominator | exactly 1 ResidentPopulationSize |
| | forLocation | only Location |
| | i72:hasUnit | value i72:population_ratio_unit |
| MinSupermarketsPopulationRatio | rdfs:subClassOf | SupermarketServiceCapacity |
| | rdfs:subClassOf | QuantityConstraint |
| | i72:hasUnit | value i72:population_ratio_unit |
| AvailableSupermarketsPopulationRatio | rdfs:subClassOf | SupermarketServiceCapacity |
| | rdfs:subClassOf | QuantityConstraint |
| | i72:term_1 | exactly 1 SupermarketsPopulationRatio |
| | i72:term_2 | exactly 1 MinSupermarketsPopulationRatio |
| | i72:hasUnit | value i72:population_ratio_unit |
| SupermarketCount | rdfs:subClassOf | i72:Cardinality |
| | i72:cardinality_of | exactly 1 SupermarketPopulation |
| SupermarketPopulation | rdfs:subClassOf | i72:Population |
| | i72:defined_by | only SupermarketSite |

### 3.2.23 Senior Services

The Senior Services Access pattern, illustrated in Figure 27, focuses on senior *care*. Other senior services (e.g., recreation) may be considered in extensions to the HPCDM (either as separate patterns, or extensions to existing patterns – e.g. community centre services). This pattern reuses and extends the general Service Accessibility pattern to capture accessibility to senior services. It defines the following classes:

- LongTermCareService: represents the services provided for seniors both through care facilities, or by support workers (in home care). Long-term care services focus on medical and physical needs of older adults but may also include support for emotional/social needs. It has the following properties:
  - providedFromSite: identifies the LongTermCareFacility from which the service is provided.
  - 5087-1:hasCapacity: identifies the capacity of the senior care service. This is specified as a LongTermCareServiceCapacity.
  - 5087-1:hasAvailableCapacity: identifies the available capacity of the senior care service. This is specified as a LongTermCareServiceCapacity.
  - 5087-1:capacityInUse: identifies the senior care service's capacity in use. This is specified as a LongTermCareServiceCapacity.
  - hasPlannedCapacity: identifies a planned capacity of the senior care service. This is specified as a LongTermCareServiceCapacity.
- LongTermCareServiceCapacity: represents the capacity (total, in-use, available or planned) of a long-term care service.
- LongTermCareResidenceService: a type of long-term care service that includes residence (i.e., in an assisted living or nursing home facility). Note that other types of services might be identified in extensions to this model, such as home care services.
- NumberOfLongTermCareBeds: is a type of LongTermCareServiceCapacity that is also a subclass of i72:Quantity. It represents the total beds (spaces) available for long term care residents.
- NumberOfLongTermCareResidents: is a type of LongTermCareServiceCapacity that is also a subclass of i72:Quantity. It represents the total number of residents in a particular long term care residence.
- NumberOfLongTermCareBedsAvailable: is a type of LongTermCareServiceCapacity that is also a subclass of DifferenceQuantity. It is defined as the difference between the NumberOfBeds and the NumberOfResidents.
- LongTermCareBedPopulation: is an i72:Population quantity that captures the population of long term care beds in a particular area (i.e., at a particular facility).
- LongTermCareResidentPopulation: is an i72:Population quantity that captures the population of long term care residents in a particular area (i.e., at a particular facility).

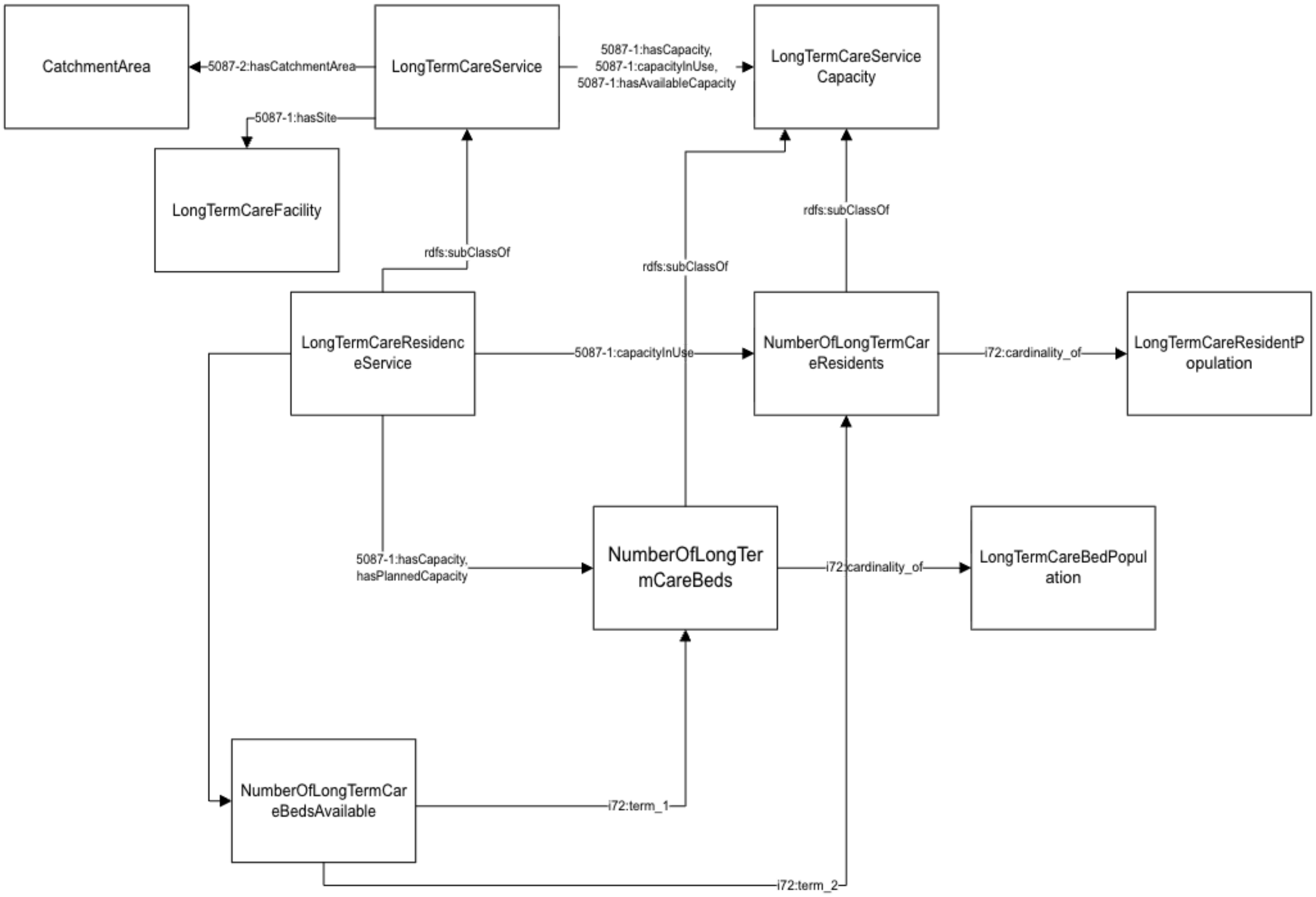

*Figure 27: Overview of the Senior Service Access pattern*

#### 3.2.23.1 Formalization

The pattern is formalized in Table 34.

*Table 34: Formalization of classes in the Senior Service Access pattern*

| Class | Property | Value Restriction |
|---|---|---|
| LongTermCareService | rdfs:subClassOf | Service |
| | providedFromSite | only LongTermCareFacility |
| | 5087-1:hasCapacity | only LongTermCareServiceCapacity |
| | 5087-1:capacityInUse | only LongTermCareServiceCapacity |
| | 5087-1:hasAvailableCapacity | only LongTermCareServiceCapacity |
| | hasPlannedCapacity | only LongTermCareServiceCapacity |
| LongTermCareResidenceService | rdfs:subClassOf | LongTermCareService |
| LongTermCareServiceCapacity | rdfs:subClassOf | i72:Quantity or QuantityConstraint |
| LongTermCareFacility | rdfs:subClassOf | 5087-1:Site |
| NumberOfLongTermCareBeds | rdfs:subClassOf | LongTermCareServiceCapacity |
| | rdfs:subClassOf | i72:Cardinality |
| | i72:cardinality_of | exactly 1 LongTermCareBedPopulation |
| NumberOfLongTermCareResidents | rdfs:subClassOf | LongTermCareServiceCapacity |
| | rdfs:subClassOf | i72:Cardinality |
| | i72:cardinality_of | exactly 1 LongTermCareResidentPopulation |
| NumberOfLongTermCareBedsAvailable | rdfs:subClassOf | LongTermCareServiceCapacity |
| | rdfs:subClassOf | DifferenceQuantity |
| | i72:term_1 | exactly 1 NumberOfLongTermCareBeds |
| | i72:term_2 | exactly 1 NumberOfLongTermCareResidents |
| LongTermCareBedPopulation | rdfs:subClassOf | i72:Population |
| LongTermCareResidentPopulation | rdfs:subClassOf | i72:Population |

### 3.2.24 Environmental Risk Pattern

Another important aspect of housing potential analysis is the consideration of environmental risks. In addition to the CQs derived from the use cases, another CQ was added within the Development Desirability theme to capture environmental risk considerations:

- Is the land subject to any climate risks (e.g. flooding)?

This pattern, illustrated in Figure 28, is informed by current research that examines the taxonomies of climate hazards (a specific class of environmental hazard), impacts, and risk factors [42]. It may be considered a kind of "stub" pattern; it serves as a high-level outline of some of the key considerations, further exploration is not required for the scope of the HPCDM, but should be pursued in future work. It introduces the following classes:

- EnvironmentalHazard: represents the actual hazard event that has the potential to cause damage and harm. It has the following property:
  - hasImpact: identifies the HazardImpact(s) that may occur as a result of the hazard.
  - hasRiskFactor: identifies a HazardRiskFactor that is known for the hazard.
- HazardImpact: represents the impact of a hazard event.

- HazardRiskFactor: a risk factor is some characteristic that is known to influence the likelihood and/or severity of a climate hazard event. It has the following property:
  - influences: identifies a HazardImpact that is influenced (i.e., increased in severity) due to the risk factor.
  - inverse (demonstratesRiskFactor): identifies an area that has the risk factor.
- Flood: a type of EnvironmentalHazard. Note that this class is included for demonstration purposes; this pattern is not intended to be exhaustive in its coverage of hazard types. It has the following properties:
  - hasRiskFactor: identifies a Floodplain. Floodplains provide an indication of the severity of impacts experienced as a result of a flood event.
- Floodplain: a type of HazardRiskFactor (included for demonstration purposes as this pattern is not intended to be exhaustive) that impacts influences the severity of impact of a Flood. It has properties such as:
  - minDistanceToWater: specifies the shortest distance between the flood plain and a body of water.
  - forWatershed: identifies the watershed that the flood plain is defined for.
  - 5087-1:hasLocation: identifies the geographical location of the Floodplain.
  - 5087-1:hasPart: identifies a FloodplainSegment that is part of the Floodplain. Floodplains may be segmented in different ways to describe different locations.
- FloodplainSegment: a type of HazardRiskFactor that describes a (geographical) part of a floodplain. It has the following properties:
  - minDistanceToWater: specifies the shortest distance between the flood plain segment and a body of water.
  - forWatershed: identifies the watershed that the flood plain segment is defined for.
  - 5087-1:hasLocation: identifies the location of the Floodplain Segment.
  - 5087-1:hasPart: identifies a FloodplainSegment that is part of the FloodplainSegment. Floodplains may be segmented in different ways to describe different locations.

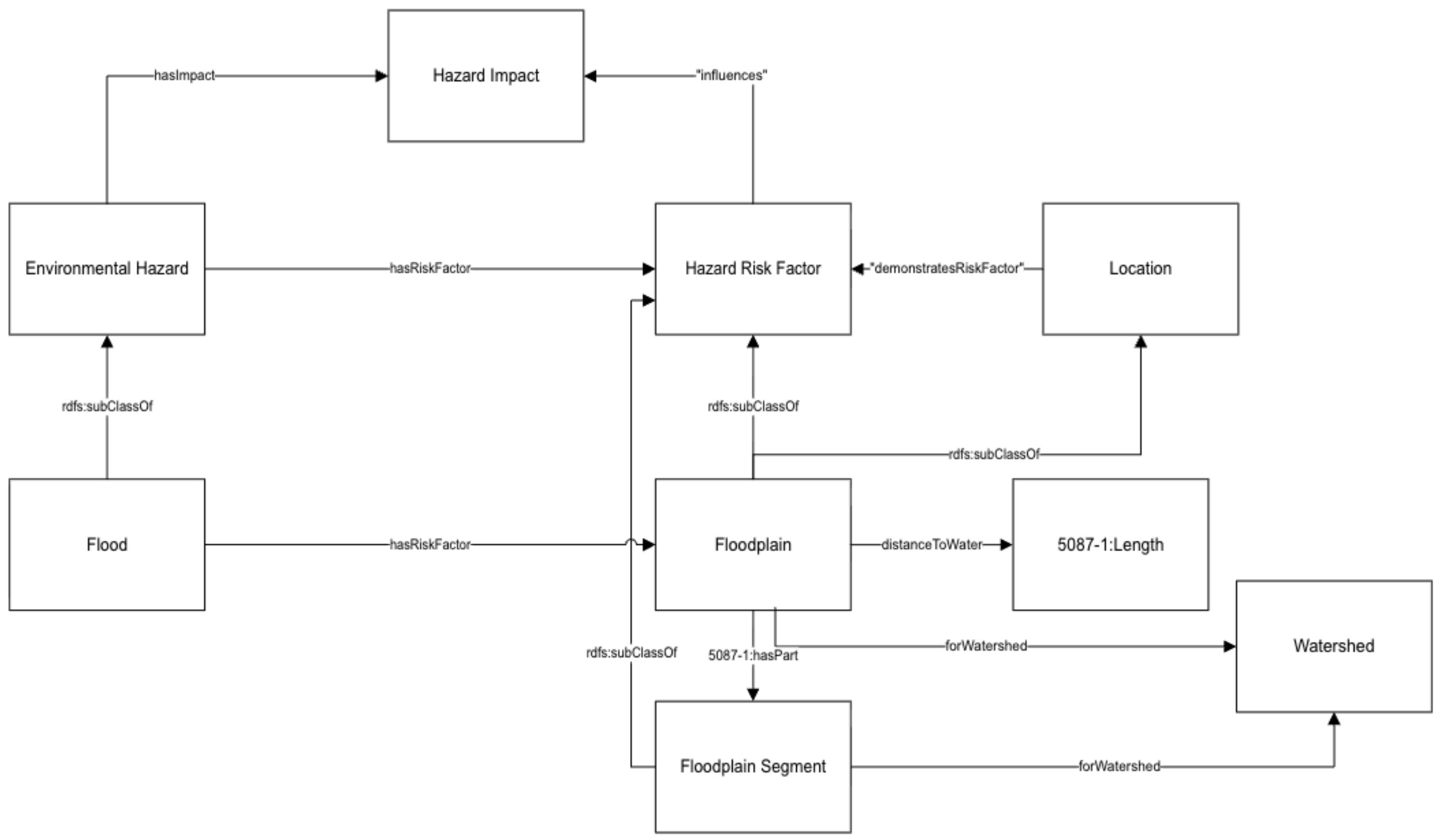

*Figure 28: Overview of the Environmental Risk pattern*

#### *3.2.24.1 Formalization*

The pattern is formalized in Table 35.

*Table 35: Formalization of classes in the Environmental Risk pattern*

| Class | Property | Value Restriction |
|---|---|---|
| EnvironmentalHazard | hasImpact | some HazardImpact |
| | hasRiskFactor | only HazardRiskFactor |
| HazardImpact | inverse(hasImpact) | some EnvironmentalHazard |
| | inverse(influencesImpact) | only HazardRiskFactor |
| HazardRiskFactor | influencesImpact | some HazardImpact |
| Flood | rdfs:subClassOf | EnvironmentalHazard |
| | hasRiskFactor | some Floodplain |
| Floodplain | rdfs:subClassOf | HazardRiskFactor |
| | minDistanceToWater | only 5087-1:Length |
| | forWatershed | only Watershed |
| | 5087-1:hasLocation | only 5087-1:Location |
| | inverse(5087-1:partOf) | only FloodplainSegment |
| FloodplainSegment | rdfs:subClassOf | HazardRiskFactor |
| | minDistanceToWater | only 5087-1:Length |
| | forWatershed | only Watershed |
| | 5087-1:hasLocation | only 5087-1:Location |
| | inverse(5087-1:partOf) | only FloodplainSegment |

## 3.3 Data Mappings

Key datasets from three Canadian cities have been mapped into the HPCDM. These mappings demonstrate the robustness and coverage of the HPCDM. They also provide the basis for the data pipeline that transforms the data into the City Digital Twin knowledge graph.

Mappings are specified with respect to a particular data source (a set of fields and expected values). Therefore, a single mapping may apply to any number of datasets *using the same data model*. For some data sources, e.g. the Census of Population, this means that a single mapping will be suitable across all Canadian municipalities. Naturally, the relevant data sources must be identified to define the mappings, so the data catalogue developed in Phase 1 is a key resource for this section. At the time of this report, the Canadian Urban Data Catalogue contains 364 housing potential-related entries, however it does not necessarily contain the required data elements to address the use cases. A continued data curation effort and a targeted review of the catalogue were conducted to identify datasets with the potential to answer the CQs for the municipalities of Toronto, Vancouver and Halifax.

The complete set of mapping specifications is included in Appendix A. Where zoning datasets are limited or incomplete, examples have been drawn from unstructured text. For Toronto synthetic datasets were generated where gaps exist to illustrate the intended use of the model and support implementation in the digital twin. The defined mappings identify mappings for datasets in green, and mappings manually extracted from documents or generated from synthetic datasets in pink. Placeholders for values drawn from dataset fields are indicated with curly brackets (e.g., {FIELD_NAME}).

### 3.3.1 Data Availability

Out of the total 71 CQs, 9 were found to have no suitable data sources (i.e., complete data gaps). On the other hand, 19 of the 71 CQs could be answered completely using the available data – all other cases required some additional input. Most requirements demonstrated partial data gaps. Key areas in which data was found to be lacking[14] are:

- Service capacities (limit and current use)
- Parcel-level service
- Zoning data:
    - Detailed use classifications

[14] Note that this assessment refers to the availability of (publicly accessible) datasets. The extraction of data from other, unstructured sources (e.g. drawings and text documents) is not considered at this time.

- Setback, shape, and capacity restrictions
- Integration of datasets, e.g. buildings and parcels
- Ownership and occupancy data
- Plans (e.g., service changes, zoning)

While some, such as parcel-level details, may present challenges related to data privacy, others represent opportunities to improve upon the data available for housing potential analysis. A complete listing of the available datasets and data gaps identified for the Toronto implementation is provided in Appendix C. More details on the datasets used may be found in the mapping specifications in Appendix A. Approaches to generating synthetic data are detailed in Appendix D.

## 3.4 Evaluation

### 3.4.1 Verification

Detailed results of both the Competency Question (CQ) formalization and CQ evaluation are documented in CQ_EvalSummary.xslx, included in the project repository[15]. The CQ evaluation was performed using the knowledge graph-based City Digital Twin, implementation of which is discussed in Section 4.

#### *3.4.1.1 Overview of CQ formalization results*

To verify the scope of the model is sufficient in both breadth and depth, and important step in ontology development is to evaluate the ontology against the CQs. This involves the formalization of the CQs in an implementation-level language (in this case, SPARQL). The results of the evaluation are summarized in Table 36. In general, most of the queries that were not formalized were omitted from the original scope following a design decision to omit market-specific considerations, and the identification of several services that were not central to the model. Of the remaining CQs that could not be formalized, two concern building shape, and one involves identifying an "amenity score." These queries have been omitted because the underlying concepts are not sufficiently well defined to be included in the HPCDM (though they could be introduced in user- extensions of the model). Any interpretation used in the CQs would therefore be arbitrary and not useful for evaluation of the model.

*Table 36: Summary of CQ formalization*

| **Theme** | **CQs formalized** | **CQs not expressed** | **Total CQs** |
|---|---|---|---|
| Land Use and Zoning | 31 | 2 | 33 |
| Development Feasibility | 20 | 0 | 20 |
| Development Desirability | 17 | 1 | 18 |

[15] https://github.com/csse-uoft/hpcdm-dashboard/blob/main/CQ_EvalSummary.xlsx

| | | | |
|---|---|---|---|
| **Total** | **68** | **3** | **71** |

Notes on CQ interpretation:

- Service availability is interpreted by catchment area or distance to site as appropriate. These parameters may be adjusted as needed.
- Walking distance is interpreted as 400m or less.
- Spatial operations: in most of the formalizations we use geo:sfIntersects for simplicity, however in practice it may be necessary to distinguish between an intersection vs containment (as partial overlap may correspond to cases that require special consideration).
- In cases where the question is about a particular object (e.g. "the parcel"), a random instance is used in the CQ encoding as a placeholder.

#### *3.4.1.2 Overview of CQ evaluation results*

After formalizing the CQs and loading the datasets (transformed according to the specified mappings) into the City Digital Twin, a second stage of verification is performed in which the CQs are tested in the implementation to ensure there are no logical or syntactic errors. All query formalizations were demonstrated to be functional, and the HPCDM has been successfully verified from the perspective of representing the required data to address the identified requirements. The verification exercise initially identified some issues in the implementations of the data mappings, which were addressed and corrected in Phase 4 of the project. The results of the evaluation are summarized in Table 37.

*Table 37: Summary of CQ Evaluation Results*

| Theme | Pass | Pass with missing data | Pass with fault | Incomplete (System error) |
|---|---|---|---|---|
| Land Use and Zoning | 20 | 6 | 0 | 5 |
| Development Feasibility | 14 | 3 | 2 | 1 |
| Development Desirability | 12 | 3 | 2 | 0 |

The results are divided into four categories to support a clear description of the outcome in the context of the current state of the overall project:

- **Pass** indicates that the query runs successfully and performs as expected.
- **Pass with missing data** means the query executes successfully but the required data is incomplete, so there was no assessment of the expected answers; it is important to note that in many of these cases, CQs of a similar form have been run and executed successfully with data, and the model and CQ formalism are correct—the question raised in these cases is whether the missing data points should be added (or synthesized) for the Phase 4 demonstration.

Regarding missing data details, several datasets were not found and no synthetic data was created, including setback regulations, plans (planned zoning, planned land use, planned greenspace), capacity regulations (max units), building occupants, planned service capacities, and actual use (and historical actual use). Plans and missing regulation types were generally omitted because the form and content were expected to match the patterns of the actual data (e.g., hasCapacity vs. hasPlannedCapacity), and it was not worthwhile to generate synthetic data for this purpose. Actual use has been approximated through queries of permitted use, and although it would be possible to attempt to approximate use by integrating building and organization data, this would require considerable effort with minimal benefit to the evaluation of the model. Other datasets have been identified, and mapping implementation or correction is in progress, including height regulations, federal government building ownership, federal government building occupancy/use, and population centre locations; in these cases, the datasets were identified too recently to be included in the mapping implementation, or some errors in the implementation were identified and are now being corrected.

In addition to resolving CQ encoding mapping implementation bugs, verification served to test the initial knowledge graph set-up to prepare for Phase 4 of the project. It also provided some insights into considerations for implementation, discussed in Section 5.

### 3.4.2 Validation

A survey was conducted to solicit feedback as a means of validating the model with subject matter experts. Specifically, the goal of the validation exercise was to convey the way in which the HPCDM is used to answer the required questions (CQs) identified by the use cases and solicit feedback on whether the questions were interpreted correctly.

Feedback was solicited from the advisory panel as well as from some suggested external reviewers. The advisory panel had previous exposure to the project and were involved in the definition of the use cases and specification of requirements. The external reviewers provide a complementary view as they have no previous exposure to the project and so may offer distinct views on the model and its requirements.

The survey was conducted as follows:

- To keep the length of the task reasonable and encourage participation, a selection of examples from the use cases (i.e., rather than a detailed review of the entire model) that convey a breadth of concepts from the model was provided.
- Each example breaks down the way in which a question is captured by the model and identifies the concepts that are used to answer it.
- We then ask, for each CQ:
    - “Do you agree with the interpretation of the question?”
    - “Is the right data being used to answer the question?”
    - “Any other feedback?”
- At the end of the survey, participants are given the opportunity to provide general feedback on the contents of the model’s content and scope, as well as thoughts on future extensions.

#### *3.4.2.1 Results summary*

Seven survey responses were received, one of which is from an external participant. The feedback received on the model is summarized in Table 38. The responses were positive overall.

The participants identified some useful considerations for the implementation, and challenges for future work. Key themes that arose in the responses were:

- The need to include contextual information – e.g., in the context of a particular regulation: what other regulations are in place in the area? What factors might influence the regulation?
- Transportation capacity is complex and it may not be appropriate to try to address it with a single query.
- Zoning isn't a reliable proxy for the actual use/activities in an area; other types of data are required.

*Table 38: Summary of validation responses.*

| **CQ** | **Interpretation Agreement** | **Data Use Agreement** | **Actions Taken** |
|---|---|---|---|
| CQ 8d. How closely does recent development conform to height restrictions? | Strong agreement; suggest considering context in the query (e.g., what factors into the height restriction) | Mostly yes; radius too large; missing contextual datasets. | Reduce radius in query; consider including additional elements in CQs (e.g., presenting multiple restrictions at once) in Phase 4. |
| CQ 24b. What density policies apply to a parcel of land? | High agreement. | Mostly yes; note that there are other policies that contribute to density – FAR is insufficient alone | Inclusion of additional elements in CQs (e.g., presenting multiple restrictions at once) in Phase 4. |
| CQ 57. What is the building zoned for? | Agreement; suggest distinction between zoning (land use) and "permitted use" | Generally yes; but site-specific rules unclear.<br>Note the importance of *actual* use vs zoned use. | Plan to include terminology for "permitted use" in the next version of the HPCDM. |
| CQ 6a. Is the parcel serviced by water? | Full agreement. | Full agreement. | - |
| CQ 10d. What is the capacity for transport to absorb increases in population? | Concerns about the question – suggest it is a difficult question and may not make sense to ask in this way because there are so many factors involved. | Concerns about unclear capacity definitions and availability of data. | Agree that transport capacity is difficult to capture – the purpose of the mapping in this case was to illustrate one possible way to represent the capacity. The model is designed to |
| | | | support multiple different representations of capacities, as appropriate. Future work should explore effective ways of communicating these different measures. |
| CQ 12a. Is it 15 minutes to public transit? | Strong agreement.<br>Suggest that the type of transit service should also be considered. | Yes; note that "15 minutes" should be measured as network distance. | Plan to include ways of categorizing public transit (e.g. by mode, frequency) in the next version.<br>Investigated feasibility of computing network distance in an implementation. Not necessary for the dashboard application in Phase 4, but should be reasonable to implement for other applications. |
| CQ 66. Is the vacant land close to employment opportunities? | Mixed agreement – query should include other factors beyond zoning. Suggestion that "close to" radius could account for locations that are accessible by transit. | Zoning not reliable proxy for employment. Suggest question should also query for building uses (not just nearby zoning); the volume and type of employment would also be useful. | Updated the query to include building use (though this data is currently not widely available). |
| CQ 11a. What is the population of the neighbourhood? | Strong agreement. | Mostly yes; Note that census-only may be insufficient. | - |
| CQ 30a. What are the plans for greenspace in the area? | Strong agreement. | Mixed; Suggest that this data is difficult to capture (multiple sources, in nonstandard/unstructured | Plan to identify (and provide example mappings) for a broader range of data sources in the |
| | | formats); Suggest to include tree canopy plans | next version of the HPCDM. |

# 4 Housing Potential Common Data Model Implementation

To move beyond a theoretical framework and demonstrate the utility of the HPCDM, Phase 4 of the project focused on implementation. This is described in two parts: the first is in the development of the general-purpose CDT, the second is in the development of the pilot housing potential dashboard which interfaces with the CDT.

## 4.1 CDT Architecture

The general-purpose CDT is a knowledge graph-based tool that employs the HPCDM as an *interlingua* – a common intermediary language – to represent and integrate relevant data sources. It is also the data model for the CDT knowledge graph. The architecture of the CDT is summarized in Figure 29. Relevant data (e.g. zoning bylaws, population demographics) are transformed via data mapping pipelines that re-encode the data according to the terms defined in the HPCDM. The transformed data is then integrated and linked (the same places, organizations, etcetera that are identified differently in the various datasets are unified to have a single common identifier) in the CDT knowledge graph, where they all share the same underlying data model. Once the knowledge graph that integrates the housing potential-relevant dataset is completed, housing potential algorithms can access the data to perform analysis in support of question answering and visualization.

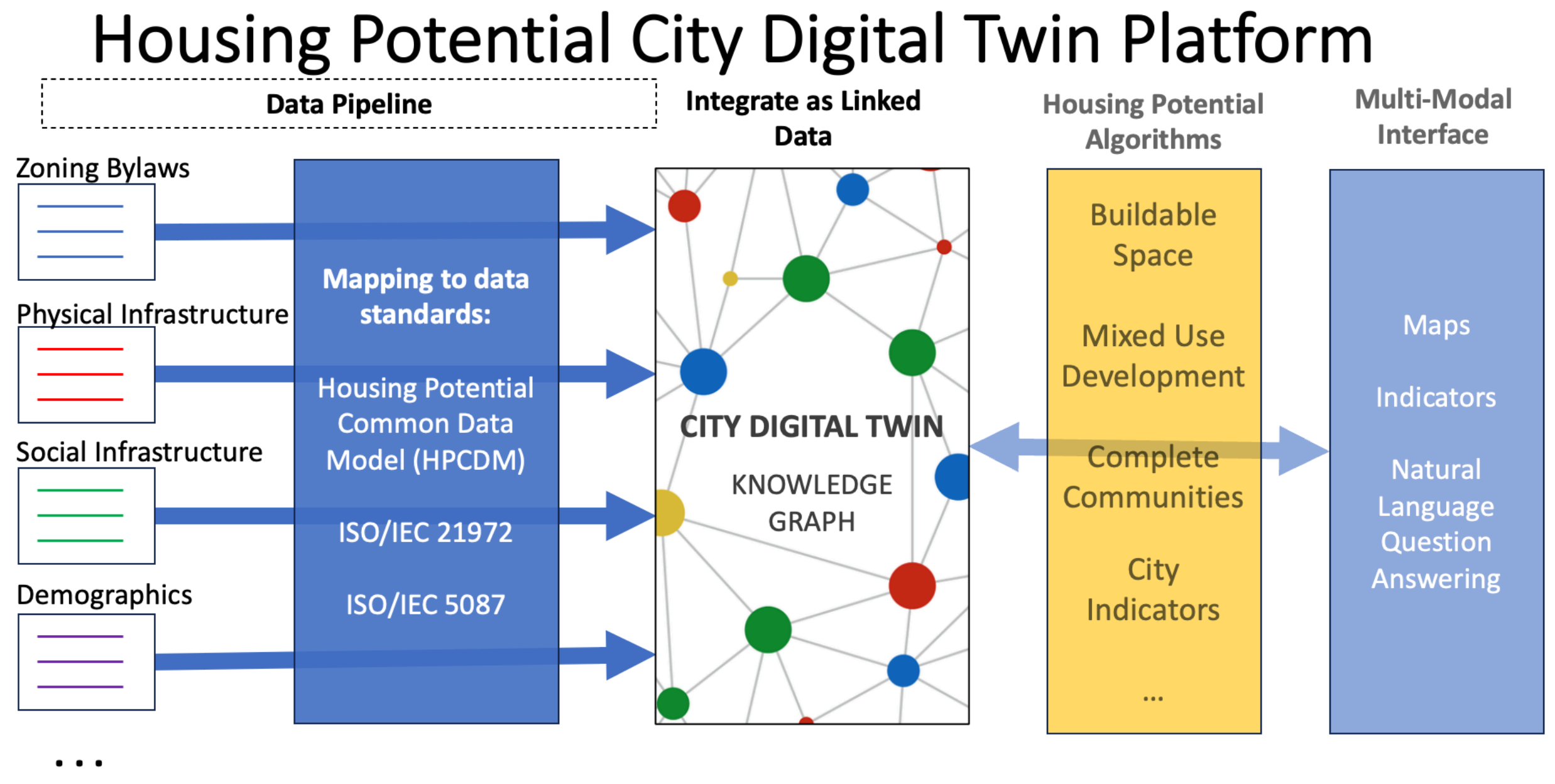

*Figure 29: Housing Potential City Digital Twin overview. Algorithms and interface illustrate how the digital twin may be utilized but are not part of the digital twin itself.*

This architecture may be implemented with various toolsets. The instance of the CDT created for this project was implemented using GraphDB 11.3[16], a graph database designed to support RDF (the Resource Description Framework[17], a W3C standard data representation language) and SPARQL[18] (a W3C standard query language for linked data).

The CDT knowledge graph is currently deployed at https://cdt.project.urbandatacentre.ca/?repositoryId=CDT_HPCDM_Demo, running on an an 8-core Intel Xeon virtual server equipped with 30GB of RAM and a total of 564GB of storage. The repository is currently running with an OWL2-RL [43] reasoner to allow for inference with property chains, with GeoSPARQL [44] indexing enabled to support more efficient geospatial queries.

It has been populated with the Toronto datasets (real and synthetic where actual data does not exist) and mappings. The data mapping pipeline implementation has been verified through evaluation SPARQL formalizations of the Competency Questions (CQs), as described in the Phase 3 report. The implementation files are available in the City Digital Twin repository: https://github.com/csse-uoft/city-digital-twin-ontology.

---

[16] https://graphdb.ontotext.com/documentation/11.3/

[17] https://w3.org/RDF/

[18] https://w3.org/TR/sparql11-query/

### 4.1.1 Data mapping pipeline

The data mapping pipeline is the process by which data is transformed from its original format into a representation consistent with the HPCDM, integrated and accessible in the CDT knowledge graph. This includes pre-processing steps such as data cleaning as well as a reformulation of the data in accordance with the HPCDM. The mechanism by which a dataset is transformed may vary depending on the implementation. For this project, the mappings specified for the Toronto datasets have been implemented in Python.

The mapping scripts serve to complete the mapping but also to address any preprocessing requirements. For example, since there is a geospatial component to many of the datasets, there may be requirements for conversion of location coordinate systems. The GraphDB system implements geospatial functions that require coordinates to use CRS 84, so in some instances conversion may be required. It is also important to include geometry validation checks to avoid issues arising at load time. Minor extensions were made to the Toronto mapping specifications as originally defined. In particular, annotations and service site names have been included where necessary to improve the presentation of the query results.

Once the data has been transformed, it can be imported into the CDT knowledge graph. Since GraphDB supports RDF and SPARQL, the data is transformed into turtle (or other standard RDF encoding). Each data source type is loaded into a separate named graph. This is done to facilitate updates and corrections to individual datasets. It also serves as a means of capturing rudimentary provenance for the data in the knowledge graph.

Finally, once the data has been loaded into the knowledge graph, automated inference is performed to capture additional relationships that are not explicitly represented in the source datasets. Inferences are made based on the definitions of the HPCDM, which is also loaded into the knowledge graph in its OWL encoding[19]. The following three SPARQL update queries perform additional inferences:

- zonedAsType: If an administrative area has a geospatial location that overlaps with the area that a regulation designates a zoning type for then the area is zoned as that zoning type. Defined as SPARQL_update_zonedAsType.rql in the dashboard repository.
- zonedForConstraint: If an administrative area has a geospatial location that overlaps with the area that a regulation defines a constraint for, then the area is zoned for that constraint. Defined as SPARQL_update_zonedForConstraint.rql in the dashboard repository.
- servicedBy: If an administrative area has a geospatial location that overlaps with a catchment area or is within the defined radius of a service site, then it is serviced by that service. Defined as SPARQL_update_servicedBy.rql in the dashboard repository.

## 4.2 Housing Potential Visualization Tool

To provide a concrete demonstration of how the model can be applied to develop tools to support housing potential analysis, a pilot housing potential dashboard has been developed, leveraging the general-purpose CDT knowledge graph as its source of integrated data. The HPCDM serves as a standard language that can be used for the specification of SPARQL queries to access the KG. In this way, the same implementation may easily be repurposed across municipalities.

---

[19] http://ontology.eil.utoronto.ca/HPCDM.owl

### 4.2.1 Requirements

Requirements for the tool have been defined according to stages, with Stage 1 – described here - being the initial target for development. The subsequent stages, outlined in Appendix E, identify directions to be pursued for future work. The purpose of defining this series of stages is twofold: (1) to ensure completion of a minimum viable tool within the timeline of the project, while (2) communicating the range of potential extensions.

In Stage 1, the key functional requirements to be supported by the application are:

1. The tool must support address-based parcel selection
2. The tool must enable parcel-focused queries for:
    2.1. Parcel attributes (e.g. area)
    2.2. Applicable zoning constraints (e.g., minimum lot area, maximum building height)
    2.3. Land use (defined by zoning and actual, if known)
    2.4. Zoning compliance in the area (e.g., zoning constraints vs attributes of the developments, if known)
    2.5. Accessible services and their capacities
    2.6. Neighbourhood demographics

Alongside the above, the following non-functional requirements were identified:

1. The tool should simplify the task of running each query (e.g. no SPARQL knowledge required).
2. The tool should display the results in a way that is easy to understand.

Each functional requirement corresponds to one or more of the CQs defined in the requirements for the HPCDM. In some cases, additional SPARQL queries are required to support the implementation. An overview of the queries implemented for the application is provided in Table 39. The queries implemented to support the above functionality capture 41 of the original 71 CQs defined in the use cases for the HPCDM. This discrepancy is because the original CQs were defined as requirements for a common data model for housing potential, *not* to capture the requirements for a single tool. The focus of the application development was to create a tool that could demonstrate use of the model, targeting some of the identified use cases, with the CDT knowledge graph. CQs were excluded based on three primary factors: data quality (e.g., limited data on planned future states), functional alignment (e.g., publicly owned land, which is slated for the Stage 2 search interface), and conceptual scope (e.g., building-specific attributes are better suited for a dedicated building-view rather than a parcel-view).

### 4.2.2 Design

The current design implementation is available with documentation at https://github.com/csse-uoft/hcdm-dashboard, an instance of which is running at http://206.12.97.46:7860/. The application is developed with Python 3.10, and is developed primarily with the Gradio[20] 6.9 library, as it allows for rapid prototyping, with a minimal development to create a UI on top of the python functions. This approach is focused on developing the required logic to demonstrate the functionality that can be supported by the HPCDM and the CDT knowledge graph that implements it. Future iterations can repurpose the functions developed here – in particular, the sparql_client module – to build extensions of this work such as the additional functionality outlined in the future stages, or to develop applications in other software frameworks.

[20] https://www.gradio.app/

The HPCDM plays key role in the design of the sparql_client, it provides the language for standard query templates that can be used to retrieve the required data. Using these query templates, the application connects directly to the CDT knowledge graph, pulling data on demand when a search is initialized by the user.

The application is initiated with a basic address-based search for a parcel. Using a geocoder, a point coordinate is retrieved for the address, which is then used as a parameter for a SPARQL template to retrieve the associated parcel (search results view depicted in Figure 30). Once the parcel is selected, the user can choose between the following "parcel queries":

1. Parcel Attributes: returns descriptive attributes (e.g., area) currently defined for the parcel.
2. Available Services: returns the services that are identified as available to the parcel and their associated capacities.
3. Applicable Zoning: returns the zoning regulations currently defined for the parcel.
4. Land Use: returns the current land use and allowed land uses for the parcel.
5. Neighbourhood Demographics: returns the relevant demographic information (population density, average income, and number of private dwellings) for the parcel's neighbourhood.
6. Zoning Compliance: compares nearby parcels (within 200m) against applicable zoning regulations for a user-selected attribute (e.g., building height or lot area).

Each dropdown triggers a SPARQL query(s) which is then run against the underlying CDT knowledge graph. The results are primarily returned in the form of a table, as the nature of the data is not well-suited for other visuals (e.g. graphs, charts). Where possible, relevant areas are plotted on the map to provide additional context. For example, the results of the zoning compliance query for the attribute of lot size provide a table with details such as the specified limit, the actual value, and the parcel ID, while the map displays a visual summary indicating whether each nearby parcel is compliant. A screenshot of this view is shown in Figure 31. This basic set-up allows for straightforward extension to capture other of parcel-specific data retrieval tasks. A complete set of application screenshots for each dashboard option is included in Appendix G.

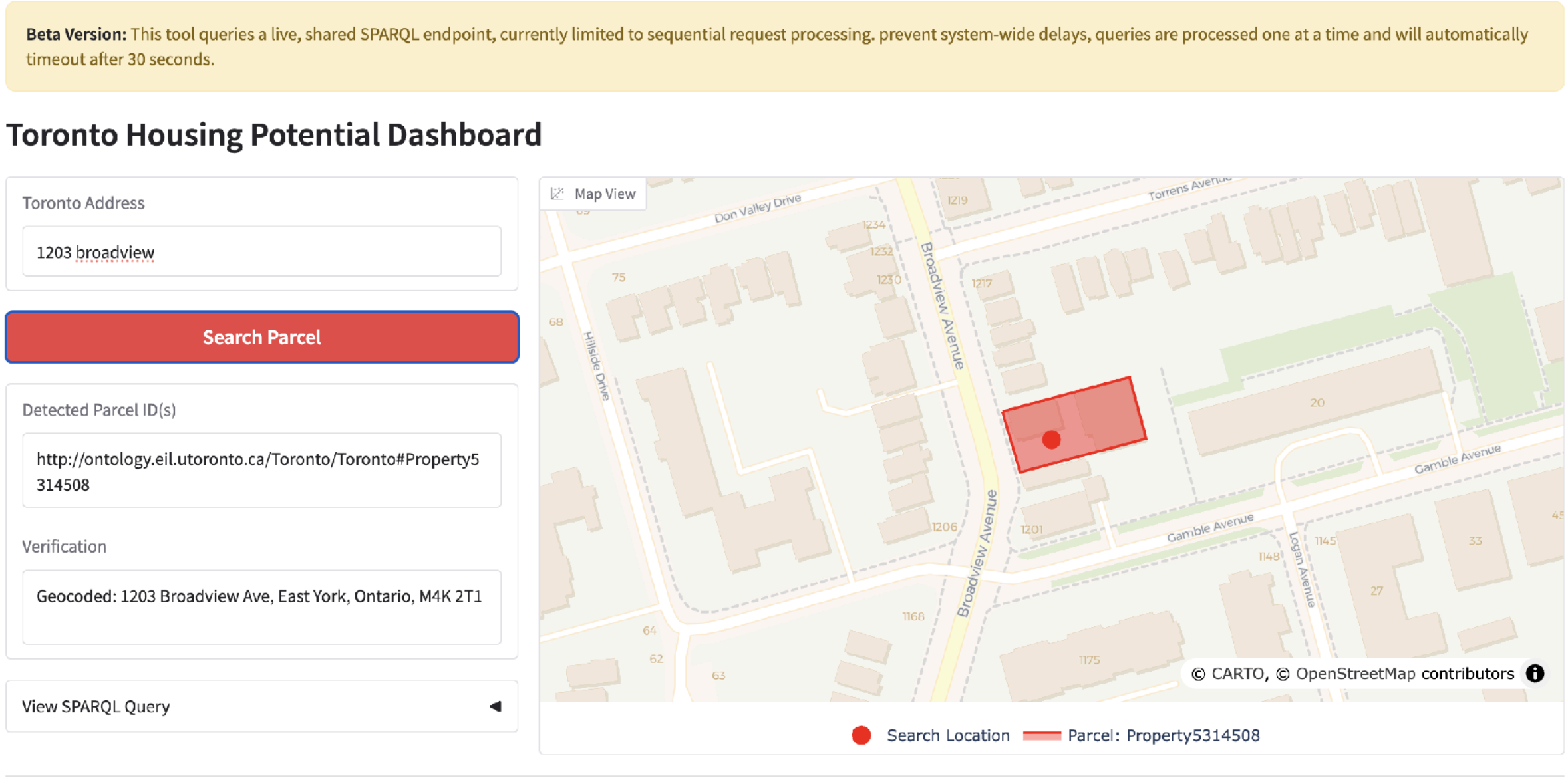

*Figure 30: Parcel search results*

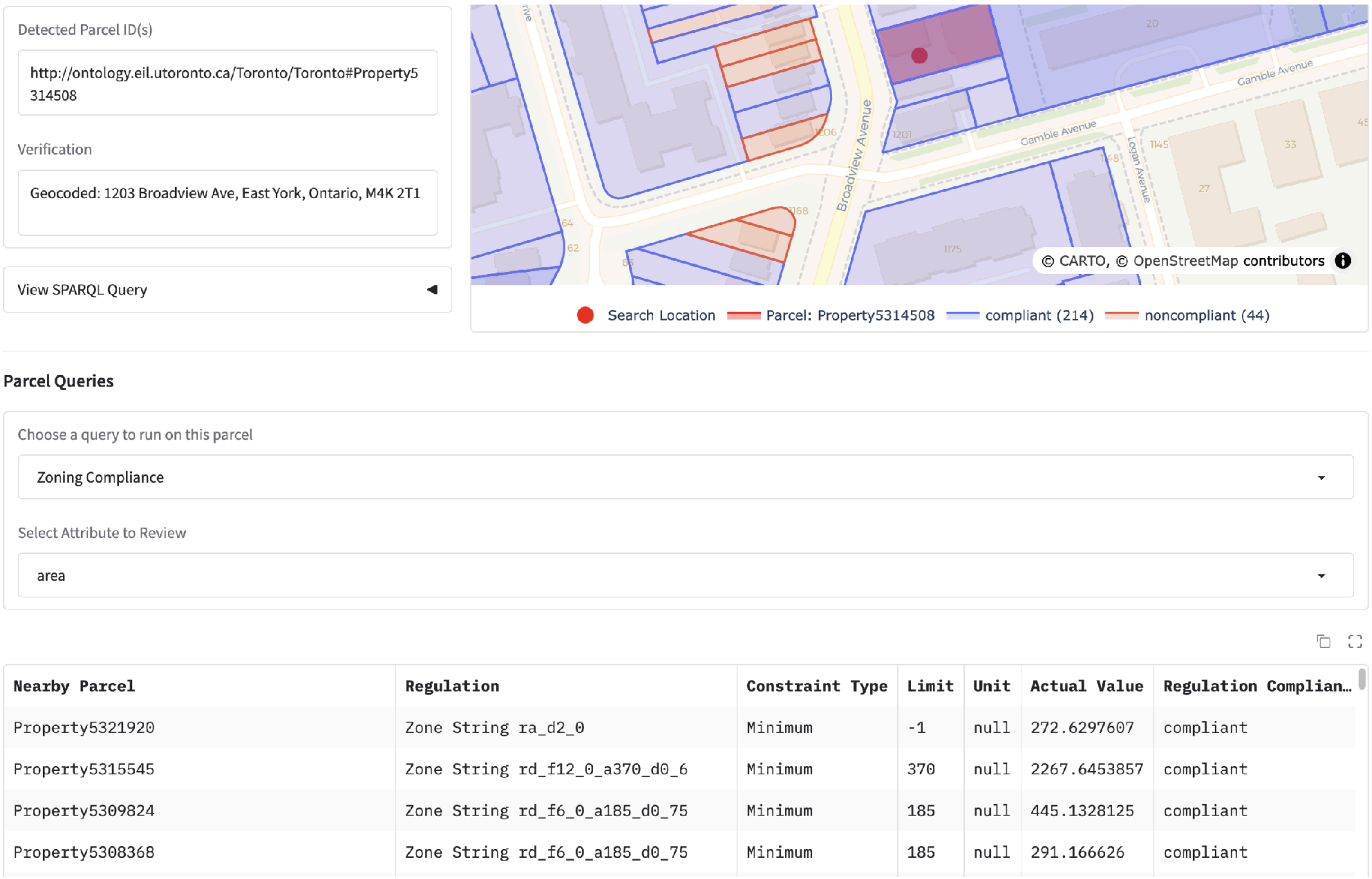

*Figure 31: Map and table results for nearby parcel area zoning compliance*

Beyond the specific dropdown queries, the application executes summary queries for: Available Services, Applicable Zoning, and Neighbourhood Demographics. These summaries return city-wide averages alongside the specific results, offering supplemental context to help users better interpret the data. The complete set of SPARQL query templates used by the application is provided in Appendix FSPARQL Queries.

As summarized in Table 39, the dropdown queries do not directly correspond to specific CQs from the set originally defined in the use cases for the HPCDM. Instead, they are defined to capture generalizations of many of the core types of queries that appear in the requirements.

*Table 39: Overview of queries supported by the application.*

| **Dropdown Option** | **Query description(s)** ***(natural language)*** | **Corresponding CQ(s) (HPCDM use case requirements)** |
|---|---|---|
| **Parcel Attributes** | What are all the known attributes of the parcel? | 1a. What is the size of the parcel?<br>1b. What is the perimeter of the parcel? |
| **Available Services** | Part 1: What types of services are defined in the model?<br>Part 2: For each type of service: Is it available to the parcel (either by catchment area or proximity), and if so what is its available capacity? | 5. Is the parcel accessible directly by road?<br>5a. Fire and emergency access?<br>6a. Is the parcel serviced by water?<br>6b. Is the parcel serviced by wastewater?<br>6c. Is the parcel serviced by electricity?<br>10a. What is the capacity for schools to absorb increases in population?<br>10b. What is the capacity for water services to absorb increases in population?<br>10c. What is the capacity for wastewater to absorb increases in population?<br>10d. What is the capacity for transport to absorb increases in population?<br>28c. Is the area currently serviced by electricity? What is the capacity?<br>28d. Is the area currently serviced by waste disposal? What is the capacity?<br>28f. Is the area currently serviced by fire and emergency services? What is the capacity?<br>35a. For a parcel of land: What public utilities currently service the area? What is the capacity? What are the future expansion plans?<br>12a. Is it 15 minutes to transportation?<br>12b. Is it 15 minutes to schools?<br>(What schools are in the area?) What is their capacity?<br>What is their planned capacity?"<br>12c. Is it 15 minutes to food?<br>12d. Is it 15 minutes to medical services?<br>(What medical facilities and pharmacies are in the area?) What is their capacity?<br>What is their planned capacity?<br>12e. Is it 15 minutes to parks?<br>(What parks are in the area?) What is their planned capacity?<br>12f. Is it 15 minutes to libraries? |
| | | (What libraries are in the area?) What is their capacity?<br>What is their planned capacity?<br>13a. Does the parcel have access to Community centre/programmes?<br>13e. Senior services/home care?<br>29b. What parks are in the area? What is their capacity?<br>"29d. What childcare facilities are in the area? What is their capacity?<br>What is their planned capacity?" |
| **Applicable Zoning** | What zoning regulations exist for the parcel? | 7a. What regulations exist on setbacks from road or adjacent parcels?<br>7b. What regulations exist on FSR (Floor Surface Ratio)?<br>7c. What height restriction exists?<br>7d. What minimum lot size restriction exists?<br>24b.What density policies apply to a parcel of land?<br>37. What is the zoned capacity for a parcel of land? |
| **Land Use** | What use is the parcel zoned for, and what is it currently used for (based on building use, if known)? | 3. What use is parcel x zoned for, e.g., residential (single family, multifamily), commercial, mixed, industrial?<br>3a. What is it currently being used for?<br>24c. Does the parcel have the potential for mixed use development?<br>56. What is the current use of the building? |
| **Neighbourhood Demographics** | What is the value of a given set of census characteristics for the census tracts in the parcel's neighbourhood? | 3b. What is the current density of the neighbourhood?<br>11a. What is the population of the neighbourhood?<br>11b. What is the annual income (average) of the neighbourhood?<br>11c. What is the average rooms per home in the neighbourhood? |
| **Zoning Compliance** | For nearby parcels - compare the known attributes (e.g. size, building height) with restrictions on those attributes defined in the applicable zoning. | 8a. How closely does recent development conform to setback regulations?<br>8b How closely does recent development conform to FSR regulations?<br>8c How closely does recent development conform to height restrictions?<br>8d. How closely does recent development conform to minimum lot size restrictions? |

## 4.3 HPCDM Extension

An extension to the HPCDM[21] has been defined to introduce Toronto-specific subclasses. This is expected usage of the common data model. By defining Toronto specialization of service subclasses (e.g. "Toronto Supermarket Service", "Toronto Library Service"), application-specific default service radii may be defined for each class. This allows the application to assess service access in the absence of more precise catchment area information.

# 5 Implementation Guidance

This section outlines a typical implementation approach – leveraging the HPCDM to construct a CDT knowledge graph, then developing tools for housing potential analysis that utilize the graph as the underlying data layer. Implementation begins with the creation of the CDT knowledge graph, then end-user value is created through the development of applications that interface with the graph.

## 5.1 Constructing a Knowledge Graph

The process of constructing a CDT knowledge graph (Figure 32) begins with the specification of requirements that the CDT is intended to support. These requirements are then used to inform the selection of the software platform that will host the CDT, followed by the design and implementation of the data mapping pipeline. Each task is described in more detail in the following.

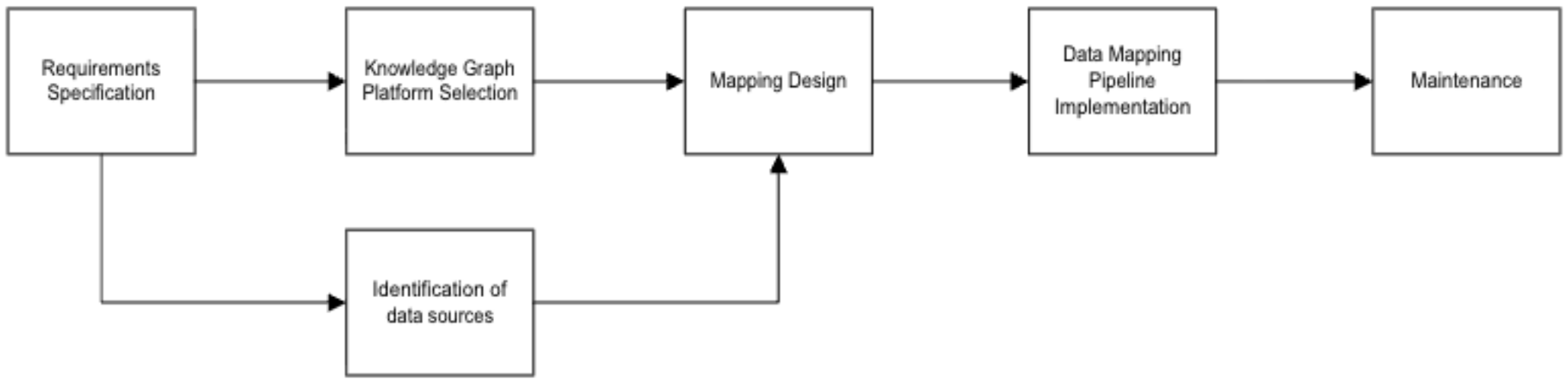

*Figure 32: Overview of the knowledge graph construction process*

### 5.1.1 Requirements specification

The construction of an integrated knowledge graph requires considerations of software resources and functional requirements. A knowledge graph should be implemented in anticipation of at least one (if not more) intended applications to be supported by the knowledge graph. These serve as use cases, requirements that are important to:

- Define the scope of instantiation: Rather than aimlessly mapping and integrating all available domain data, identifying intended applications allows for a requirement-driven instantiation. This ensures that the resulting CDT is both manageable and purpose-built, while remaining extensible for future needs.

---

[21] https://github.com/csse-uoft/hpcdm-dashboard/blob/main/TorontoHPCDM.ttl

- Understand the technical requirements: reasoning and other specialized tasks that must be supported. This will help to inform the selection of a software platform to host the knowledge graph.

While the detailed functional requirements for each application may not be known at this stage, understanding of the kinds of data that are required and the types of questions that will be posed is critical to inform the platform selection and knowledge graph population.

### 5.1.2 Platform selection

The core function of the knowledge graph tool is to serve as a platform to integrate relevant datasets, providing a single point of access to the CDT. But these tools also provide important functions such as inference and geospatial reasoning. These abilities (along with their costs and limitations) need to be considered when choosing the tool to adopt for a particular CDT implementation. While the knowledge graph set-up and its data sources can and should be designed to be general-purpose, different knowledge graph systems excel at different tasks, so it may be that one system may be better for a certain class of CDT applications than another.

There may also be considerations of organizational resources and requirements; it may be preferred to implement smaller scale knowledge graph instantiations in some cases (e.g., to minimize cost), in other instances the organizational policies may require the creation of a single, central instance of the CDT (e.g., to maintain ownership by a particular department).

In this project, the free license of GraphDB was implemented in this project to minimize costs. While it doesn't impose limits on the size of the graph, it places a restriction on graph interactions with a maximum of two concurrent queries. This is sufficient for CQ evaluation as well as to support the design and testing of the dashboard application, however it would likely not meet the requirements of a production system.

### 5.1.3 Data ingestion and maintenance

Once the tool is selected, the data mapping pipelines can be employed to load the transformed data into the knowledge graph. As discussed in Section 4.1.1, data can be transformed from its original format into an HPCDM-compliant representation with the definition of data mappings. The design and implementation of the mappings will vary between organizations, depending on the data available and its encoding, however in all cases it follows the same overall process.

#### *5.1.3.1 Mapping design*

The mapping design involves the identification and specification of correspondences between the data elements and the HPCDM:

1. For each dataset, consider what element(s) are being described. Identify what class(es) the data points correspond to.
2. For each instance of a class, a unique identifier must be defined. To achieve integrated, *linked* data, it is important that the identifiers not be arbitrary but follow a standard naming convention, making use of identifiers in place in existing systems. If different identifiers are produced for the same element in different datasets, then another mechanism will need to be undertaken to recognize and align (link) these instances in any implementations.

3. For each instance, a dataset likely defines one or more properties that relate it to another instance or specify some datatype value. These relationships must be identified and then specified using the language of the HPCDM.
4. In addition to defining the individual objects described in the data, it is important to add names and labels where possible. This makes the data easier for stakeholders to interpret and provides utility for future applications where labels are important for presentation of the data in a user interface.

#### *5.1.3.2 Data mapping pipeline*

Once the mapping is designed, it can be implemented and loaded into the CDT with various technologies. This transformation is referred to as the data mapping pipeline. This project has implemented a suite of mappings using Python to transform datasets into encodings in RDF, however there are other UI-style tools available that support mappings from tabular data into the RDF language according to a given model (e.g. sand[22], TopQuadrant[23]). Prior to testing the CDT with queries and applications, mapping errors may not be apparent. It is important to include a check of the results to ensure that the design has been correctly implemented (e.g., do the terms used correctly correspond to the terms in the HPCDM?).

The data mapping pipeline should be designed such that any datasets in the original format may be transformed and loaded into the CDT – i.e., it should be reusable when new datasets are released. This means it may also be reusable across municipalities when datasets are available in the same format.

Beyond the data itself, an encoding of the HPCDM should also be captured within the graph. This results in a CDT that is more than a store of linked data. It allows the system to support reasoning about the objects and their relationships to one another.

#### *5.1.3.3 Maintenance*

Once the CDT has been created, typical policies for data systems must be designed and adopted, e.g. for system maintenance and updates. If running SPARQL updates, efficiency should be factored into the query design. It is often the case that "streamlined", general-purpose updates may incur many more operations than is necessary. For example, if there's been an update to the data on libraries, rather than running the generic SPARQL update to infer the servicedBy property (discussed in Section 4.1.1), a specialized update may be designed to apply specifically to LibraryService objects (see SPARQL_update_servicedBy_optimized.rql).

## 5.2 Developing an application

Once the CDT has been created, users may begin to create applications that interface with it (i.e., to satisfy some of the requirements that were identified in Section 5.1.1). To connect to the knowledge graph requires a connection to the query endpoint. For the dashboard application, the SPARQL API for the Toronto CDT knowledge graph was accessed via SPARQLWrapper[24] for python. For SPARQL APIs, the two common supporting libraries to create this connection are

[22] https://github.com/usc-isi-i2/sand

[23] https://www.topquadrant.com/

[24] https://github.com/RDFLib/sparqlwrapper

Apache Jena[25] for Java and SPARQLWrapper[26] for python. Once the connection is established, queries may be posed directly to the knowledge graph with their output used to power the application. The architecture, illustrated in Figure 33,instantiates the general CDT architecture depicted in Figure 29.

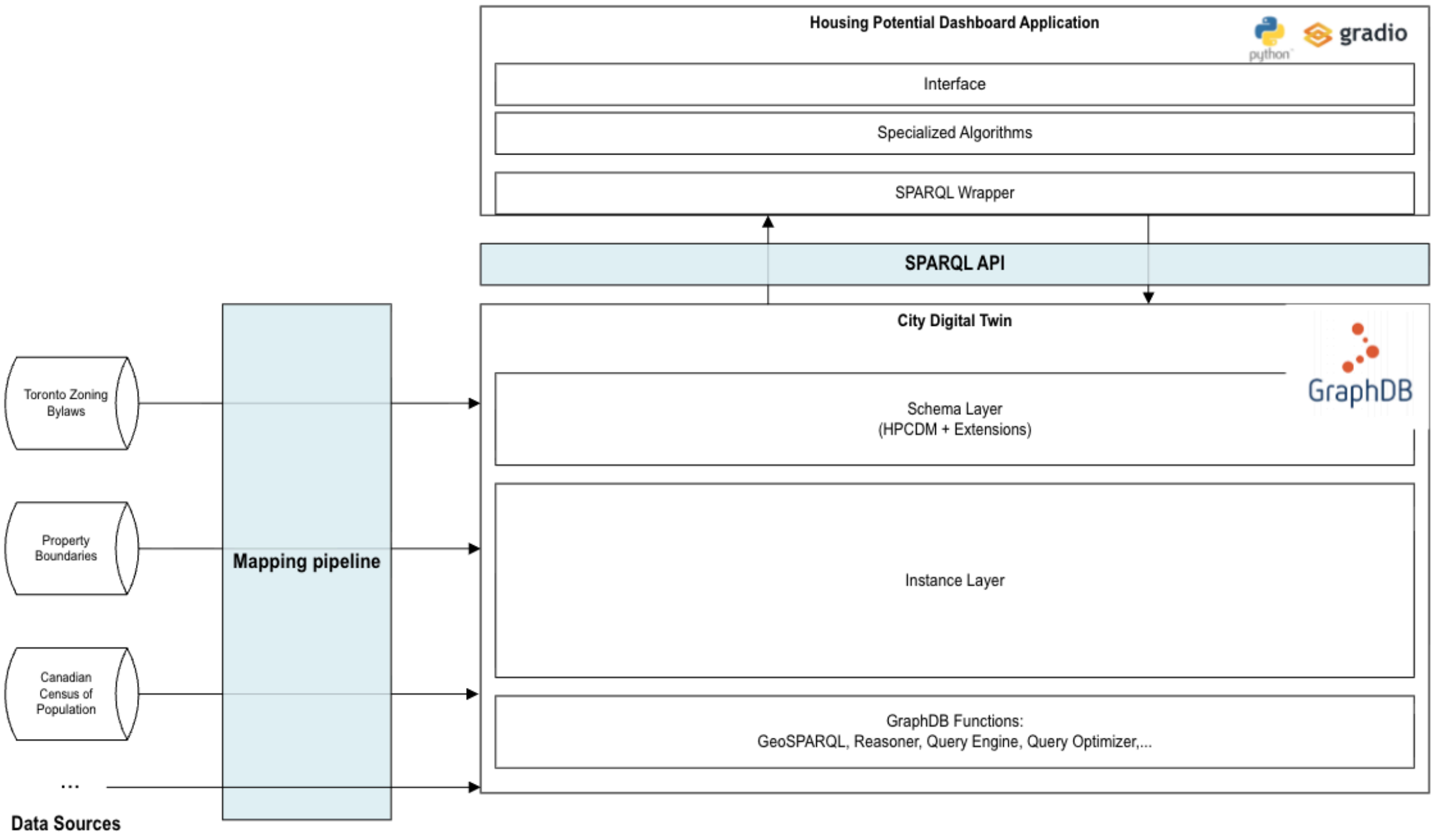

*Figure 33: Architecture of the housing potential dashboard application.*

### 5.2.1 The role of the knowledge graph and the HPCDM

The knowledge graph serves as the underlying source of data for the application (though of course, applications may also be designed to support data ingestion into the graph), with logic for user input and result processing developed on top.

Beyond providing a single point of access to integrated data for housing potential analysis – the knowledge graph and the HPCDM provide other advantages for application design. Recall that since the HPCDM is likely stored in the knowledge graph as well, the application can run queries as to the content of the model. This means that the interface does not have to be modified whenever new content is added to the graph. For example, in the housing potential dashboard, the service accessibility queries are run for each type of service currently defined in the HPCDM. If new service types are added in an extension to the model – these will automatically become part of the query (and display) logic without any changes to the application. The dropdown list of properties displayed for the “Zoning Compliance” option is pulled directly from the knowledge graph based upon zoning bylaw regulations that are currently captured in the graph, so if new data is added, the options will be updated automatically.

---

[25] https://jena.apache.org/

[26] https://github.com/RDFLib/sparqlwrapper

If the application has additional reasoning requirements, beyond what is supported by the knowledge graph, these will need to be addressed through the implementation of rules written in the language of the HPCDM. Depending on the nature of the rules, they may be implemented internally (i.e., with the addition of rules directly in the knowledge graph) or externally to the graph (i.e., in the application layer).

### 5.2.2 Defining application extensions of the HPCDM

The HPCDM provides a comprehensive foundation to represent and integrate the key types of data for housing potential however it is also designed for extension. Extensions to the HPCDM may be defined to capture different types of data (e.g. other types of services) as well as specializations of existing terms (e.g., the Toronto-specific subclasses described in Section 4.3 HPCDM Extensions from Phase 3). They may also be defined to support or streamline the queries require for some application(s).

### 5.2.3 Query template design

The design of query templates is the key step in developing an application based upon the knowledge graph. Query templates represent classes of queries that must be answered to support the application. Sometimes the queries are fully defined, but often they require some input parameters such that the query will be instantiated in different ways depending on the users' interaction. A simple example of this is the "parcel attributes" query option in the housing potential dashboard. There is a generic query structure (template) that runs a query for the attributes of a parcel – only the parcel id needs to be provided when the application runs, based on the user's parcel search input.

A practical element of query template design involves the recognition of query *patterns* – repeatable elements of a query can be reused to accomplish a common task. For example, in several queries there is a requirement to consider only the most specific ("leaf") classes in a taxonomy. The general approach to filter for these classes is reused as needed, improving the efficiency of the template design work.

Capturing and sharing these templates and patterns is a useful practice that also lowers the barrier to entry for new users and provides them with resources to implement the HPCDM effectively. The success of these initial projects then encourages wider awareness and adoption throughout the community. The provision of resources to support development is one of several mechanisms highlighted in the following section on encouraging adoption.

# 6 Considerations for Adoption

The primary goal of this project was the development of the HPCDM. For tangible benefits to be realized from its completion, adoption of the model must be promoted. Potential users of the HCPDM include municipalities and other government organizations, but also extend to stakeholders in organizations that consume, publish, or process housing potential-related data. Adoption may be promoted through several mechanisms:

- Education: promoting an understanding of the model and how it can be used
- Utility: providing HPCDM-based resources that offer immediate functionality for aligned datasets.

- Support: publication of resources for potential users. From technical guidance to mapping pipeline and query template libraries.

## 6.1 Education

To encourage adoption, stakeholders must understand the different ways in which the model may be implemented and the value it produces.

While Section 5 provided a description of a typical, comprehensive implementation of the HPCDM, adoption can occur in different ways and at different levels. The most basic level is publishing data to be HPCDM-compliant, i.e. to facilitate interoperability; beyond this, the HPCDM can be used to create an integrated knowledge graph (a housing potential CDT); with such a resource available there is the opportunity to develop applications on top of the housing potential CDT that leverage the common data model to create useful tools for the community.

Publishing data that is structured according to the HPCDM provides a foundation to enable data integration within an organization (e.g. a municipality) as well as across organizations (e.g., for national efforts) by ensuring that data can be correctly and uniformly understood. This requires an initial investment to understand, clean and reformat the data. The outcome of this investment is high-quality data that streamlines value creation by end-users. This level of adoption presents a low barrier as it does not require modification to existing systems or data schema. Instead, existing datasets can either be transformed and shared (e.g. on open data portals) or the mappings themselves may be published as complementary resources for the original datasets.

Adoption of the HPCDM improves the quality of the data that is published; however, it cannot resolve issues of errors and missing data. On the other hand, the design of mappings into the HPCDM can help to identify issues in data quality as it requires careful consideration of the meaning of the data elements. It can also help to identify gaps in available data: concepts in the model that have no corresponding datasets.

When data is integrated with the model, e.g. into a CDT, it provides the foundation upon which a wide range of applications may be built. It is important that users understand that the development of the CDT doesn’t correspond to a single application, it is an investment with continued benefit. This can be communicated with an overview of the range of CQs and knowledge categories that are supported by the model. It may also be more tangible to stakeholders through demonstrations of possible applications, such as the housing potential dashboard developed here. Creation and promotion of various demo applications is another mechanism that should be pursued to encourage adoption and implementation of the model.

## 6.2 Resources

In addition to educating stakeholders on the potential value of the HPCDM, another approach is to provide them with the value directly. Tools that provide new and valuable capabilities to users provide a strong, tangible motivation for adoption. An initial investment into the development of an HPCDM-based application would be a worthwhile means of initiating greater adoption, at which point additional opportunities (e.g., for the development of national/provincial-level tools) may emerge.

Note that an HPCDM-based application need not utilize the CDT architecture, though it is an option. Alternatives include specialized tools that HPCDM aligned data as input and produce some useful output or function. The gtfs[27] (global transit feed specification), originally developed by Google, is an example of this in practice. Municipalities are incentivized to publish their data according to the specification for it to be included in Google's applications, such as Google Maps' wayfinding functionality [45].

## 6.3 Support

Adoption of the model and implementation success should be encouraged through the provision of technical support. Toward this, this report includes implementation guidance for stakeholders. Barriers to adoption and implementation will be lowered with the creation of additional resources. A central location for documentation, best practices, and technical resources such as mapping pipeline libraries and query templates should be pursued to support stakeholders' use of the HPCDM.

# 7 Challenges, Risks and Limitations

Challenges for the HPCDM can be grouped into three categories:

- Technical (CDT implementation): Issues related to the software platforms and infrastructure used to implement the housing potential CDT.
- Administrative (HPCDM governance): Issues related to the governance, maintenance, and ongoing evolution of the model.
- Application (stakeholder use): Issues encountered by stakeholders when applying the model in specific use cases.

Each category is presented in the following, with a discussion of related risks and mitigation strategies.

## 7.1 Technical

Technical issues for a housing potential CDT can arise when there is a heavy reliance on (geospatial) reasoning. For large municipalities with many data points, this can cause performance difficulties for queries and maintenance (data updates or loading). Performance issues pose a risk to effective use of the CDT (and consequently the HPCDM) and are likely to deter potential users. One way to mitigate these issues is with the selection of alternate graph platforms – some software platforms specialize in geospatial reasoning (e.g., ArcGIS), whereas others implement on-demand reasoning. GraphDB, the platform selected for the Toronto CDT developed in this project implements forward-chaining reasoning. In other words, it materializes all inferences when the data is ingested. This greatly improves the performance of queries at runtime; however, it can cause issues for data loading and updates. As discussed in Section 5.1.1, the requirements for the CDT may serve to guide the tool selection and determine if a more appropriate option is available.

When a chosen software platform encounters efficiency issues, there are still many strategies that can be employed to address reasoning limitations. For example, in forward-chaining systems, rules can be defined to offload reasoning to data load time. The creation of "smoother" geometries may also be employed to simplify geospatial reasoning in cases where such approximations are acceptable. Where parallel options are supported, it is also often possible to break down demanding tasks into parts (e.g. batches of updates) to increase efficiency. Finally, it is also important to consider the nature of the required reasoning. In the case of the Toronto CDT, some queries employing geospatial reasoning are only necessary as a means of approximating service information. If parcel-level service data could be obtained, then this class of reasoning would no longer be required in the implementation.

[27] https://gtfs.org/

## 7.2 Administrative

Currently, lack of HPCDM governance presents a possible risk to the longevity of the model. This can be mitigated with the determination of model ownership. Roles and responsibilities must also be defined. Who determines if some extensions or revisions are required to the core model? Who publishes the changes and where?

## 7.3 Application

One challenge that may be encountered in developing applications based upon the HPCDM is the technical expertise required to develop data mappings, and (possibly) application-specific extensions to the model. If not addressed, this poses a risk as either a barrier to adoption, or a risk of misuse of the standard resulting in incorrectly defined datasets or poorly defined extensions. Any applications resulting from this work will likely be ineffective and resulting datasets won't integrate with other datasets as expected. These risks can be mitigated through the provision of resources such as training and workshop, along with the creation and fostering of an active support community.

Finally, the availability of data presents a major potential challenge for the development of housing potential applications using the HPCDM. For a given application, it's possible that not all the data required is available (in the CDT or in external sources). This poses a risk that stakeholders will find themselves unable to successfully utilize CDT to satisfy their requirements. – reducing the effectiveness/potential value provided by the HPCDM. To mitigate this, stakeholders may be encouraged to develop estimates for missing datasets where possible and appropriate. A clear separation between authoritative and estimated data is possible in the knowledge graph to maintain transparency in such cases. Separate graphs may be used to store estimates, or models of provenance could be incorporated into the representation. In cases where the data is not available and estimates are not appropriate, the HPCDM provides a clear framework with which to describe and make the case for the need to create or share the dataset to the appropriate organization.

# 8 Future Work

This project completed the design of a common data model that captures the knowledge required to assess housing potential. It takes various factors such as physical, regulatory, and social constraints into account and represents concepts in a generic, extensible way that can be applied across community contexts in Canada. Its design integrates existing city data standards, ensuring a that it shares a common, well-defined foundation with other models in the urban data landscape. The flexibility of the model has been demonstrated through the specification of example sets of mappings for a range of datasets from a range of communities in Canada. To guide and promote use of the model, its use to construct an integrated housing potential CDT has been illustrated for the City of Toronto, and an example of the possible functionality that can be supported have been demonstrated through the development of a housing potential dashboard.

At this stage, the focus should shift to promoting awareness and adoption of the HPCDM, as described in Section 6. Tool development is one area that can serve as a means of education while providing utility for adoption. One possible approach to achieving adoption is to focus on one common and important component of the HPCDM that municipalities vary greatly in representing, namely zoning. HICC should sponsor the creation of a tool that focuses specifically on zoning-based analyses. In order for municipalities to use this tool, they would have to map their zoning data into it. HICC could then sponsor educational forums, e.g., workshops, online help, etc. to develop expertise within municipalities on how to use HPCDM zoning.

In the context of the housing potential dashboard application, there are several directions that could be pursued in future iterations. Functionality of the interface may be extended, as described in Sections 4.2 and Appendix E. In addition, the same application logic may be implemented with a React-based toolset to provide a more polished interface. It would also be valuable for development to focus on an LLM question-answering interface to the CDT. While a dashboard provides a familiar, structured workflow, a Q&A system offers a more flexible and open-ended mechanism for users to interact directly with the integrated dataset. It also creates the potential to augment results with external data.

It is also important to address the administrative risk of a lack of governance. Ownership and maintenance of the HPCDM is essential to foster a community of users.

Finally, the HPCDM is not a static artifact; it is designed to evolve and expand its capacity to support Canadian communities over time. Future resources should be devoted to developing extensions to the model to address additional use cases. For example, housing market-related elements were omitted from the scope of the model in this project, but it remains an important subject for future consideration. Other topics include data provenance and trustworthiness, support for the creation of estimations, and the housing development pipeline. The engagement of stakeholders to promote the use of the model will provide an important opportunity to gather additional use cases and consider and prioritize how the model may be extended to capture other elements relevant for housing potential (and housing in general).

# References

[1] "Carsten Rönsdorf, Fabrice Servant, H.C. Gruler, Nick Giannias, Kyoungsook Kim, Zubran Soleiman, Dimitri Sarafinof," Open Geospatial Consortium, 2024.

[2] M. Katsumi, "Housing Potential Data Catalogue," Urban Data Research Centre, Toronto, 2025.

[3] International Standards Organization, "Information technology — Upper level ontology for smart city indicators".

[4] International Standards Organization, "Information technology — City data model Part 1: Foundation level concepts".

[5] International Standards Organization, "Information technology — City data model Part 2: City level concepts".

[6] M. Katsumi, "A Common Data Model for Housing Potential – Core Model," Urban Data Research Centre, Toronto, 2025.

[7] M. Katsumi, "A Common Data Model for Housing Potential," Urban Data Research Centre, Toronto, 2025.

[8] Standardization, International Organization for, "ISO/IEC 5087-1:2023 Information technology — City data model Part 1: Foundation level concepts," ISO, Geneva, 2023.

[9] Standardization, International Organization for, "ISO/IEC 5087-2:2024 Information technology — City data model Part 2: City level concepts," ISO, Geneva, 2024.

[10] Standardization, International Organization of, "ISO/IEC 21972:2020 Information technology — Upper level ontology for smart city indicators," ISO, Geneva, 2020.

[11] O. G. Consortium, "OGC City Geography Markup Language (CityGML) Part 1: Conceptual Model Standard," OGC, 2021.

[12] O. G. Consortium, "OGC® Land and Infrastructure Conceptual Model Standard (LandInfra)," OGC, 2016.

[13] 2.3.1, Temporary MIWP 2021-2024 sub-group, "D2.8.I.6 Data Specification on Cadastral Parcels – Technical Guidelines," INSPIRE Maintenance and Implementation Group (MIG), 2024.

[14] 2.3.1, Temporary MIWP 2021-2024 sub-group, "D2.8.I.7 Data Specification on Transport Networks – Technical Guidelines," INSPIRE Maintenance and Implementation Group (MIG), 2024.

[15] 2.3.1, Temporary MIWP 2021-2024 sub-group, "D2.8.III.2 Data Specification on Buildings – Technical Guidelines," INSPIRE Maintenance and Implementation Group (MIG), 2024.

[16] 2.3.1, Temporary MIWP 2021-2024 sub-group, "D2.8.III.4 Data Specification on Land Use – Technical Guidelines," INSPIRE Maintenance and Implementation Group (MIG), 2024.

[17] Temporary MIWP 2021-2024 sub-group 2.3.1, "D2.8.III.6 Data Specification on Utility and Government Services Technical Guidelines," INSPIRE Maintenance and Implementation Group (MIG), 2024.

[18] R. L. M. P. M. &. D. L. García‐Castro, "The ETSI SAREF ontology for smart applications: a long path of development and evolution.," in *Energy Smart Appliances: Applications, Methodologies, and Challenges*, 2023 , pp. 183-215.

[19] Standardization, International Organization for, "ISO 19152-1:2024 Geographic information — Land Administration Domain Model (LADM) Part 1: Generic conceptual model," ISO, Geneva, 2024.

[20] Standardization, International Organization for, "ISO 19152-2:2024 – Geographic information – Land Administration Domain Model – Part 2: Land administration services," ISO, Geneva, 2024.

[21] Standardization, International Organization for, "ISO 19152-4:2024 – Geographic information – Land Administration Domain Model – Part 4: Valuation information," ISO, 2024, Geneva.

[22] International Organization for Standardization, "ISO 19152-5:2024 – Geographic information – Land Administration Domain Model – Part 5: Spatial units," ISO, Geneva.

[23] International Federation of Surveyors (FIG), "The Land Administration Domain Model An Overview," FIG.

[24] International Organization for Standardization, "ISO 37105:2019 – Descriptive framework for cities and communities," ISO, Geneva, 2019.

[25] City Protocol Society, "CPA-PR_003: City Anatomy Ontology," City Protocol Society, 2015.

[26] P. B. M. B. R. N. P. &. R. N. Bellini, "Km4City ontology building vs data harvesting and cleaning for smart-city services," *Journal of Visual Languages & Computing,* vol. 25, no. 6, pp. 827-839, 2014.

[27] H. Silvennoinen, A. Chadzynski, F. Farazi, A. Grišiūtė, Z. Shi, A. von Richthofen, S. Cairns, M. Kraft, M. Raubal and P. Herthogs, "A semantic web approach to land use regulations in urban planning: The OntoZoning ontology of zones, land uses and

programmes for Singapore," *Journal of Urban Management,* vol. 12, no. 2, pp. 151-167, 2023.

[28] M. R. P. H. Ayda Grisiute, "3D Land Use Planning: Making Future Cities Measurable with Ontology-Driven Representations of Planning Regulations," *AGILE: GIScience Series,* vol. 6, no. 3, 2025.

[29] M. S. Fox, "An Education Ontology for Global City Indicators (ISO 37120)," Enterprise Integration Laboratory, University of Toronto, Toronto, 2014.

[30] K. a. F. M. Komisar, "An Energy Ontology for Global City Indicators (ISO 37120)"," Enterprise Integration Lab, University of Toronto, 2017.

[31] N. a. F. M. Rauch, "A Fire and Emergency Ontology for Global City Indicators (ISO 37120)," Enterprise Integration Lab, University of Toronto, 2017.

[32] J. a. F. M. Falodi, "A Health Ontology for Global City Indicators (ISO 37120)," Enterprise Integration Lab, University of Toronto, 2018.

[33] K. a. F. M. Khazai, "A Public Safety Ontology for Global City Indicators (ISO 37120)," Enterprise Integration Laboratory, University of Toronto, 2017.

[34] T. a. F. M. Abdulai, "Recreation Ontology for Global City Indicators (ISO 37120)," Enterprise Integration Laboratory, University of Toronto, 2017.

[35] Y. a. F. M. Wang, "A Shelter Ontology for Global City Indicators (ISO 37120)," Enterprise Integration Laboratory, University of Toronto, 2015.

[36] X. a. F. M. Sun, "Solidwaste Ontology for Global City Indicators (ISO 37120)," Enterprise Integration Laboratory, University of Toronto, 2020.

[37] W. a. F. M. Yousif, "A Transportation Ontology for Global City Indicators (ISO 37120)," Enterprise Integration Laboratory, University of Toronto, 2018.

[38] R. a. F. M. Navalta, "A Water and Sanitation Ontology for Global City Indicators (ISO 37120)," Enterprise Integration Laboratory, University of Toronto, 2019.

[39] A. Wong, M. Fox and M. Katsumi, "Semantically Interoperable Census Data: Unlocking the Semantics of Census Data Using Ontologies and Linked Data," *International Journal of Population Data Science,* vol. 9, no. 1, 2024.

[40] N. a. F. M. Rauch, "A Fire and Emergency Ontology for Global City Indicators (ISO 37120), Working Paper," [Online]. Available: http://eil.mie.utoronto.ca/wp-content/uploads/2015/06/Fire-and-Emergency-response-V10.pdf. [Accessed 2025].

[41] T. a. F. M. Abdulai, "Recreation Ontology for Global City Indicators (ISO 37120), Working Paper," Enterprise Integration Laboratory, University of Toronto, Toronto, 2017.

[42] J. F. M. a. R. D. Stroock, "Hazards, Impacts, and Risk in Urban Climate Resilience Planning: A Taxonomic Approach," 2025.

[43] "OWL 2 Web Ontology Language Profiles (Second Edition)," W3C Recommendation, 2012.

[44] "OGC GeoSPARQL - A Geographic Query Language for RDF Data," Open Geospatial Consortium, 2024.

[45] B. McHugh, "Pioneering open data standards: The GTFS story," in *Beyond transparency: open data and the future of civic innovation*, 2013, pp. 125-135.

## Appendix A Supplementary Examples

### ISO/IEC 5087-2 Code Pattern Example

The Code Pattern, reused from ISO/IEC 5087-2, provides a general approach to formalizing enumerated sets that supports integration and extension with multiple classification systems. It defines the Code class, as specified in Table 40, and introduces the hasCode property that can be used to associate an object with a Code. Codes allow for the specification of, not only the code (identifier) for the value, but a description of its meaning along with other metadata on the classification system such as the source (e.g., definedBy the Government of Canada, or some other organization), and the location of its definition (specification). Use of the Code Pattern to represent residential land use is illustrated in Figure 34. In the example, the residential land use may be described with a code from a Toronto-based system (with values for the other properties not pictured) or from the Land Based Classification Standards (LBCS) system.

*Table 40: Code class definition*

| Class | Property | Value Restriction |
|---|---|---|
| Code | definedBy | max 1 5087-1:Organization |
| | specification | only xsd:string |
| | 5087-1:hasIdentifier | max 1 xsd:string |
| | 5087-1:hasName | only xsd:string |
| | 5087-1:hasDescription | only xsd:string |

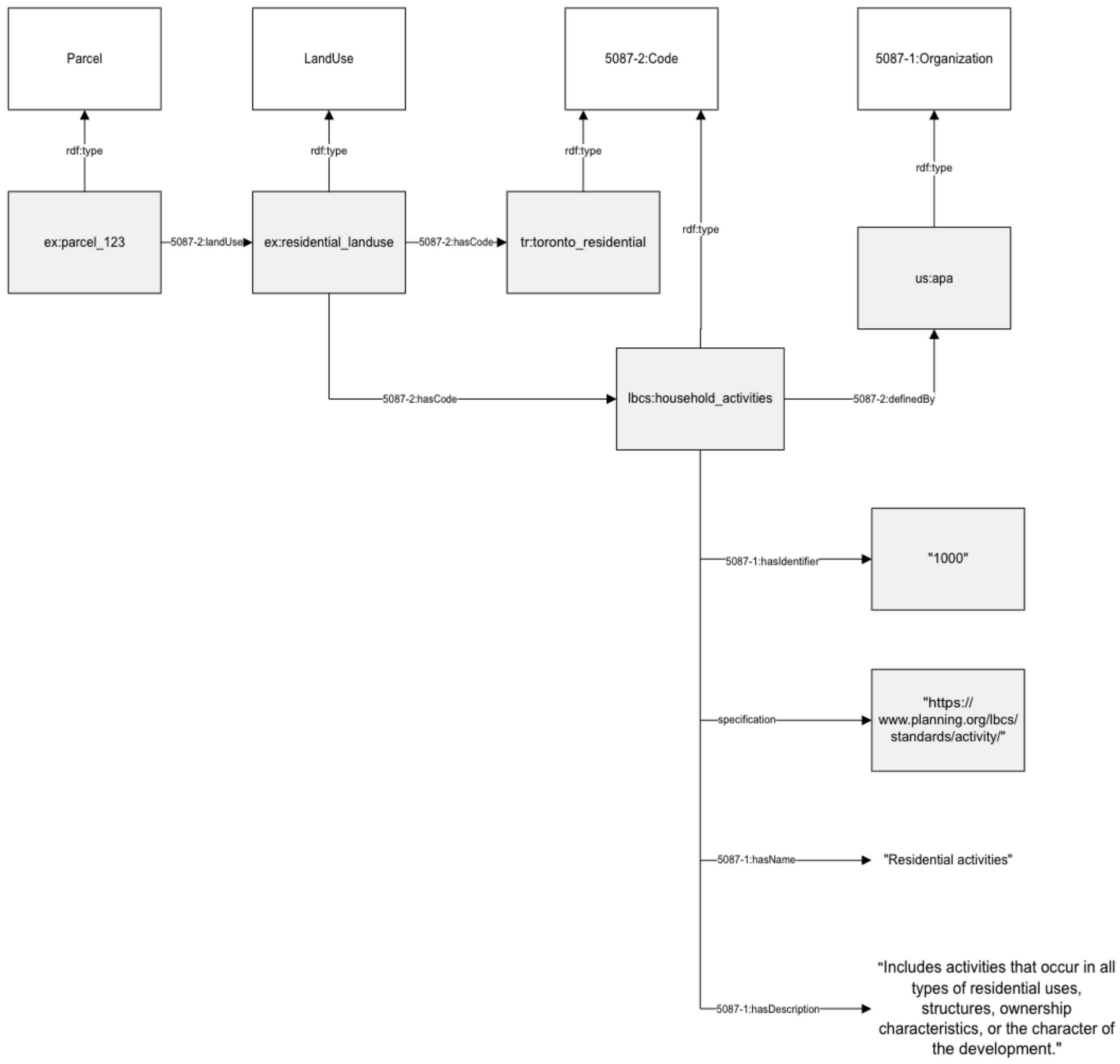

*Figure 34: Example use of Code class to integrate different classification systems defining the same type of land use.*

## ISO/IEC 21972 Population Example

The ability to refer to a population is important to specify the definition of many types of indicators – for example, the "average household income in a neighbourhood" refers to a parameter (average) of a particular population: not persons, but households in a neighbourhood. The approach taken by ISO/IEC 21972 supports a connection to the classes needed to define the conditions for membership in a population. The Population class is not restricted to residents of a particular area but extend to refer to any group of *things* being studied. A Population is *defined by* some this set of things, the membership condition for the population is captured with the definition of the associated class (in this case, "Household"). The definition of the 21972:Population class, extended with the definition of a HouseholdPopulation is outlined in Table 41.

*Table 41: Household population definition*

| Class | Property | Value Restriction |
|---|---|---|
| 21972:Population | 21972:defined_by | only owl:Thing |
| | 21972:located_in | only geo:Feature |
| | 21972:for_time_interval | only time:DateTimeInterval |
| HouseholdPopulation | rdfs:subClassOf | 21972:Population |
| | 21972:defined_by | Household |

## Extension Examples

The following extensions are drawn from the Toronto Zone Category Mapping outlined in Appendix B.1. Examples of their definitions (as extensions to the HPCDM) are specified below.

- tor:Frontage: a length quantity that identifies the frontage of the Lot in Toronto. While there may not be any additional axioms added for the subclass, in some cases it is important to distinguish the quantity when its interpretation varies between municipalities and applications.

| Class | Property | Value Restriction |
|---|---|---|
| tor:LotFrontage | rdfs:subClassOf | Frontage |
| tor:Lot | rdfs:subClassOf | Lot |
| | hasFrontage | only tor:LotFrontage |

- tor:hasNumDwellings: a Quantity that represents the cardinality of the dwelling population in a particular area. Since characteristic represents the count of a population, it is defined more precisely with the introduction of the population it is a count of (i.e., “tor:DwellingUnitPopulationInALot”).

| Class | Property | Value Restriction |
|---|---|---|
| tor:Lot | rdfs:subClassOf | Lot |
| | tor:hasNumDwellings | only tor:NumDwellings |
| | … | … |
| tor:NumDwellings | rdfs:subClassOf | 21972:Cardinality |
| | 21972:cardinality_of | tor:DwellingUnitPopulationInALot |
| tor:DwellingUnitPopulationInALot | rdfs:subClassOf | 21972:Population |
| | 21972:located_in | exactly 1 Lot |
| | 21972:defined_by | DwellingUnit |

# Appendix B Data Mapping Specifications

Prefixes are defined as follows:

- bdg: <https://standards.iso.org/iso-iec/5087/-2/ed-1/en/ontology/Building/>
- cacensus: <http://ontology.eil.utoronto.ca/tove/cacensus#>
- code: <https://standards.iso.org/iso-iec/5087/-2/ed-1/en/ontology/Code/>
- genprop: <https://standards.iso.org/iso-iec/5087/-1/ed-1/en/ontology/GenericProperties/>
- geo: <http://www.opengis.net/ont/geosparql#>
- geoext: <http://rdf.useekm.com/ext#>
- geof: <http://www.opengis.net/def/function/geosparql/>
- hp: <http://ontology.eil.utoronto.ca/HPCDM/>
- i72: <http://ontology.eil.utoronto.ca/ISO21972/iso21972#>
- loc_old: <http://ontology.eil.utoronto.ca/5087/1/SpatialLoc/>
- loc: <https://standards.iso.org/iso-iec/5087/-1/ed-1/en/ontology/SpatialLoc/>
- opr: <http://www.theworldavatar.com/ontology/ontoplanningregulation/OntoPlanningRegulation.owl#>
- owl: <http://www.w3.org/2002/07/owl#>
- oz: <http://www.theworldavatar.com/ontology/ontozoning/OntoZoning.owl#>
- rdf: <http://www.w3.org/1999/02/22-rdf-syntax-ns#>
- rdfs: <http://www.w3.org/2000/01/rdf-schema#>
- res: <https://standards.iso.org/iso-iec/5087/-1/ed-1/en/ontology/Resource/>
- service: <https://standards.iso.org/iso-iec/5087/-2/ed-1/en/ontology/CityService/>
- time: <http://www.w3.org/2006/time#>
- tor: <http://ontology.eil.utoronto.ca/Toronto/Toronto#>
- uom: <http://www.opengis.net/def/uom/OGC/1.0/>

Note on time indexing of services

- Generally omitted except in cases where start/end times of datasets are known/provided. Most current use cases don't address the question of past/present/future values, so this isn't generally captured in the data/queries however in practice it is something that would be needed to capture updates over time.

Note on test (artificial) data

- Mappings are highlighted in pink
- Mappings are stored in separate named graphs to support complete testing while maintaining separation between real and synthetic/estimated data

The mapping specifications and implementation may also be found in the City Digital Twin repository[28].

[28] https://github.com/csse-uoft/city-digital-twin-ontology/tree/main/Housing%20Potential%20Python

## B.1 Zoning and Land Use Data Mappings

This section presents dataset mappings for Zoning and Land Use data from Toronto, Vancouver, and Halifax. The mappings will be extended to include other datasets in the next phase of the project when the HPCDM is extended to address the remaining requirements. The mappings are specified with a table overview that identifies the fields in the dataset and corresponding statements (triples) constructed with HPCDM classes and properties to capture the semantics of the value. The mappings also introduce example namespaces as required to define instance data (e.g., to introduce a URI for an individual parcel of land) or class extensions. These namespaces are introduced for demonstration purposes only and are not intended to be prescriptive in the context of the HPCDM. Supplementary diagrams are included to complement some of the table specifications.

### B.1.1 Toronto Zoning Category

https://data.urbandatacentre.ca/catalogue/city-toronto-zoning-by-law

This dataset defines zones with different land use regulations, along with specialized limitations on lot dimensions such as frontage and land area, and development density. Mappings from the dataset's fields are grouped into the following categories: (1) definition of bylaw references, (2) identification of applicable zoning types, and (3) definition of the zoning type.

The fields identified in Table 42 capture different parts of the bylaw that are referenced by the dataset. In particular, the section and chapter identify the relevant parts for the identified zoning type. These references are useful for referencing the relevant parts of a document for a restriction.

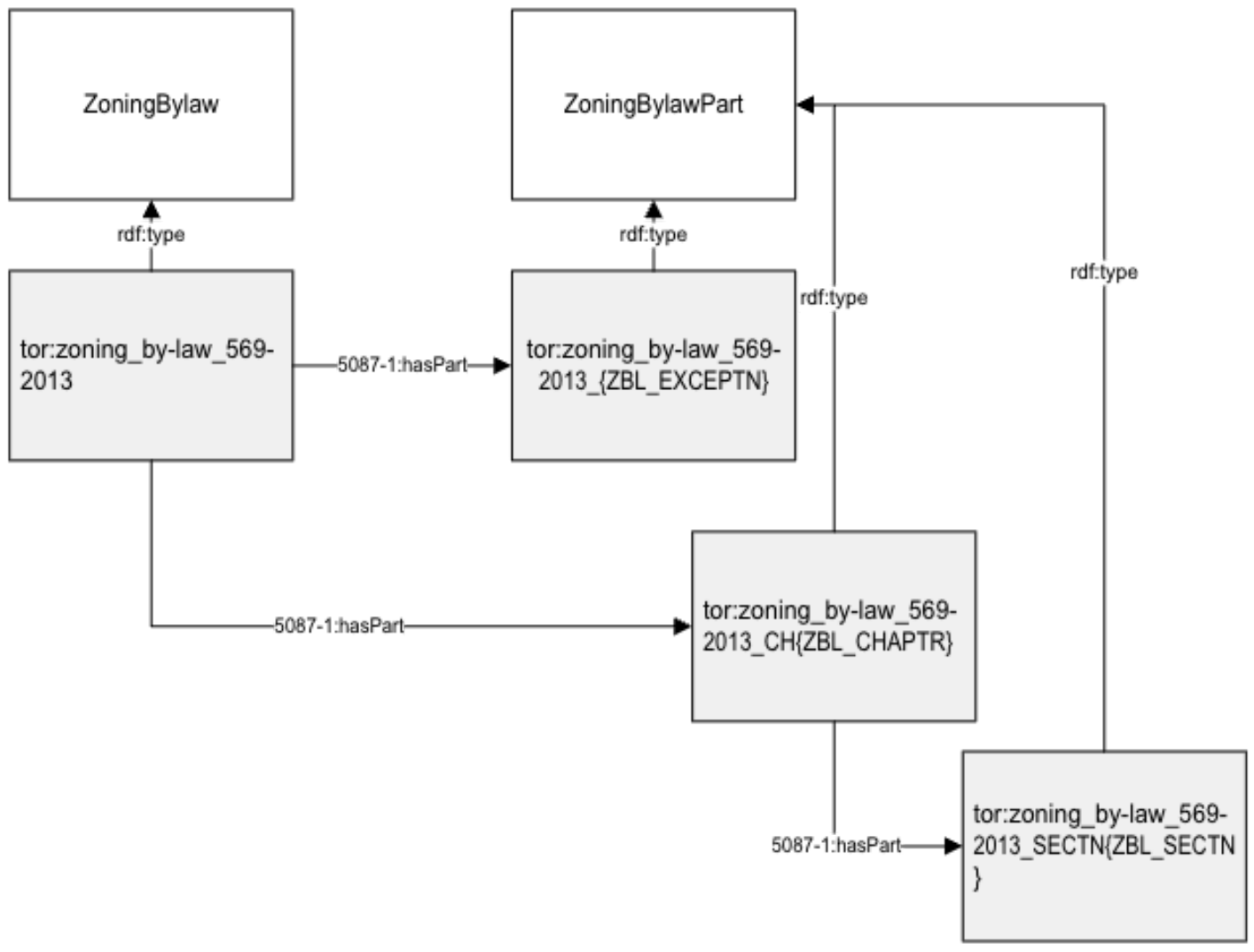

*Figure 35: Diagram of bylaw reference mapping result*

*Table 42: Mapping bylaw references from Toronto "Zone Categories" data*

| "Zone Categories" Field | Mapping to HPCDM | | |
|---|---|---|---|
| | **Subject** | **Property** | **Object** |
| ZONING_BY-LAW_569-2013 | tor:zoning_by-law_569-2013 | rdf:type | hp:ZoningBylaw |
| | | bylaw:legislationIdentifier | "ZONING_BY-LAW_569-2013" |
| ZBL_EXCPTN = (By-law text section number) | tor:zoning_by-law_569-2013 | mer:hasProperPart | tor:zoning_by-law_569-2013_{ZBL_EXCEPTN} |
| | tor:zoning_by-law_569-2013_{ZBL_EXCEPTN} | rdf:type | hp:ZoiningBylawPart |
| | tor:zoning_by-law_569-2013_{ZBL_EXCEPTN} | genprop:hasIdentifier | {ZBL_EXCPTN} |
| ZBL_CHAPTR = (By-law text chapter number) | tor:zoning_by-law_569-2013 | mer:hasProperPart | tor:zoning_by-law_569-2013_CH{ZBL_CHAPTR} |
| | tor:zoning_by-law_569-2013_CH{ZBL_CHAPTR} | rdf:type | hp:ZoiningBylawPart |
| | tor:zoning_by-law_569-2013_CH{ZBL_CHAPTR} | genprop:hasIdentifier | {ZBL_CHAPTR} |
| ZBL_SECTN = (By-law text section number) | tor:zoning_by-law_569-2013_CH{ZBL_CHAPTR} | mer:hasProperPart | tor:zoning_by-law_569-2013_SECTN{ZBL_SECTN} |
| | tor:zoning_by-law_569-2013_SECTN{ZBL_SECTN} | rdf:type | hp:ZoiningBylawPart |
| | tor:zoning_by-law_569-2013_SECTN{ZBL_SECTN} | genprop:hasIdentifier | {ZBL_SECTN} |

Each row in the table associates a zoning type with a particular area. This mapping is outlined in Table 43. Note that each of GEN_ZONE, ZN_ZONE, and ZN_STRING are considered zoning types (at different levels of abstraction) that apply to the same area. A hold (when applied) is interpreted as a different zoning type.

This allows for additional specification on the nature of the hold (e.g., what developments may be permitted/prohibited). On the other hand, exceptions are identified with respect to the ZN_STRING zoning type, as they contribute to its interpretation.

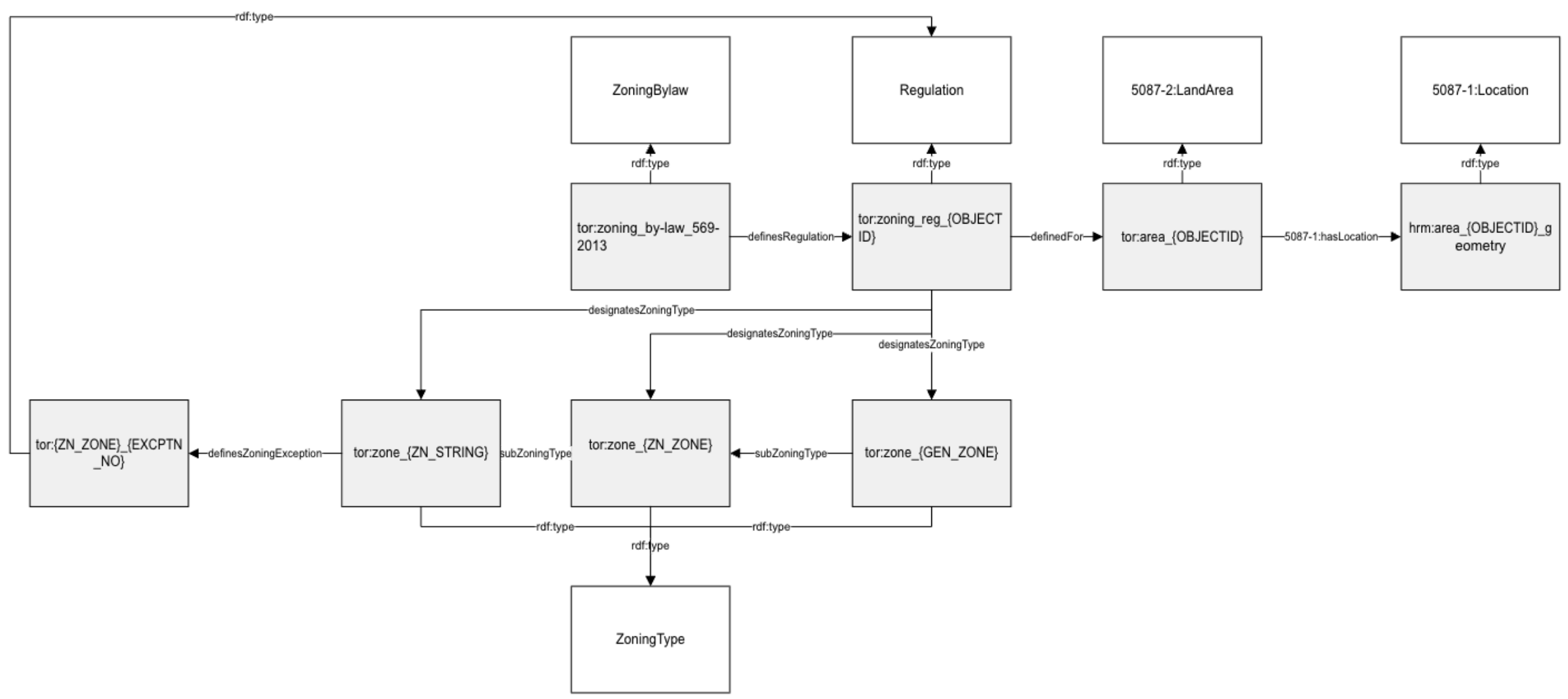

*Figure 36: Diagram of zoning type assignment mapping result*

*Table 43: Mapping zoning type assignments in Toronto*

| "Zone Categories" Field | Mapping to HPCDM | | |
|---|---|---|---|
| | **Subject** | **Property** | **Object** |
| OBJECTID = (Unique system identifier) | tor:zoning_by-law_569-2013 | hp:definesRegulation | tor:zoning_reg_{OBJECTID} |
| | tor:zoning_reg_{OBJECTID} | rdf:type | hp:Regulation |
| | tor:zoning_reg_{OBJECTID} | hp:definedIn | tor:zoning_by-law_569-2013 |
| | tor:zoning_reg_{OBJECTID} | hp:definedFor | tor:area_{OBJECTID} |
| | tor:area_{OBJECTID} | rdf:type | hp:AdministrativeArea |
| | tor:area_{OBJECTID} | loc:hasLocation | tor:area_{OBJECTID}_geometry |
| geometry | tor:area_{OBJECTID}_geometry | geo:asWKT | {geometry} |
| GEN_ZONE = (The land use category of the lands within the | tor:zoning_reg_{OBJECTID} | hp:designatesZoningType | tor:zone_{GEN_ZONE} |
| zone boundary. Each "zone category" has its own Chapter in the text of By-law 569-2013.) | | | |
| | tor:zone_{GEN_ZONE} | hp:subZoningType | tor:zone_{ZN_ZONE} |
| ZN_ZONE = (The land use category of the lands within the zone boundary. Each "zone category" has its own Chapter in the text of By-law 569-2013.) [Zoned destination of the zone limited by GEN_ZONE. | tor:zoning_reg_{OBJECTID} | hp:designatesZoningType | tor:zone_{ZN_ZONE} |
| | tor:zone_{ZN_ZONE} | hp:subZoningType | tor:zone_{ZN_STRING} |
| | tor:zone_{ZN_ZONE} | hp:definedIn | tor:zoning_by-law_569-2013_SECTN{ZBL_SECTN} |
| ZN_STRING = (Complete label of the zone.) | tor:zoning_reg_{OBJECTID} | hp:designatesZoningType | tor:zone_{ZN_STRING} |
| ZN_EXCPTN = (This indicates whether a zone has an Exception.) [Yes (Y) or No (N)] | | | |
| EXCPTN_NO = (This is the Exception Number for the zone if one exists. The exception number is prefaced by the letter "x" in the zone label. Each zone has its own series of exception numbers, starting at 1, so the exception number | tor:zone_{ZN_STRING} | hp:definesZoningException | tor:{ZN_ZONE}_{EXCPTN_NO} |
| must be read in conjunction with the respective zone symbol.) | | | |
| ZBL_EXCPTN = (By-law text section number) | tor:{ZN_ZONE}_{EXCPTN_NO} | hp:definedIn | tor:zoning_by-law_569-2013_{ZBL_EXCEPTN} |
| ZN_HOLDING = (To indicate whether there is a HOLDING status for the zone. The zone label will be prefaced by the letter (H). These are not common, and when used, a Holding Zone is most often applied to specific sites.) [Yes (Y) or No (N)] | tor:holding_reg_{OBJECTID} | rdf:type | hp:Regulation |
| | tor:holding_reg_{OBJECTID} | hp:definedFor | tor:area_{OBJECTID} |
| | tor:holding_reg_{OBJECTID} | hp:designatesZone | tor:holding_zone |
| HOLDING_ID (Holding Number if it exists.) | tor:holding_reg_{OBJECTID} | genprop:hasIdentifier | {HOLDING_ID} |

The zoning type is defined with the instantiation of any applicable regulations. Here, we outline example mappings for frontage, unit, and density regulations defined for a zoning type. Table 44 specifies the mapping to capture any frontage regulations for the zoning type. A new Regulation is introduced that applies to the zoning, the regulation specifies a Quantity Constraint (requirement) on the minimum frontage (hasFrontage Variable) within a population of lots. The constraint itself is specified with the associated “FRONTAGE” value provided in the data. The mapping encodes the intended unit of measure (metres), along with the population that the minimum describes. In this case, it is the population of Lots in Toronto, with the specified zoning type. In practice, the required classes could be identified and defined in a separate extension to the HPCDM, such that they could be reused in any mapping implementation.

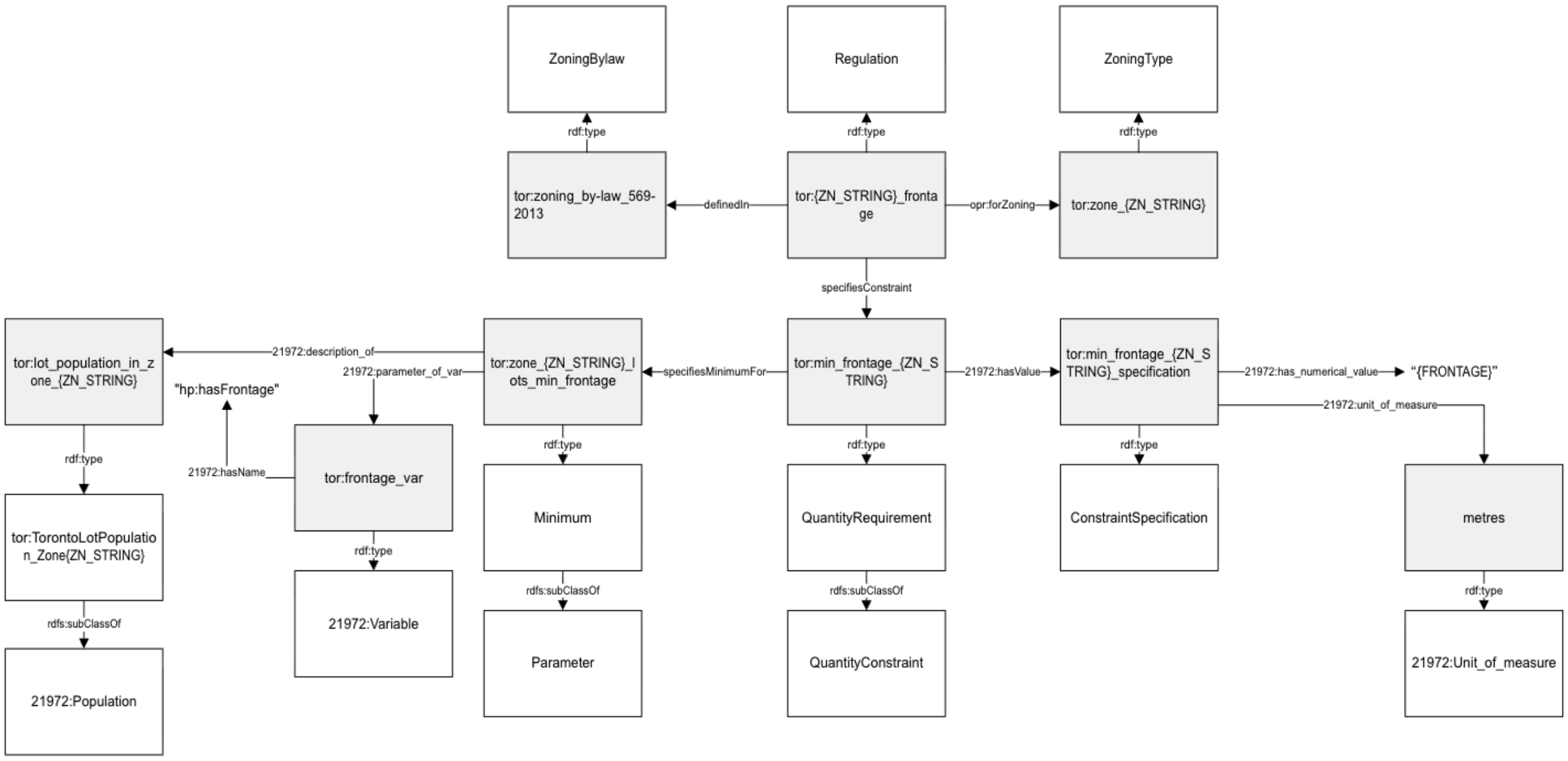

*Figure 37: Diagram of Frontage restriction mapping result*

*Table 44: Mapping the defined Frontage restriction for lots in the zone in Toronto*

| "Zone Categories" fields | Mapping to HPCDM | | |
|---|---|---|---|
| | **Subject** | **Property** | **Object** |
| FRONTAGE = (The required minimum Lot Frontage in the zone, and is a numeric value prefaced by the letter "f" within a residential zone label.) [Unit = metres.] | tor:{ZN_STRING}_regulation_constraints | rdf:type | hp:Regulation |
| | tor:{ZN_STRING}_regulation_constraints | hp:definedIn | tor:zoning_by-law_569-2013 |
| | tor:{ZN_STRING}_regulation_constraints | opr:forZoning | tor:zone_{ZN_STRING} |
| | tor:{ZN_STRING}_regulation_constraints | hp:specifiesConstraint | tor:min_frontage_{ZN_STRING} |
| | tor:min_frontage_{ZN_STRING} | rdf:type | hp:QuantityRequirement |
| | tor:min_frontage_{ZN_STRING} | i72:hasValue | tor:min_frontage_{ZN_STRING}_specification |
| | tor:min_frontage_{ZN_STRING}_specification | i72:hasNumericalValue | {FRONTAGE} |
| | tor:min_frontage_{ZN_STRING}_specification | i72:hasUnit | hp:metres |
| | tor:min_frontage_{ZN_STRING} | hp:specifiesMinimumFor | tor:zone_{ZN_STRING}_lots_min_frontage |
| | tor:zone_{ZN_STRING}_lots_min_frontage | rdf:type | hp:Minimum |
| | tor:zone_{ZN_STRING}_lots_min_frontage | i72:parameter_of_var | tor:frontage_var |
| | tor:zone_{ZN_STRING}_lots_min_frontage | i72:minimum_of | tor:lot_population_in_zone_{ZN_STRING} |
| | tor:frontage_var | i72:hasName | "hp:hasFrontage" |
| | tor:lot_population_in_zone_{ZN_STRING} | rdf:type | tor:TorontoLotPopulation |

Table 45 specifies the mapping to formalize the maximum number of dwelling units per lot for a particular zone. It is defined as part of the same Regulation that applies to the identified zoning type. The allowance is specified as a constraint on the maximum number of dwelling units per lot within the zone. The mapping involves an extension of the HPCDM core to capture the property "hasNumDwellings" for Toronto. This property defines the number of dwellings (a Cardinality of the Dwelling population) in a given lot.

*Table 45: Mapping the maximum number of units per lot in the zone in Toronto*

| **"Zone Categories" fields** | **Mapping to HPCDM** | | |
|---|---|---|---|
| | **Subject** | **Property** | **Object** |
| UNITS = (The permitted maximum number of Dwelling Units allowed on a lot in the zone, and is a numeric value prefaced by the letter "u" in a residential zone.) | tor:{ZN_STRING}_regulation_constraints | rdf:type | hp:Regulation |
| | tor:{ZN_STRING}_regulation_constraints | hp:definedIn | tor:zoning_by-law_569-2013 |
| | tor:{ZN_STRING}_regulation_constraints | opr:forZoning | tor:zone_{ZN_STRING} |
| | tor:{ZN_STRING}_regulation_constraints | hp:specifiesConstraint | tor:max_units_{ZN_STRING} |
| | tor:max_units_{ZN_STRING} | rdf:type | hp:QuantityAllowance |
| | tor:max_units_{ZN_STRING} | i72:hasValue | tor:max_units_{ZN_STRING}_specification |
| | tor:max_units_{ZN_STRING}_specification | i72:hasNumericalValue | {UNITS} |
| | tor:max_units_{ZN_STRING} | hp:specifiesMaximumFor | tor:zone_{ZN_STRING}_lots_max_dwelling |
| | tor:zone_{ZN_STRING}_lots_max_dwelling | rdf:type | hp:Maximum |
| | tor:zone_{ZN_STRING}_lots_max_dwelling | i72:parameter_of_var | tor:num_dwellings_var |
| | tor:zone_{ZN_STRING}_lots_max_dwelling | i72:maximum_of | tor:lot_population_in_zone_{ZN_STRING} |
| | tor:num_dwellings_var | i72:hasName | "tor:hasNumDwellings" |
| | tor:lot_population_in_zone_{ZN_STRING} | rdf:type | tor:TorontoLotPopulation |

Table 46 specifies the mapping to represent a constraint on density in the zone. This constraint specifies a limit on floor space index (FSI) and so is defined for the maximum "hasFSI" variable for the population of lots in the zone.

*Table 46: Mapping of density regulation in a zone in Toronto*

| **"Zone Categories" fields** | **Mapping to HPCDM** | | |
|---|---|---|---|
| | **Subject** | **Property** | **Object** |
| DENSITY = (The permitted maximum Density in the zone by FSI (floor space index), and is a numeric value prefaced by the letter "d" in residential zones.) | tor:{ZN_STRING}_regulation_constraints | rdf:type | hp:Regulation |
| | tor:{ZN_STRING}_regulation_constraints | hp:definedIn | tor:zoning_by-law_569-2013 |
| | tor:{ZN_STRING}_regulation_constraints | opr:forZoning | tor:zone_{ZN_STRING} |
| | tor:{ZN_STRING}_regulation_constraints | hp:specifiesConstraint | tor:max_density_{ZN_STRING} |
| | tor:max_density_{ZN_STRING} | rdf:type | hp:QuantityAllowance |
| | tor:max_density_{ZN_STRING} | i72:hasValue | tor:max_density_{ZN_STRING}_specification |
| | tor:max_density_{ZN_STRING}_specification | i72:hasNumericalValue | {DENSITY} |
| | tor:max_density_{ZN_STRING}_specification | i72:hasUnit | i72:Unit_division |
| | tor:max_density_{ZN_STRING} | hp:specifiesMaximumFor | tor:zone_{ZN_STRING}_lots_max_density |
| | tor:zone_{ZN_STRING}_lots_max_density | rdf:type | hp:Maximum |
| | tor:zone_{ZN_STRING}_lots_max_density | i72:parameter_of_var | tor:density_var |
| | tor:zone_{ZN_STRING}_lots_max_density | i72:maximum_of | tor:lot_population_in_zone_{ZN_STRING} |
| | tor:density_var | i72:hasName | "hp:hasFSI" |
| | tor:lot_population_in_zone_{ZN_STRING} | rdf:type | tor:TorontoLotPopulation_Zone{ZN_STRING} |
| | tor:TorontoLotPopulation_Zone{ZN_STRING} | rdfs:subClassOf | tor:TorontoLotPopulation |

| | |
|---|---|
| **Data:** | Zoning Height Overlay<br>https://open.toronto.ca/dataset/zoning-by-law/ |
| **Notes** | The same general approach can be defined for any area-based regulations defined independently of zoning types (not incorporated in the zoning area layer) |

| **Subject** | **Property** | **Object** | **Notes** |
|---|---|---|---|
| _id | tor:height_zone{_id} | rdf:type | hp:Regulation |
| | tor:zoning_by-law_569-2013 | hp:definesRegulation | tor:height_zone{_id} |
| geometry | tor:height_zone{_id} | hp:definedFor | tor:height_zone{_id}Area |
| | tor:height_zone{_id}Area | rdf:type | hp:AdministrativeArea |
| | tor:height_zone{_id}Area | loc:hasLocation | tor:height_zone{_id}AreaLoc |
| | tor:height_zone{_id}AreaLoc | geo:asWKT | {geometry} |
| | tor:height_zone{_id} | hp:specifiesConstraint | tor:height_zone{_id}HeightConstraint |
| | tor:height_zone{_id}HeightConstraint | rdf:type | hp:QuantityAllowance |
| | tor:height_zone{_id}HeightConstraint | i72:hasValue | tor:height_zone{_id}HeightConstraintValue |
| HT_LABEL | tor:height_zone{_id}HeightConstraintValue | i72:hasNumericalValue | {HT_LABEL} |
| | tor:height_zone{_id}HeightConstraintValue | i72:hasUnit | i72:metre |
| | tor:height_zone{_id}HeightConstraint | hp:specifiesMaximumFor | tor:height_zone{_id}MaxHeight |
| | tor:height_zone{_id}MaxHeight | rdf:type | hp:Maximum |
| | tor:height_zone{_id}MaxHeight | hp:maximumOf | tor:buildingPopulationHeightZone{_id} |
| | tor:buildingPopulationHeightZone{_id} | rdf:type | tor:BuildingPopulation |
| | tor:height_zone{_id}MaxHeight | i72:parameter_of_var | tor:height_zone{_id}BuildingHeight |
| | tor:height_zone{_id}BuildingHeight | i72:hasName | "hp:hasBuildngHeight" |
| | tor:BuildingPopulation | rdfs:subClassOf | i72:Population |

**Data:** zoning type land use (approximation) zoning_landuse_test.csv

**Notes** synthetic data, approximated (via chatGPT)

| Field | Subject | Property | Object |
|---|---|---|---|
| Zone Symbol | tor:zone_{Zone Symbol} | rdf:type | hp:ZoningType |
| Zone Category | tor:zone_{Zone Symbol} | genprop:hasName | {Zone Category} |
| Allowed Use | tor:zone_{Zone Symbol} | oz:allowsUse | tor:{Allowed Use} |
| | tor:{Allowed Use} | genprop:hasName | {Allowed Use} |

### B.1.2 Vancouver Zoning Districts & Labels

https://data.urbandatacentre.ca/catalogue/zoning-districts-and-labels

This dataset provides an assignment of zoning types to geometries. This can be mapped into both a definition of zoning regulations, as well as the identification of zoning types, as specified in Table 47. Since multiple levels of zoning types are identified in the table, this information can also be mapped to capture the definition of the subZoning relationships between zoning types, and the application of generalized zoning types to the area (though this could also be inferred, if not defined).

*Table 47: Mapping zone type assignments for Vancouver*

| **"Zoning Districts and Labels" fields** | **Mapping to HPCDM** | | |
|---|---|---|---|
| | **Subject** | **Property** | **Object** |
| | van:bylaw3575 | rdf:type | hp:ZoningBylaw |
| | van:bylaw3575 | 5087-2:legislationIdentifier | "By-law 3575" |
| OBJECT ID | <zoning_reg_{OBJECT ID}> | rdf:type | hp:Regulation |
| | <zoning_reg_{OBJECT ID}> | hp:definedIn | van:bylaw3575 |
| | <zoning_reg_{OBJECT ID}> | definedFor | van:area_{OBJECT ID} |
| | van:area_{OBJECT ID} | rdf:type | 5087-2:LandArea |
| | van:area_{OBJECT ID} | 5087-1:hasLocation | van:area_{OBJECTID}_geometry |
| GEOMETRY | van:area_{OBJECTID}_geometry | 5087-1:hasGeometry | {GEOMETRY} |
| ZONING DISTRICT | <zoning_reg_{OBJECT ID}> | designatesZoningType | van:zone_{ZONING DISTRICT} |
| ZONING CATEGORY | <zoning_reg_{OBJECT ID}> | designatesZoningType | van:zone_{ZONING CATEGORY} |
| | van:zone_{ZONING CATEGORY} | hp:subZoningType | van:zone_{ZONING DISTRICT} |
| ZONING CLASSIFICATION | <zoning_reg_{OBJECT ID}> | designatesZoningType | van:zone_{ZONING_CLASSIFICATION} |
| | van:zone_{ZONING_CLASSIFICATION} | hp:subZoningType | van:zone_{ZONING CATEGORY} |

No other datasets defining zoning regulations are available in Vancouver. Excerpts of some of the zoning documents are used here as examples to demonstrate the mapping by extracting the relevant information and formalizing it using the HPCDM. The first example specifies a representation of use data extracted from the schedule for the RM-1 zone. An excerpt of the document is shown in Figure 38. The Zoning Type identifier defined from the "Zoning Districts and Labels" dataset is referenced and additional information describing regulations for the zone are added to describe the allowed types of use. Note that in cases where integration across Canada is of interest, the relationship between land use types and codes defined in each municipality may be reviewed and formally defined in an extension of the ontology.

**RM-1**

## 2 USE REGULATIONS

### 2.1 Outright and Conditional Approval Uses

All outright and conditional approval uses are subject to all other provisions of this by-law, including **Section 2**, **Section 10** and **Section 11**, and compliance with the regulations of this schedule including section **2.2**.

The uses identified in the table below as outright approval uses are permitted in these districts and will be issued a permit.

The uses identified in the table below as conditional approval uses may be approved in these districts by the Director of Planning, with or without conditions, if the Director of Planning considers:

(a) the intent of this schedule and all applicable Council policies and guidelines; and

(b) the submission of any advisory group, property owner or tenant.

Uses are listed under their general land use category. Applicable use-specific regulations in section **2.2** of this schedule are cross-referenced in the third column. Cross-references to applicable use-specific regulations are provided for information purposes only and do not form part of this by-law.

| Use | Approval | Use-Specific Regulations |
|---|---|---|
| **Agricultural Uses** | | |
| Urban Farm - Class A | Conditional | |

*Figure 38: Excerpt of use regulations for the RM-1 zone in Vancouver.*

*Table 48: Example mapping of RM-1 zone regulations in Vancouver*

| | Mapping to HPCDM | | |
|---|---|---|---|
| | **Subject** | **Property** | **Object** |
| | van:zone_{ZONING DISTRICT} | rdf:type | hp:ZoningType |
| | van:zone_{ZONING DISTRICT} | hp:definedIn | van:bylaw3575 |
| | van:zone_{ZONING DISTRICT} | 5087-1:hasIdentifier | "RM-1" |
| | van:zone_{ZONING DISTRICT} | 5087-1:hasName | "RM-1" |
| | van:zone_{ZONING DISTRICT} | 5087-1:hasDescription | "Residential, multiple dwelling" |
| | van:zone_{ZONING DISTRICT} | oz:allowsUse | van:mcd2_landuse |
| | van:mcd2_landuse | rdf:type | hp:LandUse |
| | van:mcd2_landuse | 5087-2:hasCode | van:mcd2_landusecode |
| | van:mcd2_landusecode | 5087-1:hasName | "Multiple Conversion Dwelling, containing 2 dwelling units" |
| | van:mcd2_landusecode | 5087-2:specification | "https://bylaws.vancouver.ca/zoning/zoning-by-law-section-2.pdf" |
| | … | … | … |
| | van:zone_{ZONING DISTRICT} | oz:mayAllowUse | van:ufa_landuse |
| | van:ufa_landuse | rdf:type | hp:LandUse |
| | van:ufa_landuse | 5087-2:hasCode | van:ufa_landuse_code |
| | van:ufa_landuse | 5087-1:hasName | "Urban Farm - Class A" |
| | van:ufa_landuse | 5087-2:specification | "https://bylaws.vancouver.ca/zoning/zoning-by-law-section-2.pdf" |

### B.1.3 Halifax Bylaw Areas and Zoning Boundaries

https://data.urbandatacentre.ca/catalogue/by-law-areas

https://data.urbandatacentre.ca/catalogue/zoning-boundaries

Halifax publishes two key datasets on zoning regulations. The first is a Zoning Boundaries dataset that identifies and associates zoning types with the geometries they apply to. It also specifies a bylaw identifier and, in this way, provides a reference for the zoning regulation. The mapping for this dataset is outlined in Table 49. The second is the By-law Areas dataset. In Halifax, zones are defined in the context of land use bylaw (LUB) areas, so different zoning types are defined in different by-laws. This second dataset is used to define the LUBs for the zoning types. The mapping is outlined in Table 50, it primarily instantiates the areas *impacted by* a particular bylaw. In the absence of any more detailed zoning regulation data, an example mapping of data manually extracted from the bylaw text is also included, specified in Table 51. The mapping illustrates an example specification of permitted types of land use for the "RR" zoning type, along with a specification of the minimum permitted lot area (5 acres).

*Table 49: Mapping of zone type assignments in Halifax*

| "Zoning Boundaries" fields | Mapping to HPCDM | | |
|---|---|---|---|
| | **Subject** | **Property** | **Object** |
| BYLAW_ID | hrm:bylaw_{BYLAW_ID} | hp:definesZoningType | hrm:zone_{BYLAW_ID}_{ZONE} |
| ZONE | hrm:zone_{BYLAW_ID}_{ZONE} | rdf:type | ZoningType |
| | hrm:zone_{BYLAW_ID}_{ZONE} | 5087-1:hasIdentifier | "{ZONE}" |
| OBJECTID | hrm:bylaw_{BYLAW_ID} | hp:definesRegulation | hrm:zoning_reg_{OBJECTID} |
| | hrm:zoning_reg_{OBJECTID} | designatesZoningType | hrm:zone_{BYLAW_ID}_{ZONE} |
| | hrm:zoning_reg_{OBJECTID} | hp:definedFor | hrm:area_{OBJECTID} |
| | hrm:area_{OBJECTID} | rdf:type | 5087-2:LandArea |
| | hrm:area_{OBJECTID} | 5087-1:hasLocation | hrm:area_{OBJECTID}_geometry |
| geometry | hrm:area_{OBJECTID}_geometry | 5087-1:hasGeometry | "{geometry}" |

*Table 50: Mapping of bylaws and associated areas in Halifax*

| “Bylaw Areas” fields | Mapping to HPCDM | | |
|---|---|---|---|
| | **Subject** | **Property** | **Object** |
| Bylaw ID | hrm:bylaw_{BYLAW_ID} | rdf:type | hp:ZoningBylaw |
| Legal Name | hrm:bylaw_{BYLAW_ID} | 5087-1:hasName | {Legal Name} |
| | hrm:bylaw_{BYLAW_ID} | 5087-2:legislation Identifier | "LAND USE BY-LAW BEDFORD" |
| geometry | hrm:bylaw_{BYLAW_ID} | 5087-2:impacts | _:some_land_area |
| | _:some_land_area | rdf:type | 5087-2:LandArea |
| | _:some_land_area | 5087-1:hasLocati on | hrm:area_{OBJECTID}_geom etry |
| | hrm:area_{OBJECTID}_ge ometry | 5087-1:hasGeom etry | {geometry} |

*Table 51: Example zoning type definition extracted from Halifax (Bedford) bylaw document*

| | Mapping to HPCDM | | |
|---|---|---|---|
| | **Subject** | **Property** | **Object** |
| | hrm:zone_{BYLAW_ID}_{Z ONE} | 5087-1:hasName | "Residential Reserve Zone" |
| | hrm:zone_{BYLAW_ID}_{Z ONE} | oz:allowsUse | hrm:single_unit_dwelling |
| | hrm:single_unit_dwelling | rdf:type | LandUse |
| | hrm:zone_{BYLAW_ID}_{Z ONE} | oz:allowsUse | hrm:neighbourhood_parks |
| | hrm:neighbourhood_parks | rdf:type | LandUse |
| | … | | |
| | hrm:bylaw_{BYLAW_ID} | hp:definesRegula tion | hrm:rr_reg_lot_area |
| | hrm:rr_reg_lot_area | opr:forZoningType | hrm:Bedford_RR |
| | hrm:rr_reg_lot_area | hp:specifiesConstraint | hrm:req_area_bedford_rr |
| | hrm:req_area_bedford_rr | rdf:type | hp:QuantityRequirement |
| | hrm:req_area_bedford_rr | i72:hasValue | hrm:req_area_bedford_rr_specification |
| | hrm:req_area_bedford_rr_specification | i72:hasNumericalValue | "5" |
| | hrm:req_area_bedford_rr_specification | i72:hasUnit | hp:acres |
| | hrm:req_area_bedford_rr | hp:specifiesMinimumFor | hrm:bedford_rr_min_area |
| | hrm:bedford_rr_min_area | rdf:type | hp:Minimum |
| | hrm:bedford_rr_min_area | i72:parameter_of_var | hrm:area_var |
| | hrm:bedford_rr_min_area | i72:minimum_of | hrm:lot_population_in_bedford_rr_zone |
| | hrm:area_var | i72:hasName | "hp:hasArea" |
| | hrm:lot_population_in_bedford_rr_zone | rdf:type | hrm:BedfordLotPopulation_Bedford_RR |
| | hrm:BedfordLotPopulation_Bedford_RR | rdfs:subClassOf | hrm:BedfordLots |
| | hrm:BedfordLots | rdfs:subClassOf | i72:Population |
| | hrm:BedfordLots | i72:located_in | https://www.geonames.org/5897321/bedford.html |
| | hrm:BedfordLots | i72:defined_by only | hp:Lot |

## B.2 Parcels

### B.2.1 Toronto

**Dataset:** Toronto Property Boundaries (for parcel geometries)

https://open.toronto.ca/dataset/property-boundaries/

**Notes:** *data on property (parcel) geometries;*

| Field | Subject | Property | Object |
|---|---|---|---|
| PARCELID | tor:Property{PARCELID} | rdf:type | hp:Parcel |
| STATEDAREA | tor:Property{PARCELID} | hp:hasArea | tor:PropertyArea{PARCELID} |
| | tor:PropertyArea{PARCELID} | i72:hasValue | tor:PropertyAreaMeasure{PARCELID} |
| | tor:PropertyAreaMeasure{PARCELID} | i72:hasNumericalValue | "{STATEDAREA}" |
| | tor:PropertyAreaMeasure{PARCELID} | i72:hasUnit | i72:square_metre |
| geometry | tor:Property{PARCELID} | loc:hasLocation | tor:PropertyLoc{PARCELID} |
| | tor:PropertyLoc{PARCELID} | geo:asWKT | "{geometry}" |
| Perimeter* | tor:Property{PARCELID} | hp:hasPerimeter | tor:PropertyPerimeter{PARCELID} |
| | tor:PropertyPerimeter{PARCELID} | i72:hasValue | tor:PropertyPerimeterMeasure{PARCELID} |
| | tor:PropertyPerimeterMeasure{PARCELID} | i72:hasNumericalValue | {Perimeter} |
| | tor:PropertyPerimeterMeasure{PARCELID} | i72:hasUnit | i72:metre |

**Dataset:** Fake ownership and PIN

Synthetic data fakeowners_1of2.csv, fakeowners_2of2.csv

**Notes:** *fake attributes generated for completion*

| Field | Subject | Property | Object |
|---|---|---|---|
| PARCEL ID | tor:Property{PARCELID} | hp:ownership | tor:{PARCELID}Ownership{FakeOwner} |
| Fake Owner | tor:{PARCELID}Ownership{Fake Owner} | genprop:hasName | "{Fake Owner}" |

**Dataset**: Government land ownership

provincial lands.csv, gtha lower tier toronto.csv, gtha upper tier.csv

**Notes**: via: https://experience.arcgis.com/experience/1ab05066121a4dbfad3abdbd58c420f9/page/Page

| **Field Name** | **Subject** | **Property** | **Object** |
|---|---|---|---|
| OBJECTID_1 | tor:{data_source}property{OBJECTID_1} | rdf:type | hp:Parcel |
| Tier or myp_tier | tor:{data_source}property{OBJECTID_1} | hp:ownership | hp:{Tier}Org* |
| | hp:{Tier}Org* | rdf:type | org_city:GovernmentOrganization |
| | tor:{data_source}property{OBJECTID_1} | loc:hasLocation | tor:{data_source}property{OBJECTID_1}Loc |
| geometry | tor:{data_source}property{OBJECTID_1}Loc | geo:asWKT | {geometry} |

**Dataset:** Federal Property Structures https://www.tbs-sct.gc.ca/dfrp-rbif/opendata-eng.aspx DFRP Data Extract (zip)

**Notes:** Filter for Toronto-only buildings; join with Property boundaries to infer building-parcel association Shapefile attributes don't match xml data dictionary; need to use a combination of xml file for attributes and shapefile to infer parcel occupation

| **Field** | **Subject** | **Property** | **Object** |
|---|---|---|---|
| Structure_Number | tor:{Structure_Number}Building | rdf:type | hp:Building |
| | tor:{Structure_Number}Building | genprop:hasIdentifier | {Structure_Number} |
| Structure_Name_E | tor:{Structure_Number}Building | genprop:hasName | {Structure_Name_E} |
| Address_E | tor:{Structure_Number}Building | contact:hasAddress | tor:{Structure_Number}Address |
| | tor:{Structure_Number}Address | contact:hasStreetNumber | parsed from {Address_E} if available |
| | tor:{Structure_Number}Address | contact:hasStreet | parsed from {Address_E} if available |
| | tor:{Structure_Number}Address | contact:hasStreetType | parsed from {Address_E} if available |
| Floor_Area | tor:{Structure_Number}Building | hp:hasFloorArea | tor:{Structure_Number}BuildingFloorArea |
| | tor:{Structure_Number}BuildingFloorArea | i72:hasValue | tor:{Structure_Number}BuildingFloorAreaMeasure |
| | tor:{Structure_Number}BuildingFloorAreaMeasure | i72:hasNumericalValue | {Floor_Area} |
| unitofMeasure | tor:{Structure_Number}BuildingFloorAreaMeasure | i72:hasUnit | i72:square_metre or hp:square_foot |
| Construction_Year | tor:{Structure_Number}Building | bdg:yearOfConstruction | tor:Year{Construction_Year} |
| | tor:Year{Construction_Year} | time:year | {Construction_Year} |
| Condition_E | tor:{Structure_Number}Building | hp:hasCondition | tor:{Structure_Number}BuildingCondition |
| | tor:{Structure_Number}BuildingCondition | rdf:type | hp:BuildingCondition |
| | tor:{Structure_Number}BuildingCondition | code:hasCode | tor:{Structure_Number}BuildingConditionCode |
| code | tor:{Structure_Number}BuildingConditionCode | genprop:hasIdentifier | {code} |
| | tor:{Structure_Number}BuildingConditionCode | genprop:hasName | {Condition_E} |
| Location | tor:{Structure_}Building | loc:hasLocation | tor:{Structure_}BuildingLoc |
| | tor:{Structure_}BuildingLoc | geo:asWKT | POINT({Longitude} {Latitude}) |
| Tenants | tor:{Structure_}Building | hp:occupiedBy | tor:{Tenant code}tenant |
| | tor:{Tenant code}tenant | rdf:type | org_city:Organization |
| | tor:{Tenant code}tenant | genprop:hasName | {Name_E} |
| UseTypes | tor:{Structure_}Building | bdg:use | tor:{Structure_}BuildingUse{code} |
| | tor:{Structure_}BuildingUse{code} | code:hasCode | tor:{Structure_}BuildingUse{code}Code |
| | tor:{Structure_}BuildingUse{code}Code | genprop:hasIdentifier | {code} |
| | tor:{Structure_}BuildingUse{code}Code | genprop:hasName | {Use_Name_E} |
| PARCELID | tor:{Structure_}Building | hp:occupies | tor:Property{PARCELID} |

### B.2.2 Vancouver Property Parcel Polygons

Shape: https://opendata.vancouver.ca/explore/dataset/property-parcel-polygons/

| Field | Subject | Property | Object |
|---|---|---|---|
| | van:{SITE_ID}Parcel | rdf:type | hp:Parcel |
| SITE_ID | van:{SITE_ID}Parcel | genprop:hasIdentifier | {SITE_ID} |
| TAX_COORD | van:{SITE_ID}Parcel | genprop:hasIdentifier | {TAX_COORD} |
| Geom | van:{SITE_ID}Parcel | loc:hasLocation | van:{SITE_ID}ParcelLoc |
| | van:{SITE_ID}ParcelLoc | geo:asWKT | {Geom} |

Ownership: https://opendata.vancouver.ca/explore/dataset/property-tax-report/

Note: this dataset does not directly provide owner details, however the PIN can be used to lookup owner information

| Field | Subject | Property | Object |
|---|---|---|---|
| PID | van:{PID}Parcel | rdf:type | hp:Parcel |
| | van:{PID}Parcel | genprop:hasIdentifier | {PID} |
| LAND_COORDINATE | van:{PID}Parcel | genprop:hasIdentifier | {LAND_COORDINATE} |
| | van:{PID}Parcel | hp:ownership | - |
| ZONING_DISTRICT | van:{PID}Parcel | hp:hasZone | van:{PID}ParcelZone |
| | van:{PID}ParcelZone | rdf:type | hp:ZoningType |
| | van:{PID}ParcelZone | genprop:hasName | {ZONING_DISTRICT} |

### B.2.3 Halifax

**Dataset:** Halifax civic addresses

https://data-hrm.hub.arcgis.com/datasets/HRM::civic-addresses/about

**Notes:** *data on parcels;*
*geometry is point-based (no polygons);*
*ownership information may be determined with PID (given access to land ownership database*

| Field | Subject | Property | Object |
|---|---|---|---|
| PID | hrm:{PID}Parcel | rdf:type | hp:Parcel |
| | hrm:{PID}Parcel | genprop:hasIdentifier | {PID} |
| LOT_NUM | hrm:{PID}Parcel | hp:createdFromLot | hrm:{LOT_NUM}Lot |
| | hrm:{LOT_NUM}Lot | rdf:type | hp:Lot |
| | hrm:{LOT_NUM}Lot | genprop:hasIdentifier | {LOT_NUM} |
| geometry | hrm:{PID}Parcel | loc:hasLocation | hrm:{PID}ParcelLoc |
| | hrm:{PID}Parcel Loc | geo:asWKT | {geometry} |

## B.3 Buildings

Building data is drawn from the Open Database of Buildings (ODB) published by StatCan. It provides an integrated collection of building data from various sources throughout Canada.

| | |
|---|---|
| Dataset | Open Database of Buildings<br><br>original dataset from: https://www150.statcan.gc.ca/n1/pub/34-26-0001/342600012018001-eng.htm (parts 1 and<br><br>merged datasets on arcgis at https://utoronto.maps.arcgis.com/home/item.html?id=9d123bc3e0da4555abf5c88fd8bb7b1b and https://utoronto.maps.arcgis.com/home/item.html?id=1d271ca5c49e406ea4a25f32aa15e066 |
| Notes | Some fields are incomplete |

| Field Name | Subject | Property | Object |
|---|---|---|---|
| id | tor:Building{id} | rdf:type | hp:Building |
| name | tor:Building{id} | genprop:hasName | "{name}" |
| address | tor:Building{id} | contact:hasAddress | tor:BuildingAddress{id} |
| | tor:BuildingAddress{id} | contact:hasStreetNumber | number parsed from {address} if available |
| | tor:BuildingAddress{id} | contact:hasStreet | street name parsed from {address} if available |
| | tor:BuildingAddress{id} | contact:hasStreetType | street type parsed from {address} if available |
| type | tor:Building{id} | bdg:use | tor:BuildingUse{type} |
| | tor:BuildingUse{type} | code:hasCode | tor:BuildingUseCode{type} |
| | tor:BuildingUseCode{type} | genprop:hasName | "{type}" |
| height | tor:Building{id} | hp:hasHeight | tor:BuildingHeight{id} |
| | tor:BuildingHeight{id} | i72:hasValue | tor:BuildingHeightMeasure{id} |
| | tor:BuildingHeightMeasure{id} | i72:hasNumericalValue | "{height}" |
| | tor:BuildingHeightMeasure{id} | i72:hasUnit | i72:metre |
| floors | tor:BuildingHeight{id} | i72:hasValue | tor:BuildingFloorsMeasure{id} |
| | tor:BuildingFloorsMeasure{id} | i72:hasNumericalValue | "{floors}" |
| | tor:BuildingFloorsMeasure{id} | i72:hasUnit | hp:storeys |
| sq_ft | tor:Building{id} | hp:hasFloorArea | tor:BuildingFloorArea{id} |
| | tor:BuildingFloorArea{id} | i72:hasValue | tor:BuildingFloorAreaMeasure{id} |
| | tor:BuildingFloorAreaMeasure{id} | i72:hasNumericalValue | "{sq_ft}" |
| | tor:BuildingFloorAreaMeasure{id} | i72:hasUnit | i72:square_metre |
| year_built | tor:Building{id} | bdg:yearOfConstruction | tor:Year{year_built} |
| | tor:Year{year_built} | time:year | "{year_built}" |
| geometry | tor:Building{id} | loc:hasLocation | tor:Building{id}Loc |
| | tor:Building{id}Loc | geo:asWKT | "{geometry}" |
| PARCELID | tor:Building{id} | hp:occupies | tor:Property{PARCELID} |

## B.4 Demographics

The demographics data is drawn from the Canadian Census of Population (2016). More recent census data may be transformed with a similar mapping (with adjustments to account for any changes to characteristics).

Dataset: Canadian Census of Population 2016 (Toronto Data)

https://www12.statcan.gc.ca/census-recensement/2016/dp-pd/prof/details/page.cfm?Lang=E&Geo1=CSD&Geo2=PR&Code2=01&SearchType=Begins&SearchPR=01&TABID=1&B1=All&type=0&Code1=3520005&SearchText=toronto

Notes: Generalized mapping applies to all census characteristics

| Field Name | Subject | Property | Object |
|---|---|---|---|
| jurisdictional area | cacensus: {jurisdictional area}CensusProfile2016 | rdf:type | census:CensusProfile |
| | toronto:{jurisdictional area} | census:hasCensusProfile | cacensus: {jurisdictional area}CensusProfile |
| | cacensus: {jurisdictional area}CensusProfile2016 | census:hasLocation | toronto:{jurisdictional area} |
| | cacensus: {jurisdictional area}CensusProfile2016 | census:hasTime | cacensus: censusProfile2016DateTimeInterval |
| Characteristic | cacensus: {jurisdictional area}CensusProfile2016 | census:hasCharacteristic | cacensus:{jurisdictional area}{Characteristic} |
| | cacensus: {jurisdictional area}{Characteristic} | rdf:type | cacensus:{appropriate subclass of Characteristic2016} (e.g., 0-14Years2016 for the “0 to 14 years” characteristic as seen in the 2016 census datasets) |
| | cacensus:{appropriate subclass of Characteristic2016} | rdfs:subClassOf | i72:{appropriate subclass of Parameter or Statistic} |
| | cacensus: {jurisdictional area}{Characteristic} | i72:hasValue | cacensus: {jurisdictional area}{Characteristic}Measure |
| Value of the characteristic | cacensus: {jurisdictional area}{Characteristic}Measure | rdf:type | i72:Measure |
| | cacensus: {jurisdictional area}{Characteristic}Measure | i72:hasNumericalValue | “{value of the characteristic}” |
| | cacensus: {jurisdictional area}{Characteristic}Measure | i72:hasUnit | i72:{appropriate instance of the Unit class} |

Dataset: Census 2016 Population Centres

https://www12.statcan.gc.ca/census-recensement/2011/geo/bound-limit/bound-limit-2016-eng.cfm

Notes: Other fields available that might be useful in future work (e.g. popctr class etc)

| **Field** | **Subject** | **Property** | **Object** |
|---|---|---|---|
| PCUID | tor:{PCUID}Popctr | rdf:type | cacensus:PopulationCentre |
| PCNAME | tor:{PCUID}Popctr | genprop:hasName | {PCNAME} |
| Shape | tor:{PCUID}Popctr | loc:hasLocation | tor:{PCUID}PopctrLoc |
| | tor:{PCUID}PopctrLoc | geo:asWKT | {Shape} |

## B.5 Fire & Emergency Services

### B.5.1 Toronto Fire Services

Toronto Dataset: Fire Services Run Areas https://data.urbandatacentre.ca/catalogue/city-toronto-toronto-fire-services-run-areas

| **Field Name** | **Subject** | **Property** | **Object** |
|---|---|---|---|
| | tor:fire_service{RUN_AREA} | rdf:type | hp:FireEmergencyService |
| RUN_AREA | tor:fire_service{RUN_AREA} | service:hasCatchmentArea | tor:fire_catchment_{AREA_ID} |
| geometry | tor:fire_catchment_{AREA_ID} | geo:asWKT | {geometry} |
| AREA_ID | | genprop:hasIdentifier | "{AREA_ID}" |

Dataset: Fire Facility Locations
https://data.urbandatacentre.ca/catalogue/city-toronto-fire-station-locations

| Field Name | Subject | Property | Object |
|---|---|---|---|
| STATION | tor:fire_service{STATION} | hp:providedFromSite | tor:fire_station_{STATION} |
| ADDRESS_POINT_ID | tor:fire_station_{STATION} | org:siteAddress | tor:fire_station_address_{ADDRESS_POINT_ID} |
| | tor:fire_station_address_{ADDRESS_POINT_ID} | rdf:type | contact:Address |
| ADDRESS_NUMBER | tor:fire_station_address_{ADDRESS_POINT_ID} | contact:hasStreetNumber | {ADDRESS_NUMBER} |
| LINEAR_NAME_FULL | tor:fire_station_address_{ADDRESS_POINT_ID} | contact:hasStreetName | {LINEAR_NAME} |
| geometry | tor:fire_station_{STATION} | loc:hasLocation | tor:fire_station_loc_{ADDRESS_POINT_ID} |
| | tor:fire_station_loc_{ADDRESS_POINT_ID} | geo:asWKT | {geometry} |
| | tor:fire_service{STATION} | res:hasCapacity | tor:fire_service{STATION}Capacity |
| | tor:fire_service{STATION}Capacity | rdf:type | hp:MinFirefighterPerPopulation |
| | tor:fire_service{STATION}Capacity | 21972:hasValue | tor:fire_service{STATION}CapacityMeasure |
| | tor:fire_service{STATION}CapacityMeasure | i72:hasNumericalValue | 0.001 |
| | tor:fire_service{STATION}CapacityMeasure | i72:hasUnit | hp:firefighter_per_person |

Dataset: Synthetic firefighter population counts

synthetic firefighter and population counts.csv

Notes: randomly generated, "available capacity" may be computed on the fly as the difference between total and in use

| Field Name | Subject | Property | Object |
|---|---|---|---|
| RUN_AREA | tor:fire_service{RUN_AREA} | res:capacityIn Use | tor:fire_service{RUN_AREA} CapacityUse |
| | tor:fire_service{RUN_AREA} CapacityUse | rdf:type | hp:FirefighterPerPopulation |
| | tor:fire_service{RUN_AREA} CapacityUse | i72:hasValue | tor:fire_service{RUN_AREA} CapacityUseMeasure |
| Firefighters per person in run area | tor:fire_service{RUN_AREA} CapacityUseMeasure | i72:hasNumer icalValue | {Firefighters per person in run area} |
| | tor:fire_service{RUN_AREA} CapacityUseMeasure | i72:hasUnit | hp:firefighter_per_person |

### B.5.2 Vancouver Fire Services

Dataset: Fire halls

https://opendata.vancouver.ca/explore/dataset/fire-halls

Notes: No run areas provided

| Field Name | Subject | Property | Object |
|---|---|---|---|
| NAME | van:{NAME*}_fireservice | rdf:type | hp:FireEmergencyService |
| | van:{NAME*}_fireservice | hp:providedFromSit e | van:{NAME*}_fireserviceSit e |
| | van:{NAME*}_fireserviceSite | genprop:hasName | {NAME} |
| GEOM | van:{NAME*}_fireserviceSite | loc:hasLocation | van:{NAME*}_fireserviceSit eLoc |
| | van:{NAME*}_fireserviceSiteLo c | geo:asWKT | {GEOM} |

### B.5.3 Halifax Fire Services

Dataset: Fire Response Zones

https://data-hrm.hub.arcgis.com/datasets/fire-response-zones

Notes:

| Field Name | Subject | Property | Object |
|---|---|---|---|
| PRIME | hrm:{PRIME}_fireservice | rdf:type | hp:FireEmergencyService |
| | hrm:{PRIME}_fireservice | hp:providedFromSite | hrm:{PRIME}_fireserviceSite |
| | hrm:{PRIME}_fireserviceSite | genprop:hasIdentifier | {PRIME} |
| Zone | hrm:{PRIME}_fireservice | service:hasCatchmentArea | hrm:{Zone}_fireserviceCatchment |
| | hrm:{Zone}_fireserviceCatchment | genprop:hasIdentifier | {Zone} |
| Shape | hrm:{Zone}_fireserviceCatchment | loc:hasLocation | hrm:{Zone}_fireserviceCatchmentLoc |
| | hrm:{Zone}_fireserviceCatchmentLoc | geo:asWKT | {Shape} |

## B.6 Power

### B.6.1 Toronto Power

| | |
|---|---|
| | Power (Toronto Hydro) Toronto Hydro available |
| Data set: | https://data.urbandatacentre.ca/catalogue/load-capacity-map https://services8.arcgis.com/SnGTjuDV2RIxBTxw/ArcGIS/rest/services/PRD_FeederLayers/FeatureServer |
| Notes: | extracted via Toronto Hydro available capacity map; transformer stations available but not currently mapped (no area coverage defined) |

| Field Name | Subject | Property | Object |
|---|---|---|---|
| Network_id | tor:hydro_feeder_service{Network_id} | rdf:type | hp:ElectricService |
| | tor:hydro_feeder_service{Network_id} | genprop:hasIdentifier | "{Network_id}" |
| | tor:hydro_feeder_service{Network_id} | service:hasAvailableCapacity | tor:hydro_feeder_service{Network_id}CapacityAvail |
| | tor:hydro_feeder_service{Network_id}CapacityAvail | i72:hasValue | tor:hydro_feeder_service{Network_id}CapacityAvailMeasure |
| Feeder_Capacity | tor:hydro_feeder_service{Network_id}CapacityAvailMeasure | i72:hasNumericalValue | {Feeder_Capacity} |
| | tor:hydro_feeder_service{Network_id}CapacityAvailMeasure | i72:hasUnit | hp:kilovolt_ampere |
| SHAPE | tor:hydro_feeder_service{Network_id} | service:hasCatchmentArea | tor:hydro_feeder_service{Network_id}Area{OJBECTID} |
| | tor:hydro_feeder_service{Network_id}Area{OJBECTID} | geo:asWKT | {SHAPE} |

Dataset: Fake Total Load Capacity (Toronto Hydro)

Notes:

| Field Name | Subject | Property | Object |
|---|---|---|---|
| Network_id | tor:hydro_feeder_service{Network_id} | service:hasCapacity | tor:hydro_feeder_service{Network_id}Capacity |
| Fake Max Avail Capacity (kVA) | tor:hydro_feeder_service{Network_id}Capacity | i72:hasValue | tor:hydro_feeder_service{Network_id}CapacityMeasure |
| | tor:hydro_feeder_service{Network_id}CapacityMeasure | i72:hasNumericalValue | {Fake Max Avail Capacity (kVA)} |
| | tor:hydro_feeder_service{Network_id}CapacityMeasure | i72:hasUnit | hp:kilovolt_ampere |

### B.6.2 Vancouver Power

No datasets found

### B.6.3 Halifax Power

No datasets found

## B.7 Solid Waste

### B.7.1 Toronto Solid Waste

Toronto Dataset: Solid Waste Daytime Curbside Collection Areas
https://open.toronto.ca/dataset/solid-waste-daytime-curbside-collection-areas/
https://data.urbandatacentre.ca/catalogue/solid-waste-daytime-curbside-collection-areas

| Field Name | Subject | Property | Object |
|---|---|---|---|
| FID | tor:solidwaste_service{FID} | rdf:type | hp:SolidWasteService |
| AREA_ID | tor:solidwaste_service{FID} | service:hasCatchmentArea | tor:solidwaste_servicearea_{AREA_ID} |
| | tor:solidwaste_servicearea_{AREA_ID} | genprop:hasIdentifier | "AREA_ID" |
| AREA_LONG | tor:solidwaste_servicearea_{AREA_ID} | genprop:hasIdentifier | "AREA_LONG" |
| Area | tor:solidwaste_servicearea_{AREA_ID} | genprop:hasName | "Area" |
| geometry | tor:solidwaste_servicearea_{AREA_ID} | geo:asWKT | "geometry" |

Dataset: Solid Waste Capacities

swms_synthetic_capacities.csv

Notes: Synthetic dataset

| Field Name | Subject | Property | Object |
|---|---|---|---|
| FID | tor:solidwaste_service{FID} | res:hasCapacity | tor:solidwaste_service{FID}Capacity |
| | tor:solidwaste_service{FID}Capacity | i72:hasValue | tor:solidwaste_service{FID}CapacityMeasure |
| Randomized total capacity per area | tor:solidwaste_service{FID} CapacityMeasure | i72:hasNumericalValue | "{Randomized total capacity per area}" |
| | tor:solidwaste_service{FID} CapacityMeasure | i72:hasUnit | hp:tonnes_per_year |
| | tor:solidwaste_service{FID} | res:capacityInUse | tor:solidwaste_service{FID} CapacityUse |
| | tor:solidwaste_service{FID} CapacityUse | rdf:type | hp:WasteProcessingRate |
| | tor:solidwaste_service{FID} CapacityUse | i72:hasValue | tor:solidwaste_service{FID} CapacityUseMeasure |
| Estimated Capacity in use(tonnes / year) | tor:solidwaste_service{FID} CapacityUseMeasure | i72:hasNumericalValue | "{Estimated Capacity in use(tonnes / year)}" |
| | tor:solidwaste_service{FID} CapacityUseMeasure | i72:hasUnit | hp:tonnes_per_year |
| | tor:solidwaste_service{FID} | res:hasAvailableCapacity | tor:solidwaste_service{FID} CapacityAvail |
| | tor:solidwaste_service{FID} CapacityAvail | i72:hasValue | tor:solidwaste_service{FID} CapacityAvailMeasure |
| Available capacity | tor:solidwaste_service{FID} CapacityAvailMeasure | i72:hasNumericalValue | "{Available capacity}" |
| | tor:solidwaste_service{FID} CapacityAvailMeasure | i72:hasUnit | hp:tonnes_per_year |

### B.7.2 Vancouver Solid Waste

Dataset: Garbage collection schedule zones for single family homes and duplexes

https://opendata.vancouver.ca/explore/dataset/garbage-collection-schedule-zones-for-single-family-homes-and-duplexes/information/

| **Field Name** | **Subject** | **Property** | **Object** |
|---|---|---|---|
| OBJECTID | van:{OBJECTID}solidwasteservice | rdf:type | hp:SolidWasteService |
| | van:{OBJECTID}solidwasteservice | service:hasCatchmentArea | van:{OBJECTID}solidwasteserviceCatchment |
| KML_NAME | van:{OBJECTID}solidwasteserviceCatchment | genprop:hasName | {KML_NAME} |
| SHAPE | van:{OBJECTID}solidwasteserviceCatchment | loc:hasLocation | van:{OBJECTID}solidwasteserviceCatchmentLoc |
| | van:{OBJECTID}solidwasteserviceCatchmentLoc | geo:asWKT | {SHAPE} |

### B.7.3 Halifax Solid Waste

Dataset: Solid Waste Collection Areas

https://data-hrm.hub.arcgis.com/datasets/HRM::solid-waste-collection-areas/about

| **Field Name** | **Subject** | **Property** | **Object** |
|---|---|---|---|
| OBJECTID | hrm:{OBJECTID}solidwasteservice | rdf:type | hp:SolidWasteService |
| | hrm:{OBJECTID}solidwasteservice | service:hasCatchmentArea | hrm:{OBJECTID}solidwasteserviceCatchment |
| UNIQUE_ID | hrm:{OBJECTID}solidwasteserviceCatchment | genprop:hasIdentifier | {UNIQUE_ID} |
| | hrm:{OBJECTID}solidwasteserviceCatchment | loc:hasLocation | hrm:{OBJECTID}solidwasteserviceCatchmentLoc |
| SHAPE | hrm:{OBJECTID}solidwasteserviceCatchmentLoc | geo:asWKT | {SHAPE} |
| CONTRACTOR | hrm:{CONTRACTOR*}contractor_org | cdt:providesService | hrm:{OBJECTID}solidwasteservice |
| | hrm:{CONTRACTOR*}contractor_org | genprop:hasName | {CONTRACTOR} |

## B.8 Water

### B.8.1 Toronto Water

Toronto watermain locations are interpreted as service sites (sites of the distribution service) in this mapping.

Dataset: Watermains

https://open.toronto.ca/dataset/watermains/
https://data.urbandatacentre.ca/catalogue/city-toronto-watermains

Notes: specifically, "distribution" dataset, filtered by "Watermain Type" == 0 (distribution lines); note that there is other data that could be useful for extensions to this work

| Field Name | Subject | Property | Object |
|---|---|---|---|
| _id | tor:waterservice | rdf:type | hp:WaterService |
| | tor:waterservice | hp:providedFromSite | tor:waterservice_distributionpipes{_id} |
| Watermain Asset Identification | tor:waterservice_distributionpipes{_id} | genprop:hasIdentifier | "{Watermain Asset Identification}" |
| | tor:waterservice_distributionpipes{_id} | loc:hasLocation | tor:waterservice_distributionpipes_loc{_id} |
| geometry | tor:waterservice_distributionpipes_loc{_id} | geo:asWKT | "{geo}" |

Dataset: Water billing (by ward)

https://data.urbandatacentre.ca/catalogue/city-toronto-water-billing-by-ward

Notes: In this case, wards are simply used as a means of aggregating water usage.
We can interpret this data as providing usage for sub-services of the city-wide water distribution service, where the water distribution is broken down by ward
This data gives us an annual consumption rate though in theory the numbers should be available e.g. per day per region.

| Field Name | Subject | Property | Object |
|---|---|---|---|
| Ward | tor:waterservice | hp:hasSubService | tor:water_distributionservice_ward{Ward}_{Year} |
| | tor:water_distributionservice_ward{Ward}_{Year} | rdf:type | hp:WaterService |
| | tor:water_distributionservice_ward{Ward}_{Year} | service:hasCatchmentArea | tor:water_distributionservice_ward_catchment{Ward} |
| Year | tor:water_distributionservice_ward{Ward}_{Year} | change:existsAt | tor:interval_{year} |
| | tor:interval_{year} | time:hasBeginning | tor:instant_{year}_start |
| | tor:interval_{year} | time:hasEnd | tor:instant_{year}_end |
| | tor:instant_{year}_start | time:inXSDDateTimeStamp | {year}-01-01T00:00:00-05:00 |
| | tor:instant_{year}_end | time:inXSDDateTimeStamp | {year}-12-31T23:59:59-05:00 |
| Total consumption (m3) | tor:water_distributionservice_ward{Ward}_{Year} | res:capacityInUse | tor:water_distributionservice_ward{Ward}_{Year}_capacityuse |
| | tor:water_distributionservice_ward{Ward}_{Year}_capacityuse | rdf:type | hp:WaterDistributionRate |
| | tor:water_distributionservice_ward{Ward}_{Year}_capacityuse | i72:hasValue | tor:water_distributionservice_ward{Ward}_{Year}_capacityuse_measure |
| | tor:water_distributionservice_ward{Ward}_{Year}_capacityuse_measure | i72:hasNumericalValue | "Total consumption (m3)" |
| | tor:water_distributionservice_ward{Ward}_{Year}_capacityuse_measure | i72:hasUnit | hp:cubic_meter_per_year |

Dataset: Ward locations

Notes: Query to graph to define the geometry for the water service catchment areas (defined per ward)

| **Field Name** | **Subject** | **Property** | **Object** |
|---|---|---|---|
| loc | tor:water_distributionservice_ward_catchment{Ward} | loc:hasLocation | {loc} |

Dataset: Water capacity (total and available) by ward

Notes: Used to define the geometry for the water service catchment areas (defined per ward)

| Field Name | Subject | Property | Object |
|---|---|---|---|
| Synthetic Capacity | tor:water_distributionservice_ward{Ward}_{Year} | res:hasCapacity | tor:water_distributionservice_ward{Ward}_{Year}_Capacity |
| | tor:water_distributionservice_ward{Ward}_{Year}_Capacity | rdf:type | hp:WaterDistributionRate |
| | tor:water_distributionservice_ward{Ward}_{Year}_Capacity | i72:hasValue | tor:water_distributionservice_ward{Ward}_{Year}_CapacityMeasure |
| | tor:water_distributionservice_ward{Ward}_{Year}_CapacityMeasure | i72:hasNumericalValue | "{Synthetic Capacity}" |
| | tor:water_distributionservice_ward{Ward}_{Year}_CapacityMeasure | i72:hasUnit | hp:cubic_metre_per_year |
| Synthetic Available Capacity | tor:water_distributionservice_ward{Ward}_{Year} | res:hasAvailableCapacity | tor:water_distributionservice_ward{Ward}_{Year}_AvailCapacity |
| | tor:water_distributionservice_ward{Ward}_{Year}_AvailCapacity | rdf:type | hp:WaterDistributionRate |
| | tor:water_distributionservice_ward{Ward}_{Year}_AvailCapacity | i72:hasValue | tor:water_distributionservice_ward{Ward}_{Year}_AvailCapacityMeasure |
| | tor:water_distributionservice_ward{Ward}_{Year}_AvailCapacityMeasure | i72:hasNumericalValue | "{Synthetic Available Capacity}" |
| | tor:water_distributionservice_ward{Ward}_{Year}_AvailCapacityMeasure | i72:hasUnit | hp:cubic_metre_per_year |

### B.8.2 Vancouver Water

Dataset Water distribution mains

https://opendata.vancouver.ca/explore/dataset/water-distribution-mains

| Field Name | Subject | Property | Object |
|---|---|---|---|
| OBJECTID | van:waterdistributionservice | rdf:type | hp:WaterService |
| Geom | van:waterdistributionservice | hp:providedFromSite | van:{OBJECTID}waterdistributionsite |
| | van:{OBJECTID}waterdistributionsite | loc:hasLocation | van:{OBJECTID}waterdistributionsiteLoc |
| | van:{OBJECTID}waterdistributionsiteLoc | geo:asWKT | {Geom} |

### B.8.3 Halifax Water

Dataset: PPL&C Halifax Water Permits

https://data-hrm.hub.arcgis.com/datasets/HRM::pplc-halifax-water-permits/about

Filtered for Primary_Work_Scope = "Individual Service Connections", "Shared Service Connections"
AND Permit_Status = "Completed"

Notes: Can be used to infer (incomplete) information on water service connections

| Field Name | Subject | Property | Object |
|---|---|---|---|
| PID | hrm:waterdistributionservice | hp:hasSubService | hrm:{PID}waterdistributionservice |
| | hrm:{PID}waterdistributionservice | hp:providedToParcel | hrm:parcel{PID}* |

## B.9 Wastewater

Sewer locations are interpreted as service sites, and may be used to approximate service accessibility

### B.9.1 Toronto Wastewater

Dataset: Sewer Pressurized Mains (mapping to service locations)

https://open.toronto.ca/dataset/sewer-pressurized-mains/
https://data.urbandatacentre.ca/catalogue/city-toronto-sewer-pressurized-mains

Notes: Map main locations; filter where "Sewer Pressurized Trunk Sewer" = No

| Field Name | Subject | Property | Object |
|---|---|---|---|
| _id | tor:wastewaterservicepressurizedmain{_id} | rdf:type | hp:WastewaterService |
| | tor:wastewaterservicepressurizedmain{_id} | hp:providedFromSite | tor:wastewaterservice_pressurizedmain{_id}Site |
| Sewer Pressurized Asset Identification | tor:wastewaterservice_pressurizedmain{_id}Site | genprop:hasIdentifier | "{Sewer Pressurized Asset Identification}" |
| | tor:wastewaterservice_pressurizedmain{_id}Site | loc:hasLocation | tor:wastewaterservice_pressurizedmain_loc{_id} |
| geometry | tor:wastewaterservice_pressurizedmain_loc{_id} | geo:asWKT | "{geometry}" |

Dataset: Sewer Gravity Mains (mapping to service locations)

https://open.toronto.ca/dataset/sewer-gravity-mains/
https://data.urbandatacentre.ca/catalogue/city-toronto-sewer-gravity-mains

Notes: Mapping sewer main locations as approximations for service; filter where "Sewer Gravity Trunk Sewer" = No

| Field Name | Subject | Property | Object |
|---|---|---|---|
| _id | tor:wastewaterservicegravitymain{_id} | rdf:type | hp:WastewaterService |
| | tor:wastewaterservicegravitymain{_id} | hp:providedFromSite | tor:wastewaterservice_gravitymain{_id}Site |
| Sewer Gravity Asset Identification | tor:wastewaterservice_gravitymain{_id}Site | genprop:hasIdentifier | "{Sewer Gravity Asset Identification}" |
| | tor:wastewaterservice_gravitymain{_id}Site | loc:hasLocation | tor:wastewaterservice_gravitymain_loc{_id} |
| geometry | tor:wastewaterservice_gravitymain_loc{_id} | geo:asWKT | "{geometry}" |

Dataset: Wastewater Capacity (Gravity mains)

Notes: synthetic (Sewer Gravity Main_Capacity.csv)

| **Field Name** | **Subject** | **Property** | **Object** |
|---|---|---|---|
| _id | tor:wastewaterservicegravitymain{_id} | res:hasCapacity | tor:wastewaterservicegravitymain{_id}Capacity |
| | tor:wastewaterservicegravitymain{_id}Capacity | i72:hasValue | tor:wastewaterservicegravitymain{_id}CapacityMeasure |
| Est Flow Capacity (m3/year) | tor:wastewaterservicegravitymain{_id}CapacityMeasure | i72:hasNumericalValue | "{Est Flow Capacity (m3/year)}" |
| | tor:wastewaterservicegravitymain{_id}CapacityMeasure | i72:hasUnit | hp:cubic_metre_per_year |
| | tor:wastewaterservicegravitymain{_id} | res:capacityInUse | tor:wastewaterservicegravitymain{_id}CapacityUse |
| | tor:wastewaterservicegravitymain{_id}CapacityUse | rdf:type | hp:WaterProcessingRate |
| | tor:wastewaterservicegravitymain{_id}CapacityUse | i72:hasValue | tor:wastewaterservicegravitymain{_id}CapacityUseMeasure |
| Synthetic (Randomized) utilization | tor:wastewaterservicegravitymain{_id}CapacityUseMeasure | i72:hasNumericalValue | "{Synthetic (Randomized) utilization}" |
| | tor:wastewaterservicegravitymain{_id}CapacityUseMeasure | i72:hasUnit | hp:cubic_metre_per_year |
| | tor:wastewaterservicegravitymain{_id} | res:hasAvailableCapacity | tor:wastewaterservicegravitymain{_id}CapacityAvail |
| | tor:wastewaterservicegravitymain{_id}CapacityAvail | i72:hasValue | tor:wastewaterservicegravitymain{_id}CapacityAvailMeasure |
| Synthetic Available Capacity | tor:wastewaterservicegravitymain{_id}CapacityAvailMeasure | i72:hasNumericalValue | "{Synthetic Available Capacity}" |
| | tor:wastewaterservicegravitymain{_id}CapacityAvailMeasure | i72:hasUnit | hp:cubic_metre_per_year |

Dataset: Wastewater Capacity (Forced Mains)

Notes: synthetic (Sewer Pressurized Main_Capacity.csv)

| **Field Name** | **Subject** | **Property** | **Object** |
|---|---|---|---|
| _id | tor:wastewaterservicepressurizedmain{_id} | res:hasCapacity | tor:wastewaterservicepressurizedmain{_id}Capacity |
| | tor:wastewaterservicepressurizedmain{_id}Capacity | i72:hasValue | tor:wastewaterservicepressurizedmain{_id}CapacityMeasure |
| Synthetic Capacity (annual flow m3) | tor:wastewaterservicepressurizedmain{_id}CapacityMeasure | i72:hasNumericalValue | "{Synthetic Capacity (annual flow m3)}" |
| | tor:wastewaterservicepressurizedmain{_id}CapacityMeasure | i72:hasUnit | hp:cubic_metre_per_year |
| | tor:wastewaterservicepressurizedmain{_id} | res:capacityInUse | tor:wastewaterservicepressurizedmain{_id}CapacityUse |
| | tor:wastewaterservicepressurizedmain{_id}CapacityUse | rdf:type | hp:WaterProcessingRate |
| | tor:wastewaterservicepressurizedmain{_id}CapacityUse | i72:hasValue | tor:wastewaterservicepressurizedmain{_id}CapacityUseMeasure |
| Randomized Annual Use (m3) | tor:wastewaterservicepressurizedmain{_id}CapacityUseMeasure | i72:hasNumericalValue | "{Randomized Annual Use (m3)}" |
| | tor:wastewaterservicepressurizedmain{_id}CapacityUseMeasure | i72:hasUnit | hp:cubic_metre_per_year |
| | tor:wastewaterservicepressurizedmain{_id} | res:hasAvailableCapacity | tor:wastewaterservicepressurizedmain{_id}CapacityAvail |
| | tor:wastewaterservicepressurizedmain{_id}CapacityAvail | i72:hasValue | tor:wastewaterservicepressurizedmain{_id}CapacityAvailMeasure |
| Available Annual Flow | tor:wastewaterservicepressurizedmain{_id}CapacityAvailMeasure | i72:hasNumericalValue | "{Available Annual Flow}" |
| | tor:wastewaterservicepressurizedmain{_id}CapacityAvailMeasure | i72:hasUnit | hp:cubic_metre_per_year |

### B.9.2 Vancouver Wastewater

| | |
|---|---|
| Dataset | Sewer Mains |
| | https://opendata.vancouver.ca/explore/dataset/sewer-mains/ |
| Notes | Filter for EFFLUENT_TYPE="Sanitary" or "Combined" |

| Field Name | Subject | Property | Object |
|---|---|---|---|
| OBJECTID | van:wastewaterservice | rdf:type | hp:WastewaterService |
| | van:wastewaterservice | hp:providedFromSite | van:{OBJECTID}wastewaterserviceSite |
| Geom | van:{OBJECTID}wastewaterserviceSite | loc:hasLocation | van:{OBJECTID}wastewaterserviceSiteLoc |
| | van:{OBJECTID}wastewaterserviceSiteLoc | geo:asWKT | {Geom} |

### B.9.3 Halifax Wastewater

No datasets found

## B.10 Transit

The following mapping formulation applies generally to any municipality and transit service publishing data in the gtfs format. Variations will likely occur in the specification of capacity data.

| Field Name | Subject | Property | Object |
|---|---|---|---|
| stop_id | tor:{stop_id}TransitStop | rdf:type | cdt:TransitStop |
| stop_lat, stop_lon | tor:{stop_id}TransitStop | loc:hasLocation | tor:{stop_id}TransitStop_loc |
| | tor:{stop_id}TransitStop_loc | geo:asWKT | "POINT({stop_lat} {stop_lon})" |
| stop_name | cdt:{stop_id}TransitStop | genprop:hasName | "{stop_name}" |
| stop_id | cdt:{stop_id}TransitStop | genprop:hasIdentifier | "{stop_id}" |
| Operator | tor:ttc | org:hasSite | cdt:{stop_id}TransitStop |
| | tor:ttc | genprop:hasName | "Toronto Transit Commission" |
| route_id | tor:ttc | cdt:providesService | tor:{route_id}RouteService |
| | tor:{route_id}RouteService | hpcdm:providedFromSite | Site stop_id |

### B.10.1 Toronto Transit Capacity

Dataset TTC capacties (in use)

https://open.toronto.ca/dataset/ttc-ridership-all-day-weekday-for-surface-routes/
https://data.urbandatacentre.ca/catalogue/city-toronto-ttc-ridership-all-day-weekday-for-surface-routes

| **Field Name** | **Subject** | **Property** | **Object** |
|---|---|---|---|
| route_id (via TTC stops) | tor:{route_id}RouteService | genprop:hasIdentifier | "{route_short_name}" |
| | tor:{route_id}RouteService | rdf:type | hp:PublicTransitService |
| route_short_name (via TTC stops routes.txt), Route # (via ridership) | | res:capacityInUse | tor:{route_id}RouteServiceCapacityUse |
| | tor:{route_id}RouteServiceCapacityUse | rdf:type | hp:PassengerThroughputRate |
| All-Day Ridership (via ridership) | tor:{route_id}RouteServiceCapacityUse | i72:hasValue | tor:{route_id}RouteServiceCapacityUseMeausre |
| | tor:{route_id}RouteServiceCapacityUseMeausre | i72:hasNumericalValue | "{All Day Ridership}" |
| | tor:{route_id}RouteServiceCapacityUseMeausre | i72:hasUnit | hp:person_per_day |

| | |
|---|---|
| Dataset | TTC capacties (total)<br><br>estimated data (TTC_est_throughput_report.csv) |
| Notes: | We could also implement a calculation for the available capacity based on this (possibly as a rule in the KG, or a separate function based on the two files), but for now this is omitted. If needed we can always query for the difference directly. |

| **Field Name** | **Subject** | **Property** | **Object** |
|---|---|---|---|
| route_id | tor:{route_id}RouteService | res:hasCapacity | tor:{route_id}RouteServiceCapacityTotal |
| daily_passenger_throughput | tor:{route_id}RouteServiceCapacityTotal | i72:hasValue | tor:{route_id}RouteServiceCapacityTotalMeausre |
| | tor:{route_id}RouteServiceCapacityTotalMeausre | rdf:type | hp:PassengerThroughputRate |
| | | | |
| | tor:{route_id}RouteServiceCapacityTotalMeausre | i72:hasNumericalValue | "{daily_passenger_throughput}" |
| | tor:{route_id}RouteServiceCapacityTotalMeausre | i72:hasUnit | hp:person_per_day |

## B.11 Transportation

### B.11.1 Toronto Transportation Network

| | |
|---|---|
| Dataset | ORN<br><br>https://geohub.lio.gov.on.ca/datasets/mnrf::ontario-road-network-orn-road-net-element |

https://data.urbandatacentre.ca/catalogue/gov-canada-c8719fa7-f09c-46fa-a928-f13dd3b613e5

| Field Name | Subject | Property | Object |
| --- | --- | --- | --- |
| OGF_ID | orn:roadLink_{OGF_ID} | genprop:hasIdentifier | {OGF_ID} |
| | orn:roadLink_{OGF_ID} | rdf:type | cdt:RoadLink |
| SPEED_LIMIT | orn:typicalRoadLinkUser | rdf:type | cdt: RoadLinkUser |
| | cdt:typicalRoadLinkUser | cdt:speedLimit | orn:speed_{OGF_ID} |
| | orn:speed_{OGF_ID} | rdf:type | cityunits:Speed |
| | orn:speed_{OGF_ID} | i72:hasValue | orn: speedMeasure_{OGF_ID} |
| | orn: speedMeasure_{OGF_ID} | rdf:type | i72: Measure |
| | orn: speedMeasure_{OGF_ID} | i72: hasNumericalValue | {SPEED_LIMIT} |
| | orn: speedMeasure_{OGF_ID} | i72:hasUnit | hp:kilometre_per_hour |
| ROAD_CLASS | orn:roadLink_{OGF_ID} | cdt:hasRoadClass | orn:roadClass_{ROAD_CLASS} |
| | orn:roadClass_{ROAD_CLASS} | rdf:type | cdt:RoadClass |
| | orn:roadClass_{ROAD_CLASS} | code:hasCode | orn:roadClass_ Code_ {ROAD_CLASS} |
| | orn:roadClass_ Code_ {ROAD_CLASS} | genprop:hasName | {ROAD_CLASS} |
| NUMBER_OF_LANES | orn:roadLink_{OGF_ID} | cdt:numLanes | {NUMBER_OF_ LANES} |
| geometry | orn:roadLink_{OGF_ID} | loc:hasLocation | orn:roadLinkLocation_{OGF_ID} |
| | orn:roadLinkLocation_{OGF_ID} | geo:asWKT | {geometry} |

**Dataset** ORN Capacities

**Notes** (synthetic data)

| **Subject** | **Property** | **Object** |
|---|---|---|
| orn:roadLink_{OGF_ID}Service | rdf:type | hp:TransportationNetworkService |
| orn:roadLink_{OGF_ID}Service | hp:providedFromSite | orn:roadLink_{OGF_ID} |
| orn:roadLink_{OGF_ID} | res:hasCapacity | orn:roadLinkCapacity_{OGF_ID} |
| orn:roadLink_{OGF_ID}Service | res:hasCapacity | orn:roadLinkCapacity_{OGF_ID} |
| orn:roadLinkCapacity_{OGF_ID} | rdf:type | hp:VehicleThroughputRate |
| orn:roadLinkCapacity_{OGF_ID} | i72:hasValue | orn:roadLinkCapacityMeasure_{OGF_ID} |
| orn:roadLinkCapacityMeasure_{OGF_ID} | i72:hasNumericalValue | {SPEED_LIMIT}*{NUMBER_OF_LANES}*{C_DENSITY({ROAD_CLASS})} |
| orn:roadLinkCapacityMeasure_{OGF_ID} | i72:hasUnit | hp:vehicles_per_hour |
| orn:roadLink_{OGF_ID} | res:capacityInUse | orn:roadLinkCapacityUse_{OGF_ID} |
| orn:roadLink_{OGF_ID}Service | res:capacityInUse | orn:roadLinkCapacityUse_{OGF_ID} |
| orn:roadLinkCapacityUse_{OGF_ID} | rdf:type | hp:VehicleThroughputRate |
| orn:roadLinkCapacityUse_{OGF_ID} | i72:hasValue | orn:roadLinkCapacityUseMeasure_{OGF_ID} |
| orn:roadLinkCapacityUseMeasure_{OGF_ID} | i72:hasNumericalValue | {Capacity} * random.uniform(0.5, 0.95) |
| orn:roadLinkCapacityUseMeasure_{OGF_ID} | i72:hasUnit | hp:vehicles_per_hour |
| orn:roadLink_{OGF_ID} | res:hasAvailable Capacity | orn:roadLinkCapacityAvail_{OGF_ID} |
| orn:roadLink_{OGF_ID}Service | res:hasAvailable Capacity | orn:roadLinkCapacityAvail_{OGF_ID} |
| orn:roadLinkCapacityAvail_{OGF_ ID} | rdf:type | hp:AvailableVehicleThroughputRate |
| orn:roadLinkCapacityAvail_{OGF_ ID} | i72:hasValue | orn:roadLinkCapacityAvailMeasure_{OGF_ID} |
| orn:roadLinkCapacityAvailMeasur e_{OGF_ID} | i72:hasNumerica lValue | {Capacity} - {Capacity In Use} |
| orn:roadLinkCapacityAvailMeasur e_{OGF_ID} | i72:hasUnit | hp:vehicles_per_hour |

### B.11.2 National Road Network (NRN)

Note that the ORN mapping was already defined in the CDT, however moving forward mapping for the NRN may be the preferred approach as it applies throughout Canada. Regional and national road network datasets may be integrated with the NID identifier.

| | |
|---|---|
| Dataset | National Road Network (NRN) |
| | https://open.canada.ca/data/en/dataset/3d282116-e556-400c-9306-ca1a3cada77f |
| Notes | Should use this mapping instead of ORN moving forward |

| **Field Name** | **Subject** | **Property** | **Object** |
|---|---|---|---|
| NID | nrn:roadLink_{NID} | genprop:hasIdentifier | {NID} |
| | nrn:roadLink_{NID} | rdf:type | cdt:RoadLink |
| SPEED | nrn:typicalRoadLinkUser | rdf:type | cdt: RoadLinkUser |
| | cdt:typicalRoadLinkUser | cdt:speedLimit | nrn:speed_{NID} |
| | nrn:speed_{NID} | rdf:type | cityunits:Speed |
| | nrn:speed_{NID} | i72:hasValue | nrn: speedMeasure_{NID} |
| | nrn: speedMeasure_{NID} | rdf:type | i72: Measure |
| | nrn: speedMeasure_{NID} | i72: hasNumericalValue | {SPEED} |
| | nrn: speedMeasure_{NID} | i72:hasUnit | hp:kilometre_per_hour |
| ROADCLASS | nrn:roadLink_{OGF_ID} | cdt:hasRoadClass | nrn:roadClass_{ROADCLASS} |
| | nrn:roadClass_{ROADCLASS} | rdf:type | cdt:RoadClass |
| | nrn:roadClass_{ROADCLASS} | code:hasCode | nrn:roadClass_ Code_ {ROADCLASS} |
| | nrn:roadClass_ Code_ {ROADCLASS} | genprop:hasName | {ROADCLASS} |
| NBRLANES | nrn:roadLink_{NID} | cdt:numLanes | {NBRLANES} |
| geometry | nrn:roadLink_{NID} | loc:hasLocation | nrn:roadLinkLocation_{NID} |
| | nrn:roadLinkLocation_{NID} | geo:asWKT | {geometry} |

## B.12 Childcare

### B.12.1 Toronto Licensed Childcare Centres

Dataset: Licensed Childcare Centres

https://open.toronto.ca/dataset/licensed-child-care-centres/
Child care centres - 4326_fake occupancy.xslx

| **Field Name** | **Subject** | **Property** | **Object** |
|---|---|---|---|
| _id | tor:childcareservice_toronto{_id} | rdf:type | hp:ChildcareService |
| | tor:childcareservice_toronto{_id} | hp:providedFromSite | tor:childcareservice_toronto{_id}Site |
| LOC_ID | tor:childcareservice_toronto{_id}Site | genprop:hasIdentifier | {LOC_ID} |
| LOC_NAME | tor:childcareservice_toronto{_id}Site | genproper:hasName | {LOC_NAME} |
| geometry | tor:childcareservice_toronto{_id}Site | loc:hasLocation | tor:childcareservice_toronto{_id}SiteLoc |
| | tor:childcareservice_toronto{_id}SiteLoc | geo:asWKT | {geometry} |
| TOTSPACE | tor:childcareservice_toronto{_id} | res:hasCapacity | tor:childcareservice_toronto{_id}Capacity |
| | tor:childcareservice_toronto{_id}Capacity | rdf:type | hp:ChildcareEnrollmentSpaces |
| | tor:childcareservice_toronto{_id}Capacity | i72:hasValue | tor:childcareservice_toronto{_id}CapacityMeasure |
| | tor:childcareservice_toronto{_id}CapacityMeasure | i72:hasNumericalValue | {TOTSPACE} |
| | tor:childcareservice_toronto{_id}CapacityMeasure | i72:hasUnit | i72:population_cardinality_unit |
| FAKE OCCUPANCY | tor:childcareservice_toronto{_id} | res:capacityInUse | tor:childcareservice_toronto{_id}CapacityUse |
| | tor:childcareservice_toronto{_id}CapacityUse | rdf:type | hp:ChildcareEnrollmentSize |
| | tor:childcareservice_toronto{_id}CapacityUse | i72:hasValue | tor:childcareservice_toronto{_id}CapacityUseMeasure |
| | tor:childcareservice_toronto{_id}CapacityUseMeasure | i72:hasNumericalValue | {FAKE OCCUPANCY} |
| | tor:childcareservice_toronto{_id}CapacityUseMeasure | i72:hasUnit | i72:population_cardinality_unit |

### B.12.2 Open Street Map (OSM) Childcare Listings

Childcare services may be found in the Business license listing in Vancouver's open data portal but there is not a clear category for identification. No datasets were found for Halifax. Open

Street Map[29] (OSM) presents an alternative, though less official source for childcare services across Canada.

Dataset: Childcare (OSM)

using node["amenity"="childcare"]

Notes: The same mapping may be applied to OSM data for other municipalities

| Field Name | Subject | Property | Object |
|---|---|---|---|
| id | van:childcare_osm{id} | rdf:type | hp:ChildcareService |
| | van:childcare_osm{id} | hp:providedFromSite | van:childcare_osm{id}serviceSite |
| name | van:childcare_osm{id}serviceSite | genprop:hasName | "{name}" |
| lat, lon | van:childcare_osm{id}serviceSite | loc:hasLocation | van:childcare_osm{id}serviceSiteLoc |
| | van:childcare_osm{id}serviceSiteLoc | geo:asWKT | "POINT({lat} {lon})" |

## B.13 Community Centre

### B.13.1 Toronto Community Centres

Dataset: Parks and Recreation Facilities (focus on Recreation Facilities as parks are already mapped)

https://open.toronto.ca/dataset/parks-and-recreation-facilities/
https://data.urbandatacentre.ca/catalogue/toronto-parks-and-recreation-facilities

Notes: Filter for TYPE="Community Centre";
Other attributes that could be mapped but not required for the current use cases;
Look into integrating parks identified here with those already captured in the CDT (via OSM)

| Field Name | Subject | Property | Object |
|---|---|---|---|

[29] https://www.openstreetmap.org/

| | | | |
|---|---|---|---|
| _id | tor:communitycentre_service{_id} | rdf:type | hp:CommunityCentreService |
| ASSET_ID | tor:communitycentre_service{_id} | hp:providedFromSite | tor:communitycentresite{ASSET_ID} |
| | tor:communitycentresite{ASSET_ID} | genprop:hasIdentifier | "{ASSET_ID}" |
| ASSET_NAME | | genprop:hasName | "{ASSET_NAME}" |
| geometry | | loc:hasLocation | tor:communitycentresite{ASSET_ID}_location |
| | tor:communitycentresite{ASSET_ID}_location | geo:asWKT | "{geometry}" |
| FAKE CAPACITY | tor:communitycentre_service{_id} | res:hasCapacity | tor:communitycentre_service{_id}Capacity |
| | tor:communitycentre_service{_id}Capacity | rdf:type | hp:CommunityCentreClientSpaces |
| | tor:communitycentre_service{_id}Capacity | i72:hasValue | tor:communitycentre_service{_id}CapacityMeasure |
| | tor:communitycentre_service{_id}CapacityMeasure | i72:hasNumericalValue | {FAKE CAPACITY} |
| | tor:communitycentre_service{_id}CapacityMeasure | i72:hasUnit | i72:population_cardinality_unit |
| | tor:communitycentre_service{_id} | res:capacityInUse | tor:communitycentre_service{_id}CapacityUse |
| | tor:communitycentre_service{_id}CapacityUse | rdf:type | hp:CommunityCentreClientSize |
| | tor:communitycentre_service{_id}CapacityUse | i72:hasValue | tor:communitycentre_service{_id}CapacityUseMeasure |
| | tor:communitycentre_service{_id}CapacityUseMeasure | i72:hasNumericalValue | 13,903 |
| | tor:communitycentre_service{_id}CapacityUseMeasure | i72:hasUnit | i72:population_cardinality_unit |

### B.13.2 Vancouver Community Centres

Dataset: Community centres

https://opendata.vancouver.ca/explore/dataset/community-centres

| Field Name | Subject | Property | Object |
|---|---|---|---|
| OBJECTID | van:{OBJECTID}communitycentreservice | rdf:type | hp:CommunityCentreService |
| | van:{OBJECTID}communitycentreservice | hp:providedFromSite | van:{OBJECTID}communitycentreserviceSite |
| NAME | van:{OBJECTID}communitycentreserviceSite | genprop:hasName | {NAME} |
| Geom | van:{OBJECTID}communitycentreserviceSite | loc:hasLocation | van:{OBJECTID}communitycentreserviceSiteLoc |
| | van:{OBJECTID}communitycentreserviceSiteLoc | geo:asWKT | {Geom} |

### B.13.3 Halifax Community Centres

Dataset: Parks & Rec

https://data-hrm.hub.arcgis.com/datasets/HRM::hrm-park-recreation-features-2/

Notes: Requires appropriate filter; unclear whether dataset is complete with respect to community centres

| Field Name | Subject | Property | Object |
|---|---|---|---|
| Fac_ID | hrm:{Fac_ID}commcentre_service | rdf:type | hp:CommunityCentreService |
| | hrm:{Fac_ID}commcentre_service | hp:providedFromSite | hrm:{Fac_ID}commcentre_serviceSite |
| | hrm:{Fac_ID}commcentre_serviceSite | loc:hasLocation | hrm:{Fac_ID}commcentre_serviceSiteLoc |
| SHAPE | hrm:{Fac_ID}commcentre_serviceSiteLoc | geo:asWKT | {SHAPE} |
| Park Facility Name | hrm:{Fac_ID}commcentre_serviceSite | genprop:hasName | {Park Facility Name} |

## B.14 Library

### B.14.1 Toronto Libraries

Dataset: Library

https://data.urbandatacentre.ca/catalogue/city-toronto-toronto-public-library-branch-locations

Notes: KG includes mappings for both Toronto-published and OSM library data

| **Field Name** | **Subject** | **Property** | **Object** |
|---|---|---|---|
| _id | tor:library_{_id} | genprop:hasIdentifier | {_id} |
| | tor:library_{_id} | rdf:type | cdt:Library |
| | tor:addLibrary_{_id} | rdf:type | contact:Address |
| | tor:library_site_{_id} | rdf:type | cdt:LibrarySite |
| | tor:library_{_id} | org:hasSite | tor:library_site_{_id} |
| | tor:library_site_{_id} | org:siteAddress | tor:addLibrary_{_id} |
| | tor:library_{_id} | cdt:providesService | tor:library_service{_id} |
| | tor:library_service{_id} | rdf:type | hp:LibraryService |
| | tor:library_service{_id} | hp:providedFromSite | tor:library_site_{_id} |
| geometry | tor:library_site_{@id} | loc:hasLocation | tor:location_{@id} |
| | loc:location_{@id} | rdf:type | loc:Location |
| | loc:location_{@id} | geo:asWKT | {geometry} |
| SquareFootage | tor:library_site_{_id} | hp:hasFloorArea | tor:library_site_{_id}FloorArea |
| | tor:library_site_{_id}FloorArea | rdf:type | hp:FloorArea |
| | tor:library_site_{_id}FloorArea | i72:hasValue | tor:library_site_{_id}FloorAreaMeasure |
| | tor:library_site_{_id}FloorAreaMeasure | i72:hasNumericalValue | {SquareFootage} |
| | tor:library_site_{_id}FloorAreaMeasure | i72:hasUnit | hp:square_foot |
| | tor:library_service{_id} | res:hasCapacity | tor:library_service{_id}Capacity |
| | tor:library_service{_id}Capacity | rdf:type | hp:MinLibraryAreaPopulationRatio |
| | tor:library_service{_id}Capacity | i72:hasValue | tor:library_service{_id}Capacity Measure |
| | tor:library_service{_id}Capacity Measure | i72:hasNumericalValue | 1 |
| | tor:library_service{_id}Capacity Measure | i72:hasUnit | hp:square_metre_per_person |
| | tor:library_service{_id}Capacity Measure | res:capacityInUse | tor:library_service{_id}CapacityUse |
| | tor:library_service{_id}CapacityUse | rdf:type | hp:LibraryAreaPopulationRatio |
| | tor:library_service{_id}CapacityUse | i72:hasValue | tor:library_service{_id}CapacityUseMeasure |
| | tor:library_service{_id}CapacityUseMeasure | i72:hasNumericalValue | {SquareFootage in metres} / 55613 |
| | tor:library_service{_id}CapacityUseMeasure | i72:hasUnit | hp:square_metre_per_person |

### B.14.2 Vancouver Libraries

Dataset: Libraries

https://opendata.vancouver.ca/explore/dataset/libraries/information/

| Field Name | Subject | Property | Object |
|---|---|---|---|
| OBJECTID | van:{OBJECTID}libraryservice | rdf:type | hp:LibraryService |
| | van:{OBJECTID}libraryservice | hp:providedFromSite | van:{OBJECTID}libraryserviceSite |
| NAME | van:{OBJECTID}libraryserviceSite | genprop:hasName | {NAME} |
| Geom | van:{OBJECTID}libraryserviceSite | loc:hasLocation | van:{OBJECTID}libraryserviceSiteLoc |
| | van:{OBJECTID}libraryserviceSiteLoc | geo:asWKT | {Geom} |

### B.14.3 Halifax Libraries

No datasets found. Libraries might be identified in the published Building data, however no attributes are defined that enable direct filtering and identification for this.

**Dataset:** Libraries (OSM)

**Notes:** May be applied generally to any municipality

| Field Name | Subject | Property | Object |
|---|---|---|---|
| @id | hrm:library_{@id} | genprop:hasIdentifier | {@id} |
| | hrm:library_{@id} | rdf:type | cdt:Library |
| | hrm:library_site_{@id} | rdf:type | cdt:Site |
| | hrm:library_{@id}Service | rdf:type | hp:LibraryService |
| | hrm:library_{@id} | cdt:providesService | hrm:library_{@id}Service |
| | hrm:library_{@id}Service | hp:providedFromSite | hrm:library_site_{@id} |
| geometry | hrm:library_site_{@id} | loc:hasLocation | hrm:location_{@id} |
| | hrm:location_{@id} | rdf:type | loc:Location |
| | hrm:location_{@id} | geo:asWKT | {geometry} |

## B.15 School

### B.15.1 Toronto (Ontario) Schools

Dataset Public school contact information

https://data.ontario.ca/en/dataset/ontario-public-school-contact-information
https://data.urbandatacentre.ca/catalogue/gov-canada-fb3a7c18-90af-453e-bc0a-a76ecc471862

| **Field Name** | **Subject** | **Property** | **Object** |
|---|---|---|---|
| School Number | tor:{School Number}School | rdf:type | cdt:ElementarySchool or cdt:SecondarySchool, as indicated by the "School Level" column |
| | tor:{School Number}School | cdt:providesService | tor:{School Number}SchoolService |
| | tor:{School Number}SchoolService | rdf:type | hp:SchoolService |
| | tor:{School Number}School | genprop:hasIdentifier | "{School Number}" |
| | tor:{School Number}School | genprop:hasName | "{School Name}" |
| Board Nu | tor:{Board Number}S | rdf:type | cdt:Organization |
| mber | choolBoard | | |
| | tor:{Board Number}SchoolBoard | org:hasSubOrganization | tor:{School Number}School |
| | tor:{Board Number}SchoolBoard | | |
| | tor:{Board Number}SchoolBoard | genprop:hasIdentifier | "{Board Number}" |
| Board Name | tor:{Board Number}SchoolBoard | genprop:hasName | "{Board Name}" |
| | tor:{School Number}SchoolService | rdf:type | hpcdm:ElementarySchoolService or hpcdm:SecondarySchoolService as appropriate (based on the "School Level" column) |
| *calculated value* | tor:{School Number}School | org:hasSite | tor:{School Number}SchoolSite |
| | tor:{School Number}SchoolService | hp:providedFromSite | tor:{School Number}SchoolSite |
| | tor:{School Number}SchoolSite | loc:hasLocaton | tor:{School Number}SchoolSiteLocation |
| | tor:{School Number}SchoolSiteLocation | geo:asWKT | {calculated coordinates based on address} |
| | tor:{School Number}SchoolSite | org:siteAddress | tor:{School Number}SchoolAddress |
| Street | tor:{School Number}SchoolAddress | rdf:type | contact:Address |
| | tor:{School Number}SchoolAddress | contact:hasStreetNumber | Needs to be extracted from {Street} |
| | tor:{School Number}SchoolAddress | contact:hasStreet (where appropriate, the information in addr:street is separated and represented in more detail using additional properties from the contact ontology e.g., contact:hasStreetType) | Needs to be extracted from {Street} |
| Postal Code | tor:{School Number}SchoolAddress | contact:hasPostalCode | "{Postal Code}" |

Dataset School Enrollment

https://data.ontario.ca/en/dataset/ontario-public-schools-enrolment
https://data.urbandatacentre.ca/catalogue/gov-canada-d89271cf-c5b7-4537-9d8b-5905766d93c6

Fake data from: enrolment_by_school_2324_en_fakeadded.xlsx

| **Field Name** | **Subject** | **Property** | **Object** |
|---|---|---|---|
| School Number | tor:{School Number}SchoolService | res:capacityInUse | tor:{School Number}SchoolServiceCapacityUse |
| | tor:{School Number}SchoolServiceCapacityUse | rdf:type | hp:SchoolEnrollmentSize |
| | tor:{School Number}SchoolServiceCapacityUse | i72:hasValue | tor:{School Number}SchoolServiceCapacityUseMeasure |
| Enrolment | tor:{School Number}SchoolServiceCapacityUseMeasure | i72:hasNumericalValue | "{Enrolment}" |
| | tor:{School Number}SchoolServiceCapacityUseMeasure | i72:hasUnit | i72:population_cardinality_unit |
| {time interval} | tor:{School Number}SchoolService | change:existsAt | tor:{time interval camelcase}Interval |
| | tor:{time interval camelcase}Interval | time:hasBeginning | tor:{time interval camelcase}BeginningTimeInstant |
| | tor:{time interval camelcase}BeginningTimeInstant | time:inXSDDateTimeStamp | {start of time interval converted to xsd datetime stamp} |
| | tor:{time interval camelcase}Interval | time:hasEnd | tor:{time interval camelcase}EndTimeInstant |
| | tor:{time interval camelcase}EndTimeInstant | time:inXSDDateTimeStamp | {end of time interval converted to xsd datetime stamp} |
| Fake Capacity | tor:{School Number}SchoolService | res:hasCapacity | tor:{School Number}SchoolServiceCapacity |
| | tor:{School Number}SchoolServiceCapacity | rdf:type | hp:SchoolEnrollmentSpaces |
| | tor:{School Number}SchoolServiceCapacity | i72:hasValue | tor:{School Number}SchoolServiceCapacityMeasure |
| | tor:{School Number}SchoolServiceCapacityMeasure | i72:hasNumericalValue | {Fake Capacity} |
| | tor:{School Number}SchoolServiceCapacityMeasure | i72:haUnit | i72:population_cardinality_unit |

### B.15.2 Vancouver Schools

Dataset: Schools

https://opendata.vancouver.ca/explore/dataset/schools/

| **Field Name** | **Subject** | **Property** | **Object** |
|---|---|---|---|
| OBJECTID | van:{OBJECTID}schoolservice | rdf:type | hp:SchoolService |
| | van:{OBJECTID}schoolservice | hp:providedFromSite | van:{OBJECTID}schoolserviceSite |
| SCHOOL_NAME | van:{OBJECTID}schoolserviceSite | genprop:hasName | {SCHOOL_NAME} |
| | van:{OBJECTID}schoolserviceSite | loc:hasLocation | van:{OBJECTID}schoolserviceSiteLoc |
| Geom | van:{OBJECTID}schoolserviceSiteLoc | geo:asWKT | {Geom} |

### B.15.3 Halifax Schools

Dataset: HRCE School Boundary - English Program Elementary

https://data-hrm.hub.arcgis.com/datasets/HRM::hrce-school-boundary-english-program-elementary/

Notes: Several other similar datasets for French language and other levels of school

| Field Name | Subject | Property | Object |
|---|---|---|---|
| School Code | hrm:{School Code}SchoolService | rdf:type | hp:SchoolService |
| | hrm:{School Code}SchoolService | hp:providedFromSite | hrm:{School Code}SchoolServiceSite |
| School Name | hrm:{School Code}SchoolServiceSite | genprop:hasName | {School Name} |
| Shape | hrm:{School Code}SchoolServiceSite | loc:hasLocation | hrm:{School Code}SchoolServiceSiteLoc |
| | hrm:{School Code}SchoolServiceSiteLoc | geo:asWKT | {Shape} |

## B.16 Parks

The current mapping utilizes data from Open Street Maps. This applies generally across municipalities; however, the dataset is not authoritative. Future work may implement similar mapping for government-published datasets.

**Dataset:** Parks (OSM)

**Note:** Applies generally to any municipality

| Field Name | Subject | Property | Object |
|---|---|---|---|
| Park | tor: {OSM ID}ParkOrg | rdf:type | org:Organization |
| | tor: {OSM ID}ParkService | rdf:type | hp:ParkService |
| | tor: {OSM ID}ParkOrg | cdt:providesService | tor: {OSM ID}ParkService |
| catchment area (manually defined) | tor: {OSM ID}ParkService | service:hasCatchmentArea | tor: {OSM ID}Catchment |
| | tor: {OSM ID}Catchment | geo:asWKT | 800m radius from {geometry} |
| geometry | tor: {OSM ID}ParkOrg | org:hasSite | tor: {OSM ID} ParkSite |
| | tor: {OSM ID}ParkService | hp:providedFromSite | tor: {OSM ID} ParkSite |
| | tor: {OSM ID} ParkSite | rdf:type | cdt:Park |
| | tor: {OSM ID} ParkSite | loc:hasLocation | tor: {OSM ID} ParkSiteLoc |
| | tor: {OSM ID} ParkSiteLoc | geo:asWKT | “{geometry}” (the geometry here is converted to WKT format) |
| id | tor: {OSM ID} ParkSite | genprop:hasIdentifier | “{OSM ID}” |
| name | tor: {OSM ID} ParkSite | genprop:hasName | “{name}” |
| | tor: {OSM ID} ParkSite | org:siteAddress | |
| Address<br><br>*Generally, the address information in OpenStreetMap is represented* | tor: {OSM ID}ParkAddress | rdf:type | contact:Address |
| *using multiple properties such as addr:housenumber, addr:street, and addr:postcode.* | | | |
| | tor: {OSM ID}ParkAddress | contact:hasStreetNumber | "{addr:housenumber}" |
| | tor: {OSM ID}ParkAddress | contact:hasStreet (where appropriate, the information in addr:street is separated and represented in more detail using additional properties from the contact ontology e.g., contact:hasStreetType) | "{addr:street}" |
| | tor: {OSM ID}ParkAddress | contact:hasPostalCode | "{addr:postcode}" |
| operator | tor:{operator}Operator | rdf:type | cdt:Organization |
| Name of the operator in charge of operating the entity. The name is converted to CamelCase when it is used as an IRI. | tor:{operator}Operator | org:hasSubOrganization | tor: {OSM ID}ParkOrg |
| Surface Area | tor: {OSM ID}ParkArea | rdf:type | cityunits:Area |
| The surface area information of the feature is not directly provided in the dataset but was calculated using the geospatial geometries. | tor: {OSM ID} ParkSite | cityunits:hasArea | tor: {OSM ID}ParkArea |
| | tor: {OSM ID}ParkArea | i72:hasValue | tor: {OSM ID}ParkAreaMeasure |
| | tor: {OSM ID}ParkAreaMeasure | rdf:type | i72:Measure |
| | tor: {OSM ID}ParkAreaMeasure | 21972:numerical_value | “{surface area}” |

**Dataset:** Synthetic park capacity

**Notes:** computed via OSM data

| Field Name | Subject | Property | Object |
|---|---|---|---|
| | tor: {OSM ID}ParkService | res:capacityInUse | tor: {OSM ID}ParkServiceCapacityUse |
| | tor: {OSM ID}ParkServiceCapacityUse | rdf:type | hp:RecreationAreaPopulationRatio |
| | tor: {OSM ID}ParkServiceCapacityUse | i72:hasValue | tor: {OSM ID}ParkServiceCapacityUseMeasure |
| | tor: {OSM ID}ParkServiceCapacityUseMeasure | i72:hasNumericalValue | {surface area} / 8855 |
| | tor: {OSM ID}ParkServiceCapacityUseMeasure | i72:hasUnit | hp:square_metres_per_person |
| | tor: {OSM ID}ParkService | res:hasCapacity | tor: {OSM ID}ParkServiceCapacity |
| | tor: {OSM ID}ParkServiceCapacity | rdf:type | MinRecreationAreaPopulationRatio |
| | tor: {OSM ID}ParkServiceCapacity | i72:hasValue | tor: {OSM ID}ParkServiceCapacityMeasure |
| | tor: {OSM ID}ParkServiceCapacityMeasure | i72:hasNumericalValue | 20 |
| | tor: {OSM ID}ParkServiceCapacityMeasure | i72:hasUnit | hp:square_metres_per_person |

## B.17 Medical (hospitals)

The current mapping utilizes data from Open Street Maps. This applies generally across municipalities; however, the dataset is not authoritative. Future work may implement similar mapping for government-published datasets.

**Dataset:** Hospitals (OSM)

**Notes:** Applies generally, across municipalities

| Field Name | Subject | Property | Object |
|---|---|---|---|
| Hospital | tor: {OSM ID}Hospital | rdf:type | gcih:Hospital (or appropriate subclass of sc:Hospital from the GCIH ontology) |
| emergency | tor: {OSM ID}HospitalEmergencyDepartment | rdf:type | org:OrganizationalUnit |
| Indicates whether the hospital provides emergency services | tor: {OSM ID}Hospital | org:hasUnit | tor: {OSM ID}HospitalEmergencyDepartment |
| | tor: {OSM ID}HospitalEmergencyService | rdf:type | cdt:HospitalEmergencyService |
| | tor: {OSM ID}HospitalEmergencyService | rdf:type | hp:HospitalEmergencyService |
| | tor: {OSM ID}Hospital | cdt:providesService | tor: {OSM ID}HospitalService |
| | tor: {OSM ID}HospitalEmergencyDepartment | cdt:providesService | tor: {OSM ID}HospitalEmergencyService |
| geometry | cdt: {OSM ID} Hospital | org:hasSite | tor: {OSM ID} HospitalSite |
| | tor: {OSM ID}HospitalService | hpcdm:providedFromSite | tor: {OSM ID} HospitalSite |
| | tor: {OSM ID}HospitalService | rdf:type | HospitalService |
| | tor: {OSM ID} HospitalSite | rdf:type | cdt:Site |
| | tor: {OSM ID} HospitalSite | loc:hasLocation | tor: {OSM ID} HospitalSiteLocation |
| | tor: {OSM ID} HospitalSiteLocation | | "{geometry}" (the geometry here is converted to WKT format) |
| id | tor: {OSM ID}Hospital | genprop:hasIdentifier | "{OSM ID}" |
| name | tor: {OSM ID}Hospital | genprop:hasName | "{name}" |
| Address<br><br>*Generally, the address information in OpenStreetMap is represented using multiple properties such as addr:housenumber, addr:street, and addr:postcode.* | tor: {OSM ID}HospitalAddress | rdf:type | contact:Address |
| | tor: {OSM ID}Hospital | org_city:orgAddress | tor: {OSM ID}HospitalAddress |
| | tor: {OSM ID} HospitalSite | org:siteAddress | tor: {OSM ID}HospitalAddress |
| | tor: {OSM ID}HospitalAddress | contact:hasStreetNumber | "{addr:housenumber}" |
| | tor: {OSM ID}HospitalAddress | contact:hasStreet (where appropriate, the information in addr:street is separated and represented in more detail using additional properties from the contact ontology e.g., contact:hasStreetType) | "{addr:street}" |
| | tor: {OSM ID}HospitalAddress | contact:hasPostalCode | "{addr:postcode}" |
| operator | tor:{operator} | rdf:type | cdt:Organization |
| Name of the operator in charge of operating the entity. The name is converted to CamelCase when it is used as an IRI. | tor:{operator} | org:hasSubOrganization | tor: {OSM ID}Hospital |

**Dataset:** Hospital utilization

Canadian Institute for Health Information. Hospital Occupancy Rate. Accessed November 19, 2025. via https://www.cihi.ca/en/access-data-and-reports/indicator-library/download-indicator-data

No bed count available for beds per capita ratio, but we can use the utilization indicator published by CIHI Note: most of the data in the CIHI occupancy rate is for hospital organizations

(that include several hospital sites).

Future implementations should look into a breakdown of these values

**Notes:** (at minimum a definition of the hospital services as subactivities of the larger organizations' services)

| Field Name | Subject | Property | Object |
|---|---|---|---|
| | tor: {OSM ID}Hospital[30] | res:capacityInUse | tor: {OSM ID}HospitalCapacityUse |
| | tor: {OSM ID}HospitalCapacityUse | i72:hasValue | tor: {OSM ID}HospitalCapacityUseMeasure |
| Metric value | tor: {OSM ID}HospitalCapacityUseMeasure | i72:hasNumericalValue | "{Metric value}" |
| | tor: {OSM ID}HospitalCapacityUseMeasure | i72:hasUnit | hp:avg_inpatients_daily_per_bed |

[30] Aligned to Hospital identifier via manual inspection

| | | | |
|---|---|---|---|
| | tor: {OSM ID}Hospital | res:hasCapacity | tor: {OSM ID}HospitalCapacity |
| | tor: {OSM ID}HospitalCapacity | i72:hasValue | tor: {OSM ID}HospitalCapacityMeasure |
| | tor: {OSM ID}HospitalCapacityMeasure | i72:hasNumericalValue | 1 |
| | tor: {OSM ID}HospitalCapacityMeasure | i72:hasUnit | hp:avg_inpatients_daily_per_bed |

## B.18 Food (supermarkets)

**Dataset:** Food (supermarkets)

**Notes:** via OSM, applies generally across municipalities

| Field Name | Subject | Property | Object |
|---|---|---|---|
| Supermarket | cdt: {OSM ID}Supermarket | rdf:type | cdt:Supermarket |
| | cdt: {OSM ID}SupermarketService | rdf:type | hp:SupermarketService |
| | cdt: {OSM ID}SupermarketService | service:hasCatcmentArea | cdt: {OSM ID}SupermarketServiceCatchmentLoc |
| | cdt: {OSM ID}SupermarketServiceCatchmentLoc | geo:asWKT | computed polygon 5000m boundary from {geometry} |
| geometry | cdt: {OSM ID} Supermarket | org:hasSite | cdt: {OSM ID} SupermarketSite |
| | cdt: {OSM ID}SupermarketService | hp:providedFromSite | cdt: {OSM ID} SupermarketSite |
| | cdt: {OSM ID}SupermarketSite | rdf:type | cdt:Site |
| | cdt: {OSM ID}SupermarketSite | loc:Location | "{geometry}" (the geometry here is converted to WKT format) |
| id | cdt: {OSM ID}Supermarket | genprop:hasIdentifier | "{OSM ID}" |
| name | cdt: {OSM ID}Supermarket | genprop:hasName | "{name}" |
| | cdt: {OSM ID}SupermarketSite | org:siteAddress | cdt: {OSM ID}SupermarketAddress |
| Address<br><br>*Generally, the address information in OpenStreetMap is represented using multiple properties such as addr:housenumber, addr:street, and addr:postcode.* | cdt: {OSM ID}SupermarketAddress | rdf:type | contact:Address |
| | cdt: {OSM ID}SupermarketAddress | contact:hasStreetNumber | "{addr:housenumber}" |
| | cdt: {OSM ID}SupermarketAddress | contact:hasStreet (where appropriate, the information in addr:street is separated and represented in more detail using additional properties from the contact ontology e.g., contact:hasStreetType) | "{addr:street}" |
| | cdt: {OSM ID}SupermarketAddress | contact:hasPostalCode | "{addr:postcode}" |
| operator | cdt:{operator} | rdf:type | cdt:Organization |
| | cdt:{operator} | org:hasSubOrganization | cdt: {OSM ID}Supermarket |

**Dataset**: Supermarket capacities

(none used; based on very rough estimates)

| Field Name | Subject | Property | Object |
|---|---|---|---|
| | cdt: {OSM ID}SupermarketService | res:hasCapacity | cdt: {OSM ID}SupermarketServiceCapacity |
| | cdt: {OSM ID}SupermarketServiceCapacity | i72:hasValue | cdt: {OSM ID}SupermarketServiceCapacity Measure |
| | cdt: {OSM ID}SupermarketServiceCapacity Measure | i72:hasNumerical Value | 0.001 |
| | cdt: {OSM ID}SupermarketServiceCapacity Measure | i72:hasUnit | hp:sites_per_person |
| | cdt: {OSM ID}SupermarketService | res:capacityInUse | cdt: {OSM ID}SupermarketServiceCapacity Use |
| | cdt: {OSM ID}SupermarketServiceCapacity Use | i72:denominator | cdt: {OSM ID}SupermarketServiceCatchmentPopSize |
| | cdt: {OSM ID}SupermarketServiceCatchmentPopSize | i72:hasNumerical Value | 22139 |

## B.19 Senior Care

### B.19.1 Toronto Senior Care

**Dataset:** Long-Term Care Locations, City Operated

https://open.toronto.ca/dataset/long-term-care-locations-city-operated/
https://data.urbandatacentre.ca/catalogue/city-toronto-long-term-care-locations-city-operated

**Notes:** Extended with fake data in: long_term_care_locations_wgs84_withfakeoccupancy.xlsx

| **Field Name** | **Subject** | **Property** | **Object** |
|---|---|---|---|
| FID | tor:seniorcare_service{FID} | rdf:type | hp:SeniorCareService |
| | tor:seniorcare_service{FID} | hp:providedFromSite | tor:seniorcare_service_site{FID} |
| ID | tor:seniorcare_service_site{FID} | genprop:hasIdentifier | "{ID}" |
| NAME | | genprop:hasName | "{NAME}" |
| BEDS | tor:seniorcare_service{FID} | res:hasCapacity | tor:seniorcare_service{FID}Capacity |
| | tor:seniorcare_service{FID}Capacity | rdf:type | hp:NumberOfLongTermCareBeds |
| | tor:seniorcare_service{FID}Capacity | i72:hasValue | tor:seniorcare_service{FID}CapacityMeasure |
| | tor:seniorcare_service{FID}CapacityMeasure | i72:hasNumericalValue | "{BEDS}" |
| | tor:seniorcare_service{FID}CapacityMeasure | i72:hasUnit | i72:population_cardinality_unit |
| geometry | tor:seniorcare_service_site{FID} | loc:hasLocation | tor:seniorcare_service_site_location{FID} |
| | tor:seniorcare_service_site_location{FID} | geo:asWKT | "{geometry}" |
| none (synthetic) | tor:seniorcare_service{FID} | res:capacityInUse | tor:seniorcare_service{FID}CapacityUse |
| | tor:seniorcare_service{FID}CapacityUse | rdf:type | hp:NumberOfLongTermCareResidents |
| | tor:seniorcare_service{FID}CapacityUse | i72:hasValue | tor:seniorcare_service{FID}CapacityUseMeasure |
| Fake occupancy | tor:seniorcare_service{FID}CapacityUseMeasure | i72:hasNumericalValue | {Fake occupancy} |
| | tor:seniorcare_service{FID}CapacityUseMeasure | i72:hasUnit | i72:population_cardinality_unit |

### B.19.2 Vancouver Senior Care

No datasets found

### B.19.3 Halifax Senior Care

No datasets found

## B.20 Environmental Risk

### B.20.1 Toronto Floodplains

**Dataset:** Floodline polygons

https://trca-camaps.opendata.arcgis.com/search?tags=flood%2520plain%2CFloodplain%2520Mapping

| Field Name | Subject | Property | Object |
|---|---|---|---|
| OBJECTID | tor:Floodplain{OBJECTID} | rdf:type | hp:FloodplainSegment |
| Watershed | tor:Floodplain{OBJECTID} | hp:forWatershed | tor:Watershed{Watershed} |
| | tor:Watershed{Watershed} | genprop:hasName | "{Watershed}" |
| | tor:FloodlineAreaLocation{OBJECTID} | hp:demonstratesRiskFactor | tor:Floodplain{OBJECTID} |
| | tor:FloodlineAreaLocation{OBJECTID} | loc:hasLocation | tor:FloodLinePolygon{OBJECTID} |
| geometry | tor:FloodLinePolygon{OBJECTID} | geo:asWKT | {geometry} |

### B.20.2 Vancouver Floodplains

Dataset: Designated Floodplain

https://opendata.vancouver.ca/explore/dataset/designated-floodplain/

Notes: Omit year 2100 scenarios

| Field Name | Subject | Property | Object |
|---|---|---|---|
| OBJECTID | van:{OBJECTID}Flooplain | rdf:type | hp:FloodplainSegment |
| NAME | van:{OBJECTID}Flooplain | genprop:hasName | {NAME} |
| DESCRIPTION | van:{OBJECTID}Flooplain | genprop:hasDescription | {DESCRIPTION |
| Geom | van:{OBJECTID}Flooplain | loc:hasLocation | van:{OBJECTID}FlooplainPolygon |
| | van:{OBJECTID}FlooplainPolygon | geo:asWKT | {Geom} |

### B.20.3 Halifax Floodplains

Dataset: Flood Extents HRM Wide

https://data-hrm.hub.arcgis.com/documents/0d8154fd916e4688bc2f760e746f30ee/explore

Notes: Contains both coastal, pluvial and fluvial; Data availability TBD (some layers appear empty)

| Field Name | Subject | Property | Object |
|---|---|---|---|
| Flood Extent ID | hrm:{Flood Extent ID} | rdf:type | hp:FloodplainSegment |
| | hrm:{Flood Extent ID} | genprop:hasIdentifier | {Flood Extent ID} |
| Shape | hrm:{Flood Extent ID} | loc:hasLocation | hrm:{Flood Extent ID}Loc |
| | hrm:{Flood Extent ID}Loc | geo:asWKT | {Shape} |

# Appendix C Toronto Data Overview

The following tables identify, for each CQ, the dataset(s) used and data gaps identified for the Toronto implementation.

*Table 52: Datasets and data gaps for Land Use and Zoning CQs*

| CQ | Toronto Datasets | Toronto Data Gaps |
|---|---|---|
| **Theme: Land Use and Zoning** | | |
| 1. Where in the city does there exist vacant parcels of land? | Toronto property boundaries,<br>Open Database of Buildings (ODB);<br>Approximated parcel-building relationship with spatial join merged datasets on arcgis at https://utoronto.maps.arcgis.com/home/item.html?id=9d123bc3e0da4555abf5c88fd8bb7b1b<br>and<br>https://utoronto.maps.arcgis.com/home/item.html?id=1d271ca5c49e406ea4a25f32aa15e066<br>Merged to approximate building-parcel relationship | Infrastructure-parcel association |
| 1a. What is the size of the parcel? | Toronto property boundaries | |
| 1b. What is the perimeter of the parcel? | Toronto property boundaries | |
| 2. Who owns parcel x? | Integrated government ownership dataset | Complete ownership data |
| 3. What use is parcel x zoned for, e.g., residential (single family, multifamily), commercial, mixed, industrial? | Toronto Zoning By-law | Use breakdown (beyond zoning type} |
| 3a. What is it currently being used for? | None | No authoritative data on current use (potential to approximate or infer based on other datasets e.g. |

| CQ | Toronto Datasets | Toronto Data Gaps |
|---|---|---|
| | | locations of buildings, organizations) |
| 3b. What is the current density of the neighbourhood? | Census and property boundaries | |
| 7a. What regulations exist on setbacks from road or adjacent parcels? | None | Setback regulations not encoded |
| 7b. What regulations exist on FSR (Floor Surface Ratio)? | Toronto Zoning By-law | |
| 7c. What height restriction exists? | Toronto Zoning By-law | |
| 7d. What minimum lot size restriction exists? | Toronto Zoning By-law | |
| 7e. Are there any (building) shape restrictions on the parcel? | None | Shape restrictions not encoded |
| 8a. How closely does recent development conform to setback regulations? | Open Database of Buildings, Missing setback regulations | Setback regulations not encoded; Setback not defined in building datasets |
| 8b How closely does recent development conform to | Open Database of Buildings, Toronto Zoning Bylaw | Actual FSI not published |

| CQ | Toronto Datasets | Toronto Data Gaps |
|---|---|---|
| FSR regulations? | | |
| 8c How closely does recent development conform to height restrictions? | Open Database of Buildings, Toronto Zoning Bylaw | |
| 8d. How closely does recent development conform to minimum lot size restrictions? | Open Database of Buildings, Toronto Zoning Bylaw | |
| 8e. How closely does recent development conform to shape restrictions? | Open Database of Buildings, Missing shape restrictions | Shape restrictions; Actual building shapes |
| 16. How many dwellings occupy the parcel? | Estimated via Census and property boundaries | No parcel-level data (estimates only) |
| 17. How many residents occupy the parcel? | Estimated via Census and property boundaries | No parcel-level data (estimates only) |
| 24b.What density policies apply to a parcel of land? | Toronto Zoning By-law | |
| 24c. Does the parcel have the potential for mixed use development ? Note: this is a variation on CQ-3 | Toronto Zoning By-law | incomplete data on association of infrastructure to parcels (requires estimation through spatial joins) |

| CQ | Toronto Datasets | Toronto Data Gaps |
|---|---|---|
| 31. What publicly owned parcels of land are available for current or future development? | Integrated government ownership dataset,<br>Toronto property boundaries,<br>Open Database of Buildings (ODB);<br>Approximated parcel-building relationship with spatial join<br>merged datasets on arcgis at<br>https://utoronto.maps.arcgis.com/home/item.html?id=9d123bc3e0da4555abf5c88fd8bb7b1b<br>and<br>https://utoronto.maps.arcgis.com/home/item.html?id=1d271ca5c49e406ea4a25f32aa15e066<br>Merged to approximate building-parcel relationship | |
| 32c. What is the planned zoning for the parcel? What is the planned land use for the parcel? | None | Plans |
| 37. What is the zoned capacity for a parcel of land? | Toronto Zoning By-law | Capacity is represented but not populated in the dataset |
| 40. What is the zoning of nearby parcels (note, this is a minor deviation from #3) | Toronto Zoning By-law | Use breakdown (beyond zoning type} |
| 41. Is the parcel currently occupied by any building(s)? | Toronto property boundaries,<br>Open Database of Buildings (ODB);<br>Approximated parcel-building relationship with spatial join<br>merged datasets on arcgis at<br>https://utoronto.maps.arcgis.com/home/item.html?id=9d123bc3e0da4555abf5c88fd8bb7b1b<br>and<br>https://utoronto.maps.arcgis.com/home/item.html?id=1d271ca5c49e406ea4a25f32aa15e066<br>Merged to approximate building-parcel relationship | |

| CQ | Toronto Datasets | Toronto Data Gaps |
|---|---|---|
| 41a. If so, what is the age and condition of the building? | Open Database of Buildings<br>original dataset from:<br>https://www150.statcan.gc.ca/n1/pub/34-26-0001/342600012018001-eng.htm<br>(parts 1 and 2 contain Toronto data) | Age not completely populated; no condition data |
| 41b. Does the building have occupants? | None | No occupant data (partial data might be inferred by integrating datasets on organization locations) |
| 50. What vacant land does the federal government own? Which part of government owns it? (variation on CQ1) | Integrated government ownership dataset,<br>Toronto property boundaries,<br>Open Database of Buildings (ODB);<br>Approximated parcel-building relationship with spatial join<br>merged datasets on arcgis at<br>https://utoronto.maps.arcgis.com/home/item.html?id=9d123bc3e0da4555abf5c88fd8bb7b1b<br>and<br>https://utoronto.maps.arcgis.com/home/item.html?id=1d271ca5c49e406ea4a25f32aa15e066<br>Merged to approximate building-parcel relationship | incomplete data on association of infrastructure to parcels (requires estimation through spatial joins) |
| 53. What buildings does the federal government own? | None;<br>Data is available but was missed:<br>https://www.tbs-sct.gc.ca/dfrp-rbif/opendata-eng.aspx<br>Use will require integration with Open Database of Buildings | |
| 54. Is the building vacant? Underutilized? | None;<br>Available for federal government buildings but missed (combine "occupancy" and "tenants" fields) | |
| 56. What is the current use of the building? | None;<br>Available for federal government buildings but missed | Available for federal government |

| CQ | Toronto Datasets | Toronto Data Gaps |
|---|---|---|
| | | buildings but missed |
| 57. What is the building zoned for? | Toronto Zoning By-law | |

| Table 53: Datasets and data gaps for Development Feasibility CQs**CQ** | **Toronto Datasets** | **Toronto Data Gaps** |
|---|---|---|
| **Development Feasibility** | | |
| 4. Does there exist any issues with the parcel, such as prior use (e.g., gas station, industrial waste) requiring remediation? | None | Development classifications; historical use |
| 5. Is the parcel accessible directly by road? | ORN (access approximated) | Parcel-level road access |
| 5a. Fire and emergency access? | Fire Services Run Areas<br>Fire Facility Locations | Parcel-level accessibility |
| 6a. Is the parcel serviced by water? | Watermain locations | Parcel-level service |
| 6b. Is the parcel serviced by wastewater? | Sewer main locations | Parcel-level service |
| 6c. Is the parcel serviced by electricity? | Feeder station coverage areas | Parcel-level service |
| 9. How close are burdensome facilities (industrial plants, burdensome services, roads of supra-local significance)? | ORN, Toronto Bylaw (land use) | Identification of burdensome facilities/services |
| 10a. What is the capacity for schools to absorb increases in population? | School enrollment, Synthetic data on maximum capacities | School capacity (maximum) |
| 10b. What is the capacity for water services to absorb increases in population? | Water billing (use), Synthetic data on maximum capacities | Water service capacities |
| 10c. What is the capacity for wastewater to absorb increases in population? | Synthetic data on capacities (max, in-use) | Wastewater capacities |
| 10d. What is the capacity for transport to absorb increases in population? | Synthetic data on vehicle througput | Transportation network capacities |
| 11a. What is the population of the neighbourhood? | 2016 Census | No neighbourhood-level data |
| 11b. What is the annual income (average) of the neighbourhood? | 2016 Census | No neighbourhood-level data |
| 11c. What is the average rooms per home in the neighbourhood? | 2016 Census | No neighbourhood-level data |
| 11d. What is the average land per home in the neighbourhood? | Toronto propery boundaries | |
| 28c. Is the area currently serviced by electricity? What is the capacity? | Toronto Hydro available capacity map: provides ranges (used to generate service areas and synthetic available capacity data) | Parcel-level service; (max) capacity and capacity in-use |
| 28d. Is the area currently serviced by waste disposal? What is the capacity? | Solid Waste Daytime Curbside Collection Areas; Synthetic capacity data | Solid waste capacities |
| 28f. Is the area currently serviced by fire and emergency services? What is the capacity? | Fire Services Run Areas Fire Facility Locations; Synthetic capacities | Fire and emergency capacities |
| 35a. For a parcel of land: What public utilities currently service the area? What is the capacity? What are the future expansion plans? | Water billing (use), Synthetic data on maximum capacities; Toronto Hydro available capacity map: provides ranges (used to generate service areas and synthetic available capacity data); … | Utility capacities; Utility expansion plans |
| 37a. How do stated zoning capacities differ with the capacities of built projects over the past X years? | Census to approximate number of dwellings per parcel | Zoning bylaw max dwelling units not populated;<br>No parcel-level capacity data |

*Table 54: Datasets and data gaps for Development Desirability CQs*

| CQ | Toronto Datasets | Toronto Data Gaps |
|---|---|---|
| **Development desirability (quality)** | | |
| 12. What is the amenity score for the parcel? | Amenity locations : OSM, Toronto Licensed Childcare Centres, Parks and Recreation Facilities , Toronto Libraries,… | |
| 12a. Is it 15 minutes to transportation? | ORN | |
| 12b. Is it 15 minutes to schools? (What schools are in the area?) What is their capacity? What is their planned capacity? | Ontario Public school contact information, includes enrollment | Capacity data (max) |
| 12c. Is it 15 minutes to food? | OSM Supermarket locations | |
| 12d. Is it 15 minutes to medical services? (What medical facilities and pharmacies are in the area?) What is their capacity? What is their planned capacity? | OSM Hospital locations, CIHI hospital occupancy rates. | Capacity data (incomplete) |
| 12e. Is it 15 minutes to parks? (What parks are in the area?) What is their planned capacity? | OSM Park locations | Capacity data |
| 12f. Is it 15 minutes to libraries? (What libraries are in the area?) What is their capacity? What is their planned capacity? | Toronto library locations | Capacity data |
| 13. What support services exist for lower income families? | Library and Community Centre locations | |
| 13a. Community centre/programmes | Toronto community centre locations | |
| 13e. Senior services/home care | Long-Term Care Locations, City Operated | Capacity data |
| 29b. What parks are in the area? What is their capacity? | OSM Park locations, synthetic capacity data | Capacity data |
| 29d. What childcare facilities are in the area? What is their capacity? What is their planned capacity? | None | Planned capacity data |
| 30a. What are the plans for greenspace in the area? | None | Land use plans |
| 30b. What is the planned density for the area? | None | Zoning plans |
| 64. Is the vacant land close to a population centre? | Census Population Centres | Vacant land |
| 66. Is the vacant land close to employment opportunities? | Toronto zoning bylaw (land use) | Use breakdown (beyond zoning type} |
| 68. Is the building close to employment opportunities? | Toronto zoning bylaw (land use) | Use breakdown (beyond zoning type} |
| new: Is the land subject to any environmental risks (e.g. flooding)? | TRCA Floodline polygons | |

# Appendix D Synthetic Datasets

The approaches taken to synthetic data generation range from approximations to simplistic fabricated values, depending on the data available. For example, no data was available on parcel ownerships in Toronto, so this dataset has been created with fabricated persons using the Python Faker library. On the other hand, Toronto Hydro publishes ranges of available capacity data for areas serviced by feeder stations, so in this case these ranges could be used as the basis for (randomized) synthetic capacity values. In this project, the primary role of these datasets is to support the implementation and evaluation of the CQs and to demonstrate the functionality of the City Digital Twin tool (Phase 4). The impact of approximated or synthetic data on the validity and trustworthiness of responses is an important consideration, but not one that is relevant at this stage. In future work, it will be important to consider how the gaps in available data may be filled, and how estimated data can be used to complement official data sources without confusing users or compromising the trustworthiness of the system.

In this section, we outline the approaches taken to generate the required synthetic data. Where possible, values were estimated based on available data. In other cases, values generated may be subject to basic constraints but otherwise random. It should be noted that the chosen "reasonable" parameters used in these approaches were selected with the use of ChatGPT.

## D.1 Building Parcels

In Ontario, there is no (available) data that explicitly connects buildings to parcels. To approximate this, the *occupies* relationship is defined between Building objects and Parcel objects based on a spatial join (overlaps) of the Canadian Open Building Database with the Property Boundaries published by the City of Toronto.

The outcome has limited accuracy as the property boundaries do not always completely contain a building. In some cases where a building overlaps with multiple properties, the result is a building that "occupies" multiple properties (rather than arbitrarily choosing a single property).

## D.2 Parcel Ownership

The Python Faker[31] library was used to generate names and PIN values for each parcel. A limitation on this approach is that it doesn't capture organization ownership (each property is associated with a person-owner). In addition, there is data on government-owned parcels that is also mapped into the implementation. This dataset has not been resolved with the Toronto property boundaries data, so some parcels may be captured twice (once with a fake owner, and once with real, government ownership).

## D.3 Transportation Capacities

While some data is available on vehicle flow rates for selected parts of the transportation network, the data is not complete enough to provide information on transportation capacities throughout the city (e.g., near a particular parcel). The following approach is adopted to calculate randomized, synthetic data to represent vehicle flow rate as a measure of the capacity on some part of the road network.

---

[31] https://faker.readthedocs.io/en/master/

**Total Capacity**

A basic approach to calculate the vehicle flow for each road link as a measure of total capacity:

$$c_{total} = s \times n \times k$$

Where:

- $c_{total}$ is the capacity of the road link as vehicle flow (vehicles/hour)
- s is the speed limit of the road link (km/h)
- n is the number of lanes and
- k is the critical density of the road link (vehicles/km/lane)

Using some estimated critical densities (via ChatGPT):

"Freeway",26,

"Expressway / Highway",24,

"Arterial",20,

"Collector",18,

"Ramp",22,

"Local / Street",12,

"Local / Strata",12,

"Local / Unknown",12,

"Service",10,

"Alleyway / Laneway",10,

"Resource / Recreation",16,

"Rapid Transit",28,

"Winter",18,

20 /* default if unmatched */

**Capacity In-Use**

Some data on actual vehicle throughput is available; however, it is incomplete. To generate a complete set of synthetic data to represent vehicle throughput for each road link, a random factor drawn from a uniform distribution in the 0.5-0.95 range is applied to represent actual vehicle flow as a measure of utilization.

$$c_{use} = c_{total} \times x$$

$$where\ x \sim U(0.5, 0.95)$$

**Available Capacity**

Calculated as the difference between the total capacity and the capacity in use.

$$c_{avail} = c_{total} - c_{use}$$

Note that when the road is over capacity its vehicle flow will also be below its capacity (max rate), but due to congestion as opposed to under-use. Therefore, an in-use capacity (flow rate) below the total capacity could be indicative of over- or under-use of the road. To support interpretation of a flow rate as a measure of available capacity (and capacity in use), the average road speeds should ideally also be captured.

## D.4 Transit Capacity

The maximum daily ridership of a given transit route, as a measure of its total capacity, is estimated based on the number of trips scheduled (on a typical day) and the vehicles' capacities on the routes. Without precise knowledge of which vehicles styles (with varying capacity) are used on which routes, the following rough estimate is used based on the type of vehicle (e.g., in Toronto: bus, streetcar or subway):

- Streetcar: 130
- Subway: 1000
- Bus: 60

A simple calculation is then performed to provide an estimate of the maximum ridership per day for each route, based on the number of trips per day and the capacity of the vehicle type in the following format:

| route_id | route_name | route_type | vehicle_capacity | daily_trip_count_monday | daily_passenger_throughput |
|---|---|---|---|---|---|
| 75209 | LINE 1 (YONGE-UNIVERSITY) | 1 | 1000 | 601 | 601000 |
| 74986 | VAN HORNE | 3 | 60 | 42 | 2520 |
| … | | | | | |

The results are stored in TTC_est_throughput_report.csv. The actual use of the transit routes is available in published data on daily ridership.

## D.5 Water Capacity

Data on capacity in use is available in the form of annual water usage by ward, however beyond this no data is publicly available. The following approach is taken to generate random, plausible values of total capacity based on the reported usage numbers:

- The total service capacity for the ward is calculated based on usage data, using a random multiplier to approximate capacity between 10-30% above current use.
- The available capacity is then calculated as the difference between the total capacity and the actual capacity in use data.

The synthetic data is specified as an addition to the actual use data in "Water_Consumption_Capacity_2020.xlsx".

## D.6 Wastewater Capacity

No data is available on capacity (use or total) of the service in specific areas of the City of Toronto. In general, it's more reasonable to expect to have estimates for capacity in a specific service area, however in the absence of this information, synthetic values may be generated based on the attributes of the wastewater pipes (e.g. diameter). The resulting capacity data is then generated on a per-sewer main basis.

### D.6.1 For gravitational sewer mains

The approach uses Manning's formula to calculate a full flow rate based on pipe diameter and slope to generate an annual (max) flow capacity based on the following equation:

$$Q = 0.312 \times \frac{D^{8/3} \times S^{1/2}}{n}$$

Where: Q is the flow capacity (m3/year):

- D is the pipe diameter
- S is the slope of the pipe
- n is Manning's *n* (assumed to be constant 0.013, though for an improved approximation it should be based on the material).

A random, diameter-specific percent utilization is applied to the flow capacity to generate an annual capacity usage number for each pipe (via ChatGPT):

| Diameter range (mm) | Utilization range (%) | Typical sewer type |
|---|---|---|
| < 300 | 20–50 % | Local / residential laterals |
| 300–999 | 30–60 % | Main collectors |
| ≥ 1000 | 50–70 % | Trunks / interceptors |

### D.6.2 For pressurized sewer mains

Flow capacity is estimated based on the formula for flow rate:

$$Q = A \times V$$

Where:

- Q = flow rate (m3/s)
- A = pipe cross-sectional area
- V = flow velocity (m/s)

Utilizing the following values for typical flow velocity (via ChatGPT):

| Diameter (mm) | Typical velocity (m/s) | Notes |
|---|---|---|
| 75–150 | 0.6–1.2 | Small lines, short pumps |
| 200–400 | 0.8–1.5 | Common municipal force mains |
| 450–800 | 1.0–1.8 | Large trunk mains |
| 900–1200 | 1.0–2.0 | Interceptors / regional mains |
| >1200 | 1.2–2.5 | High-capacity transmission mains |

The flow rate is then multiplied by 31536000 for a measure of the maximum capacity as L/year.

A randomized utilization rate (based on the size of the pipe) is then applied to the capacity to generate a synthetic value for the capacity in use, based on the following ranges (via ChatGPT).

| Diameter range (mm) | Typical utilization range (% of full capacity) | Sewer type |
|---|---|---|
| <150 | 20 – 50 % | Small laterals / short pump runs |
| 150–300 | 30 – 60 % | Common small force mains |
| 300–450 | 40 – 70 % | Mid-size collectors |
| 450–800 | 50 – 80 % | Large force mains |
| >800 | 60 – 90 % | Trunks and interceptors |

## D.7 Solid Waste Capacity

- Capacity in use: Based on the statistic of approximately 830,000 tonnes of waste processed per year; a weight was applied (by catchment area size) to estimate the amount of waste processed per service area.
- A synthetic total capacity of the service is then estimated with a random factor between 1.1 to 1.25 (i.e., assume they try to maintain a capacity somewhere between 10 to 25% above the current use.
- The available capacity is then calculated as a difference between the two.

## D.8 Fire Services Capacity

- The synthetic population for each run (service) area is computed randomly as a value between 15000 to 50000. (Note: we can estimate this more accurately with a script based on overlap with census tract populations, however since the staffing numbers are synthetic too the utility of this would be minimal. To consider for future work.)
- A synthetic population of full-time firefighters is generated for each station as a random value between 5 and 20.
- These values are used to compute the capacity in use as a ratio of FT firefighters to population
- The available capacity is specified as a minimum of 0.0001 (based on a recommendation of 1 firefighter per 1000), though in practice this is subject to variability depending on the characteristics of the area.

## D.9 Electricity Services Capacity

- Referred to the capacity data published at Toronto Hydro
- Used the maximum of the provided range as an estimate for available capacity
- Randomly generated a value 3-6x greater than this to serve as a synthetic datapoint for total load capacity
    - Capacity in use may then be generated based on this

## D.10 Supermarket Capacity

The supermarket available capacity may be captures with the measure of SupermarketsPopulationRatio. This requires a count of supermarkets and resident population data for the supermarket's catchment area.

We define a synthetic catchment area for each supermarket as the $5km^2$ area surrounding its location. However, the true catchment area will vary depending on the store and its location (e.g. rural vs urban).

Data on the number of supermarkets within a particular supermarket's catchment area may be computed on-the-fly via a SPARQL query, however resident population counts aren't currently available for the catchment areas[32]. For the current implementation we construct a synthetic population for each supermarket's catchment area based on the catchment area size and the population density of the city.

---

[32] Note that this functionality (computing population estimates on the fly) will be pursued in the next stage of development.

- Toronto density (2021 Census): 4,427.8 people per square km
- 5 square km population estimate: 4,427.8 * 5 = 22139 people

The supermarket's capacity is estimated with the measure of MinSuperMarketsPopulationRatio. For the current implementation, a value of 1 supermarket per 1000 residents is defined for all supermarkets. This may be adjusted in the future to better reflect the requirements of a different types of areas (e.g. rural vs urban).

## D.11 Park Capacity

To define the capacity measure of RecreationAreaPopulationRatio requires values of *total* park area and the surrounding population in a given catchment area. Both values are available but not defined for the parks' catchment areas, which is what is required in order to calculate the capacity for a particular park. It is possible to define an approximation of this ratio based on a population estimate via spatial overlaps with areas of known populations, however this function has not yet been implemented.

True catchment areas of the parks are also not known, a synthetic catchment area of an 800m radius is defined, however in reality the catchment areas will vary depending on the park type. A synthetic population is estimated based on the population size, using a recent population density for the city of Toronto.

- Catchment area: 800m radius from Park location (or center of geometry); approx. $2km^2$ area
- Capacity (RecreationAreaPopulationRatio)
  - Total recreation area in the catchment area: park area (currently, assume it is the only park in the area)
    - To do: write query to sum all of the park areas within the catchment area
  - Population in the catchment area:
    - Toronto density (2021 Census): 4427.8 people per square km
    - square km population estimate: 4427.8 * 2= 8855 people
- MinRecreationAreaPopulationRatio: defined as 20 $m^2$/person as an estimate, though it is likely to be variable depending on the context.

## D.12 Senior Services Capacity

The City of Toronto provides data on the total number of beds for city operated long term care homes however no data is available on current occupancy. To complete the dataset for the purpose of testing, a synthetic occupancy dataset is created to complement the available data. The occupancy numbers are generated as a random estimate of between 95%-100% occupancy.

## D.13 Library Service Capacity

To define the capacity measure of LibraryAreaPopulationRatio requires values of libraries' floor areas and the surrounding population. The catchment area of the libraries are not defined, nor is specific population data likely to be available for any catchment area that may be defined. It is possible to define an approximation of this ratio based on a population estimate via spatial overlaps with areas of known populations, however this function has not yet been implemented.

True catchment areas of the libraries are also not defined, so a synthetic catchment area of a 2km radius is specified. Catchment areas are likely to vary depending on other factors such as the location of the library (e.g. is it an urban or rural setting). Future implementations should account for this if an improved approximation is to be used. A synthetic population is estimated based on the population size, using a recent population density for the city of Toronto.

- Catchment area: 2000m radius from Library location (or center of geometry); 12.56$km^2$ area
- Capacity (LibraryAreaPopulationRatio)
  - Total library floor area (available)
  - Population in the catchment area:
    - Toronto density (2021 Census): 4427.8 people per square km
    - square km population estimate: 4427.8 * 12.56 = 55613 people
- MinLibraryAreaPopulationRatio: defined as 1 $m^2$/person as an estimate, though it is likely to be variable depending on the context.

## D.14 School Capacity

The province of Ontario provides data on school enrollment; however, no data is available to capture the capacity of the schools. While there is some flexibility on a school's maximum enrollment (e.g. with the use of portables and timetabling to increase classroom use) there is still a limit to what is possible (without significant changes to infrastructure). Without any data on school capacities, we generate synthetic values for school capacities as follows:

Based on the Financial Accountability Office of Ontario (FAO) 2024 statistic of an average utilization rate of 87.6 per cent in 2023-2024, we apply a random factor of 87.6 (±5%).

## D.15 Child Care Capacity

The government publishes data on licensed childcare centres and the number of spaces for each; however, no data is provided on current enrollment. Synthetic data is generated to address this gap based on a random estimate of 95-100% occupancy for the centres.

## D.16 Community Centre Capacity

Data is available on the location of community centres, but no data is published related to the capacity of the centres and the services they provide. Synthetic data to approximate capacity as a ratio of community centres to service population is generated to address this gap, based on a simplistic estimated catchment area of a 2km radius. Note that future implementations should account for variations in catchment areas if an improved approximation is to be used. A synthetic population is estimated based on the population size, using a recent population density for the city of Toronto.

The total capacity is estimated based on the statistic of 1 community centre per 34,000 residents published in the Parks and Recreation Facilities Master Plan 2019-2038.

- Catchment area: 2km radius from community centre location; 3.14$km^2$ area
- Capacity in use (CommunityCentreClientSize)
  - Population in the catchment area:

- Toronto density (2021 Census): 4,427.8 people per square km
        - square km population estimate: 4,427.8 * 3.14 = 13,903 people
- Total capacity (CommunityCentreClientSpaces):
    - 34,000 plus or minus a random factor up to 20%

# Appendix E Future Dashboard Development Stages

As described in Section 4.2, while development focused on a basic dashboard for parcel-specific housing potential analysis, several future stages were conceived for future development work (and more are conceivable beyond this). The following provides an outline of extensions of the application to pursue in future work.

**Stage 2:**

Advanced filter-based parcel selection: depicted in Figure 39, instead of searching for a specific parcel based on its address an advanced search would allow the user to discover parcels based on the following set of filters:

- Vacancy (Y/N)
- Government owned (Y/N)
- Zoned for use (dropdown from existing values)
- Neighbourhood (dropdown from defined neighbourhoods)
- Parcel attributes:
  - Perimeter (range)
  - Area (Range)

A parcel may be selected from these results (Figure 40) for further queries in the current dashboard.

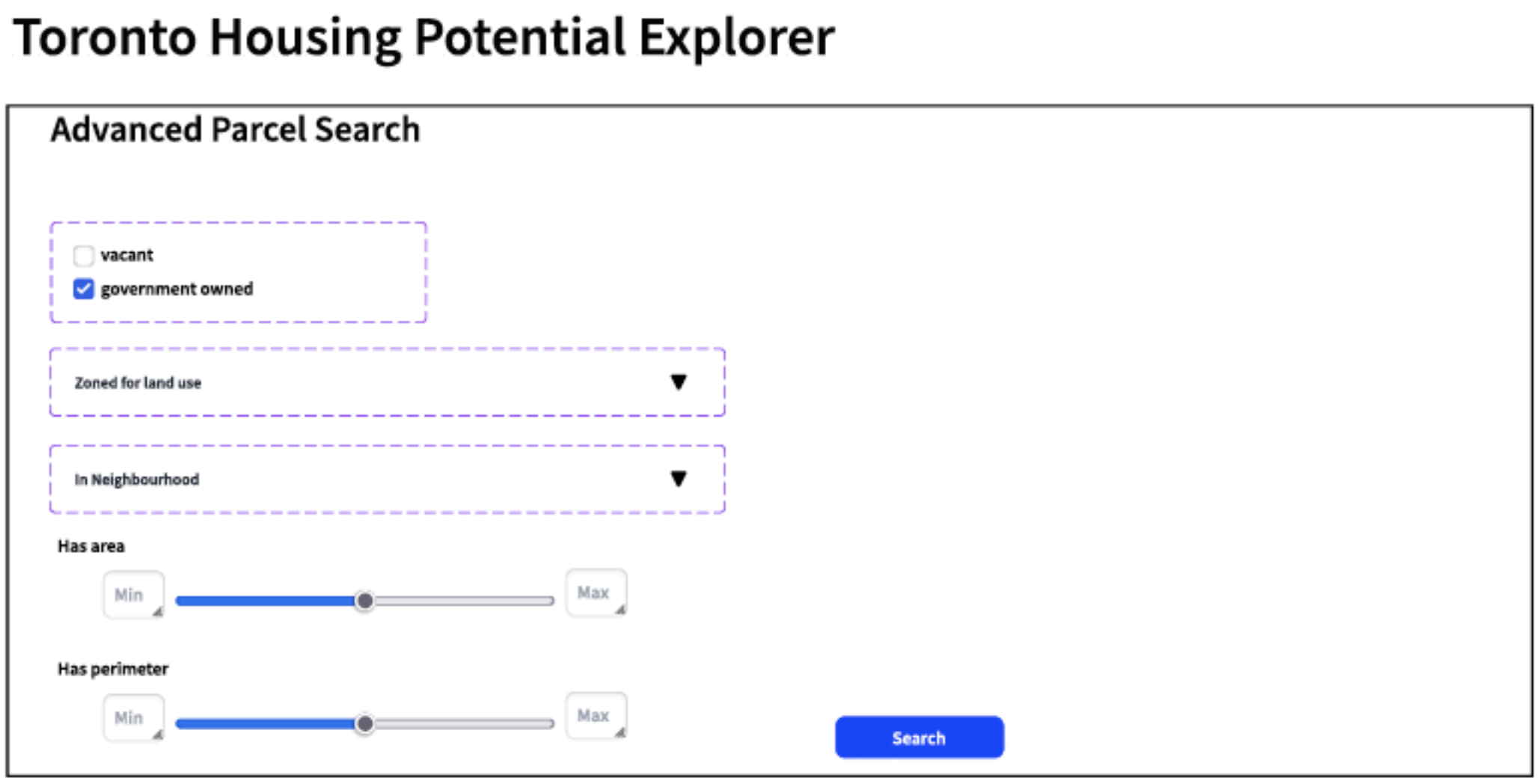

*Figure 39: View of Stage 2 advanced parcel search interface*

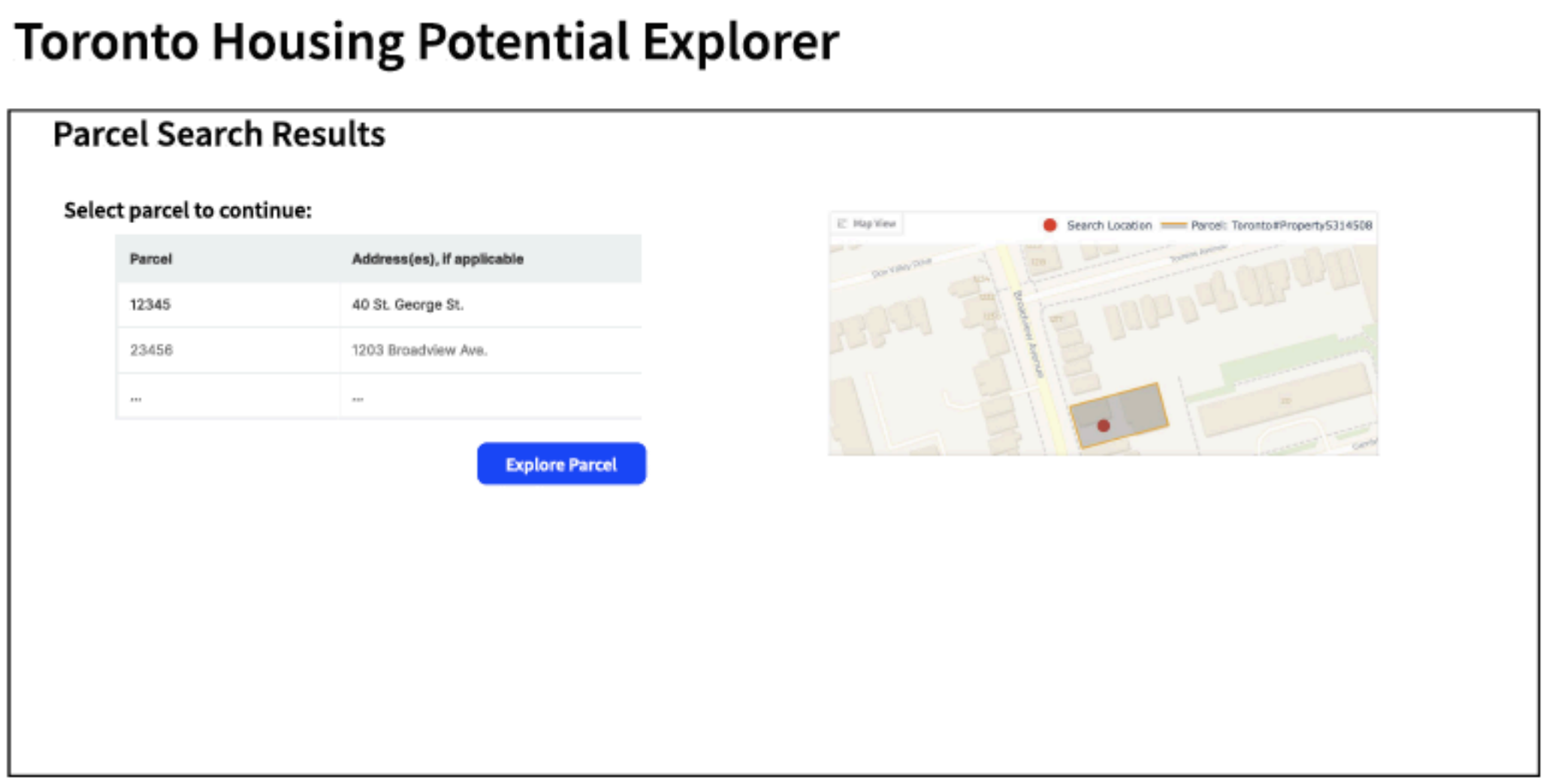

*Figure 40: View of Stage 2 parcel selection (results of advanced search)*

**Stage 3:**

Custom report generation: instead of a dropdown list to perform individual queries for a parcel (as in Stage 1), a “Dynamic report generation” interface will allow the user to pick and choose what attributes to include and then generate a report, on demand, accordingly.

**Stage 4:**

Service analysis: allow the user to investigate the capacity of a given service(s) for a parcel in more detail. Specifically, can the service (that is currently allocated to the parcel) support a population increase of X%? Services with per capita-based capacity measure could be computed directly, whereas others would require the user to input some additional estimated parameters.

# Appendix F SPARQL Queries

This section details the SPARQL queries used to generate application results. To distinguish between logic and data, queries follow a specific template syntax: single curly brackets {} identify template variables, while double curly brackets {{}} represent literal SPARQL brackets. The following prefixes are used throughout the queries specified in this section:

```
PREFIX bdg: <https://standards.iso.org/iso-iec/5087/-2/ed-1/en/ontology/Building/>
PREFIX cacensus: <http://ontology.eil.utoronto.ca/tove/cacensus#>
PREFIX code: <https://standards.iso.org/iso-iec/5087/-2/ed-1/en/ontology/Code/>
PREFIX genprop: <https://standards.iso.org/iso-iec/5087/-1/ed-1/en/ontology/GenericProperties/>
PREFIX geo: <http://www.opengis.net/ont/geosparql#>
PREFIX geoext: <http://rdf.useekm.com/ext#>
PREFIX geof: <http://www.opengis.net/def/function/geosparql/>
PREFIX hp: <http://ontology.eil.utoronto.ca/HPCDM/>
PREFIX i72: <http://ontology.eil.utoronto.ca/ISO21972/iso21972#>
PREFIX loc_old: <http://ontology.eil.utoronto.ca/5087/1/SpatialLoc/>
PREFIX loc: <https://standards.iso.org/iso-iec/5087/-1/ed-1/en/ontology/SpatialLoc/>
PREFIX opr: <http://www.theworldavatar.com/ontology/ontoplanningregulation/OntoPlanningRegulation.owl#>
PREFIX owl: <http://www.w3.org/2002/07/owl#>
PREFIX oz: <http://www.theworldavatar.com/ontology/ontozoning/OntoZoning.owl#>
PREFIX rdf: <http://www.w3.org/1999/02/22-rdf-syntax-ns#>
PREFIX rdfs: <http://www.w3.org/2000/01/rdf-schema#>
PREFIX res: <https://standards.iso.org/iso-iec/5087/-1/ed-1/en/ontology/Resource/>
PREFIX service: <https://standards.iso.org/iso-iec/5087/-2/ed-1/en/ontology/CityService/>
PREFIX time: <http://www.w3.org/2006/time#>
PREFIX tor: <http://ontology.eil.utoronto.ca/Toronto/Toronto#>
PREFIX uom: <http://www.opengis.net/def/uom/OGC/1.0/>
```

## Parcel Search

```
SELECT ?p ?wkt WHERE {{
  ?p a hp:Parcel ;
     loc:hasLocation ?loc .
   ?loc geo:asWKT ?wkt.
   BIND("{wkt_point}"^^geo:wktLiteral AS ?pwkt)
  ?loc geo:sfIntersects ?pwkt
}} LIMIT 1
```

## Parcel Attributes

```
SELECT ?attribute ?value ?unit WHERE {{

   <{pid}> a hp:Parcel;

       ?att ?q.

   ?q i72:hasValue [i72:hasNumericalValue ?value;

                          i72:hasUnit ?u].

   ?att rdfs:label ?attribute.

   ?u rdfs:label ?unit.

 # Filter out ?attribute if there exists a more specific sub-property (?sub) that defines the value for the parcel

 FILTER NOT EXISTS {{

   ?sub rdfs:subPropertyOf+ ?att .

   <{pid}> ?sub ?q .
```

```
    FILTER (?sub != ?att)
  }}
}}
```

## Land Use

```
SELECT ?allowed_use ?current_use WHERE {{
    #if we are looking for the neighbourhood of a specific parcel
    <{pid}> hp:zonedAsType ?zt.
    ?zt oz:allowsUse ?u.
    ?u genprop:hasName ?allowed_use.
    OPTIONAL {{
        ?x bdg:use [code:hasCode [genprop:hasName ?current_use]].
        ?x hp:occupies ?p.
    }}
}}
```

## Neighbourhood Demographics

Note: the following query is constructed iteratively to allow for varying lists of census characteristics to be specified. It takes the following form:

```
SELECT ?xlabel ?neighbourhood_name ?population ?unit_label ?unit ?ct ?cwkt WHERE {{
{stage_1}
    {{
{union of all characteristic blocks}
    }}
}}
```

Where stage 1 is "neighbourhood discovery"

```
    {{
        SELECT ?n ?neighbourhood_name WHERE {{
            <{pid}> loc:hasLocation ?ploc.
            ?n a tor:Neighborhood;
              loc_old:hasLocation ?nloc;
              rdfs:comment ?neighbourhood_name.
            ?ploc geo:sfWithin ?nloc.
        }}
    }}
```

And census characteristic blocks for the query are dynamically constructed based on a provided list (e.g. population density, average income, …), where each block takes the form:

```
{{
    ?x a {characteristic};
      cacensus:hasLocation ?characteristic_area;
      i72:hasValue [i72:hasNumericalValue ?population;
                    i72:hasUnit ?unit];
       rdfs:comment ?xlabel.   #indicator label
    ?characteristic_area tor:inNeighbourhood ?n;
       loc_old:hasLocation [geo:asWKT ?cwkt];
       rdfs:label ?ct. #census tract label
    OPTIONAL {{ ?unit rdfs:label ?unit_label. }}
  }}
```

## Available Services

Available services are queried in two parts. The first part queries for all types of services defined in the repository, and the second queries – for each service type – for its availability and capacity.

Query for service types:

```
SELECT distinct ?servicetype WHERE {{

    #service types (TBD: what level should we capture?)
    ?servicetype rdfs:subClassOf* hp:Service.
  #filter any classes that have subclasses
   FILTER NOT EXISTS {{
  ?sub rdfs:subClassOf* ?servicetype .
  FILTER (?sub != ?servicetype && ?sub !=owl:Nothing)
 }}
}}
```

Query for service details:

```
SELECT ?servicelabel ?servicename ?cap_type ?cap_avail ?cap_unit ?swkt WHERE {{
{{
#services with suitable catchment areas
  <{pid}> hp:servicedBy ?s;
    a hp:AdministrativeArea.
```

```
    ?s a <{servicetype}>;
       a hp:Service.
    <{servicetype}> rdfs:label ?servicelabel.
    #service site name, if defined
    OPTIONAL {{
       ?s hp:providedFromSite ?site.
    ?site genprop:hasName ?servicename.}}
    #TBD instead of showing site location (may not exist) return catchment area as swkt?
    #?s service:hasCatchmentArea [geo:asWKT ?swkt].
    #service capacity
    ?s res:hasAvailableCapacity ?cap.
    ?cap i72:hasValue [i72:hasNumericalValue ?cap_avail;
                                  i72:hasUnit ?u];
                   rdf:type ?cap_type_class.
    ?cap_type_class rdfs:label ?cap_type.
    ?u rdfs:label ?cap_unit.
    # Filter out the "Generics" (owl:Thing and owl:Nothing)
    FILTER(?cap_type_class != owl:Thing && ?cap_type_class != owl:Nothing)
    FILTER(!isBlank(?cap_type_class))

    # The Leaf Constraint:
    # Ensure there isn't another type on this node that is a SUBCLASS of our candidate.
    FILTER NOT EXISTS {{
       ?cap rdf:type ?moreSpecific .
       ?moreSpecific rdfs:subClassOf+ ?cap_type_class .

       # Standard safety filters
       FILTER(?moreSpecific != ?cap_type_class)
       FILTER(?moreSpecific != owl:Nothing)
    }}
}}
UNION
{{
    #services with suitable service radius
    #parcel location
    <{pid}> loc:hasLocation [geo:asWKT ?pwkt];
       a hp:AdministrativeArea.
```

```
#service site location(s)
?s a <{servicetype}>;
   a hp:Service;
   hp:providedFromSite ?site.
OPTIONAL {{?site genprop:hasName ?servicename.}}
?site loc:hasLocation ?sloc.
?sloc geo:asWKT ?swkt.
<{servicetype}> rdfs:label ?servicelabel.

#service-defined radius, in metres
?s hp:hasServiceRadius [i72:hasValue [i72:hasNumericalValue ?max_d;
                                      i72:hasUnit i72:metre]].

#(shortest) distance between the edge of the parcel and the service network
BIND(geof:distance(?pwkt, ?swkt, uom:metre) AS ?distance)
#limit distance to within the defined service radius
FILTER (?distance <= ?max_d)

#service capacity
?s res:hasAvailableCapacity ?cap.
?cap i72:hasValue [i72:hasNumericalValue ?cap_avail;
                   i72:hasUnit ?u];
          rdf:type ?cap_type_class.
?u rdfs:label ?cap_unit.
?cap_type_class rdfs:label ?cap_type.
# Filter out the "Generics" (owl:Thing and owl:Nothing)
FILTER(?cap_type_class != owl:Thing && ?cap_type_class != owl:Nothing)
FILTER(!isBlank(?cap_type_class))

# The Leaf Constraint:
# Ensure there isn't another type on this node that is a SUBCLASS of our candidate.
FILTER NOT EXISTS {{
   ?cap rdf:type ?moreSpecific .
   ?moreSpecific rdfs:subClassOf+ ?cap_type_class .

   # Standard safety filters
   FILTER(?moreSpecific != ?cap_type_class)
   FILTER(?moreSpecific != owl:Nothing)
```

```
    }}
}}
}}
```

## Land Use

The Land Use dropdown runs two separate queries – one to retrieve the allowed uses, and other to retrieve current use (based on building data, if available).

Allowed use:

```
SELECT ?allowed_use WHERE {{
    #if we are looking for the neighbourhood of a specific parcel
    <{pid}> hp:zonedAsType ?zt.
    ?zt oz:allowsUse ?u.
    ?u genprop:hasName ?allowed_use.
}}
```

Current use:

```
SELECT DISTINCT ?current_use WHERE {{
    #if we are looking for the neighbourhood of a specific parcel
    ?x hp:occupies <{pid}>;
        bdg:use [code:hasCode [genprop:hasName ?current_use]].

    }}
```

## Applicable Zoning

```
SELECT distinct ?reg ?zstring ?ctlabel ?constrained_property ?limit ?unit ?regwkt WHERE {{

    {{  #for efficiency first, evaluate this part
SELECT ?reg ?zstring ?ctlabel ?constrained_property ?limit ?unit ?regwkt ?ploc ?loc WHERE {{
    #if we are looking for the neighbourhood of a specific parcel
    <{pid}> hp:zonedForConstraint ?c;
        loc:hasLocation ?ploc.

    #regulations defined in law
    ?reg hp:definedIn ?source.
    ?source a hp:ZoningBylaw.
```

```
    #the regulation that designates the zoning type for an area
    ?reg a hp:Regulation;
    hp:appliesTo [loc:hasLocation ?loc];
    hp:specifiesConstraint ?c.
        ?c i72:hasValue ?v;
                    hp:constrains [i72:parameter_of_var [i72:hasName ?cp];
                            i72:description_of ?p];
                    #constraint type
                    rdf:type ?constraint_type.
    #constraint value
    ?v i72:hasNumericalValue ?limit.
    OPTIONAL {{?v i72:hasUnit [rdfs:label ?unit]}}
    OPTIONAL {{?reg genprop:hasName ?zstring. }} #zoning string label, if applicable
    #quantity constraint subtype (allowance, requirement, ...) to clarify the nature of the regulation
    ?constraint_type rdfs:subClassOf hp:QuantityConstraint;
         rdfs:label ?ctlabel.
    FILTER (?constraint_type != hp:QuantityConstraint)

    #location of the regulation
    ?loc geo:asWKT ?regwkt.
    #property label
    OPTIONAL{{?cp rdfs:label ?constrained_property}}
    }}
}}
        ?ploc geo:sfIntersects ?loc. #regulations can apply to multiple areas - we only want to display the area that the parcel is in.
}}
```

## Zoning Compliance

Similar to the Available Services dropdown, Zoning Compliance is performed in two stages. First, a query is run to retrieve all properties currently defined in zoning regulations in the graph. A subsequent query is run for each property to retrieve data on nearby parcels' compliance.

Property retrieval:

```
SELECT DISTINCT ?cp ?cp_label  WHERE {{

    #regulations defined in law
    ?reg hp:definedIn ?source.
```

```
?source a hp:ZoningBylaw.
#the regulation that designates the zoning type for an area
?reg a hp:Regulation;
hp:appliesTo [loc:hasLocation ?loc];
hp:specifiesConstraint ?c.
   ?c i72:hasValue ?v;
            hp:constrains [i72:parameter_of_var [i72:hasName ?cp];
                  i72:description_of ?p].
?cp rdfs:label ?cp_label.

}}
```

Compliance comparison:

```
SELECT DISTINCT ?nearbyp ?nearbypwkt ?zstring ?ctlabel ?limit ?vunit ?actualvalue ?actualunit ?compliancestatus
WHERE {{
   #if we are looking for the neighbourhood of a specific parcel
   <{pid}> loc:hasLocation [geo:asWKT ?pwkt].

   #identify nearby parcels and attribute of interest
    ?nearbyp a hp:Parcel ;
     loc:hasLocation [geo:asWKT ?nearbypwkt].

   FILTER(
     geof:distance(?pwkt, ?nearbypwkt, uom:metre) < 200
   )

   #zoning constraints that apply to nearby parcels
   ?nearbyp hp:zonedForConstraint ?c.
   #regulations defined in law
   ?reg hp:definedIn ?source.
   ?source a hp:ZoningBylaw.
   #the regulation that defines the constraint
   ?reg a hp:Regulation;
   hp:appliesTo [loc:hasLocation ?loc];
   hp:specifiesConstraint ?c.
     ?c i72:hasValue ?v;
              hp:constrains [i72:parameter_of_var [i72:hasName <{property}>];
                    i72:description_of ?p];
```

```
                        #constraint type
                        rdf:type ?constraint_type.
 #constraint value
 ?v i72:hasNumericalValue ?limit.
 OPTIONAL {{?v i72:hasUnit ?vunit.
    ?vunit rdfs:label ?unit.}}
 OPTIONAL {{?reg genprop:hasName ?zstring. }} #zoning string label, if applicable
 #quantity constraint subtype (allowance, requirement, ...) to clarify the nature of the regulation
 ?constraint_type rdfs:subClassOf hp:QuantityConstraint;
        rdfs:label ?ctlabel.
 FILTER (?constraint_type != hp:QuantityConstraint)

 #the actual value of the constrained attribute for the parcel, if known
 OPTIONAL {{
    ?nearbyp <{property}> [i72:hasValue [i72:hasNumericalValue ?actualvalue;
                                i72:hasUnit ?aunit]].
            ?aunit rdfs:label ?actualunit.
 }}
#or, the actual value of the constrained attribute for the building, if known
 OPTIONAL {{
    ?b hp:occupies ?nearbyp;
        <{property}> [i72:hasValue [i72:hasNumericalValue ?actualvalue;
                                i72:hasUnit ?aunit]].
    ?aunit rdfs:label ?actualunit.
 }}

 #logic to define a new column to identify whether a constraint is violated
 BIND(COALESCE(
 IF(!BOUND(?actualvalue) || !BOUND(?limit), "unknown",
    IF(BOUND(?vunit) && BOUND(?aunit) && ?vunit != ?aunit, "incompatible units",
    IF(?constraint_type = hp:QuantityAllowance,
        IF(?actualvalue > ?limit, "noncompliant", "compliant"),
        IF(?constraint_type = hp:QuantityRequirement,
        IF(?actualvalue < ?limit, "noncompliant", "compliant"),
        IF(?constraint_type = hp:QuantityEquivalence,
            IF(?actualvalue != ?limit, "noncompliant", "compliant"),
            "unknown"
        )
```

```
          )
        )
        )
    ),
    "unknown" # A safety net if the IF logic crashes
    ) AS ?compliancestatus)
}}
```

## Demographics Averages

```
SELECT ?avg_label (AVG(?val) AS ?avg) ?u_label
    WHERE {{
    {{
        # Population density
        ?x a cacensus:PopulationDensity2016 ;
        i72:hasValue [ i72:hasNumericalValue ?val ;
                        i72:hasUnit ?unit ] .
        cacensus:PopulationDensity2016 rdfs:label ?avg_label .
        OPTIONAL {{?unit rdfs:label ?u_label .}}
    }}
    UNION
    {{
        # Avg Income
        ?x a cacensus:AverageAfterTaxIncome25Sample2016 ;
        i72:hasValue [ i72:hasNumericalValue ?val ;
                        i72:hasUnit ?unit ] .
        cacensus:AverageAfterTaxIncome25Sample2016 rdfs:label ?avg_label .
        OPTIONAL {{ ?unit rdfs:label ?u_label . }}
    }}
    UNION
    {{
        # Total Private Dwellings
        ?x a cacensus:TotalPrivateDwellings2016 ;
        i72:hasValue [ i72:hasNumericalValue ?val ;
                        i72:hasUnit ?unit ] .
        cacensus:TotalPrivateDwellings2016 rdfs:label ?avg_label.
        OPTIONAL {{ ?unit rdfs:label ?u_label . }}
    }}
    }}
```

```
  GROUP BY ?avg_label ?u_label
```

## Service Capacity Averages

```
SELECT ?avg_label ?u_label (AVG(xsd:decimal(?cap)) AS ?avg)
WHERE {{
  # 1. Identify Leaf Service Types
  ?s a <{servicetype}> .
  <{servicetype}> rdfs:label ?avg_label .

  # 2. Capacity Data
  ?s res:hasAvailableCapacity ?avail_cap .
  ?avail_cap rdf:type ?avail_cap_type ;
          i72:hasValue ?valNode .
  ?valNode i72:hasNumericalValue ?cap .

  OPTIONAL {{
    ?valNode i72:hasUnit ?cap_unit .
    ?cap_unit rdfs:label ?u .
  }}

  # 3. Capacity Leaf Logic
  FILTER(!isBlank(?avail_cap_type) && ?avail_cap_type != owl:Thing)

  FILTER NOT EXISTS {{
    ?avail_cap rdf:type ?moreSpecific .
    ?moreSpecific rdfs:subClassOf ?avail_cap_type .
    FILTER(?moreSpecific != ?avail_cap_type && ?moreSpecific != owl:Nothing)
  }}
  ?avail_cap_type rdfs:label ?cap_type_label.
  BIND(CONCAT(?cap_type_label," (",?u,")") AS ?u_label)
}}
GROUP BY ?avg_label ?u_label
```

## Zoning Averages

```
SELECT ?ctlabel ?avg_label ?u_label (AVG(?limit) AS ?avg)
    WHERE {{
      # Regulations defined in law
```

```
    ?reg hp:definedIn ?source.
    ?source a hp:ZoningBylaw.

    # The regulation that designates the zoning type for an area
    ?reg a hp:Regulation;
      hp:specifiesConstraint ?c.

    ?c i72:hasValue ?v;
    hp:constrains [ i72:parameter_of_var [i72:hasName ?cp];
                i72:description_of ?p ];
    rdf:type ?constraint_type.

    # Constraint value
    ?v i72:hasNumericalValue ?limit.

    OPTIONAL {{ ?v i72:hasUnit [rdfs:label ?u_label] }}

    # Property label
    OPTIONAL {{ ?cp rdfs:label ?avg_label }}
    FILTER (?limit >=0) #ignore zoning with no limit
}}
GROUP BY ?ctlabel ?avg_label ?u_label
```

# Appendix G Application Screenshots

Screenshots of the results of each parcel query option are included below.

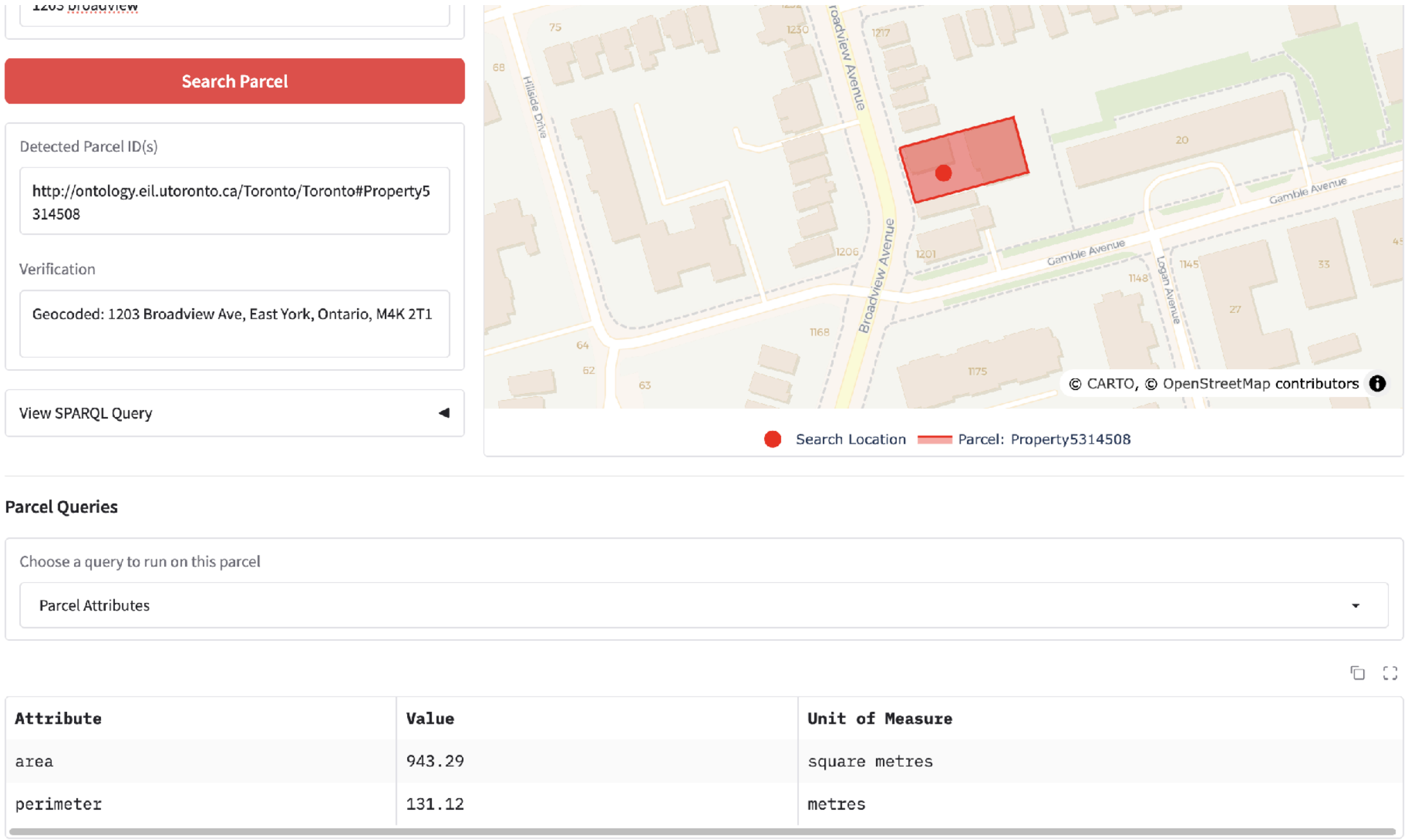

*Figure 41: Parcel Attributes result view*

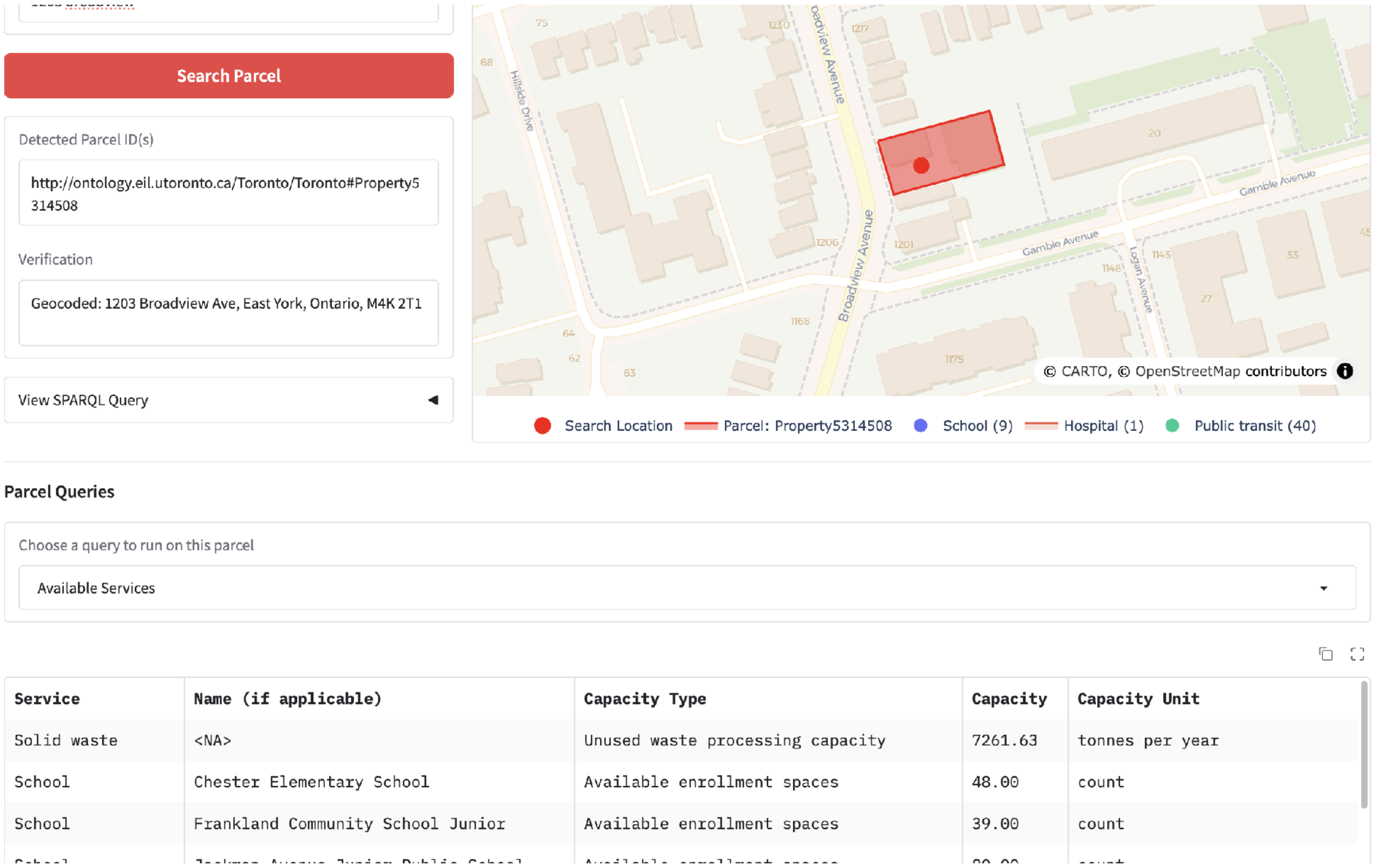

*Figure 42: Available Services result view*

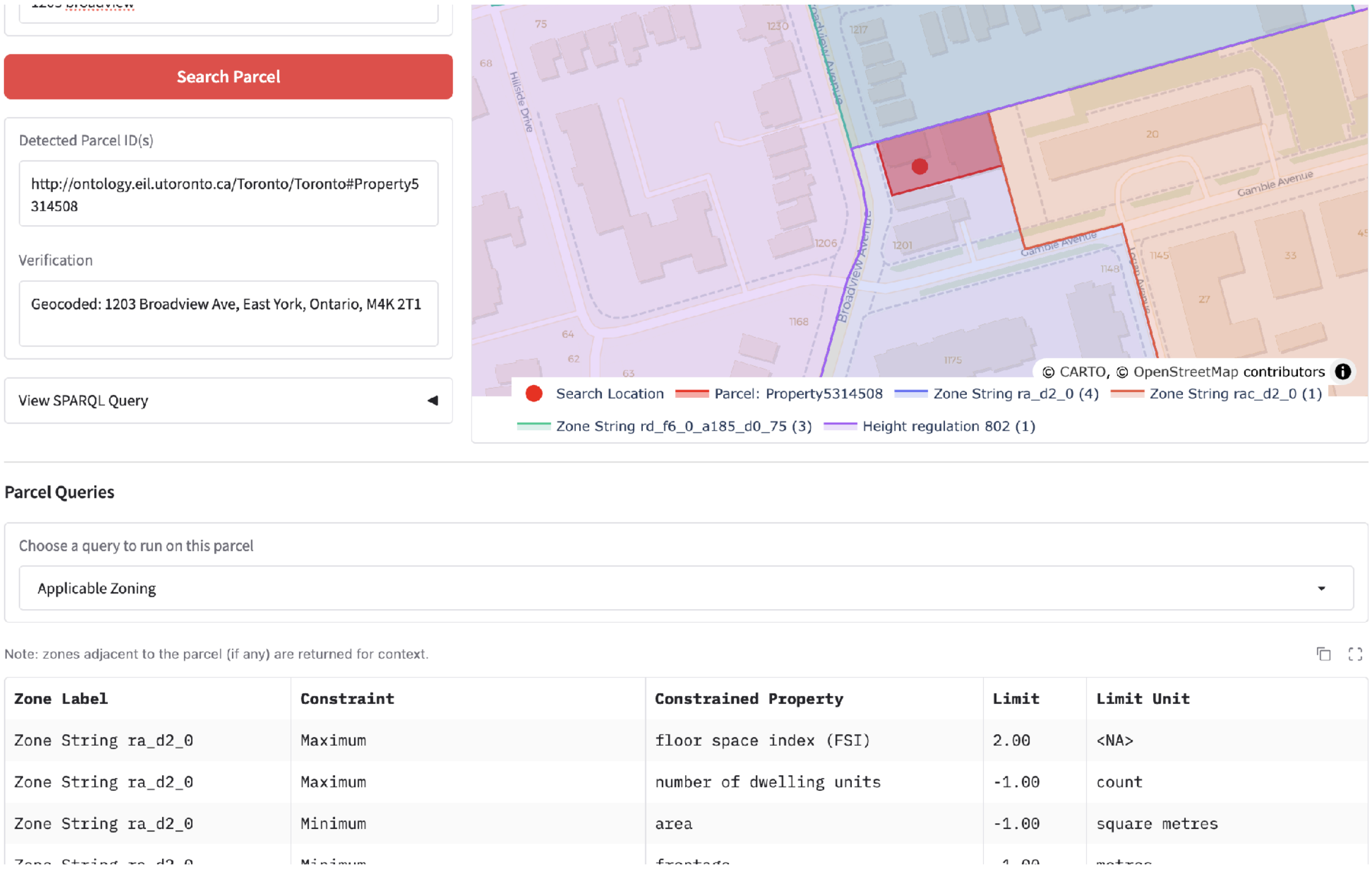

*Figure 43: Applicable Zoning result view*

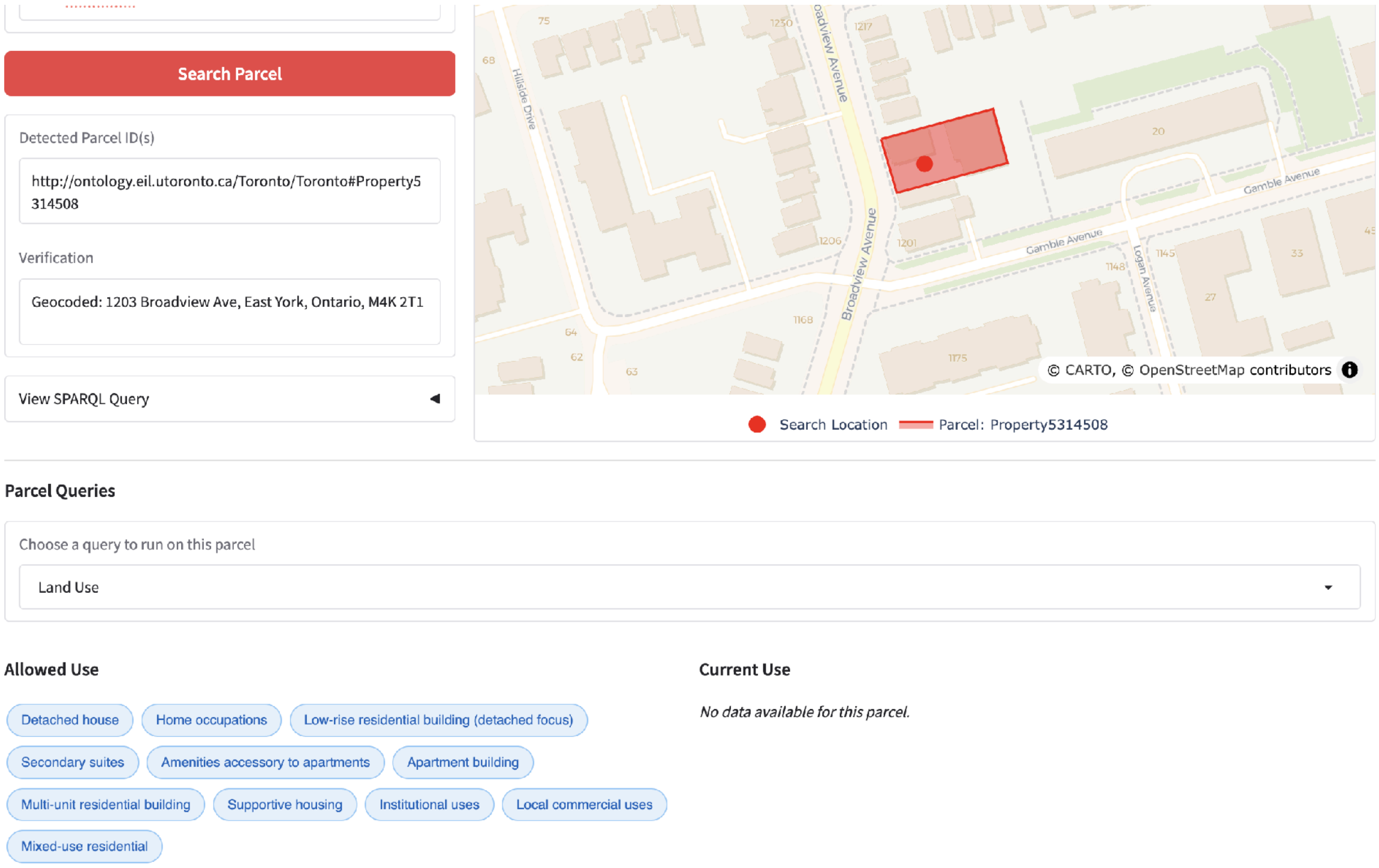

*Figure 44: Land Use result view*

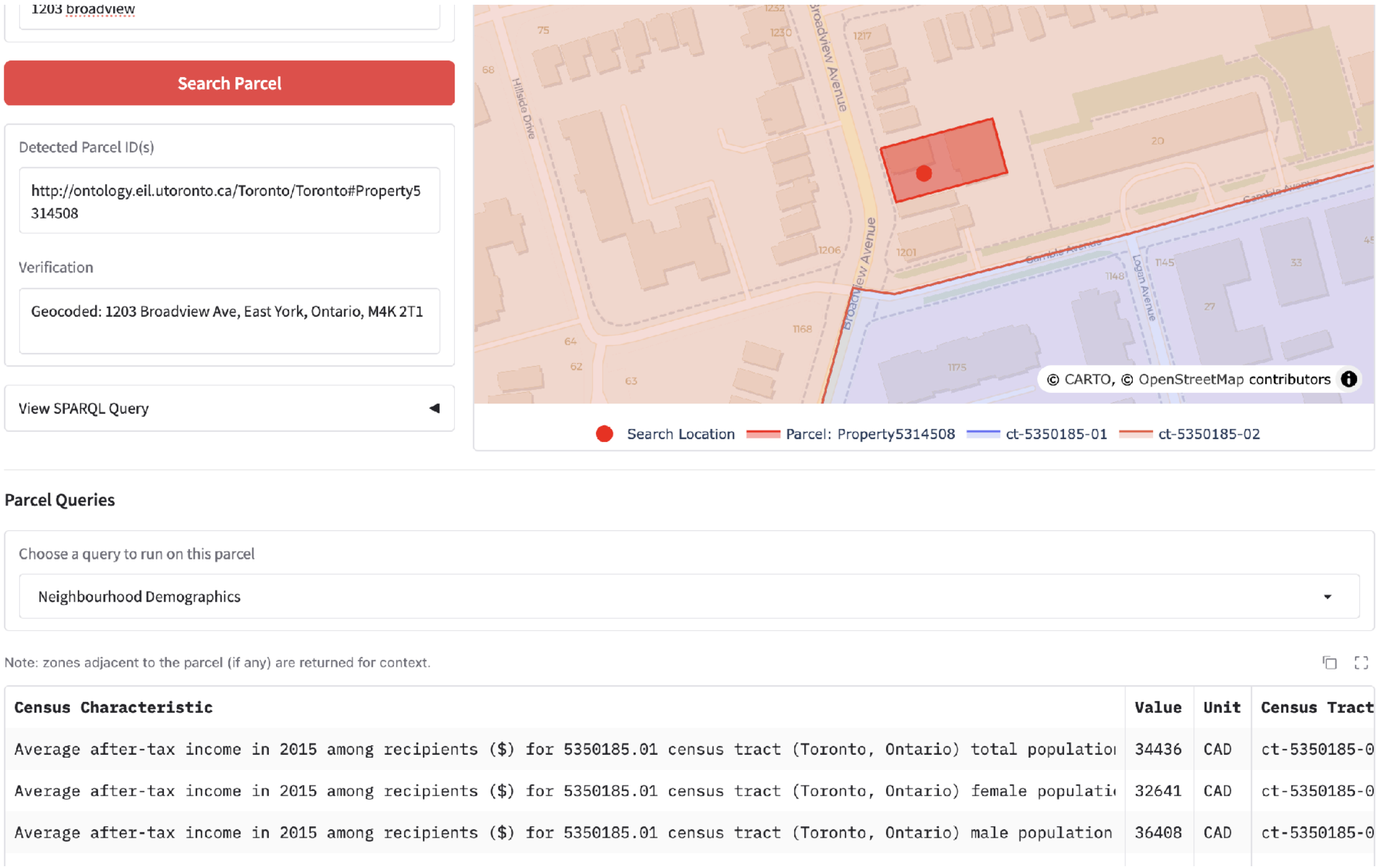

*Figure 45: Neighbourhood Demographics result view*

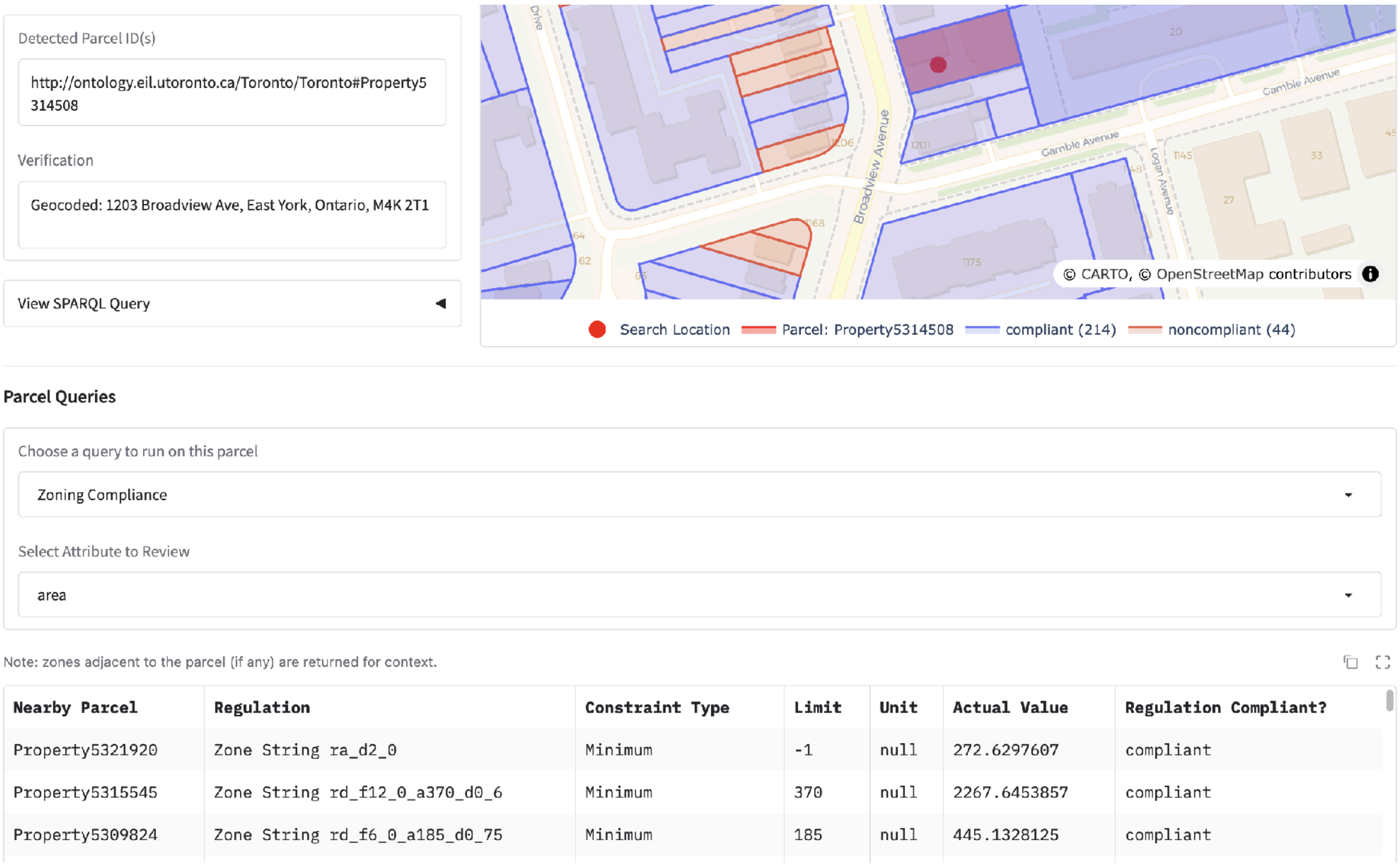

*Figure 46: Zoning compliance result view (area specific)*